\documentclass[10pt,twocolumn,letterpaper]{article}

\usepackage[pagenumbers]{cvpr} 

\usepackage{graphicx}
\usepackage{amsmath}
\usepackage{amssymb}
\usepackage{booktabs}
\usepackage{multirow}
\usepackage{color,soul}
\usepackage{xcolor,colortbl}
\usepackage[accsupp]{axessibility} 

\usepackage[pagebackref,breaklinks,colorlinks]{hyperref}

\usepackage[capitalize]{cleveref}
\crefname{section}{Sec.}{Secs.}
\Crefname{section}{Section}{Sections}
\Crefname{table}{Table}{Tables}
\crefname{table}{Tab.}{Tabs.}

\def\cvprPaperID{7818} 
\def\confName{CVPR}
\def\confYear{2022}

\usepackage{overpic}
\usepackage{enumitem} 
\usepackage{overpic} 
\usepackage{color}

\definecolor{turquoise}{cmyk}{0.65,0,0.1,0.3}
\definecolor{purple}{rgb}{0.65,0,0.65}
\definecolor{dark_green}{rgb}{0, 0.5, 0}
\definecolor{orange}{rgb}{0.8, 0.6, 0.2}
\definecolor{red}{rgb}{0.8, 0.2, 0.2}
\definecolor{darkred}{rgb}{0.6, 0.1, 0.05}
\definecolor{blueish}{rgb}{0.0, 0.3, .6}
\definecolor{light_gray}{rgb}{0.7, 0.7, .7}
\definecolor{pink}{rgb}{1, 0, 1}
\definecolor{greyblue}{rgb}{0.25, 0.25, 1}

\usepackage{blindtext}

\renewcommand{\paragraph}[1]{\vspace{1em}\noindent\textbf{#1}.}
\begin{document}
\title{CLIP-Forge: Towards Zero-Shot Text-to-Shape Generation}

\author{
Aditya Sanghi$^{1}$ \hspace{30pt} Hang Chu$^{1}$ \hspace{30pt} Joseph G. Lambourne$^{1}$ \hspace{30pt} Ye Wang$^{2}$ \\ 
Chin-Yi Cheng$^{1}$ \hspace{40pt} Marco Fumero$^{1}$ \hspace{40pt} Kamal Rahimi Malekshan$^{1}$\\
$^{1}$Autodesk AI Lab \hspace{40pt} $^{2}$Autodesk Research\\
\texttt{\scriptsize{\{aditya.sanghi,hang.chu,joseph.lambourne,ye.wang,chin-yi.cheng,marco.fumero,kamal.malekshan\}@autodesk.com}} \\
{\color{magenta} https://github.com/AutodeskAILab/Clip-Forge}
}
\twocolumn[{%
\renewcommand\twocolumn[1][]{#1}%
\maketitle
\begin{center}
\centering
\captionsetup{type=figure}
\setlength{\tabcolsep}{0pt}
\includegraphics[width=\linewidth]{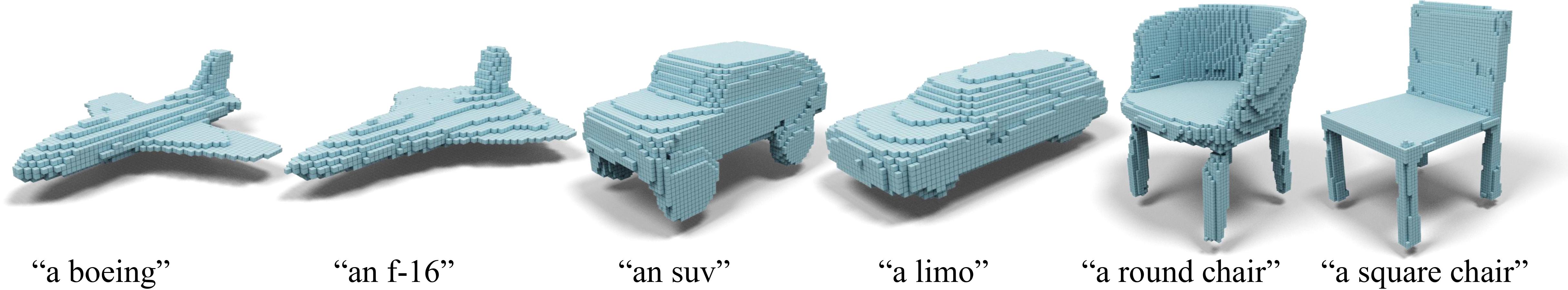}
\captionof{figure}{We propose a zero-shot text-to-shape generation method named CLIP-Forge. Without training on any shape-text pairing labels, our method generates meaningful shapes that correctly reflect the common name, (sub-)category, and semantic attribute information.}
\label{fig:teaser}
\end{center}%
}]
\begin{abstract}
Generating shapes using natural language can enable new ways of imagining and creating the things around us. While significant recent progress has been made in text-to-image generation, text-to-shape generation remains a challenging problem due to the unavailability of paired text and shape data at a large scale. We present a simple yet effective method for zero-shot text-to-shape generation that circumvents such data scarcity. Our proposed method, named CLIP-Forge, is based on a two-stage training process, which only depends on an unlabelled shape dataset and a pre-trained image-text network such as CLIP. Our method has the benefits of avoiding expensive inference time optimization, as well as the ability to generate multiple shapes for a given text. We not only demonstrate promising zero-shot generalization of the CLIP-Forge model qualitatively and quantitatively, but also provide extensive comparative evaluations to better understand its behavior.
\end{abstract}
\section{Introduction}
\label{sec:intro}

Generating 3D shapes from text input has been a challenging and interesting research problem with both significant scientific and applied value~\cite{ Han_Shang_Wang_Liu_Zwicker_2019, Lin2018GeneratingAV, Huang_Lee_Chen_Liu_2021, Huq3DSceneGenerationFromText}. 
In the artificial intelligence and cognitive science research communities, researchers have long sought to bridge the two modalities of natural language and geometric shape~\cite{tang2021part2word, Achlioptas2018LearningTR}. 
In practice, text-to-shape generation models are a key enabling component to new smart tools in creative design and manufacture as well as animation and games~\cite{chen2018text2shape}.

Significant progress has been made to connect text and image modalities~\cite{frome2013devise, socher2014grounded, karpathy2014deep, joulin2016learning, desai2021virtex,sariyildiz2020learning}. Recently, DALL-E~\cite{ramesh2021zeroshot} and its associated pre-trained visual-textual embedding model CLIP~\cite{radford2021learning} has shown promising results on the problem of text-to-image generation \cite{patashnik2021styleclip}. Notably, they have demonstrated strong zero-shot generalization while evaluated on tasks the model has not been specifically trained on. Shape generation is a more fundamental problem than image generation, because images are projections and renderings of the inherently 3D physical world. Therefore, one may wonder if the success in 2D can be transferred to the 3D domain. This turns out to be a non-trivial problem. Unlike the text-to-image case, where paired data is abundant, it is impractical to acquire huge paired datasets of texts and shapes.

\begin{figure}[t!]
\centering
\includegraphics[width=1.0\linewidth]{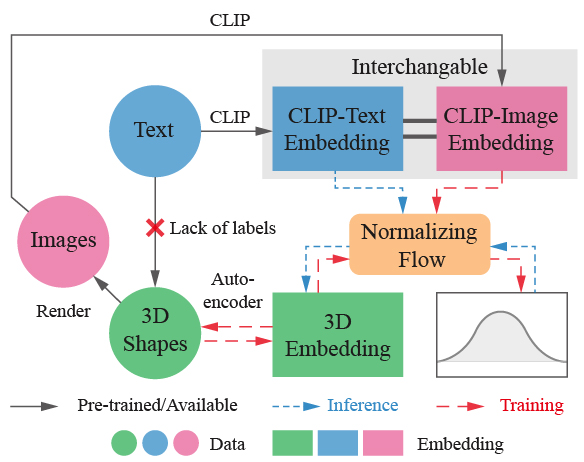}
\caption{Illustration of the main idea.  It is difficult to directly learn text-to-shape generation due to the lack of paired data. Instead, we use renderings of shapes with a pre-trained image-text joint embedding model to bridge the data gap between 3D shapes and natural language.}
\label{fig:concept}
\end{figure}

Leveraging the progress of text-to-image generation, we present CLIP-Forge. As shown in Figure~\ref{fig:concept},
we overcome the limitation of shape-text pair data scarcity via a simple and effective approach. We exploit the fact that 3D shapes can be easily and automatically rendered into images using standard graphics pipelines. We then utilize pre-trained image-text joint embedding models such as~\cite{radford2021learning, jia2021scaling},
which bring text and image embeddings in a similar latent space so that they can be used interchangeably. Hence, we can train a model using image embeddings, but at inference time replace it with text embeddings. 

In CLIP-Forge, we first obtain a latent space for shapes via training an autoencoder, and we then train a normalizing flow network~\cite{dinh2016density} to model the distribution of shape embeddings conditioned on the image features obtained from the pre-trained image encoder~\cite{radford2021learning}. 
We use the renderings of 3D shapes and hence, no labels are required for training our model.
During inference, we obtain text features of the given text query via the pre-trained text encoder. We then condition the normalizing flow network with text features to generate a shape embedding, which is converted into 3D shape through the shape decoder. In this process, CLIP-Forge requires no text labels for shapes, which means it can be extended easily to larger datasets. Since our method is fully feed-forward, it also has the advantage of avoiding the expensive inference time optimizations as employed in existing 2D approaches~\cite{frans2021clipdraw,wang2018pixel2mesh}. 

The main contributions of this paper are as follows:
\begin{itemize}
\itemsep0em 
\item We present a new method, CLIP-Forge, that generates 3D shapes directly from text as shown in Figure \ref{fig:teaser}, without requiring paired text-shape labels.

\item  Our method has an efficient generation process requiring no inference time optimization, can generate multiple shapes for a given text and can be easily extended to multiple 3D representation.
\item We provide extensive evaluation of our method in various zero-shot generation settings qualitatively and quantitatively. 
\end{itemize}

\section{Related Work}
\label{sec:related}

\noindent \textbf{Zero-Shot Learning.} Zero-shot learning is an important paradigm of machine learning, which typically aims to make predictions on classes that have never been observed during training, by exploiting certain external knowledge source.
The literature originates from the image classification problem~\cite{Palatucci2009,Lampert2009}, and has been recently extended to generative models, in particular, the task of synthesizing images from text~\cite{ramesh2021zeroshot}. 
As far as our best knowledge, our method is the first to bring this paradigm to the 3D shape domain, which enables efficient shape generation from natural language text input.

\noindent \textbf{Applications of CLIP.} A major building block of our method is CLIP~\cite{radford2021learning}, which shows groundbreaking zero-shot capability using a mechanism to connect text and image by bringing them closer in the latent space. 
Previous work such as ALIGN~\cite{jia2021scaling}, has used a similar framework on noisy datasets. Recently, pre-trained CLIP has been used for several zero-shot downstream applications~\cite{jia2021scaling, fang2021clip2video, shen2021much, patashnik2021styleclip, frans2021clipdraw, pakhomov2021segmentation}. The most similar previous work to ours is zero-shot image and drawing synthesis~\cite{patashnik2021styleclip, frans2021clipdraw}. Typically, these methods involve iteratively optimizing a random image to increase certain CLIP activations. There is still no clear way to apply them to 3D due to the significantly higher complexity. Our approach conditions a shape prior network with CLIP features, which has the advantages of significant speed-up and the capability to generate multiple shapes from a single text. 

\noindent \textbf{3D Shape Generation and Language.}
Recently, there has been tremendous progress in 3D shape generating in different data formats such as point cloud~\cite{achlioptas2018learning,li2018point,yang2019pointflow}, voxel~\cite{3DGAN2016}, implicit representation~\cite{OccupancyNetworks2019,Park_2019_CVPR,chen2018implicit_decoder} and mesh~\cite{nash2020polygen}. 
While our method is not limited to produce one 3D data format, we mainly adopt the implicit representation in this work due to their simplicity and superior quality.
More recently, methods that use text to localize objects in 3D scene have been explored~\cite{chen2020scanrefer, achlioptas2020referit3d}. 
A metric learning method for text-to-shape generation is presented in \cite{chen2018text2shape}.
The main difference and advantage of our approach is its zero-shot capability which requires no text-shape labels.

\noindent \textbf{Multi-Stage Training.} In this work we follow a multi-stage training approach, where we first learn the embeddings of the target data and then learn a probabilistic encoding model for the learned embeddings. Such an approach as been explored in image generation~\cite{oord2017neural,liu2019acceleration,esser2021taming} and 3D shape generation~\cite{achlioptas2018learning,chen2018implicit_decoder}. Concretely for CLIP-Forge, we first train a 3D shape autoencoder and then model the embeddings using normalizing flow. 

\begin{figure*}[t!]
\centering
\includegraphics[width=0.9\textwidth]{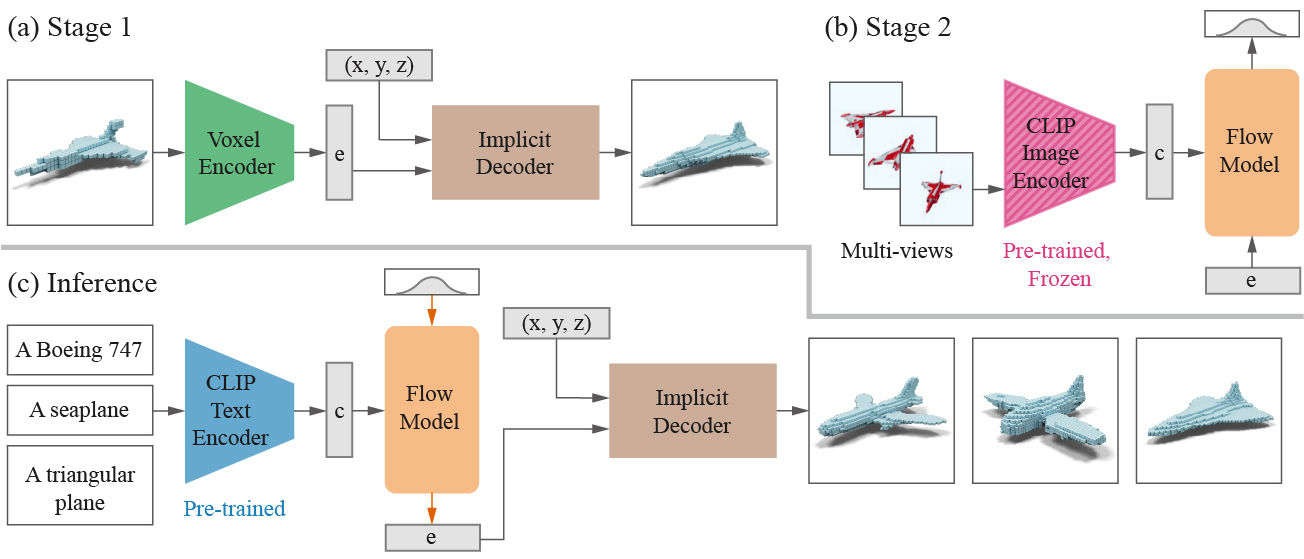}
\caption{An overview of the CLIP-Forge method. The top row shows the two training stages of shape autoencoder training, and the conditional normalizing flow training. The bottom row shows how text-to-shape inference is conducted.}
\label{fig:method}
\end{figure*}

\noindent \textbf{Normalizing Flow.} Generative models have extensive use cases such as content creation and editing. Flow-based generative networks~\cite{rezende2015variational,dinh2016density,dinh2015nice} is able to perform exact likelihood evaluation, while being efficient to sample from. 
They have been widely applied to a variety of tasks ranging from image generation~\cite{kingma2018glow}, audio synthesis~\cite{kim2019flowavenet} and video generation~\cite{kumar2020videoflow}. Recently, normalizing flow has been brought to the 3D domain enabling fast generation of point clouds~\cite{yang2019pointflow,pumarola2020cflow}. In this paper, we employ a normalizing flow model~\cite{dinh2016density} to model the conditional distribution of latent shape representations given text and image embeddings.
\section{Method}

Our method requires a collection of 3D shapes without any associated text labels, which takes the format of $\mathcal{S} = \{(\mathbf{I}_n, \mathbf{V}_n, \mathbf{P}_n, \mathbf{O}_n)\}_{n=1}^{N}$. Each shape in the collection $\mathcal{S}$ is comprised of a rendered image $\mathbf{I}_n$, a voxel grid $\mathbf{V}_n$, a set of query points in the 3D space $\mathbf{P}_n$, and space occupancies $\mathbf{O}_n$. As an overview, the CLIP-Forge training has two stages. In the first stage, we train an autoencoder with a voxel encoder and an implicit decoder. Once the autoencoder training is completed we obtain a shape embedding $\mathbf{e}_n$ for each 3D shape in $\mathcal{S}$. In the second stage, we train a conditioned normalizing flow network to model and generate $\mathbf{e}_n$, which is conditioned with image features obtained from the CLIP image encoder using $\mathbf{I}_n$. 
During inference, we first convert the text to the interchangeable text-image latent space using the CLIP text encoder. We then condition the normalizing flow network with the given text features and a random vector sampled from the uniform Gaussian distribution to obtain a shape embedding. Finally, this shape embedding is converted to a 3D shape using the implicit decoder. The overall architecture is shown in Figure~\ref{fig:method}.

\subsection{Stage 1: Shape Autoencoder}
The autoencoder consists of an encoder and a decoder.
We use an encoder $f_V$ to extract the shape embedding $\mathbf{e}_n$ for the training shape collection, using $\mathbf{V}_n$ of resolution $32^3$ as the input. 
We use a simple voxel network that comprises of a series of batch-normalized 3D convolution layers followed by linear layers. 
This can be written as:
\begin{align}
\begin{split}
\mathbf{e}_n = f_V(\mathbf{V}_n) + \epsilon, \quad \text{where}  \quad \epsilon \sim \mathcal{N}(0, 0.1) 
\end{split}
\label{equa:noise_reg}
\end{align}
where $\mathbf{e}_n$ is augmented with a Gaussian noise. We find empirically injecting this noise improves the generation quality as later shown in the ablation study. This is also theoretically verified to improve results for conditional density estimation~\cite{rothfuss2019noise}. 
We then pass $\mathbf{e}_n$ through an implicit decoder. Our decoder architecture is inspired by the Occupancy Networks~\cite{OccupancyNetworks2019}, which takes concatenated $\mathbf{e}_n$ and $\mathbf{P}_n$ as input. Our implicit decoder consists of linear layers with residual connections and predicts $\mathbf{O}_n$. We use a mean squared error loss between the predicted occupancy and the ground truth occupancy. 
Our framework is flexible and can be adapted to different forms of architectures. To showcase this,
we use a PointNet~\cite{qi2017pointnet} as the encoder and a FoldingNet~\cite{yang2018foldingnet} as the decoder that generates point clouds instead of occupancies,
which are trained with a Chamfer loss~\cite{achlioptas2018learning}. 

\subsection{Stage 2: Conditional Normalizing Flow}
We train a normalizing flow network using $\mathbf{e}_n$ and its corresponding rendered images $\mathbf{I}_n$.
Note that each $\mathbf{I}_n$ can include multiple images of the same shape from different rendering settings, such as changing camera viewpoints. 
We model the conditional distribution of $\mathbf{e}_n$ using a RealNVP network~\cite{dinh2016density} with five layers, which transforms the distribution of $\mathbf{e}_n$ into a normal distribution.
We obtain the condition vector $\mathbf{c}_n$ by passing $\mathbf{I}_n$ through the ViT~\cite{dosovitskiy2020image} based CLIP image encoder $f_{I}$, whose weights are frozen after pre-training. $\mathbf{c}_n$ is concatenated with the transformed feature vector at each scale and translation coupling layers of RealNVP: 
\begin{equation}
\mathbf{c}_n =  f_{I}(\mathbf{I}_n)  \text{,} \quad \mathbf{z}_n^{1:d} =  \mathbf{e}_n^{1:d}   \quad \text{and} 
\end{equation}
\begin{equation}
 \mathbf{z}_n^{d+1:D} =  \mathbf{e}_n^{d+1:D}  \odot \exp\big(s([\mathbf{c}_n; \mathbf{e}_n^{1:d}])\big) + t([\mathbf{c}_n; \mathbf{e}_n^{1:d}])
\end{equation}
where $s$ and $t$ stand for the scale and translation function parameterized by a neural network. 
The intuition here is we split the object embedding $\mathbf{e}_n$ into two parts where one part is modified using a neural network that is simple to invert, but still dependent on the remainder part in a non-linear manner. The splitting can be done in several ways by using a binary mask~\cite{dinh2016density}.
In particular, we investigate two strategies: \textit{checkerboard} masking and \textit{dimension-wise} masking.
The checkerboard masking has value 1 where the sum of spatial coordinates is odd, and 0 otherwise.
The dimension-wise mask has value 1 for the first half of latent vector, and 0 for the second half. 
The masks are reversed after every layer.
Finally, we impose a density estimation loss on the shape embeddings as:
\begin{align*}
\label{eq:change-variables}
 \log\left(p(\mathbf{e}_n)\right) = \log\Big(p\big(\mathbf{z}_n\big)\Big) + \log\left(\left|\det\left(\frac{\partial f(\mathbf{e}_n)}{\partial \mathbf{z}_n^T}\right)\right|\right)
\end{align*}
where $f$ is the normalizing flow model, and $\partial f(\mathbf{e}_n)/\partial \mathbf{z}_n^T$ is the Jacobian of $f$ at $\mathbf{e}_n$ \cite{dinh2016density}. We model the latent distribution $p(\mathbf{z}_n)$ as an unit Gaussian distribution. 

\subsection{Inference} 
During the inference phase, we convert a text query $\textbf{t}$ into the text embedding using the CLIP text encoder, $f_{T}$. As the CLIP image and text encoders are trained to bring the image and text embeddings in a joint latent space, we can simply use the text embedding as the condition vector for the normalizing flow model, i.e. $\mathbf{c}$=$f_{T}(\textbf{t})$. 
Once we obtain the condition vector we can sample a vector from the normal distribution and use the reverse path of the flow model to obtain a shape embedding in $p(\mathbf{e}_n)$.
The normal distribution allows us to sample multiple times to obtain multiple shape embeddings for a given text query. 
We obtain the mean shape embedding by using the mean of the normal distribution. The mean shape embedding represents the prototype for a given text query.
These shape embeddings are then converted to 3D shapes using the implicit decoder trained in stage 1.

\section{Experiments}

In this section, we first describe the experimental setup and then show qualitative and quantitative results. More results can be found in the supplementary material.

\noindent \textbf{Dataset.} For all of our experiments, we use the ShapeNet(v2) dataset~\cite{chang2015shapenet} which consists of 13 rigid object classes. We use the processed version of the data which consists of rendered images, voxel grids, query points and their occupancies from shapes as provided in \cite{choy20163d,OccupancyNetworks2019}.

\noindent \textbf{Implementation Details.} For both training stages, we use the Adam optimizer~\cite{kingma2014adam} with a learning rate of 1e-4 and a batch size of 32. We train the stage 1 autoencoder for 300 epochs whereas we train the stage 2 conditional normalizing flow model for 100 epochs. For all the experiments below we use a latent size of 128 with a BatchNorm~\cite{ioffe2015batch} based voxel encoder and a ResNet based decoder inspired by the Occupancy Network~\cite{OccupancyNetworks2019}.  We use a RealNVP~\cite{dinh2016density} based network with dimension-wise masking for the flow model. The design decisions are discussed in the ablation study section and further details are provided in the supplemental material.

\noindent \textbf{Evaluation Metrics.} To evaluate our method thoroughly, we consider four criteria and several metrics for those criteria respectively. Furthermore, for some criteria we manually define a set of 234 text queries (or prompts). These queries include direct hyponyms for the ShapeNet categories from the WordNet~\cite{wordnet} taxonomy, sub-categories and relevant shape attributes for a given category (e.g. a round chair, a square table, etc.) across the ShapeNet(v2) dataset. The text queries are listed in the appendix. The criteria are as follows:
\begin{enumerate}
\itemsep0em 
    \item \textbf{Reconstruction Quality.} This criteria is mainly used  to check the reconstruction capabilities of the stage 1 autoencoder on the test set. We use two metrics: Mean Square Error (MSE) on 30,000 sample query points and Intersection over Union (IOU) with $32^3$ voxel shapes.
    
    \item \textbf{Generation Quality.} We use this criteria to evaluate the quality of generated shapes on text queries. We consider two metrics: Fréchet inception distance (FID) \cite{heusel2017gans} and Maximum Measure Distance (MMD) using IOU . To calculate FID and MMD, we first take 224 text queries as mentioned above and generate a mean shape embedding for each text query. We then generate $32^3$ resolution 3D objects for all the text queries. For FID, we compare the generated 3D shapes with the test dataset of ShapeNet. FID depends on a pre-trained network, for which we train a voxel classifier on the 13 ShapeNet classes and use the feature vector from the fourth layer. We provide more details in the appendix. In the case of MMD, for each generated shape we match a shape in the test dataset based on the highest IOU. We then average the IOU across all the text queries. Note, MMD is a variation of the Minimum Measure Distance as described in \cite{achlioptas2018learning}, which we believe is more suitable for implicit representations as we do not need to sample the surface.

    
    \item \textbf{Diversity Across Categories.} To make sure we generate shapes across categories we design a new criteria. First, we generate the shapes based on the text queries as mentioned above. For each text query we have an assigned label. We then pass the generated voxels through the same classifier used to calculate the FID metric. We then report the accuracy  based on the assigned label. We refer to this metric as Acc. throughout the text. Also note that the FID metric  gives a good measure for diversity as we compare it with the test distribution.

    \item \textbf{Human Perceptual Evaluation.} To evaluate CLIP-Forge's ability to provide control over the generated shape using attribute, common name, and sub-category information from the text prompt, we conducted a perceptual evaluation using Amazon SageMaker Ground Truth and crowd workers from Mechanical Turk~\cite{mishra_2019}. More detail is provided in section~\ref{sec:evaluation}.

\end{enumerate}

\subsection{Comparison with Supervised Models}

\begin{table}
\centering
\setlength{\tabcolsep}{7pt}
{\small
\begin{tabular}{l|ccc}
\toprule
\textit{method}  & \textbf{FID$\downarrow$} & \textbf{MMD$\uparrow$} & \textbf{Acc.$\uparrow$}\\
\midrule
text2shape-CMA~\cite{chen2018text2shape} & 16078.05 &	0.4992 & 4.27\\ \hline 
text2shape-supervised~\cite{chen2018text2shape} & 14881.96 & 0.1418 &	6.84\\ \hline 
`CLIP-Forge (ours) &  \textbf{2425.25} & \textbf{0.6607} & \textbf{83.33} \\
\bottomrule
\end{tabular}
}
\caption{Comparing CLIP-Forge with supervised models using the text2shape dataset.}
\label{tab:baseline}
\end{table}

We compare CLIP-Forge with text-to-shape generation models that are trained with direct supervision signals. The only existing paired text-shape dataset is provided by Text2Shape~\cite{chen2018text2shape}, which contains 56,399 natural language descriptions for ShapeNet objects within the chair and table categories. We train two supervised models using the Text2Shape dataset: text2shape-CMA uses the cross-modality alignment loss described in \cite{chen2018text2shape} and text2shape-supervised uses a direct MSE loss in the embedding space. For both supervised baseline methods, we use the same CLIP text encoder and occupancy network shape encoder and decoder to ensure fair comparison. Table~\ref{tab:baseline} shows the result on our text query set. It can be seen that CLIP-Forge significantly outperforms both supervised baselines in all evaluation metrics. In particular, we observe that text2shape-CMA generates generic shapes such as boxes and spheres that do not resemble specific objects. The text2shape-supervised baseline fails to generalize and tends to generate chair- and table- like shapes that it is trained on, despite the text query is irrelevant to these two categories. 

\begin{figure*}[t!]
\begin{center}
\setlength{\tabcolsep}{2pt}
\small{
\begin{tabular}{cccccc}
\includegraphics[width=0.15\linewidth]{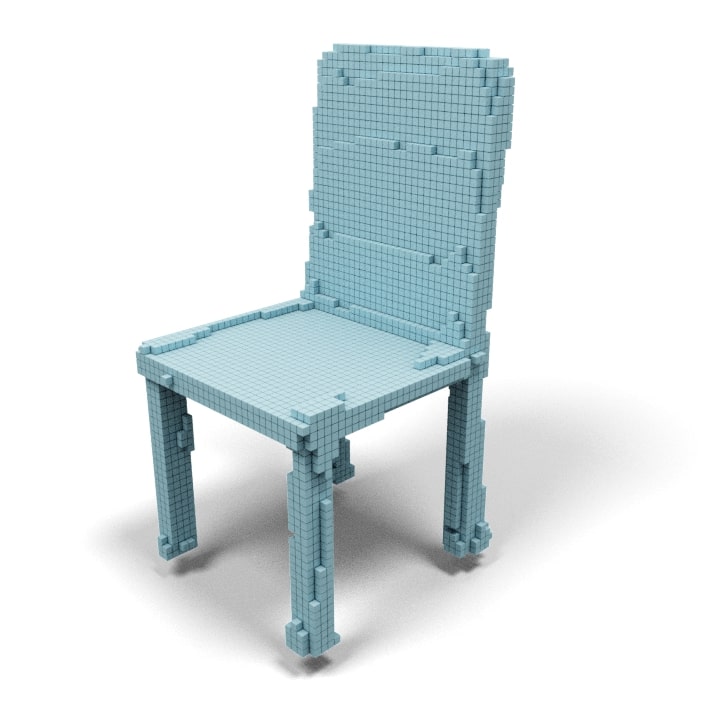} &
\includegraphics[width=0.15\linewidth]{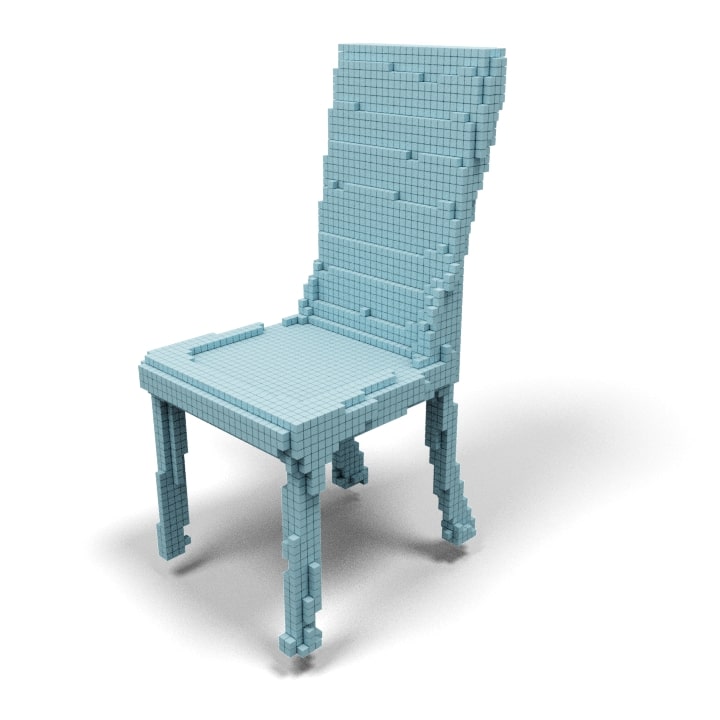} &
\includegraphics[width=0.15\linewidth]{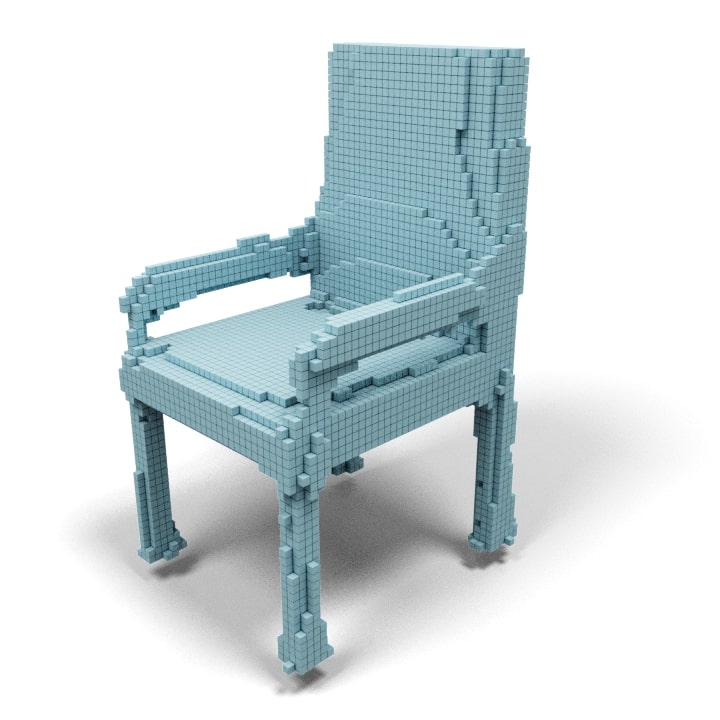} &
\includegraphics[width=0.15\linewidth]{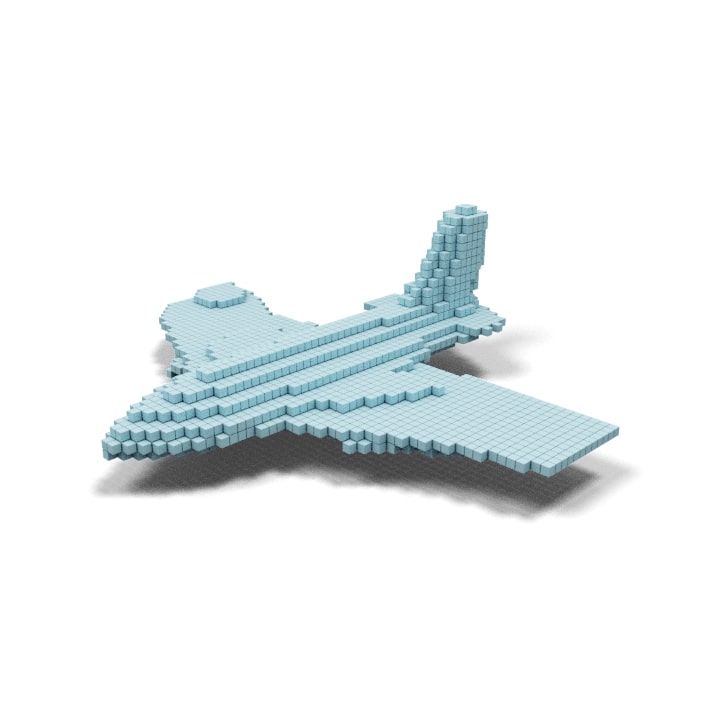} &
\includegraphics[width=0.15\linewidth]{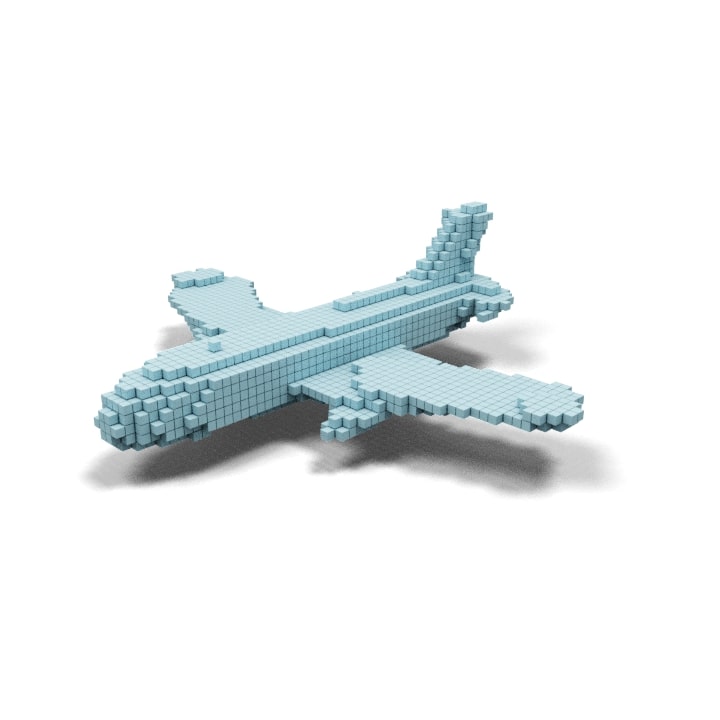} &
\includegraphics[width=0.15\linewidth]{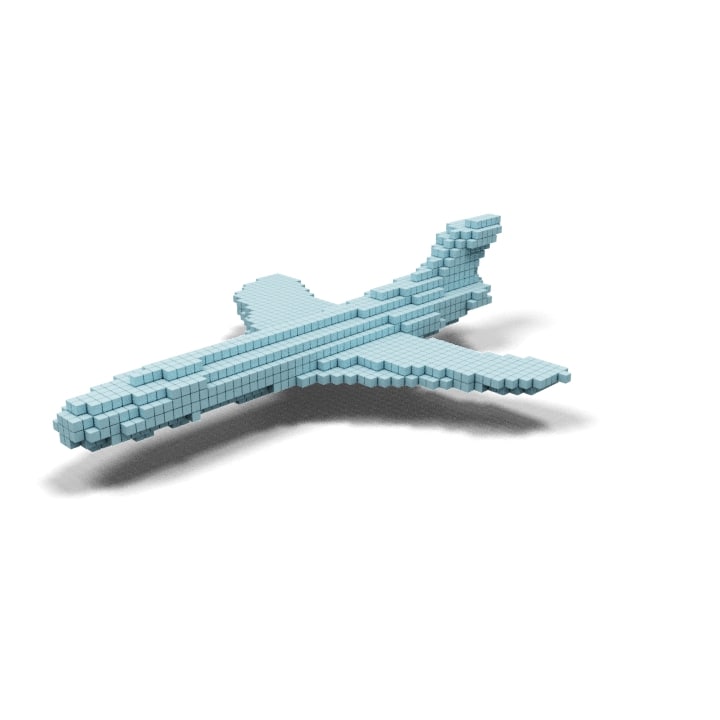}\\
\multicolumn{3}{c}{``a chair''} & \multicolumn{3}{c}{``a plane''}\\

\includegraphics[width=0.15\linewidth]{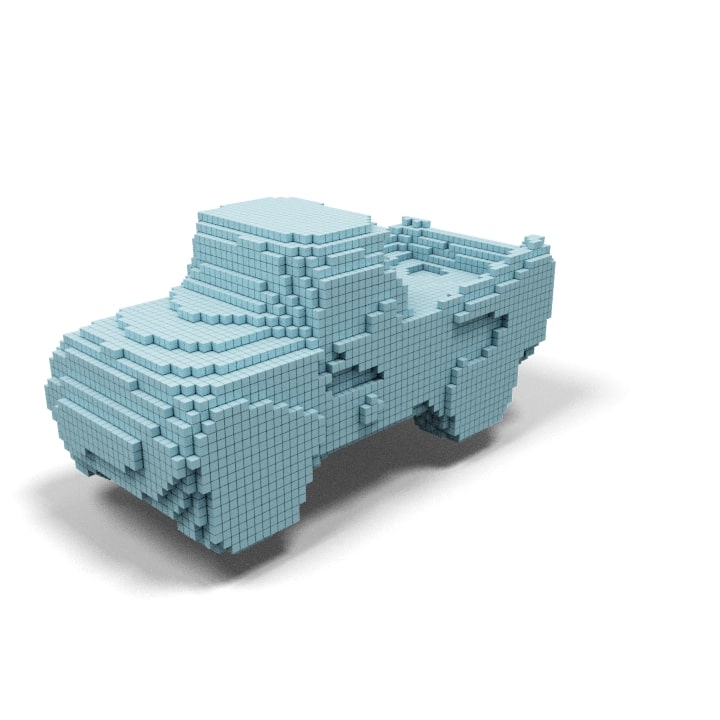} &
\includegraphics[width=0.15\linewidth]{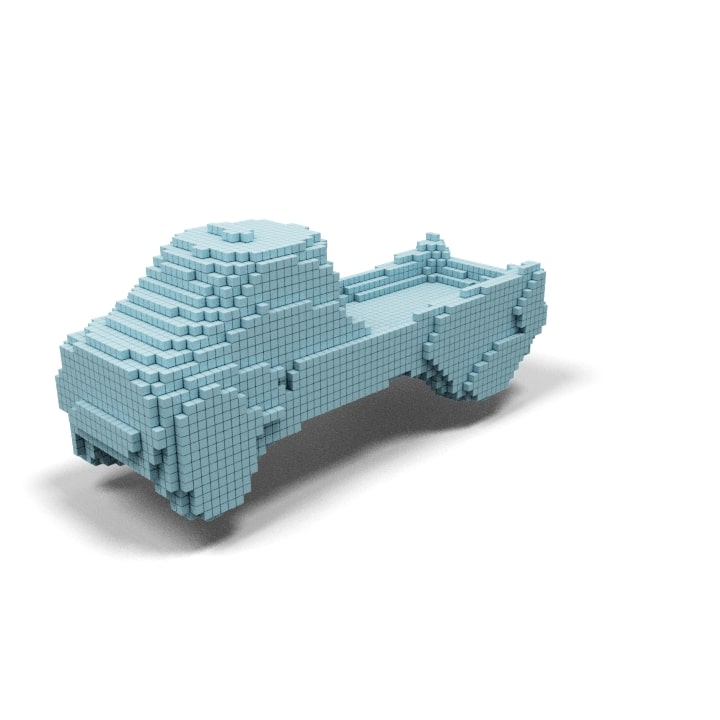} &
\includegraphics[width=0.15\linewidth]{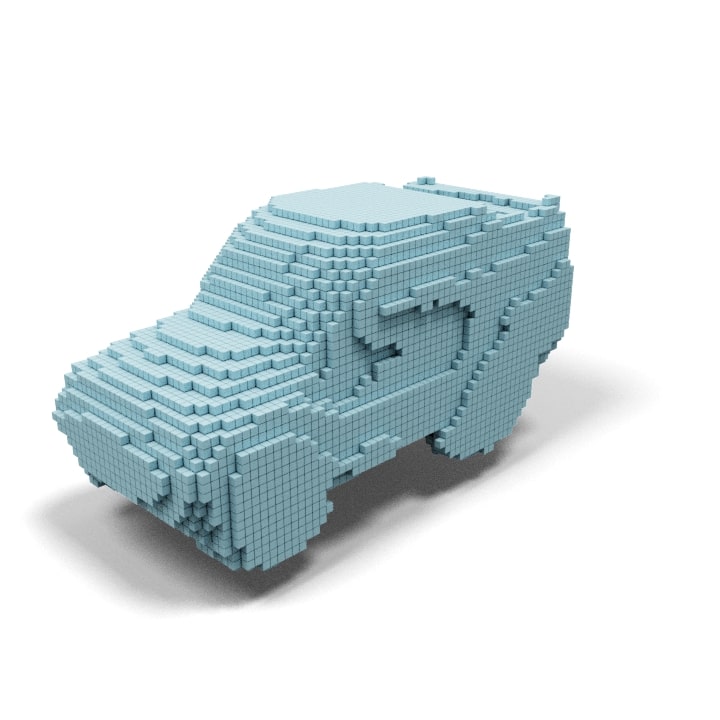} &
\includegraphics[width=0.15\linewidth]{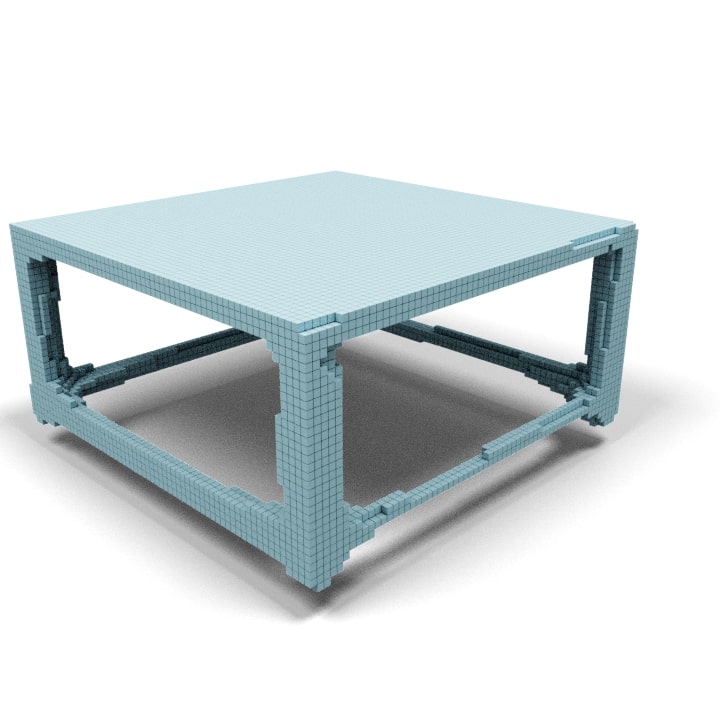} &
\includegraphics[width=0.15\linewidth]{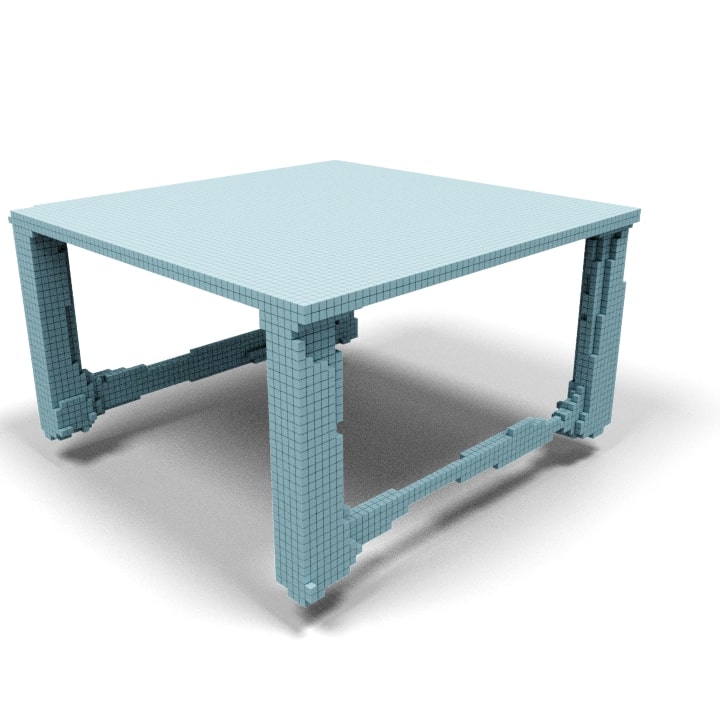} &
\includegraphics[width=0.15\linewidth]{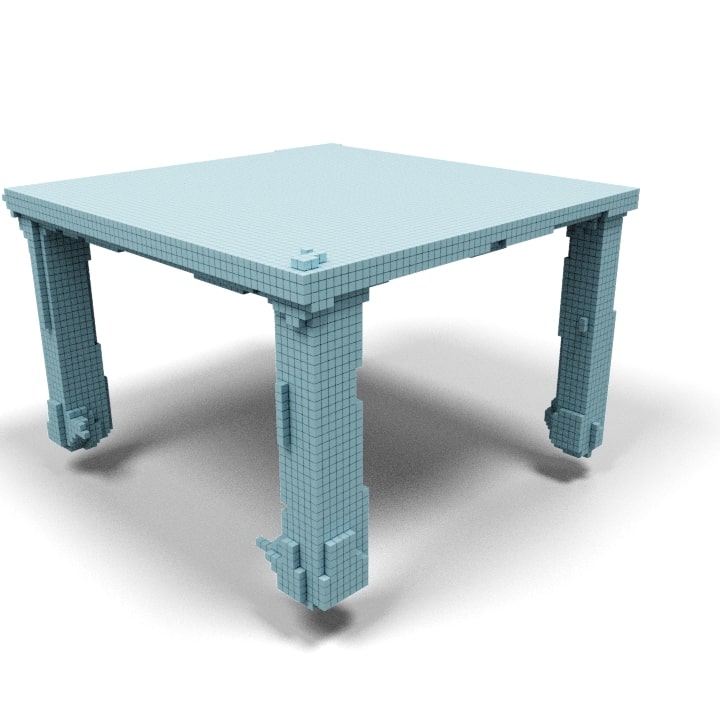}\\
\multicolumn{3}{c}{``a truck''} & \multicolumn{3}{c}{``a square table''}\\
\end{tabular}
}
\end{center}
  \caption{ Our method can generate multiple examples given a text query. In this case, we are generating 3 shapes for a given text prompt. }
\label{fig:qual_multiple}
\end{figure*}

\begin{figure*}[t!]
\begin{center}
\setlength{\tabcolsep}{2pt}
\small{
\begin{tabular}{cccccc}
\includegraphics[width=0.15\linewidth]{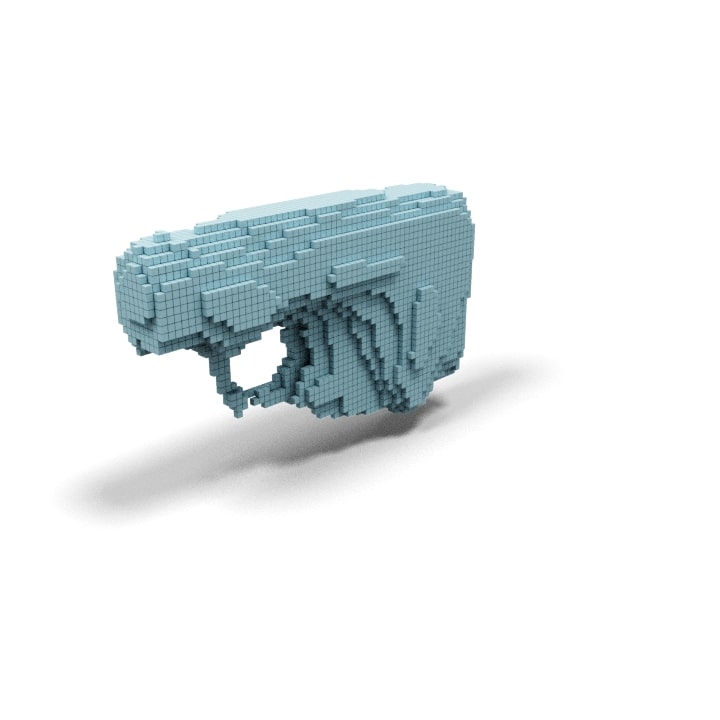} &
\includegraphics[width=0.15\linewidth]{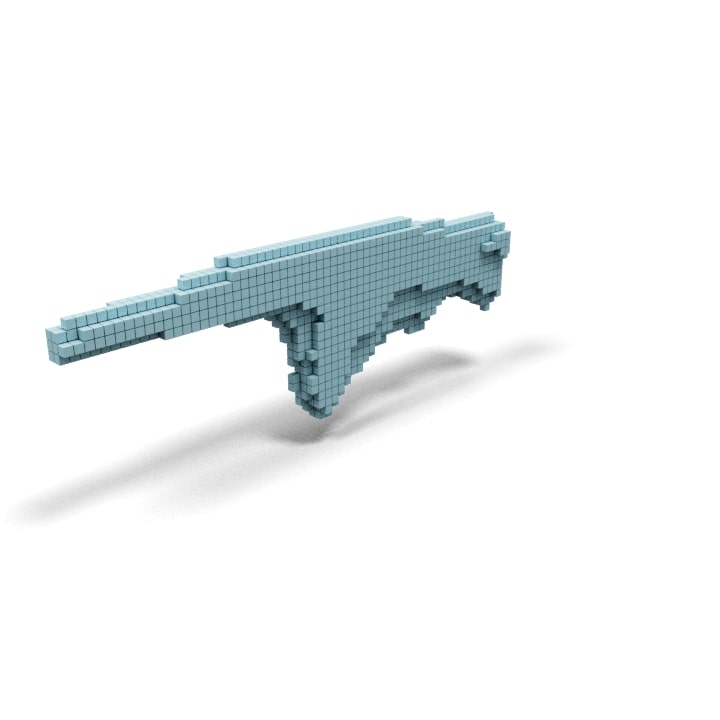} &
\includegraphics[width=0.15\linewidth]{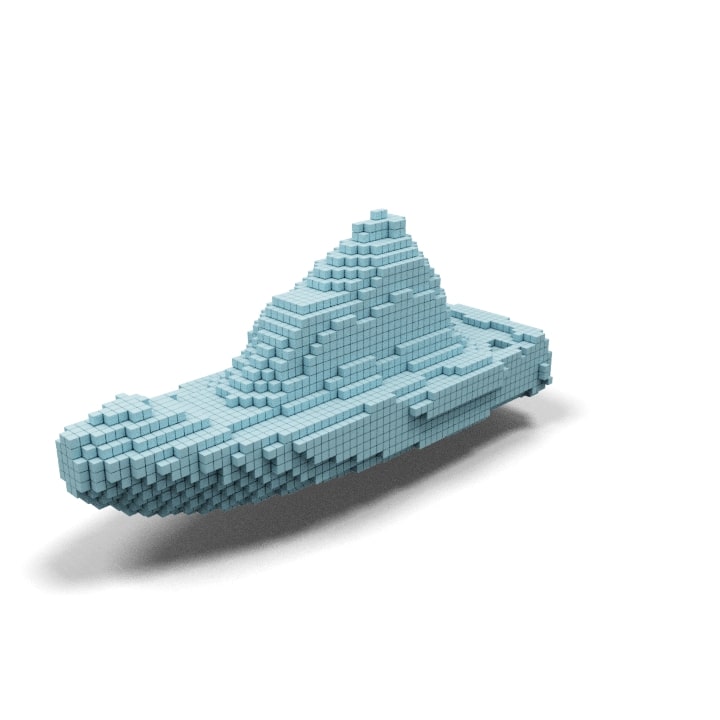} &
\includegraphics[width=0.15\linewidth]{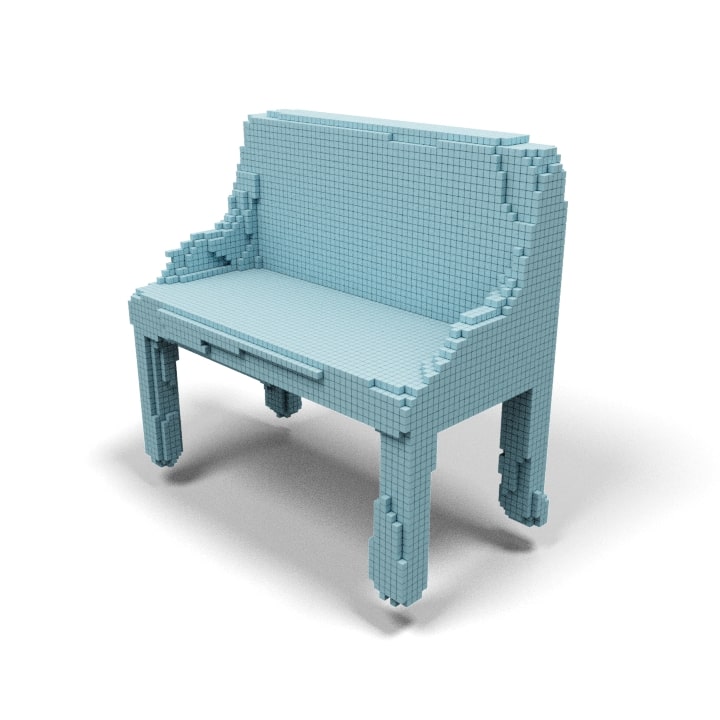} &
\includegraphics[width=0.15\linewidth]{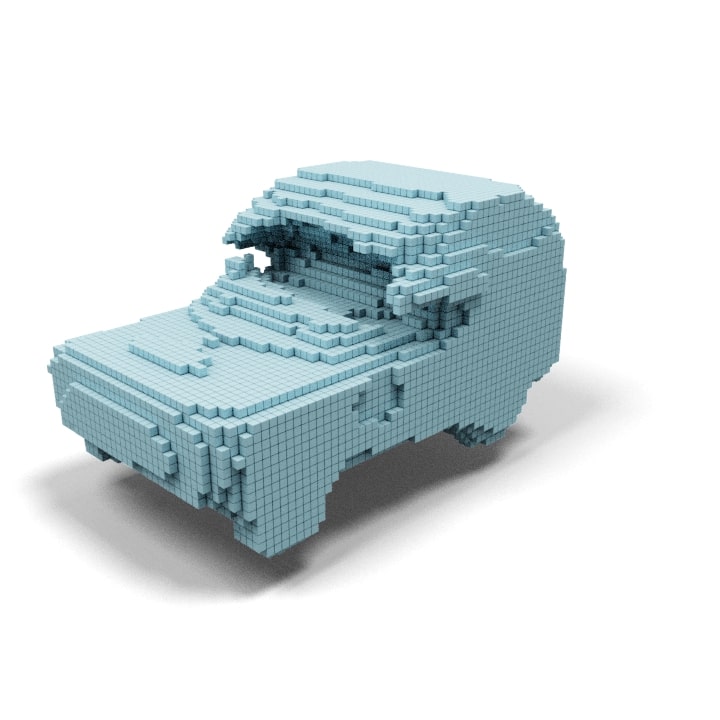} &
\includegraphics[width=0.15\linewidth]{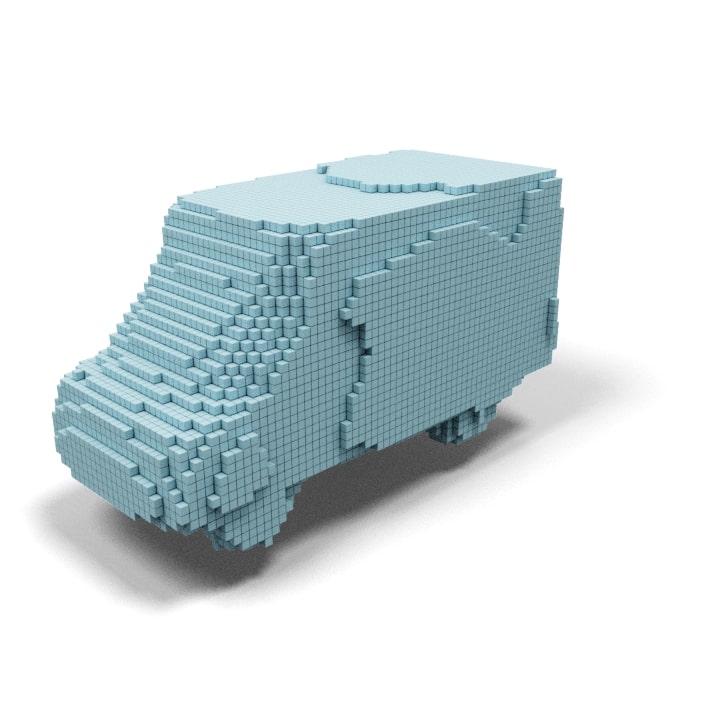}\\
``a pistol'' & ``an ak-47'' & ``a yacht'' & ``a  dressing table'' & ``a beach wagon'' & ``a van''\\
\includegraphics[width=0.15\linewidth]{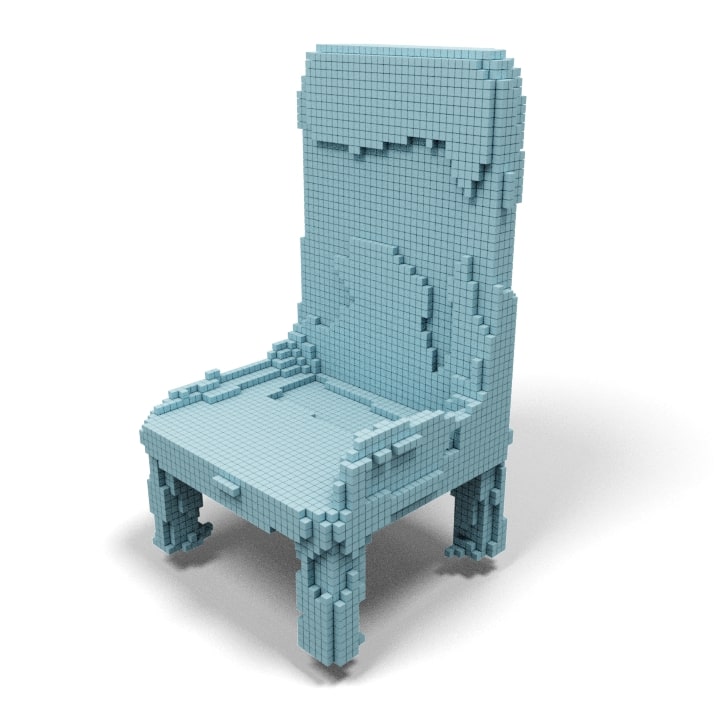} &
\includegraphics[width=0.15\linewidth]{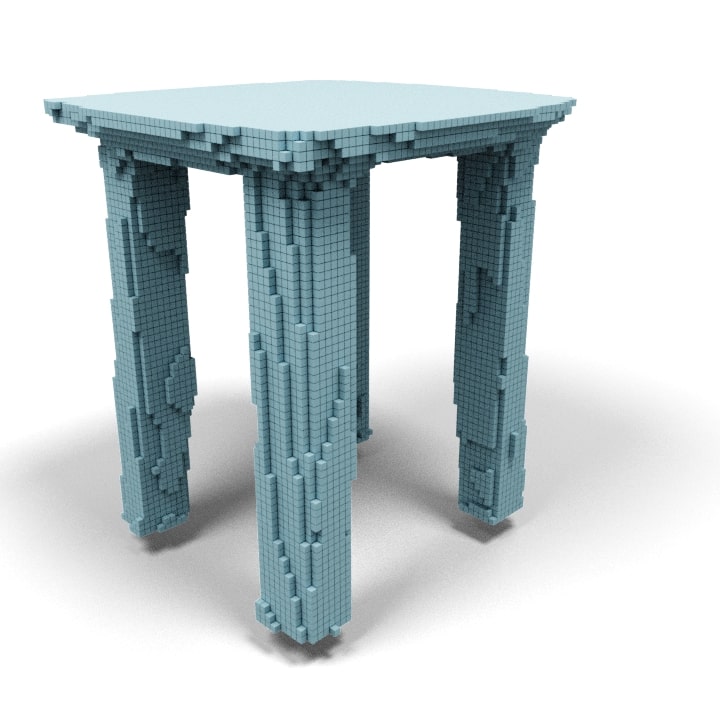} &
\includegraphics[width=0.15\linewidth]{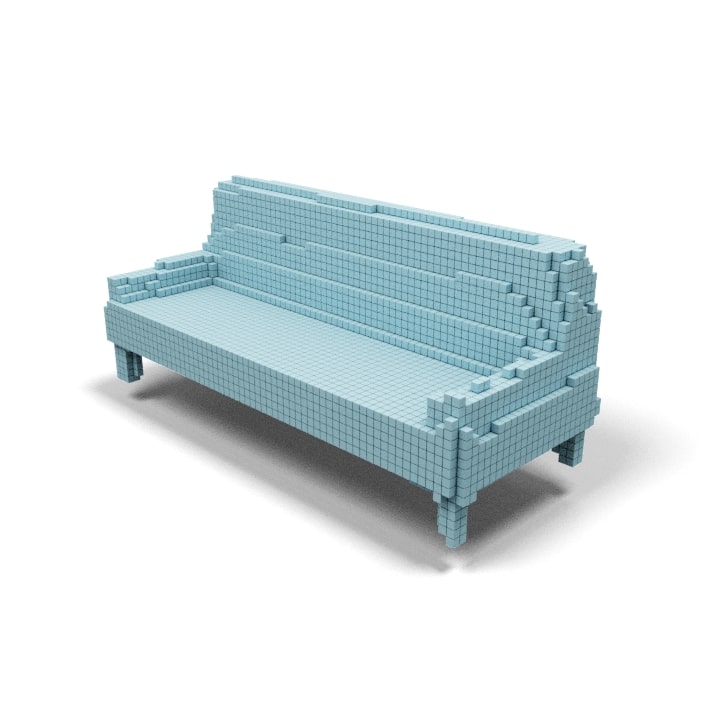} &
\includegraphics[width=0.15\linewidth]{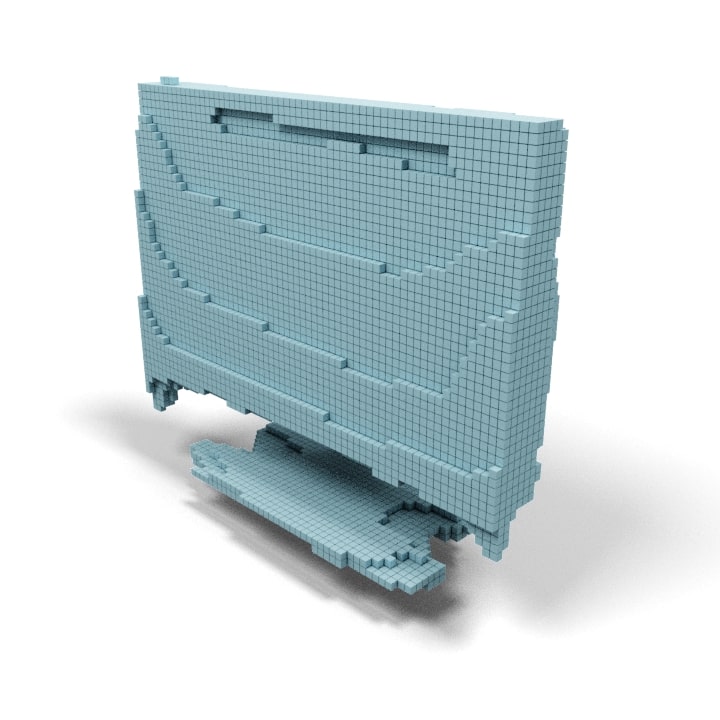} &
\includegraphics[width=0.15\linewidth]{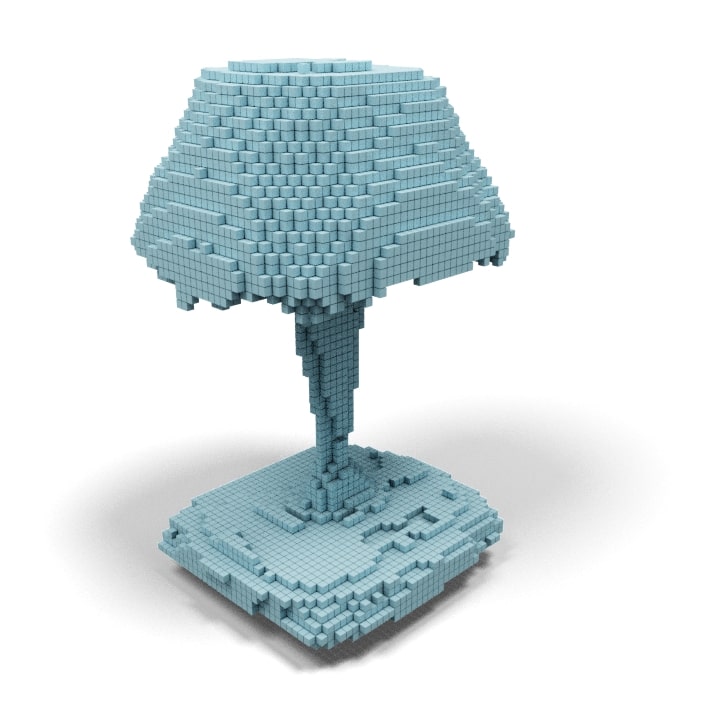} &
\includegraphics[width=0.15\linewidth]{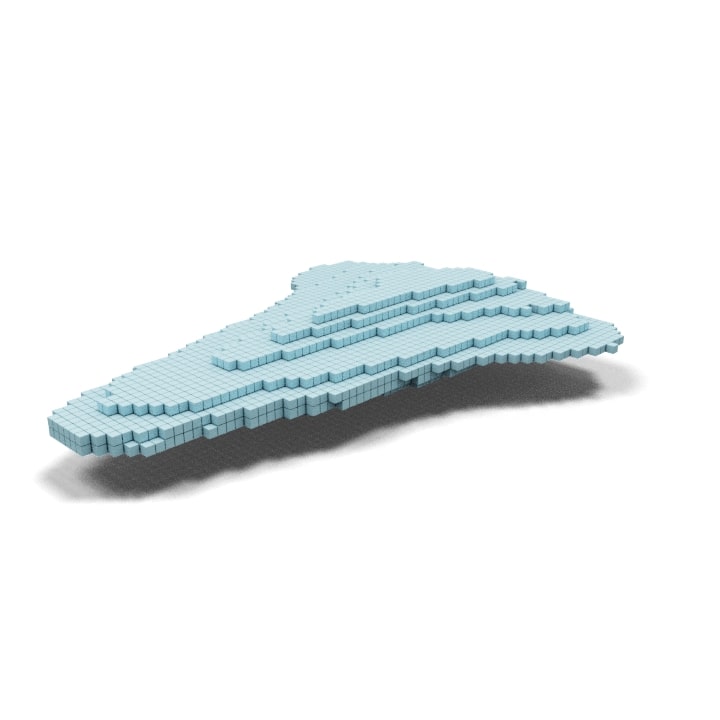}\\
``a throne'' & ``a stool'' & ``a sofa'' & ``a telly'' & ``a table lamp'' & ``a delta wing''\\
\includegraphics[width=0.15\linewidth]{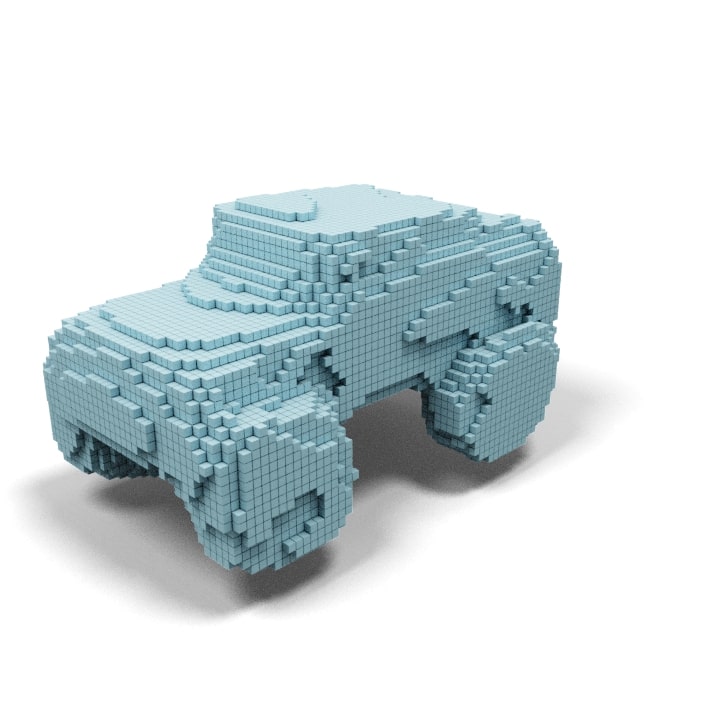} &
\includegraphics[width=0.15\linewidth]{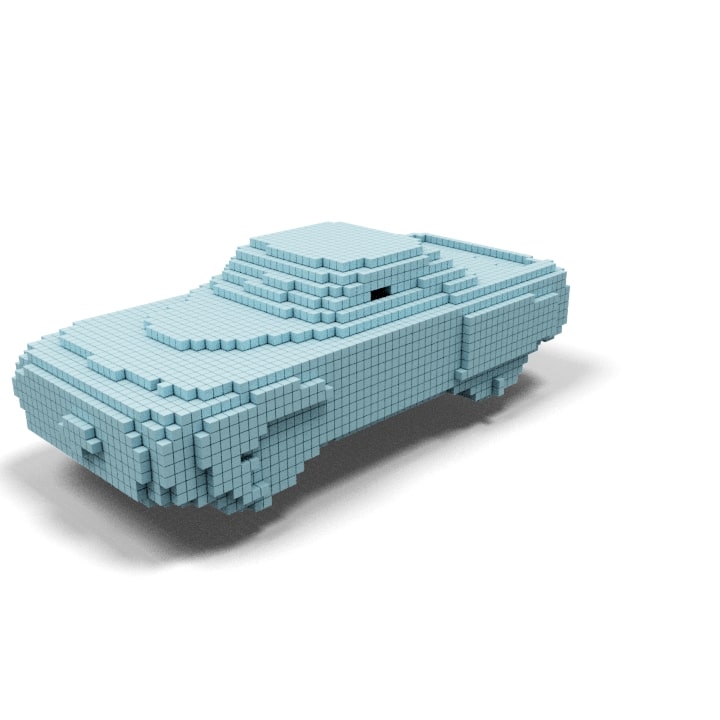} &
\includegraphics[width=0.15\linewidth]{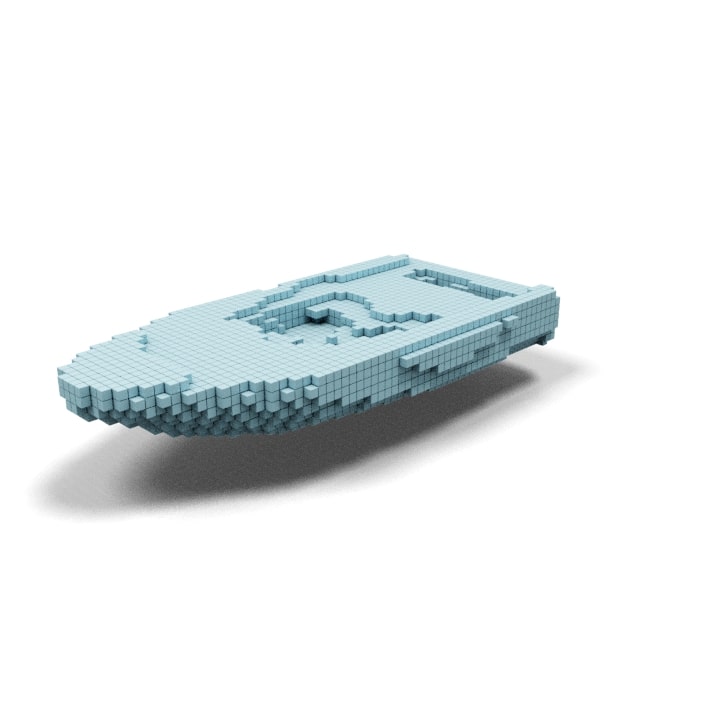} &
\includegraphics[width=0.15\linewidth]{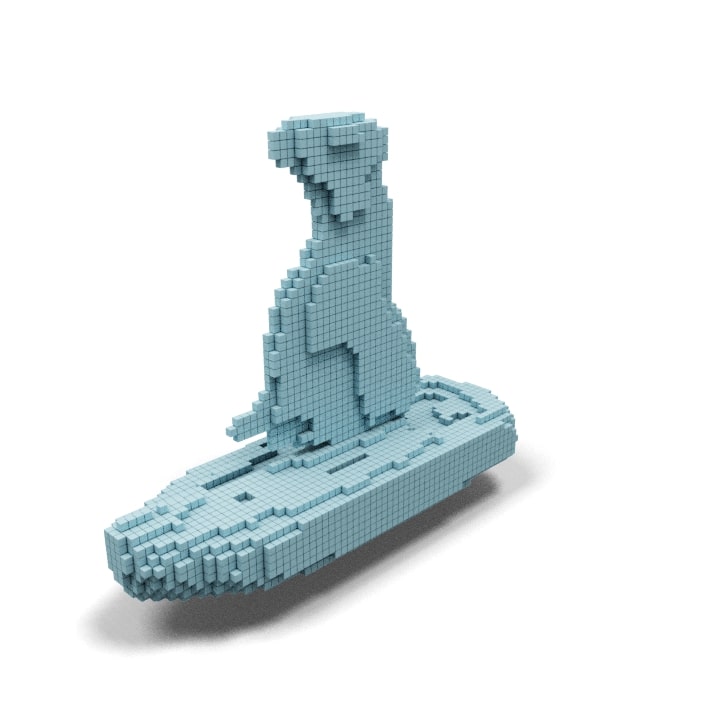} &
\includegraphics[width=0.15\linewidth]{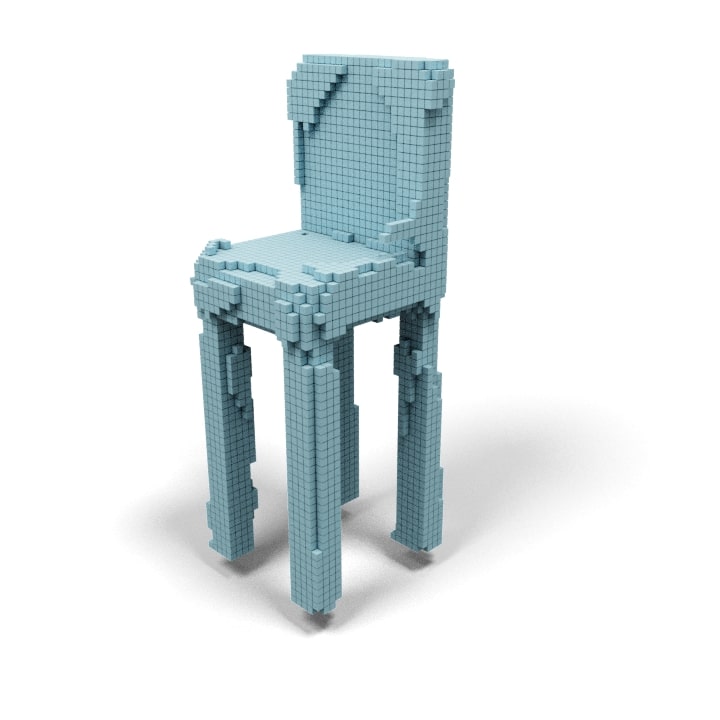} &
\includegraphics[width=0.15\linewidth]{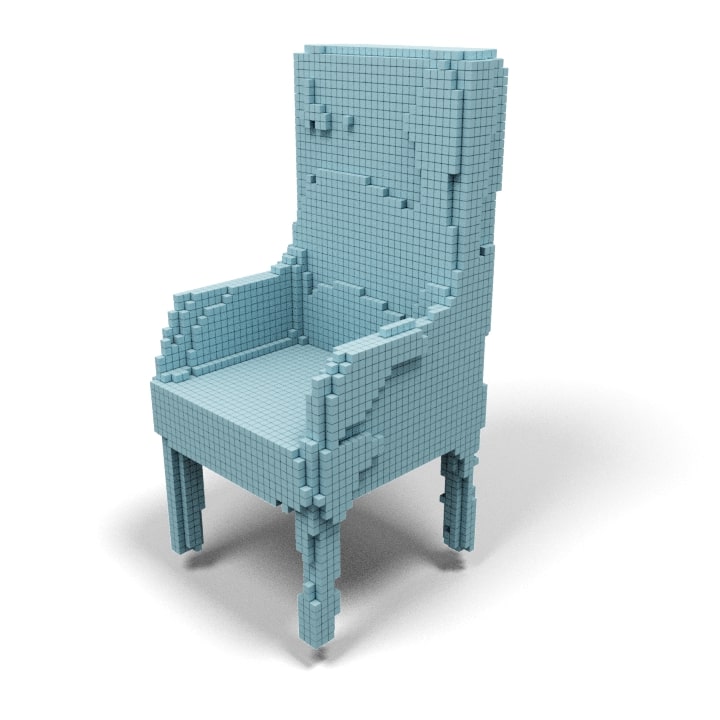}\\
``a monster truck'' & ``a muscle car'' & ``a speedboat'' & ``a sail boat'' & ``a bar stool'' & ``a wing chair''\\
\includegraphics[width=0.15\linewidth]{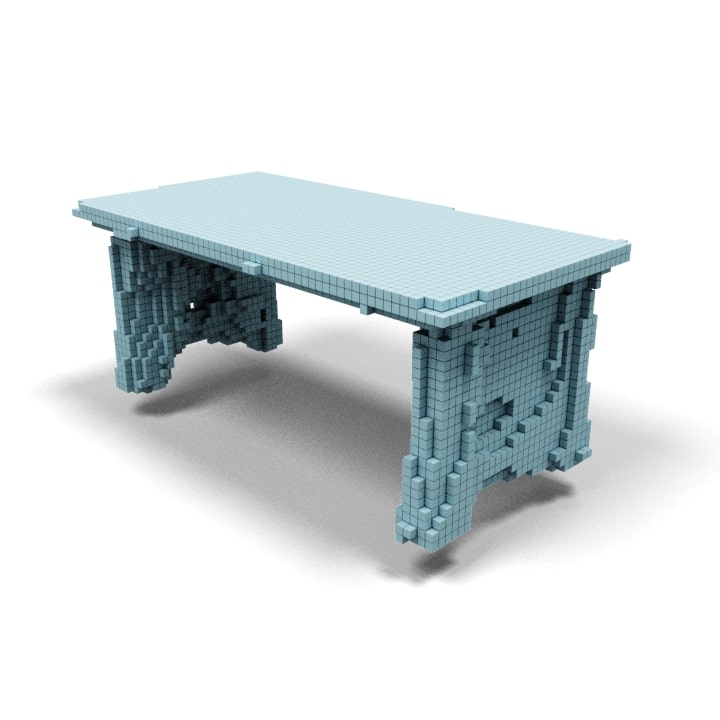} &
\includegraphics[width=0.15\linewidth]{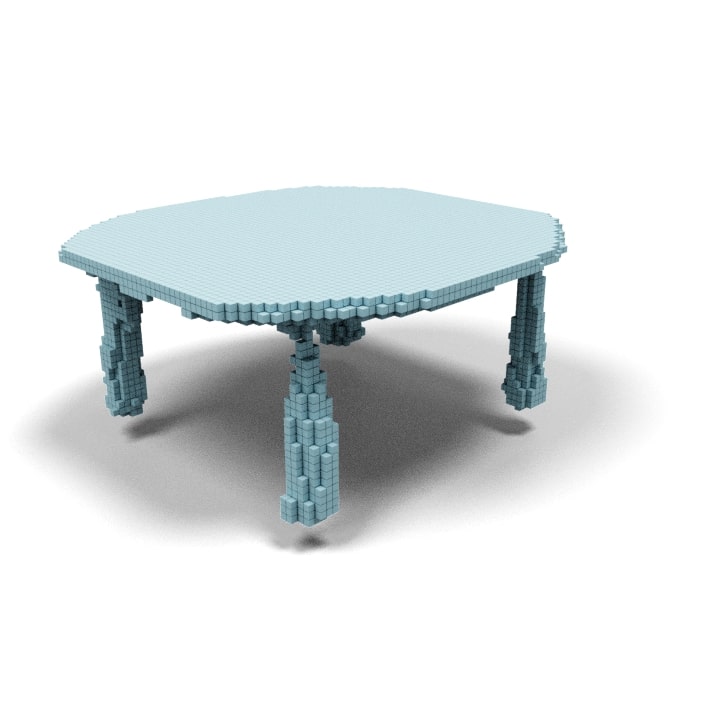} &
\includegraphics[width=0.15\linewidth]{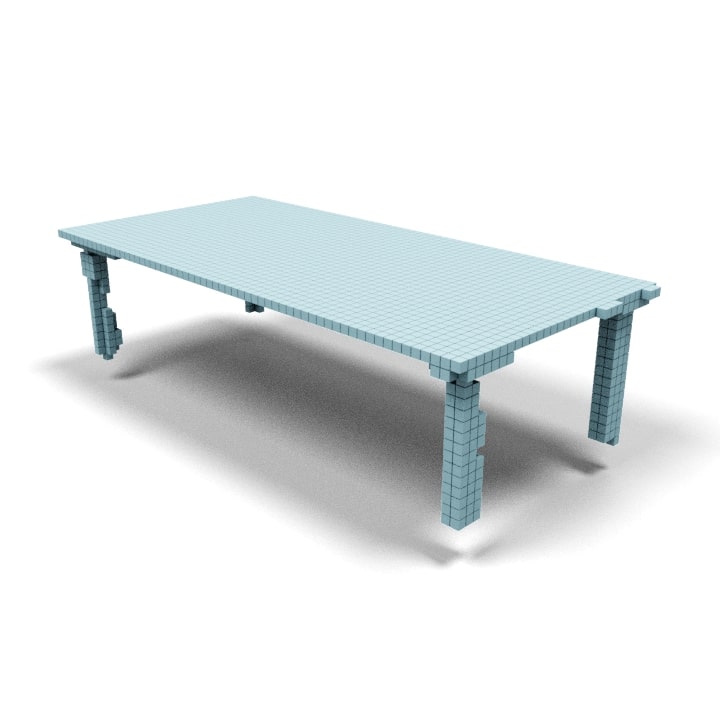} &
\includegraphics[width=0.15\linewidth]{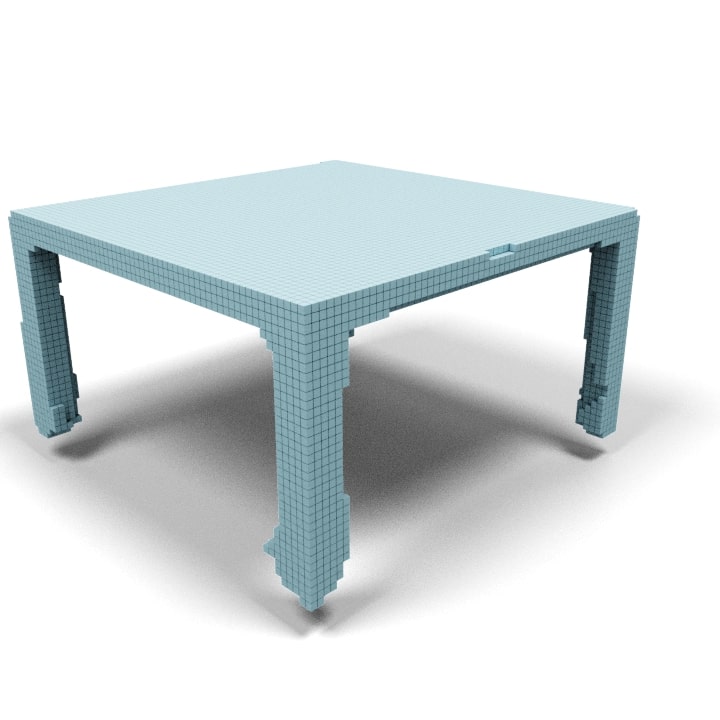} &
\includegraphics[width=0.15\linewidth]{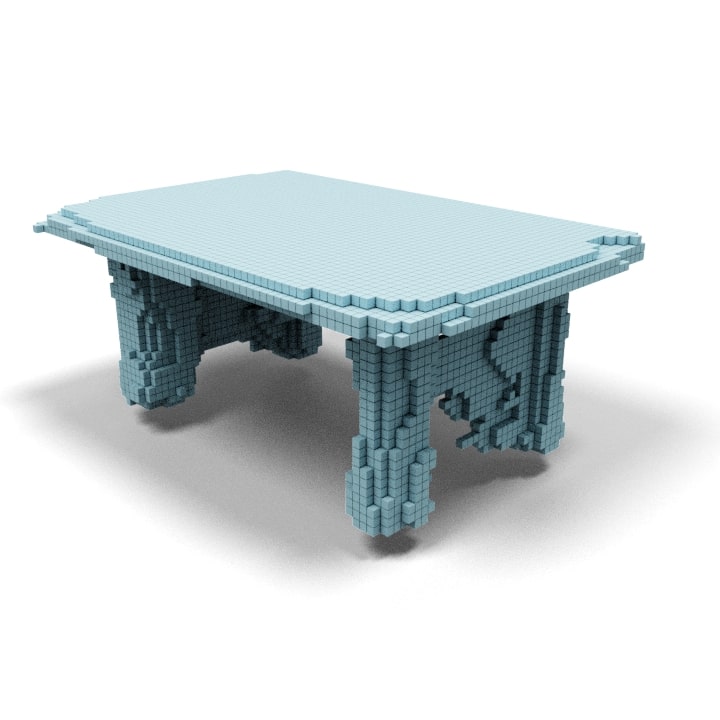} &
\includegraphics[width=0.15\linewidth]{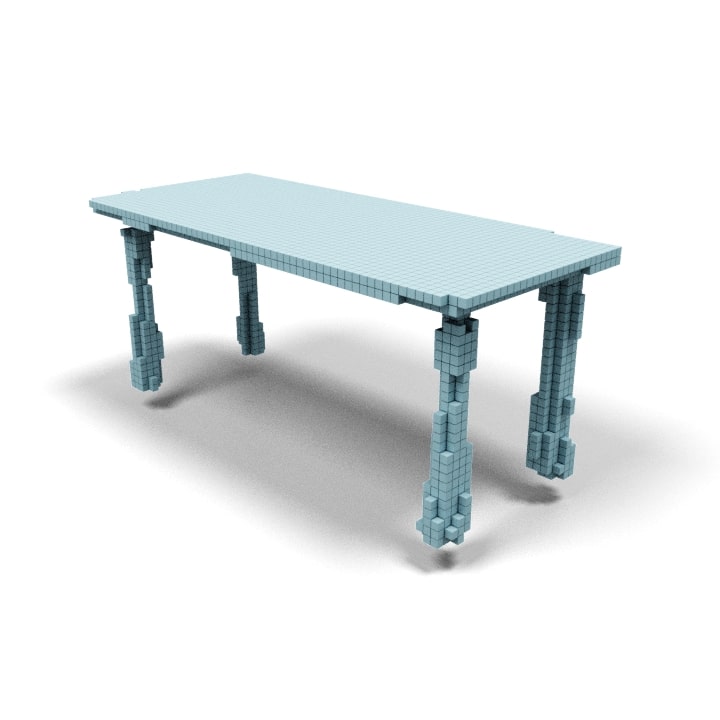}\\

``a  table'' & ``a circular table'' & ``a rectangular table'' & ``a square table'' & ``a thick table'' & ``a thin table''\\
\end{tabular}
}
\end{center}
  \caption{Illustrating our method can generate shapes via text containing common names, sub-category and attribute. The first two rows show shape generation based on using common names to describe an object. The next row shows two shapes for sub-category in car, boat and chair category. The final row illustrates the different shape attributes (circular, square, etc.) for the table class. }
\label{fig:qual_more}
\end{figure*}

\subsection{Qualitative Results}
We qualitatively evaluate generative capabilities of our method. 
First, in Figure~\ref{fig:qual_multiple} we show  that our network can generate multiple and diverse shapes using a single text query. This can be useful in a design process for imagining new variations.
Next, we show that our network can generate shapes based on category, sub-category, common semantic words, and common shape attributes as shown in Figure~\ref{fig:qual_more}. It can be seen that our network captures semantic notion of the text query.
Finally, we show the generated shapes from interpolation between two text inputs in Figure~\ref{fig:interp}. The interpolation results imply that the conditioning space is smooth.

\subsection{Human Perceptual Evaluation}
\label{sec:evaluation}
\begin{figure}[t!]
\includegraphics[width=0.47\textwidth]{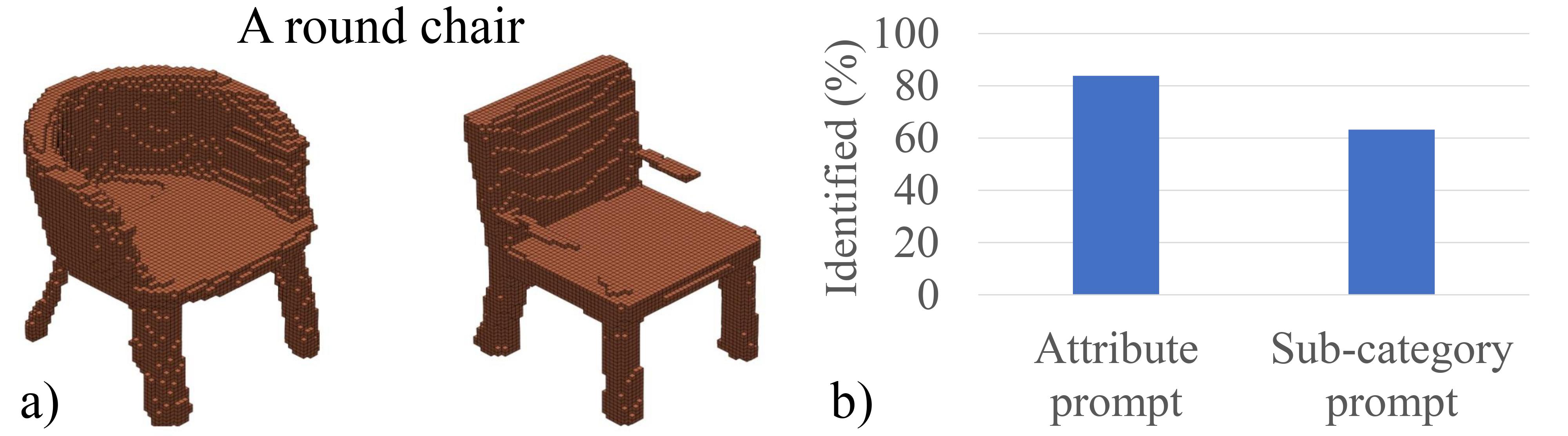}
\caption{a) An example of an image pair and text prompt shown to the human evaluators.  b) The percentage of image pairs for which the model generated by the text prompt was correctly identified.}
\label{fig:perceptual_evaluation}
\end{figure}

In this study we measure whether providing additional detail in the text prompt gives rise to semantically appropriate changes in the generated shape.   To evaluate if the shape changes are semantically correct, we used human evaluators from Amazon Mechanical Turk~\cite{mishra_2019}.  The human evaluators were presented with pairs of images as shown in Figure~\ref{fig:perceptual_evaluation}(a).  One image was generated using the ShapeNet(v2) category name (for example ``a car") while the other was generated using text which described a sub-category or shape attribute (for example ``a truck" or ``a round car").  The human evaluators were asked to identify which image best matched the sub-category or attribute text prompt. Each image pair was shown to 9 independent human evaluators.  We record the fraction of image pairs for which more than half of the evaluators selected the image generated using the sub-category or attribute augmented prompt.

The results of the perceptual study are shown in Figure~\ref{fig:perceptual_evaluation}(b).  The human evaluators correctly identified the model generated by the  detailed prompt for 70.83\% of the image pairs, showing that our method is able to utilize attribute and sub-category information in a way which is recognizable for humans.    We see the attribute prompts produced shapes which were more easily identified than those from the sub-category prompts.  One reason for this result is that the attribute augmented prompts give a clear description of how the object should look, while many of the sub-categories are less easily recognized given the quality of generation.   For example ``A circular bench" was correctly identified by 8/9 evaluators while ``A laboratory bench" was not recognized by any of the 9 humans.

\subsection{Choice of Prefix in Text Prompt}
Designing a prompt can be challenging as small changes in words can potentially have a impact on our downstream task. In this experiment, we investigate how much does prompt selection effect the performance of our method. We specifically investigate what prefix to choose before a text query.  The investigations are shown in Table \ref{tab:prompt_eng}. We find that prefix selection indeed has a  effect on generation quality and diversity. A interesting avenue of future research would be to investigate prompt engineering \cite{zhou2021learning}.

\begin{table}
\centering
\setlength{\tabcolsep}{11.5pt}
{\small
\begin{tabular}{l|ccc}
\toprule
\textit{prefix} & \textbf{FID$\downarrow$} & \textbf{MMD$\uparrow$} & \textbf{Acc.$\uparrow$}\\
\midrule
``a''/``an'' & {2425.25} & {0.6607} & {83.33}\\ \hline
``a photo of'' & \textbf{2400.36} & 0.6490 & 78.63 \\ \hline
``a photo of a'' & {2484.49} & 0.6620 & 80.77\\ \hline
``a picture of a'' & 2560.98 & \textbf{0.6681} & 81.20\\ \hline
``a rendering of'' & 3029.92 & 0.6311 &	76.50\\ \hline
``a photo of one'' & 2715.45 & 0.6597 & {82.48}\\ \hline
``one'' & 3142.07 &	{0.6608} & \textbf{87.18}\\
\bottomrule
\end{tabular}
}
\caption{Different Prompts and their effect.}
\label{tab:prompt_eng}
\end{table}

\subsection{CLIP-Forge for Point Cloud}
\begin{figure}[t!]
\begin{center}
\setlength{\tabcolsep}{6pt}
\small{
\begin{tabular}{ccc}
\includegraphics[width=0.25\linewidth]{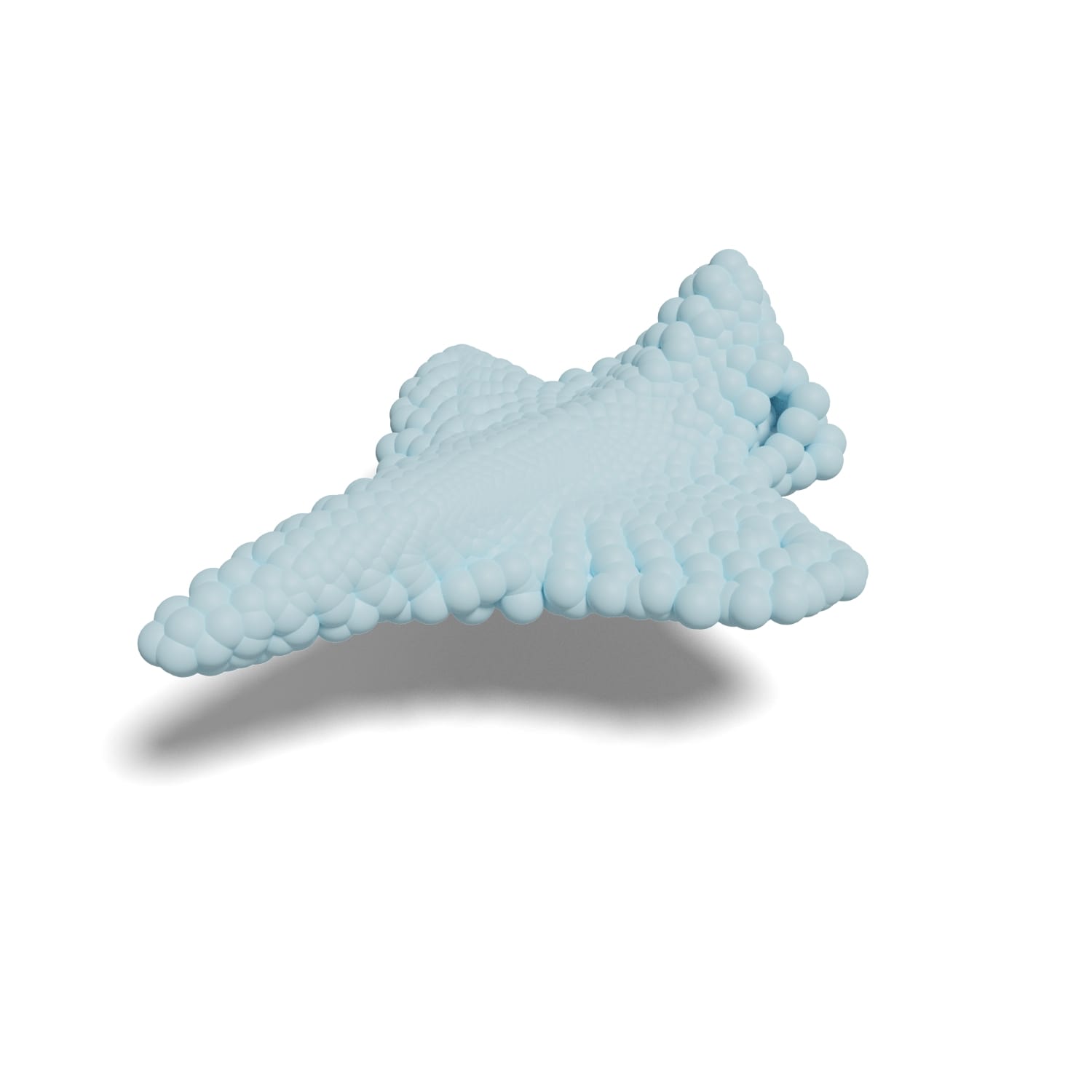} &
\includegraphics[width=0.25\linewidth]{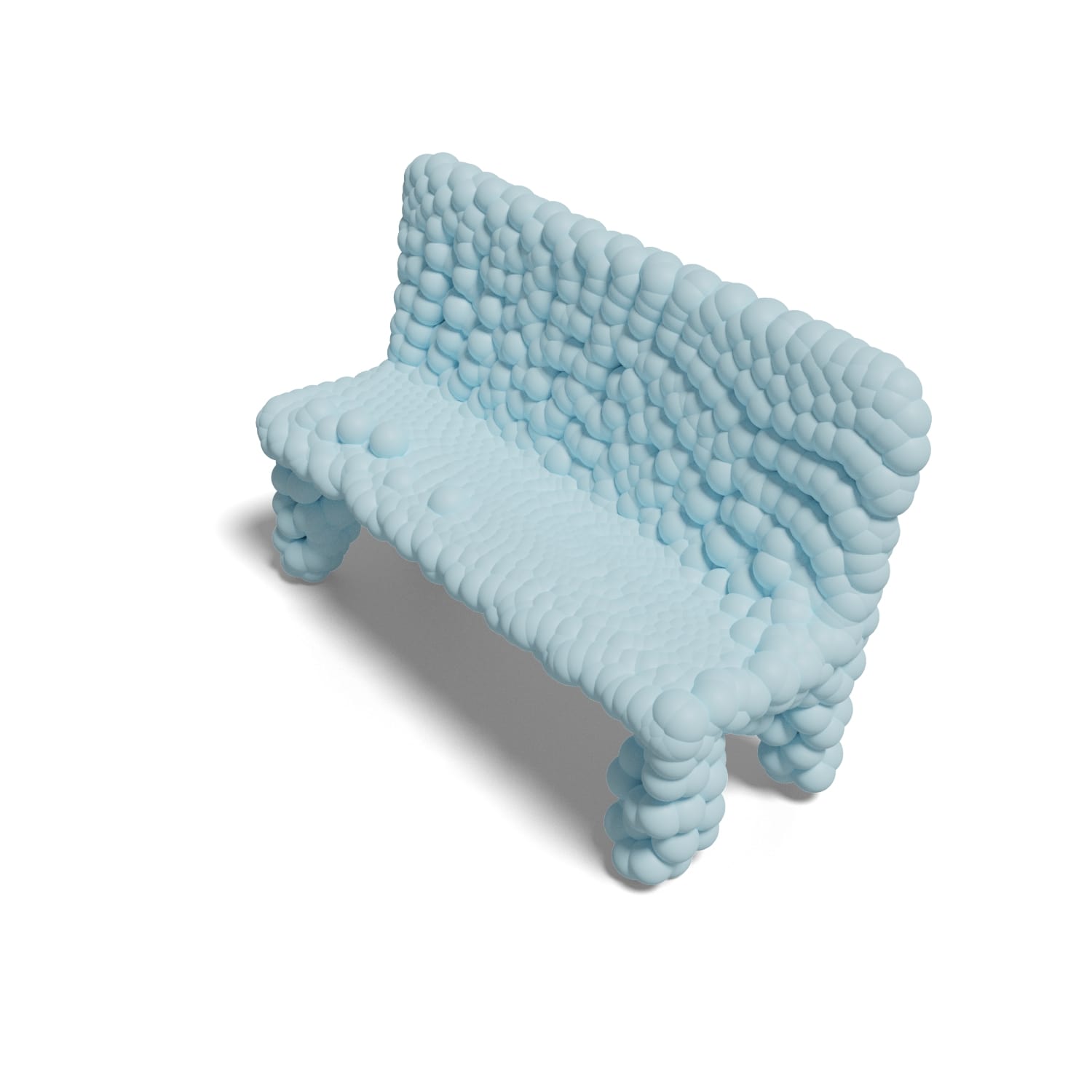} &
\includegraphics[width=0.25\linewidth]{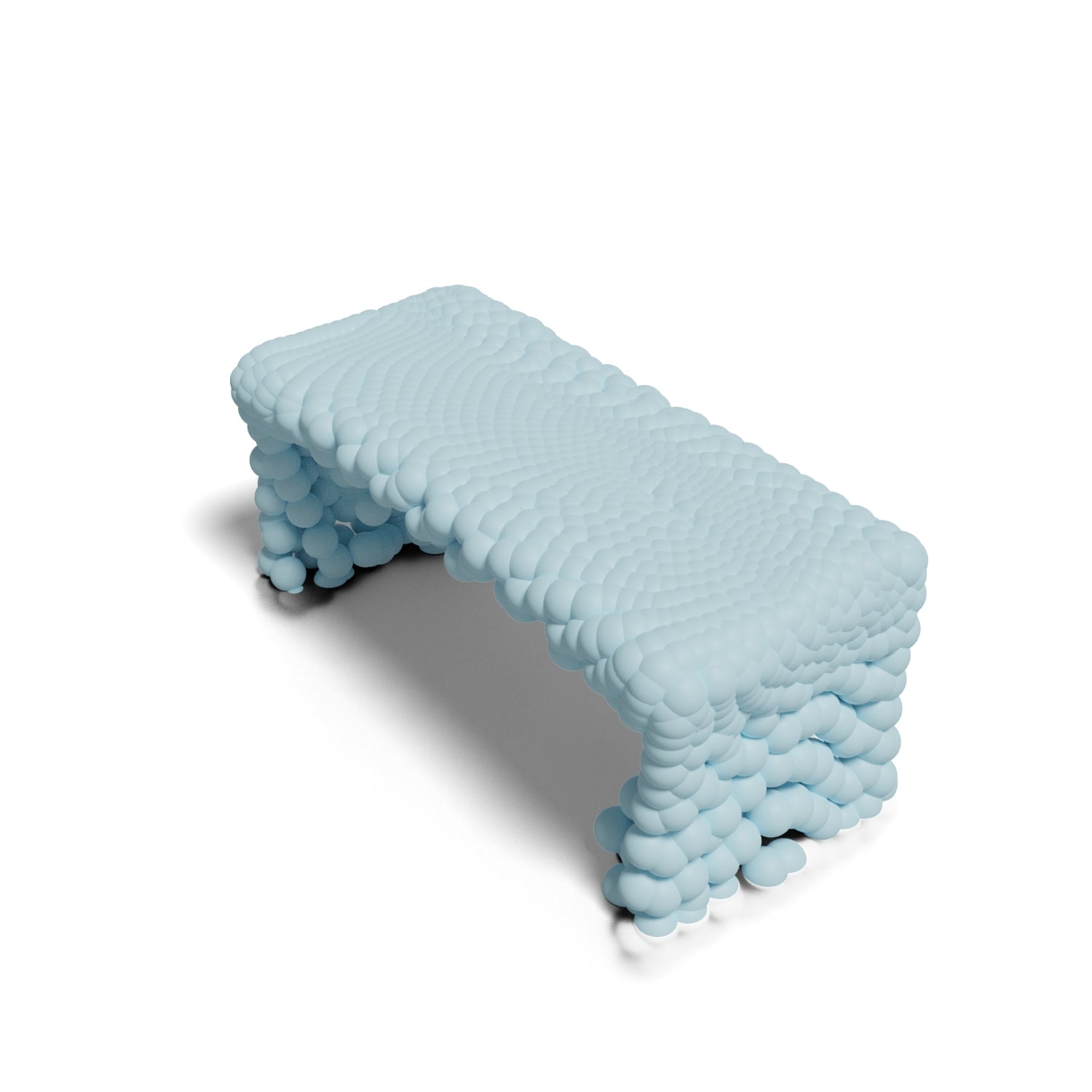} \\
``an f-16 '' & ``a bench'' & ``a desk'' \\
\includegraphics[width=0.25\linewidth]{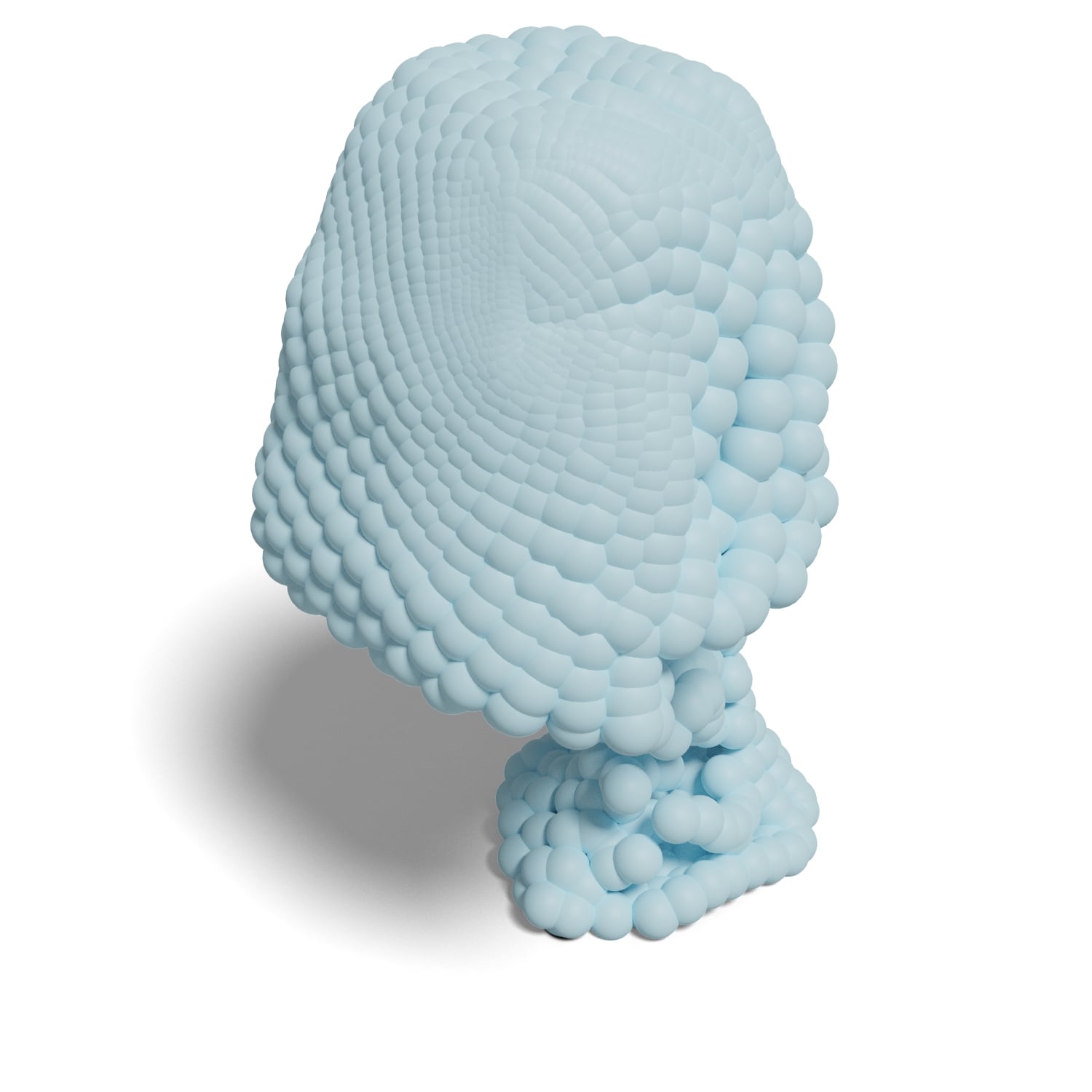} &
\includegraphics[width=0.25\linewidth]{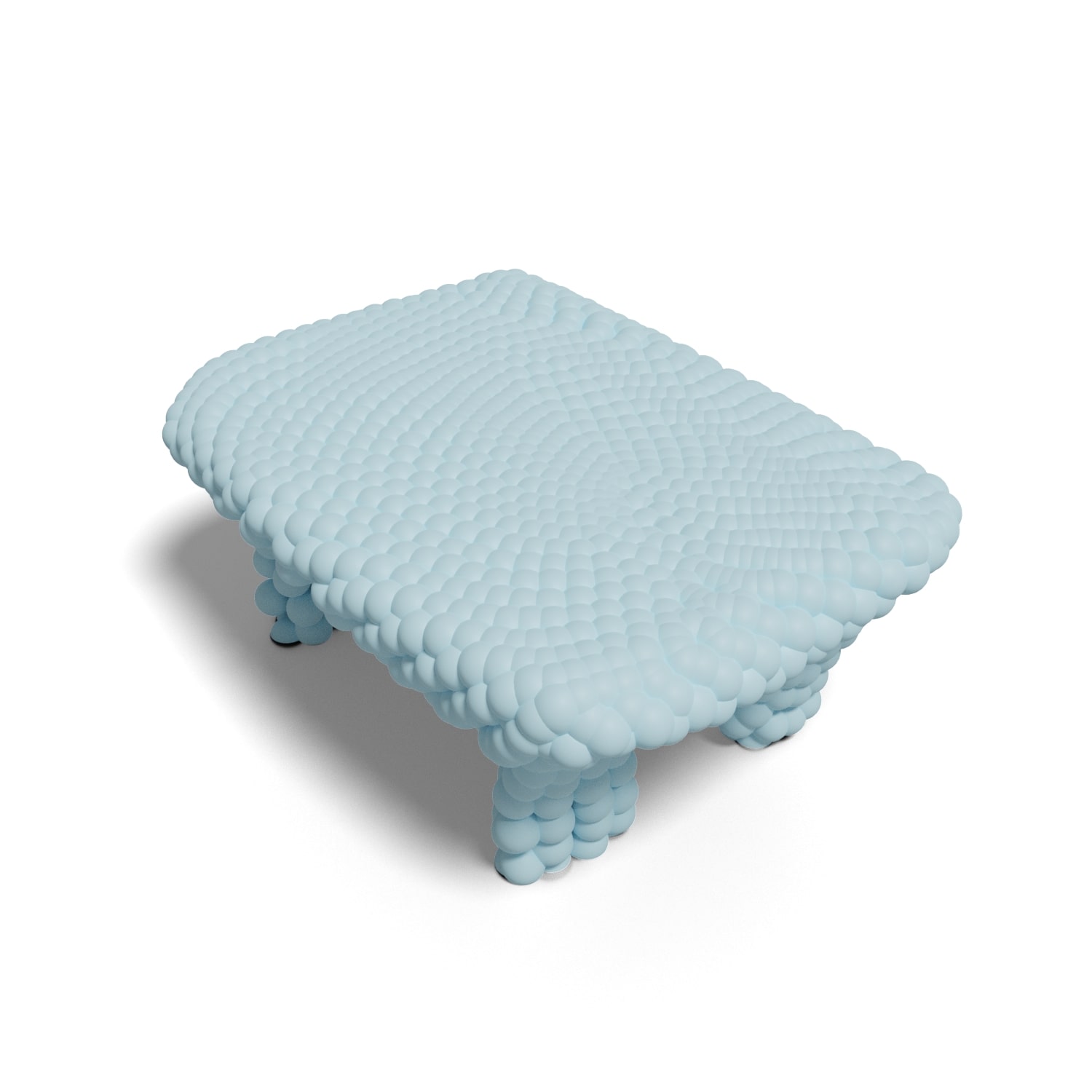} &
\includegraphics[width=0.25\linewidth]{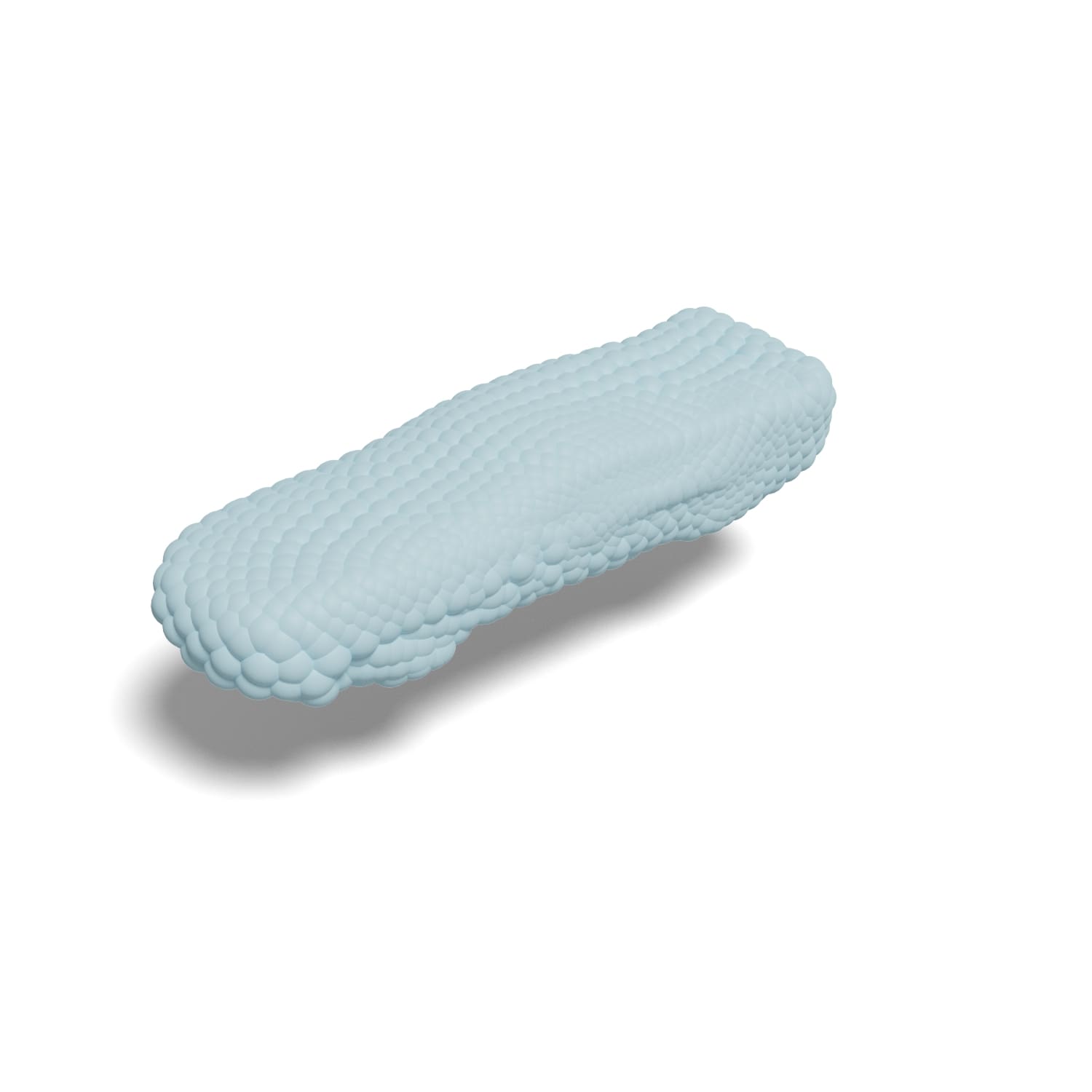}\\
``a lamp'' & ``a billiard table'' & ``a thin car''\\
\includegraphics[width=0.25\linewidth]{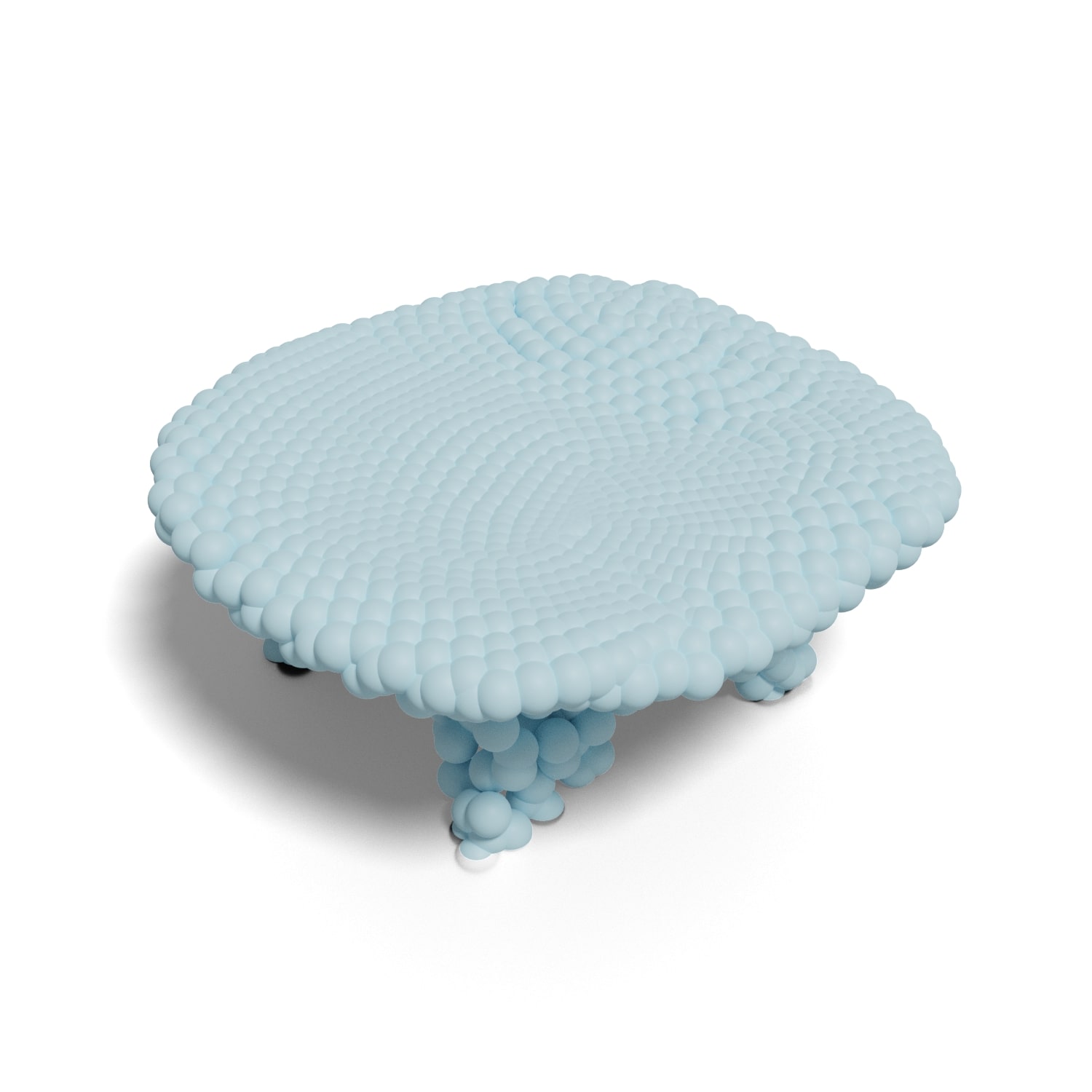} &
\includegraphics[width=0.25\linewidth]{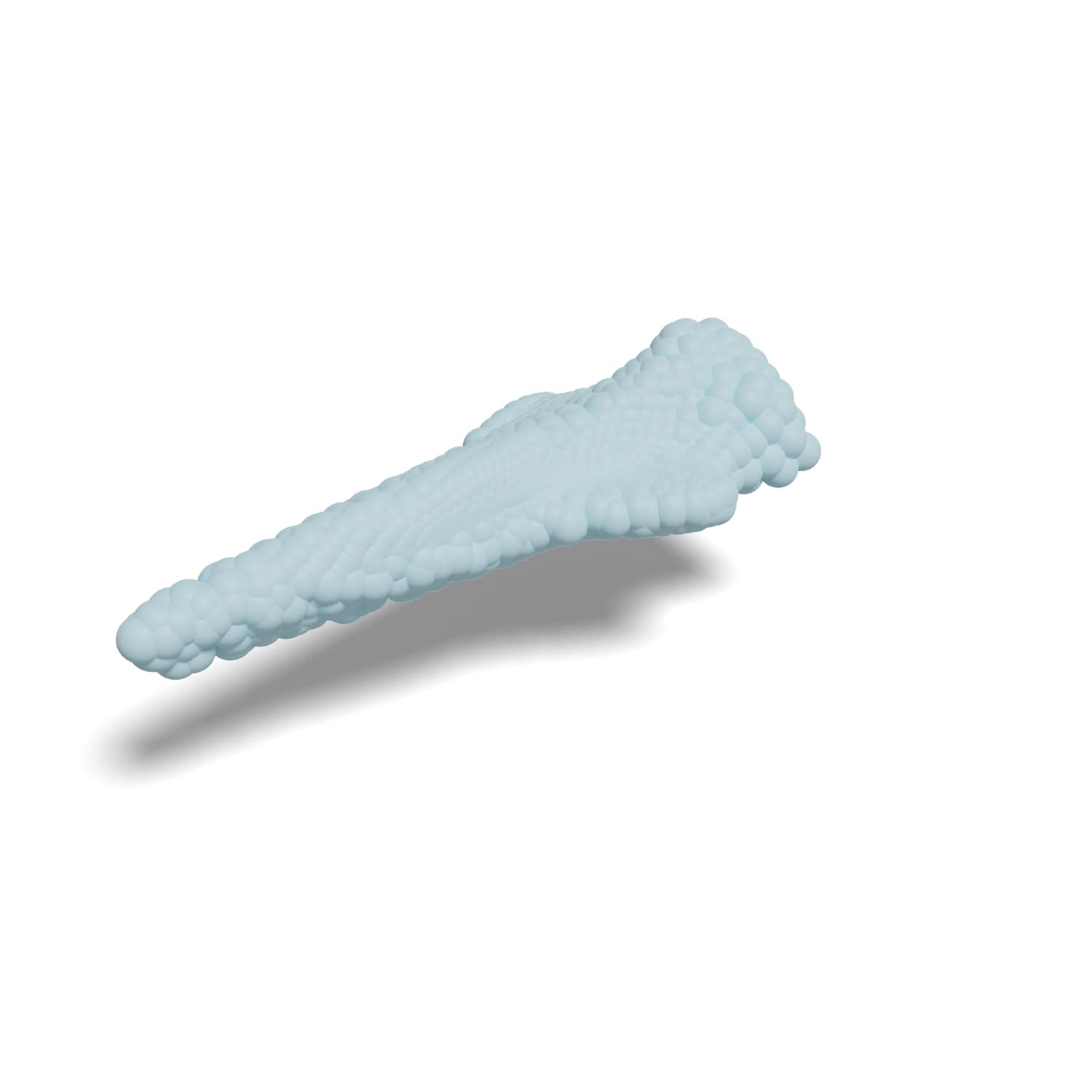} &
\includegraphics[width=0.25\linewidth]{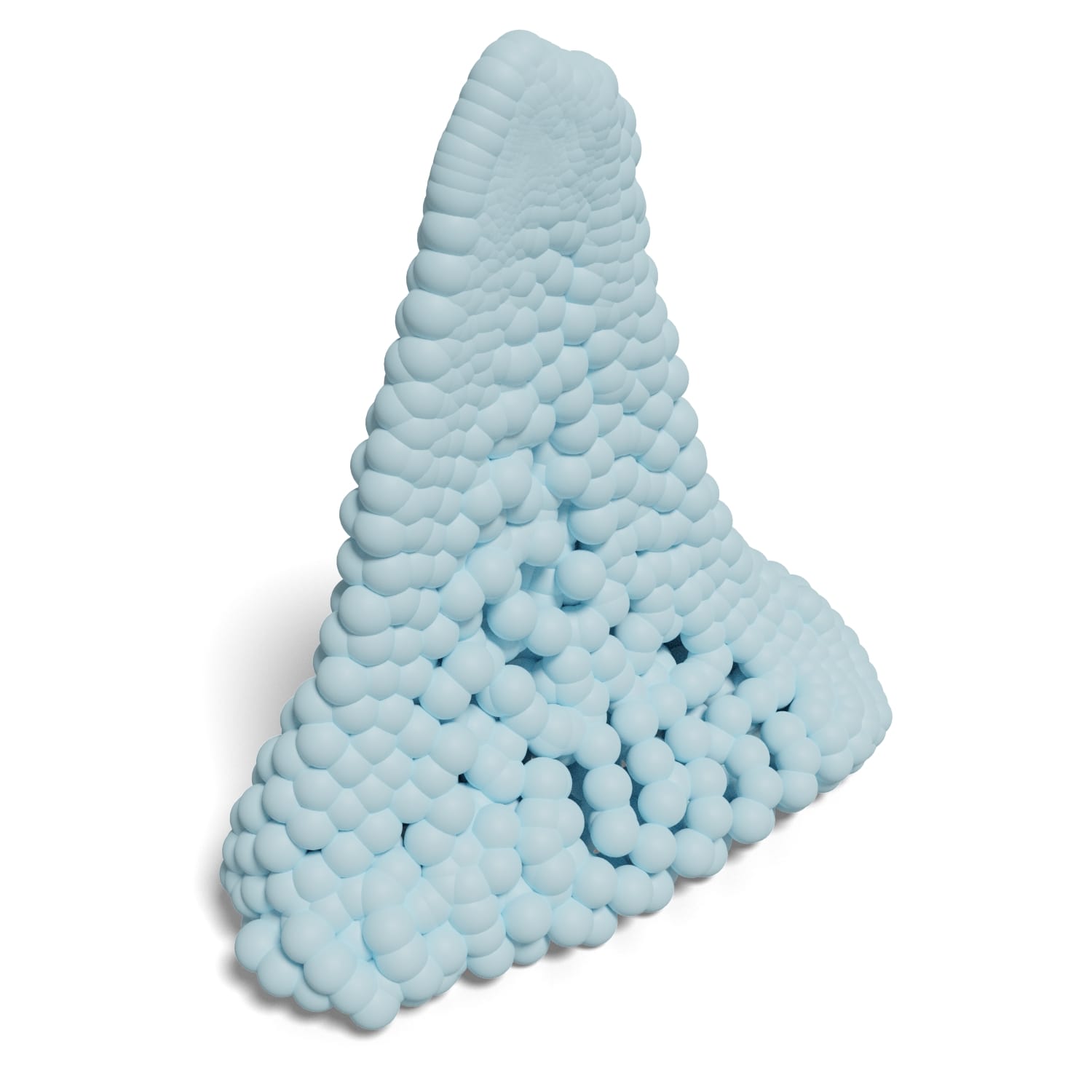}\\
``a round table'' & ``a supersonic plane'' & ``a sailing boat''\\
\end{tabular}
}
\end{center}
\caption{CLIP-Forge point cloud generations.}
\label{fig:point_clouds}
\end{figure}

In this section, we investigate if  our method  can be simply applied to a different representation, namely, a point cloud. As stated earlier we use the PointNet encoder \cite{qi2017pointnet} and FoldingNet decoder \cite{yang2018foldingnet}. We use the same flow architecture as mentioned above. We train the network on the ShapeNet(v2) dataset. The results are shown in Fig \ref{fig:point_clouds}. It can be seen that our method does a satisfactory job in generating 3D point clouds using text queries while using off the shelf point cloud encoders and decoders.

\begin{figure}[t!]
\begin{center}
\setlength{\tabcolsep}{5pt}
\small{
\begin{tabular}{ccccc}
\includegraphics[width=0.15\linewidth]{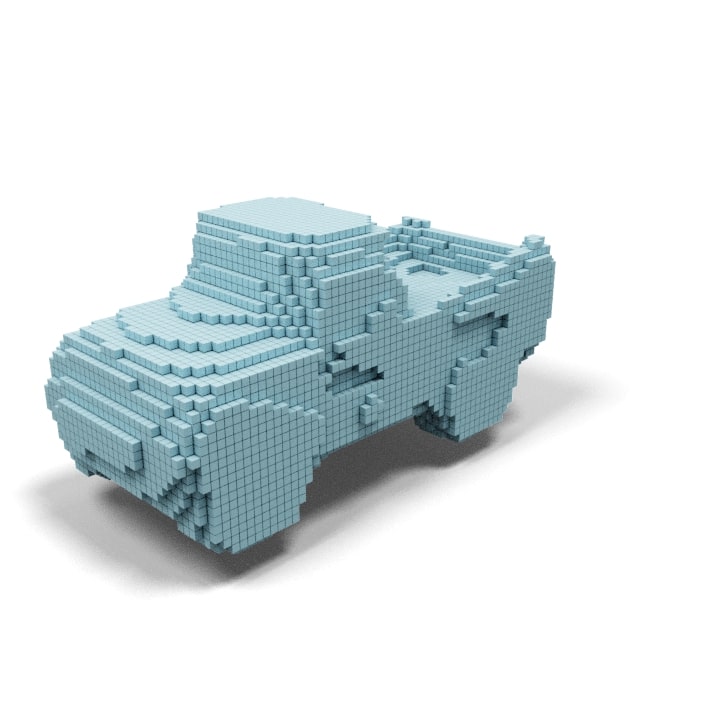} &
\includegraphics[width=0.15\linewidth]{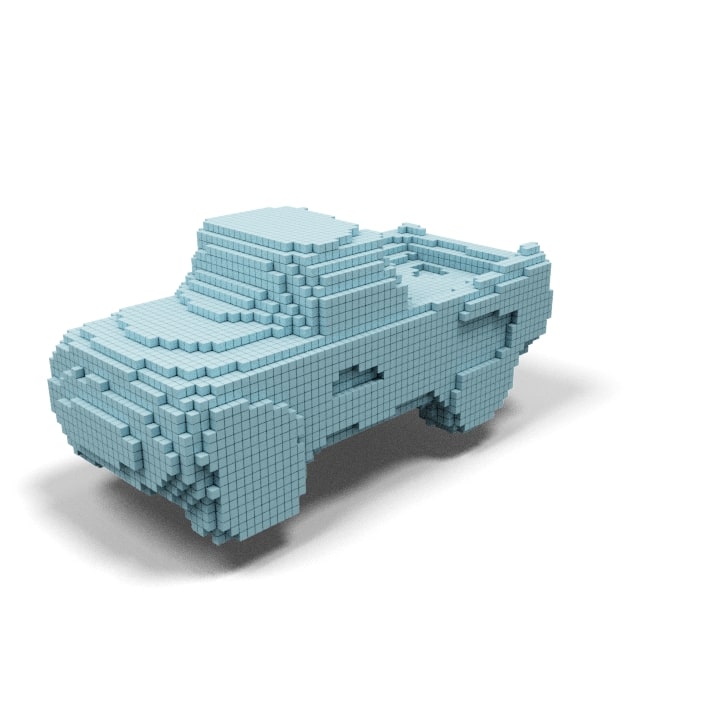} &
\includegraphics[width=0.15\linewidth]{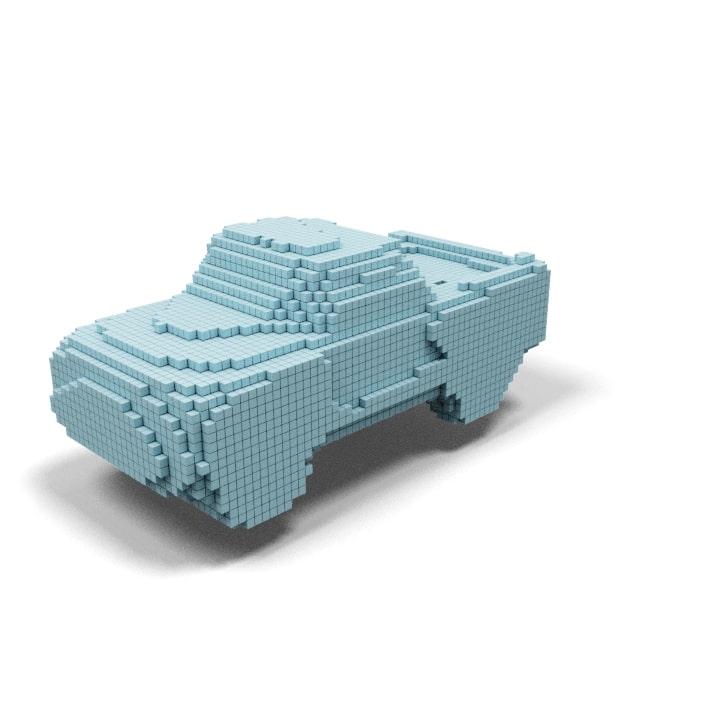} &
\includegraphics[width=0.15\linewidth]{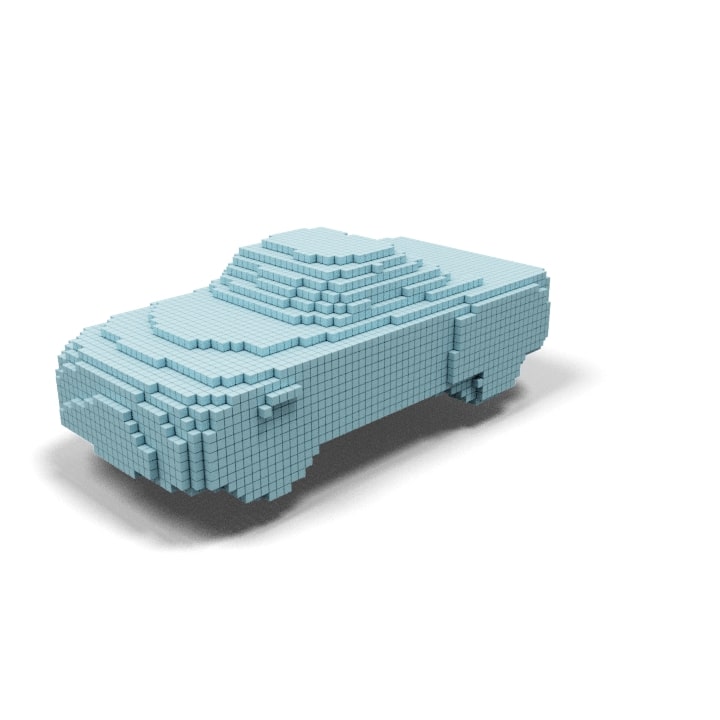} &
\includegraphics[width=0.15\linewidth]{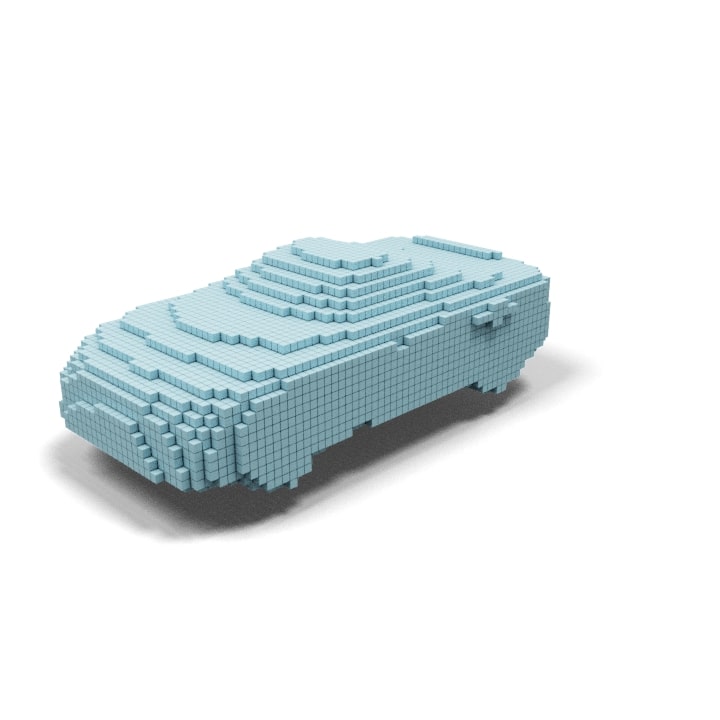}\\
\multicolumn{5}{c}{``a truck'' $\rightarrow$ ``a sports car''}\\
\includegraphics[width=0.15\linewidth]{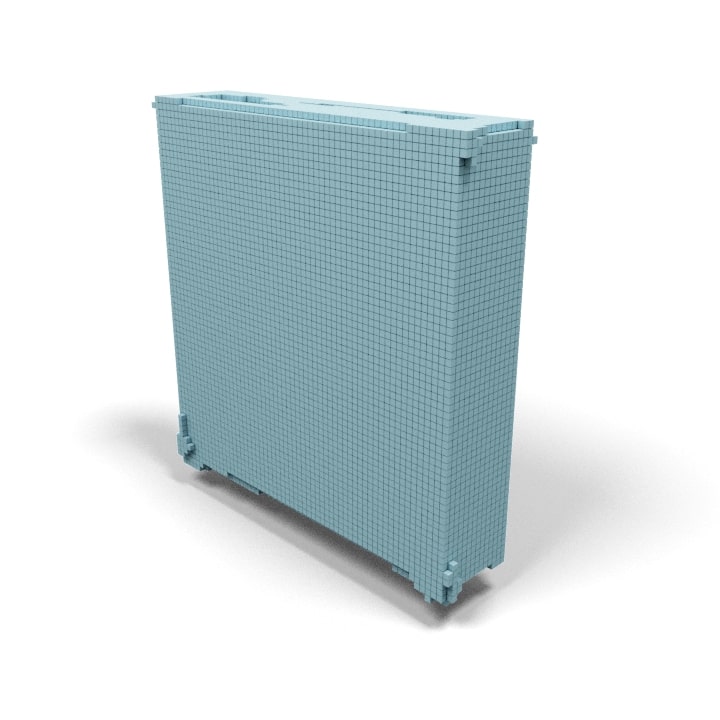} &
\includegraphics[width=0.15\linewidth]{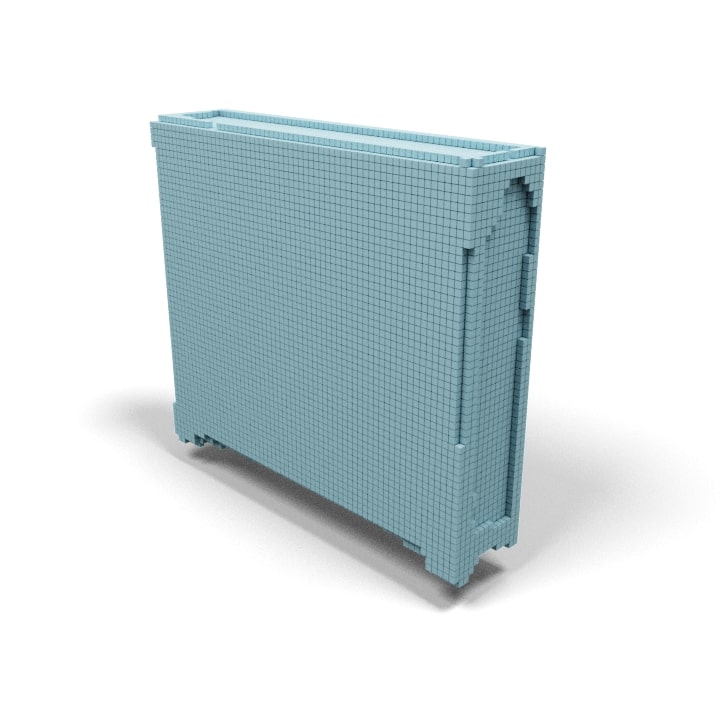} &
\includegraphics[width=0.15\linewidth]{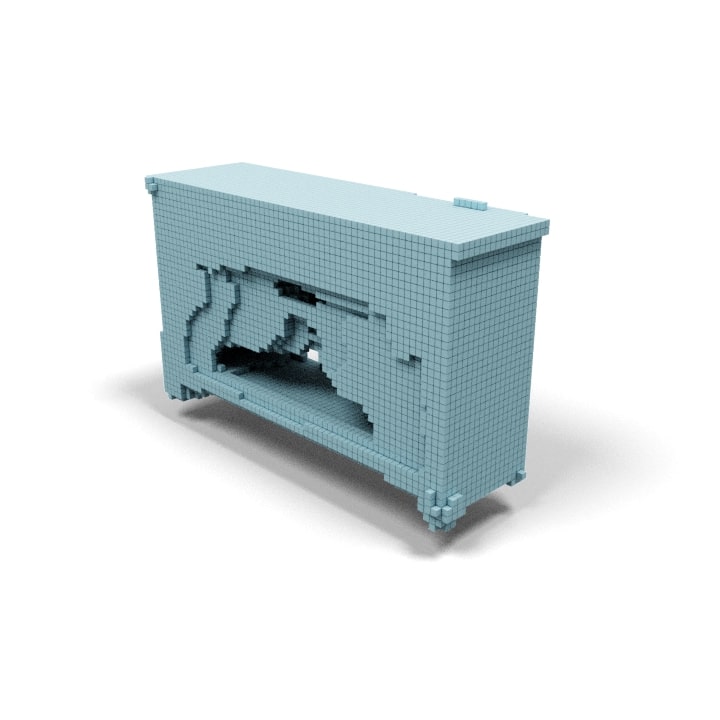} &
\includegraphics[width=0.15\linewidth]{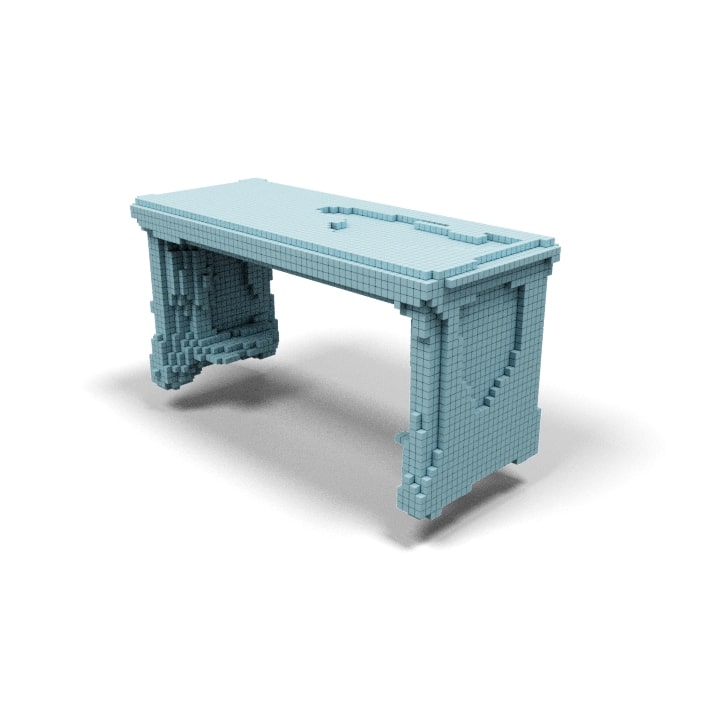} &
\includegraphics[width=0.15\linewidth]{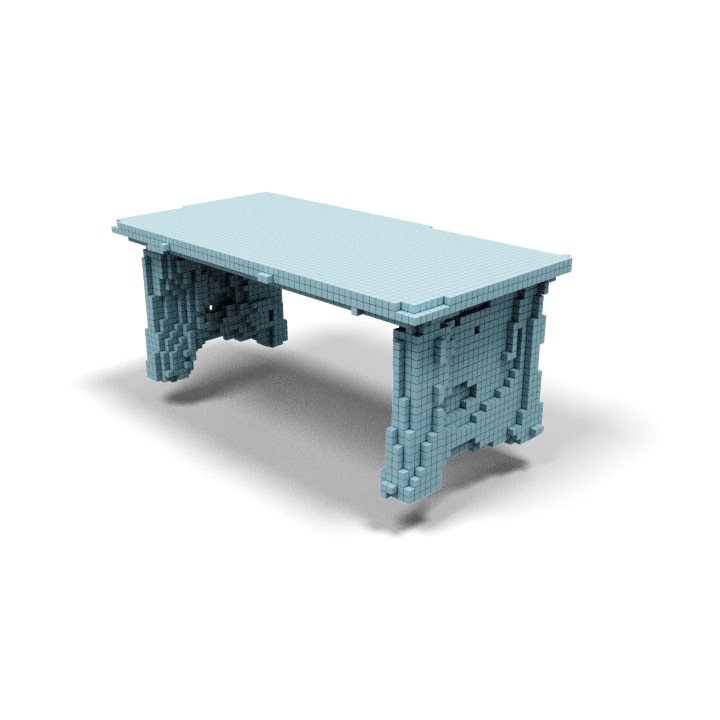}\\
\multicolumn{5}{c}{``a cabinet'' $\rightarrow$ ``a table''}\\
\end{tabular}
}
\end{center}
\caption{Interpolation results between two text queries.}
\label{fig:interp}
\end{figure}

\section{Ablation Studies}
In this section, we discuss how different components of our algorithm affect our model.  For all our ablation studies, we use the above mentioned hyperparameters for the autoencoder unless otherwise stated. For the flow model, we use RealNVP model with Checkerboard masking for most experiments unless otherwise stated.

\subsection{Stage 1 Autoencoder Design Choice}
In Table \ref{tab:autoencoder_arch_abl}, we experiment with different parts of the autoencoder architecture. The first subsection of Table \ref{tab:autoencoder_arch_abl}  investigates how adding noise in the latent space helps our model. Empirically it can be seen from the table that adding noise not only helps the reconstruction but also improves the generation and diversity of the shapes generated. Next we investigate the size of the latent space and find that our model works reasonably well while using a smaller latent size of 128. Finally, we explore different encoders and decoders for our model. The results indicate that our model can take different representations, point cloud, as the input to the encoder.  We provide more details regarding the encoder and decoder in appendix.

\begin{table*}
\centering
\setlength{\tabcolsep}{13pt}
{\small
\begin{tabular}{c|c|c|c|ccccc}
\toprule
\textit{noise} & \textit{latent} & \textit{encoder} & \textit{decoder} & \textbf{IOU$\uparrow$} & \textbf{MSE$\downarrow$} & \textbf{FID$\downarrow$} & \textbf{MMD$\uparrow$} & \textbf{Acc.$\uparrow$} \\
\midrule
$\times$ & \multirow{2}{*}{128} & \multirow{2}{*}{VoxEnc} & \multirow{2}{*}{RN-OccNet} & 0.7275 & 0.01120 & 3871.48	& 0.6559 & 71.94\\ \cline{1-1} \cline{5-9}
\checkmark & ~ & ~ & ~ & 0.7374 & 0.01159 & 2688.72 & 0.6732 & \textbf{79.34} \\ \hline
\multirow{2}{*}{\checkmark} & 256 & \multirow{2}{*}{VoxEnc} & \multirow{2}{*}{RN-OccNet} & 0.7375 & 0.01158 & 3177.92	& 0.6535 & 78.77\\ \cline{2-2} \cline{5-9}
~ & 512 & ~ & ~ & 0.7362 & 0.01155 & 3577.72 & 0.6374 &	74.50\\ \hline
\multirow{3}{*}{\checkmark} & \multirow{3}{*}{128} & PointNet & RN-OccNet & 0.7082	& 0.01051 & \textbf{2646.93} &	\textbf{0.6746} & 76.50\\ \cline{3-9}
~ & ~ & ResVoxEnc  & RN-OccNet & 0.7371 &	0.01075  & 3146.94 & 0.6509 &	74.64\\  \cline{3-9}
~ & ~ & VoxEnc & CBN-OccNet & \textbf{0.7674} & \textbf{0.01025}  & 2956.78	& 0.6645 & 78.77\\
\bottomrule
\end{tabular}
}
\caption{Effects of different autoencoder design choices in stage 1, including the usage of Gaussian noise, the latent vector size, as well as various encoder and decoder architectures.
} 
\label{tab:autoencoder_arch_abl}
\end{table*}



\subsection{Stage 2 Prior Design Choice}
In this section, we investigate the design choice for the prior network. First, we investigate different conditioning mechanisms, namely, conditioning the affine coupling layers and conditioning the prior network. From Table \ref{tab:flow_arch}, it can be seen the choice of conditioning matters and conditioning the affine layers is the most effective. This intuitively makes sense as we are conditioning multiple times as we are concatenating each coupling layer whereas we just condition the prior once. A similar phenomena is observed in the case of an architectures like \cite{Park_2019_CVPR, chen2018implicit_decoder}, where they concatenated the condition vector multiple times.

In Table \ref{tab:flow_arch}, we also investigate different masking techniques (Dimension masking and Checkered masking) \cite{dinh2016density}  and a distinct flow architecture: Masked Autoregressive Flow (MAF) \cite{papamakarios2017masked}. 
It can be seen from the table that both masking techniques are effective but Dimension masking (RealNVP-D) seems to be more effective than Checkered masking (RealNVP-C).  Furthermore, we find that MAF flow prior network is not as effective as RealNVP. For the remainder of the ablation studies we use the Dimension Masking for RealNVP.

\begin{table}
\centering
\setlength{\tabcolsep}{6pt}
{\small
\begin{tabular}{c|c|ccc}
\toprule
\textit{condition} & \textit{prior} & \textbf{FID$\downarrow$} & \textbf{MMD$\uparrow$} & \textbf{Acc.$\uparrow$}\\
\midrule
affine coupling & \multirow{2}{*}{RealNVP-C} & 2688.72 & 0.6732 & 79.34 \\ \cline{1-1} \cline{3-5}
prior network & ~ & 5227.32 & 0.6600 & 62.39 \\ \hline
\multirow{2}{*}{affine coupling} & RealNVP-D & \textbf{2591.87} & \textbf{0.6751} & \textbf{82.19}\\ \cline{2-5}
~ & MAF~\cite{papamakarios2017masked} & 6052.62 & 0.6273 & 59.40 \\
\bottomrule
\end{tabular}
}
\caption{Effects of different conditional normalizing flow design choices in stage 2.}
\label{tab:flow_arch}
\end{table}

\subsection{Number of Renderings}
Next, we evaluate if using more views helps the generation quality and diversity.  We report the results in Table \ref{tab:clip_Arch}. The views are randomly selected from the  renderings as prepared in \cite{choy20163d}. It can be seen that using more views in general help improve the generation quality and diversity. As we are using a pre-trained CLIP model which is trained on natural images from different viewpoints, training using multiple views of shape renderings allows us to better capture the output distribution of CLIP model.

\subsection{CLIP Architecture}
In this section, we evaluate using different CLIP models to see how increasing the size of CLIP model and using ResNet \cite{he2016deep} or ViT \cite{dosovitskiy2020image} based clip model effects our downstream task.  We empirically observe from Table \ref{tab:clip_Arch} that increasing the size of the model, i.e from ViT-B/32 to ViT-B/16, does not effect the text based generation too much.  A more surprising result is that ResNet based CLIP model performs inferior to Visual Transformers. We hypothesize that patch based methods such as ViT focus more on the foreground object rather than the background. This is especially helpful in the case of image renderings. 

\begin{table}
\centering
\setlength{\tabcolsep}{9pt}
{\small
\begin{tabular}{c|c|ccc}
\toprule
\textit{rendering} & \textit{CLIP} & \textbf{FID$\downarrow$}
& \textbf{MMD$\uparrow$} & \textbf{Acc.$\uparrow$}\\
\midrule
1 & \multirow{3}{*}{ViT-B/32} & 2983.72 & 0.6586 & 79.77\\ \cline{1-1} \cline{3-5}
5 & ~ & 2776.28 & 0.6655 & 80.20 \\ \cline{1-1} \cline{3-5}
10 & ~ & 2622.71 & 0.6652 & 80.63\\ \hline
\multirow{3}{*}{20} & ViT-B/32 & 2591.87 & \textbf{0.6751} & \textbf{82.19} \\ \cline{2-5}
~ & ViT-B/16 & \textbf{2515.81} & 0.6573 & 80.48 \\ \cline{2-5}
~ & RN50x16  & 2906.75 & 0.6591 & 75.93 \\
\bottomrule
\end{tabular}
}
\caption{Effects of different numbers of renderings and CLIP architectures.}
\label{tab:clip_Arch}
\end{table}


\section{Limitations and Future Work}
We believe our method can be improved in several ways. Firstly, the quality of generation is still lacking and we believe a novel future avenue would be to combine ideas from local implicit methods \cite{peng2020convolutional, chibane2020implicit}. Furthermore, our work currently focuses on geometry and it would be interesting to integrate texture to our model. Finally, we are limited by CLIP's trained data distribution and a potential future direction would be to fine tune it for a specfic dataset.

In terms of potential negative impact, language-driven 3D modeling tools enabled by CLIP-Forge might lower the technical barriers to 3D modeling and potentially reduce some tedious 3D modeling tasks for 3D modelers and animators. However, it brings a greater benefit of democratizing 3D content creation to the general public, similar to that everyone can take photos and make videos today. 

\section{Conclusion}
We presented a method, CLIP-Forge, that can efficiently generate multiple 3D shapes while preserving the semantic meaning from a given text prompt. 
Our method requires no text-shape labels as training data, offering an opportunity to leverage shape-only datasets such as ShapeNet. 
Finally, we showed that our model can generate results on other representations such as point clouds and we thoroughly studied different components of the method. 

\appendix


\twocolumn[
\centering
\Large
\vspace{0.5em}Supplementary Material \\
\vspace{1.0em}
] 
\appendix

\section{Architecture and Experiment Details}
For all our experiments in the ablation section of the main paper, we run the second stage network with 3 different seeds and report the mean in the main paper. We take the best seed for the experiment section in the main paper to report the qualitative and quantitative results. The text queries (or prompts) used for classification FID, MMD, and Acc. are shown in Table~\ref{tab:text_query}.  
Note that these text queries are mostly taken from WordNet~\cite{wordnet} with added common synonyms and shape attributes. 
In Table~\ref{tab:cat_acc}, we show category-wise accuracy results of CLIP-Forge in the main paper's Table 1.
For our visualizations, we output a shape with $64^3$ resolution and use the rendering script inspired by \cite{voxrender}. We use a set of different thresholds values and pick the threshold for different category that yields the best visual result. 

In the main paper, we refer to the batch normalization based voxel encoder as VoxEnc, whereas when we add residual connection to VoxEnc we refer it to as ResVoxEnc. For both of these encoders, we have 4 3D convolution layers followed by a linear layer. The input to these encoder is a $32^3$ voxel representation based shape. We also experiment with a point cloud based encoder which is inspired by PointNet. The PointNet encoder has 5 linear layers followed by a max pooling operation. We then use an MLP followed by a final linear layer to project it to the latent size. The input to this encoder is a point cloud with 2048 sampled points. 
For the decoder, we refer to the residual connection based network as RN-OccNet. In this model, we concatenate the query locations with the latent code and pass it through a 5 block ResNet based decoder. We also experiment with conditioning the batchnorm of the decoder instead of concatenating it, which we refer to as CBN-OccNet. Both these decoders are inspired by OccNet~\cite{OccupancyNetworks2019}. For our point cloud based generation experiments, we use a FoldingNet~\cite{yang2018foldingnet} based decoder, where we use two folding based operations with a single square grid.

Finally, we use RealNVP~\cite{dinh2016density} for the prior model. We use 5 blocks of coupling layer containing translation and scale along with batch norm, where the masking is inverted after each block. Each network comprises of a 2-layer MLP followed by a linear layer. A 1024 hidden vector size is used. For the MAF~\cite{papamakarios2017masked} model, we also use the same number of blocks and hidden vectors. 

\definecolor{lvl0}{RGB}{229,121,149}
\definecolor{lvl1}{RGB}{235,142,174}
\definecolor{lvl2}{RGB}{237,170,193}
\definecolor{lvl3}{RGB}{232,198,221}
\definecolor{lvl4}{RGB}{216,225,219}
\definecolor{lvl5}{RGB}{191,243,209}
\definecolor{lvl6}{RGB}{167,249,192}
\definecolor{lvl7}{RGB}{143,246,173}
\definecolor{lvl8}{RGB}{117,235,146}
\definecolor{lvl9}{RGB}{89,215,115}

\begin{table}
\centering
\setlength{\tabcolsep}{0.3pt}
{\tiny
\begin{tabular}{c|c|c|c|c}
\toprule
\cellcolor{lvl4}``a triangular airplane'' & ``an airplane'' & \cellcolor{lvl6}``a jet'' & \cellcolor{lvl9}``a fighter plane'' & \cellcolor{lvl4}``a biplane'' \\\hline
\cellcolor{lvl2}``a seaplane'' & \cellcolor{lvl5}``a space shuttle'' & \cellcolor{lvl3}``a supersonic plane'' & \cellcolor{lvl7}``a rocket plane'' & \cellcolor{lvl4}``a delta wing''\\\hline
\cellcolor{lvl5}``a swept wing plane'' & \cellcolor{lvl2}``a straight wing plane'' & \cellcolor{lvl5}``a propeller plane'' &
``a boeing'' & ``an airbus''\\\hline
``an f-16'' & ``a plane'' & ``an aeroplane'' & ``an aircraft'' & ``a commercial plane''\\\hline
\cellcolor{lvl7}``a square bench'' & \cellcolor{lvl7}``a round bench'' & \cellcolor{lvl8}``a circular bench'' & \cellcolor{lvl8}``a rectangular bench'' & \cellcolor{lvl7}``a thick bench''\\\hline
\cellcolor{lvl7}``a thin bench'' & ``a bench'' & \cellcolor{lvl3}``a pew'' & \cellcolor{lvl1}``a flat bench'' & \cellcolor{lvl3}``a settle''\\\hline
\cellcolor{lvl2}``a back bench'' & \cellcolor{lvl0}``a laboratory bench'' & \cellcolor{lvl3}``a storage bench'' & ``a park bench'' & \cellcolor{lvl6}``a cuboid cabinet''\\\hline
\cellcolor{lvl6}``a round cabinet'' & \cellcolor{lvl3}``a rectangular cabinet'' & \cellcolor{lvl6}``a thick cabinet'' & \cellcolor{lvl3}``a thin cabinet'' &
``a cabinet''\\\hline
\cellcolor{lvl4}``a garage cabinet'' & \cellcolor{lvl7}``a desk cabinet'' & ``a dresser'' & ``a cupboard'' & ``a container''\\\hline
``a case'' & ``a locker'' & ``a cupboard'' & ``a closet'' & ``a sideboard''\\\hline
\cellcolor{lvl6}``a square car'' & \cellcolor{lvl7}``a round car'' &
\cellcolor{lvl6}``a rectangular car'' & \cellcolor{lvl6}``a thick car'' &
\cellcolor{lvl7}``a thin car''\\\hline
``a car'' & \cellcolor{lvl6}``a bus'' & \cellcolor{lvl7}``a shuttle-bus'' & \cellcolor{lvl7}``a pickup car'' & \cellcolor{lvl7}``a truck'' \\\hline 
\cellcolor{lvl8}``a suv'' & \cellcolor{lvl4}``a sports car'' & \cellcolor{lvl7}``a limo'' & \cellcolor{lvl6}``a jeep'' & \cellcolor{lvl8}``a van''\\\hline
\cellcolor{lvl5}``a gas guzzler'' & \cellcolor{lvl5}``a race car'' & \cellcolor{lvl8}``a monster truck'' & \cellcolor{lvl8}``an armored'' & \cellcolor{lvl7}``an atv''\\\hline
\cellcolor{lvl7}``a microbus'' & \cellcolor{lvl2}``a muscle car'' & \cellcolor{lvl7}``a retro car'' & \cellcolor{lvl6}``a wagon car'' & \cellcolor{lvl8}``a hatchback''\\\hline
\cellcolor{lvl6}``a sedan'' & \cellcolor{lvl7}``an ambulance'' & \cellcolor{lvl6}``a roadster car'' &
\cellcolor{lvl9}``a beach wagon'' & ``an auto''\\\hline
``an automobile'' & ``a motor car'' & \cellcolor{lvl5}``a square chair'' & \cellcolor{lvl6}``a round chair'' & \cellcolor{lvl4}``a rectangular chair''\\\hline
\cellcolor{lvl6}``a thick chair'' & \cellcolor{lvl9}``a thin chair'' & ``a chair'' & \cellcolor{lvl5}``an arm chair'' & \cellcolor{lvl6}``a bowl chair''\\\hline
\cellcolor{lvl6}``a rocking chair'' & \cellcolor{lvl6}``an egg chair'' & \cellcolor{lvl4}``a swivel chair'' & \cellcolor{lvl8}``a bar stool'' & \cellcolor{lvl5}``a ladder back chair''\\\hline
\cellcolor{lvl6}``a throne'' & \cellcolor{lvl7}``an office chair'' & \cellcolor{lvl8}``a wheelchair'' & \cellcolor{lvl9}``a stool'' & \cellcolor{lvl6}``a barber chair''\\\hline
\cellcolor{lvl7}``a folding chair'' & \cellcolor{lvl7}``a lounge chair'' & \cellcolor{lvl4}``a vertical back chair'' & \cellcolor{lvl8}``a recliner'' & \cellcolor{lvl5}``a wing chair''\\\hline
\cellcolor{lvl4}``a sling'' & ``a seat'' & ``a cathedra'' & \cellcolor{lvl5}``a square monitor'' & \cellcolor{lvl8}``a round monitor''\\\hline
\cellcolor{lvl4}``a rectangular monitor'' & \cellcolor{lvl5}``a thick monitor'' & \cellcolor{lvl5}``a thin monitor'' & ``a monitor'' & \cellcolor{lvl1}``a crt monitor''\\\hline
``a TV'' & ``a digital display'' & ``a flat panel display'' & ``a screen'' & ``a television''\\\hline
``a telly'' & ``a video'' & \cellcolor{lvl6}``a square lamp'' & \cellcolor{lvl5}``a round lamp'' & \cellcolor{lvl5}``a rectangular lamp''\\\hline
\cellcolor{lvl8}``a cuboid lamp'' & \cellcolor{lvl5}``a circular lamp'' & \cellcolor{lvl7}``a thick lamp'' & \cellcolor{lvl4}``a thin lamp'' & ``a lamp''\\\hline
\cellcolor{lvl6}``a street lamp'' & \cellcolor{lvl1}``a fluorescent lamp'' & \cellcolor{lvl1}``a gas lamp'' & \cellcolor{lvl7}``a bulb'' & ``a lantern''\\\hline
``a table lamp'' & ``a torch'' & \cellcolor{lvl8}``a square loudspeaker'' & \cellcolor{lvl7}``a round loudspeaker'' & \cellcolor{lvl8}``a rectangular loudspeaker''\\\hline
\cellcolor{lvl4}``a circular loudspeaker'' & \cellcolor{lvl8}``a thick loudspeaker'' & \cellcolor{lvl5}``a thin loudspeaker'' & ``a loudspeaker'' & \cellcolor{lvl5}``a subwoofer speaker''\\\hline
``a speaker'' & ``a speaker unit'' & ``a tannoy'' & \cellcolor{lvl6}``a thick gun'' & \cellcolor{lvl6}``a thin gun''\\\hline
``a gun'' & \cellcolor{lvl3}``a machine gun'' & \cellcolor{lvl8}``a sniper rifle'' & \cellcolor{lvl0}``a pistol'' & \cellcolor{lvl3}``a shotgun''\\\hline
``an ak-47'' & ``an uzi'' & ``an M1 Garand'' & ``a M-16'' & ``a firearm''\\\hline
``a shooter'' & ``a weapon'' & \cellcolor{lvl6}``a square sofa'' & \cellcolor{lvl8}``a round sofa'' & \cellcolor{lvl5}``a rectangular sofa''\\\hline
\cellcolor{lvl6}``a thick sofa'' & \cellcolor{lvl8}``a thin sofa'' & ``a sofa'' & \cellcolor{lvl7}``a double couch'' & \cellcolor{lvl8}``a love seat''\\\hline
\cellcolor{lvl7}``a chesterfield'' & \cellcolor{lvl4}``a convertible sofa'' & \cellcolor{lvl6}``an L shaped sofa'' & \cellcolor{lvl5}``a settee sofa'' & \cellcolor{lvl5}``a daybed''\\\hline
\cellcolor{lvl4}``a sofa bed'' & \cellcolor{lvl4}``an ottoman'' & ``a couch'' & ``a lounge'' & ``a divan''\\\hline
``a futon'' & \cellcolor{lvl8}``a square table'' & \cellcolor{lvl7}``a round table'' & \cellcolor{lvl8}``a circular table'' & \cellcolor{lvl3}``a rectangular table''\\\hline
\cellcolor{lvl4}``a thick table'' & \cellcolor{lvl9}``a thin table'' & ``a table'' & \cellcolor{lvl4}``a dressing table'' & \cellcolor{lvl3}``a desk''\\\hline
\cellcolor{lvl5}``a refactory table'' & \cellcolor{lvl8}``a counter'' & \cellcolor{lvl2}``an operating table'' & \cellcolor{lvl3}``a stand'' & \cellcolor{lvl2}``a billiard table''\\\hline
\cellcolor{lvl1}``a pool table'' & \cellcolor{lvl3}``a ping-pong table'' & \cellcolor{lvl5}``a console table'' & ``an altar table'' & ``a worktop''\\\hline
``a workbench'' & \cellcolor{lvl7}``a square phone'' & \cellcolor{lvl7}``a rectangular phone'' & \cellcolor{lvl4}``a thick phone'' & \cellcolor{lvl7}``a thin phone''\\\hline
``a phone'' & \cellcolor{lvl9}``a desk phone'' & \cellcolor{lvl8}``a flip-phone'' & ``a telephone'' & ``a telephone set''\\\hline
``a cellular telephone'' & ``a cellular phone'' & ``a cellphone'' & ``a cell'' & ``a mobile phone''\\\hline
``an iphone'' & \cellcolor{lvl6}``a square boat'' & \cellcolor{lvl9}``a round boat'' & \cellcolor{lvl6}``a rectangular boat'' & \cellcolor{lvl7}``a thick boat''\\\hline
\cellcolor{lvl7}``a thin boat'' & ``a boat'' & \cellcolor{lvl6}``a war ship'' & \cellcolor{lvl3}``a sail boat'' & \cellcolor{lvl2}``a speedboat''\\\hline
\cellcolor{lvl3}``a cabin cruiser'' & \cellcolor{lvl8}``a yacht'' & ``a rowing boat'' & ``a watercraft'' & ``a ship''\\\hline
``a canal boat'' & ``a ferry'' & ``a steamboat'' & ``a barge'' & ~\\\midrule
\cellcolor{lvl0}0/9 & \cellcolor{lvl1}1/9 & \cellcolor{lvl2}2/9 & \cellcolor{lvl3}3/9 & \cellcolor{lvl4}4/9\\\hline
\cellcolor{lvl5}5/9 & \cellcolor{lvl6}6/9 & \cellcolor{lvl7}7/9 & \cellcolor{lvl8}8/9 & \cellcolor{lvl9}9/9\\
\bottomrule
\end{tabular}
}
\caption{The full list of text queries.    The colors show the results of the human perceptual evaluation study as described in section \ref{sec:human_eval_supp}.   Green indicates queries which gave rise to distinctive and recognisable shapes, while red indicates the shapes could not be distinguished from those generated using the ShapeNet category name.  The key at the bottom shows the fraction of the nine crowd workers who found the generated shape recognisable.   White indicates the query was not rated in the perceptual study.}
\label{tab:text_query}
\end{table}

\section{Comparison with Supervised Models}

\begin{table}
\centering
\setlength{\tabcolsep}{9pt}
{\small
\begin{tabular}{c|c|ccc}
\toprule
\textit{method} & \textit{dataset}  & \textbf{FID$\downarrow$} & \textbf{MMD$\uparrow$} & \textbf{Acc.$\uparrow$}\\
\midrule
sup. & \multirow{2}{*}{T2S~\cite{chen2018text2shape}} & 14881.96 & 0.1418 & 6.84\\ \cline{1-1} \cline{3-5}
ours & ~ & 14746.90 & 0.5412 & 30.77\\ \hline
sup. & \multirow{2}{*}{SN13~\cite{choy20163d}} & 19896.11 & 0.1805 & 14.10 \\ \cline{1-1} \cline{3-5}
ours & ~ &\textbf{2425.25} & \textbf{0.6607} & \textbf{83.33}\\
\bottomrule
\end{tabular}
}
\caption{Additional detailed comparisons with supervised models, where sup. stands for the supervised model.}
\label{tab:baseline2}
\end{table}

In this section, we provide a more detailed comparison
between CLIP-Forge and supervised methods. Note, it is
not clear how to compare our zero-shot model with supervised models. As our end goal is to generate shapes across categories and text queries, we decide to use our original text query subset (mentioned above) and Shapenet (v2) test dataset as the test set. This test set ensures we test
on commonly described words for different shape category as mentioned in WordNet \cite{wordnet}. We consider two datasets: T2S is the annotated text-shape description dataset from text2shape \cite{chen2018text2shape} which mainly contains information regarding texture, and SN13 is the ShapeNet (v2) subset containing 13 categories from \cite{choy20163d}.

T2S dataset \cite{chen2018text2shape} has text labels only for chair and tableclass, so we train a supervised baseline model that has a linear layer connecting a pre-trained CLIP text encoder \cite{radford2021learning} and a pre-trained occupancy network decoder \cite{OccupancyNetworks2019} using an L2 loss in the latent space. We use the same text encoder and shape decoder in this baseline to ensure a fair comparison. We compare the baseline model with our model which does not have use any supervision from text labels and is also trained on T2S shape dataset (chair and table only).
The results are shown in the first part of Table \ref{tab:baseline2}. It can be seen from the table that our model can generate shapes in chair and table categories based on common words with higher quality despite not using any text label information.

To test baseline models on all of ShapeNet subset
(SN13), as there is no text label data, we create a simple supervision signal by directly using the category name as the text for training the supervised model. The results are shown in second part of Table \ref{tab:baseline2}. It can be seen that our model outperforms the supervised method, demonstrating its stronger zero-shot generalization ability. This results also indicate that our model scales better with more data without requiring text-shape labels. In Figure \ref{fig:baselines}, we show
qualitative results of the supervised baselines, where the model fails to generate cars when trained on T2S, and fails to capture the details of sports car when trained with SN13.

\section{Category-wise Accuracy Results}
We also report category-wise accuracy results obtained from our classifier for our method in Table \ref{tab:cat_acc}. It can be generally noted that our method can generate shapes across all categories of Shapenet. However, accuracy across some categories such as airplane and car are higher than other categories such as boat and loudspeaker. We hypothesize that this may be due to some categories having larger data points during training compared to others.  

\begin{table*}
\centering
{\small
\setlength{\tabcolsep}{6.5pt}
\begin{tabular}{cccccccccccccc}
\toprule
Airplane  & Bench & Cabinet & Car & Chair & Monitor & Lamp & Loudspeaker & Gun & Sofa & Table & Phone & Boat \\
\midrule
95.00 &  64.29 & 87.50 &  96.88 &  96.15 &  92.86 & 93.33 &  45.45 & 92.86 & 89.47 & 75.00 & 60.00 & 61.11\\
\bottomrule
\end{tabular}
\caption{Category-wise accuracy results}. 
\label{tab:cat_acc}
}
\end{table*}

\section{Comparison with Text2Img+Img2Shape}
\begin{figure*}
\begin{center}
\setlength{\tabcolsep}{8pt}
\small{
\begin{tabular}{cccc}
\includegraphics[width=0.15\linewidth]{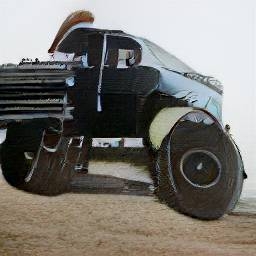} &
\includegraphics[width=0.15\linewidth]{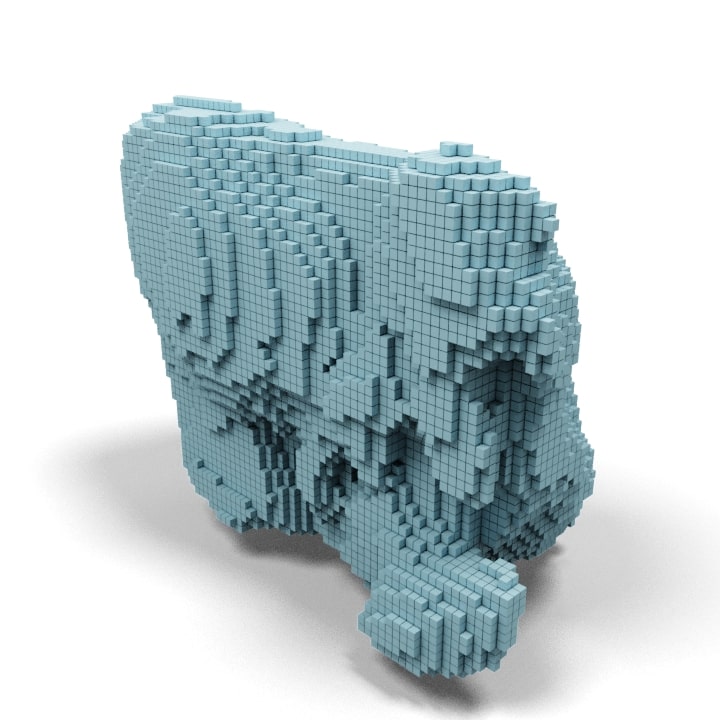} &
\includegraphics[width=0.15\linewidth]{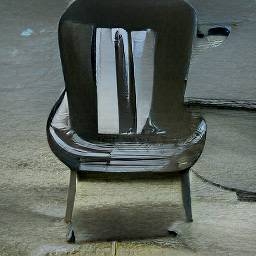} &
\includegraphics[width=0.15\linewidth]{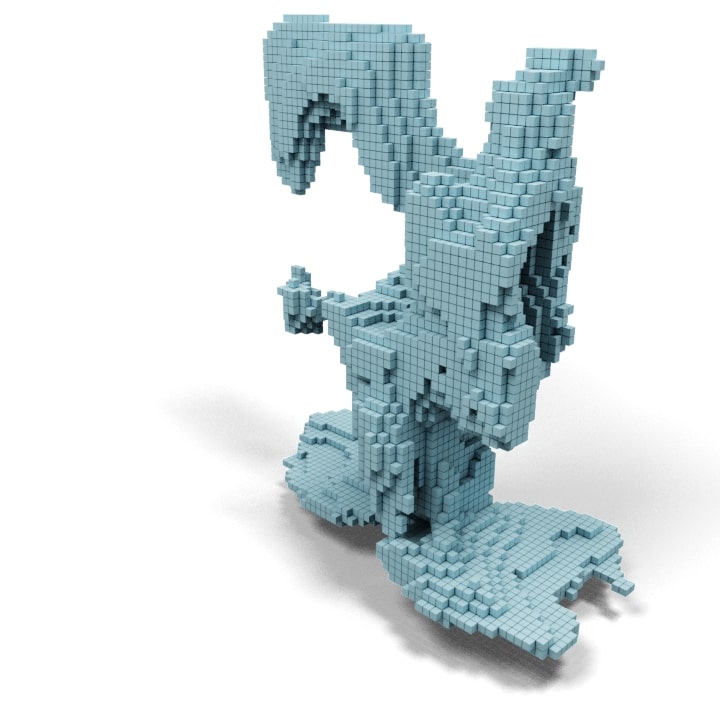}\\
\end{tabular}
}
\end{center}
\caption{Text2Img+Img2Shape baseline intermediate and final results: ``a monster truck'', ``a round chair''.}
\label{fig:dallemini}
\end{figure*}

In this section, we compare our method to off-the-shelf networks that simply generate an image from text first and then generate a 3D shape from the image. We use  pre-trained DALLE-mini for converting a text to image and use a pre-trained occupancy network with image encoder to convert an image to 3D shape. The results are shown in Fig.~\ref{fig:dallemini}. It can be seen that the resulting shapes suffer from poor quality. This is mainly due to the domain gap between generated images and natural images such as distortion artifacts and unclean background.  

\section{Effect of Threshold Parameter}
\begin{figure*}[t!]
\begin{center}
\setlength{\tabcolsep}{2pt}
\small{
\begin{tabular}{cccccc}
\includegraphics[width=0.15\linewidth]{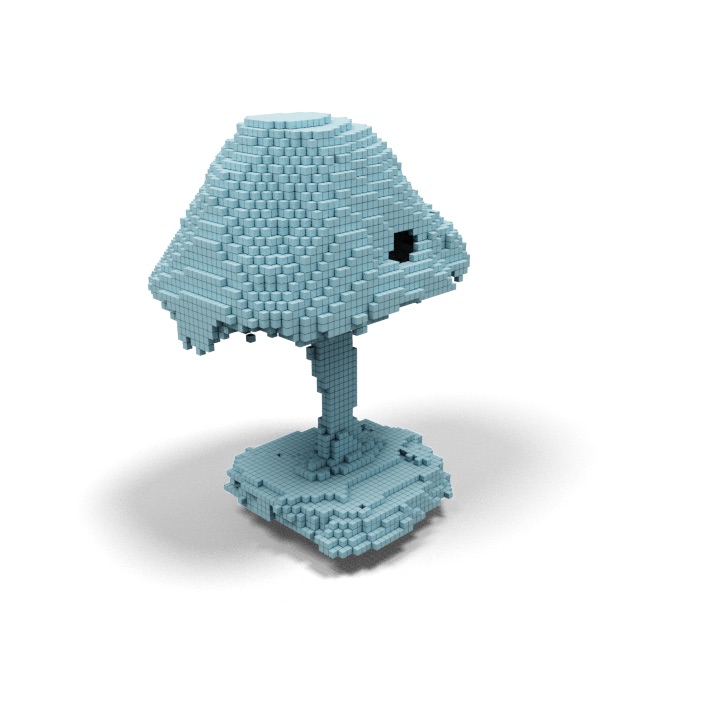} &
\includegraphics[width=0.15\linewidth]{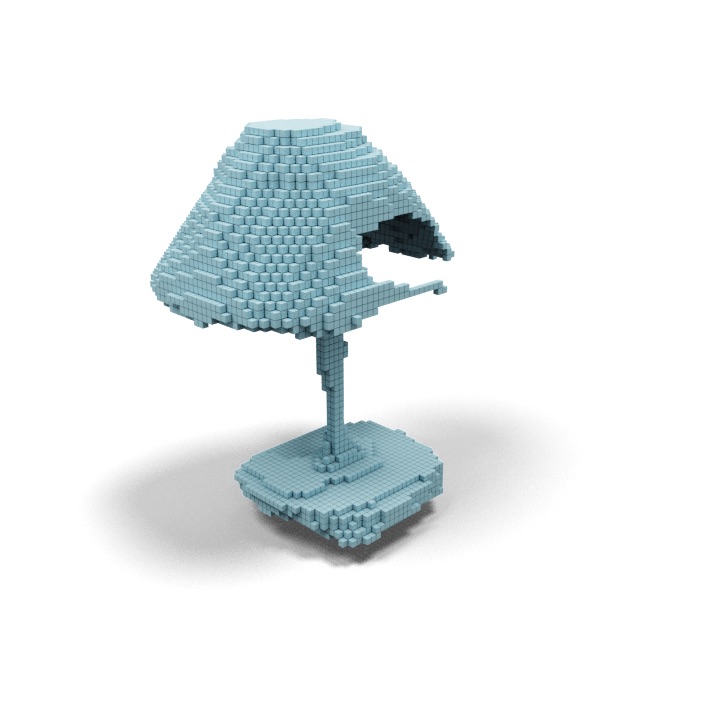} &
\includegraphics[width=0.15\linewidth]{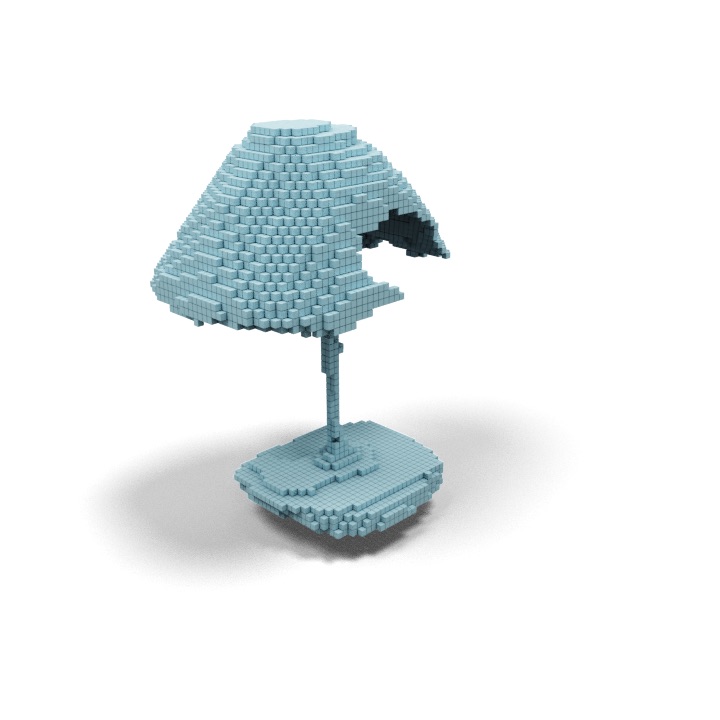} &
\includegraphics[width=0.15\linewidth]{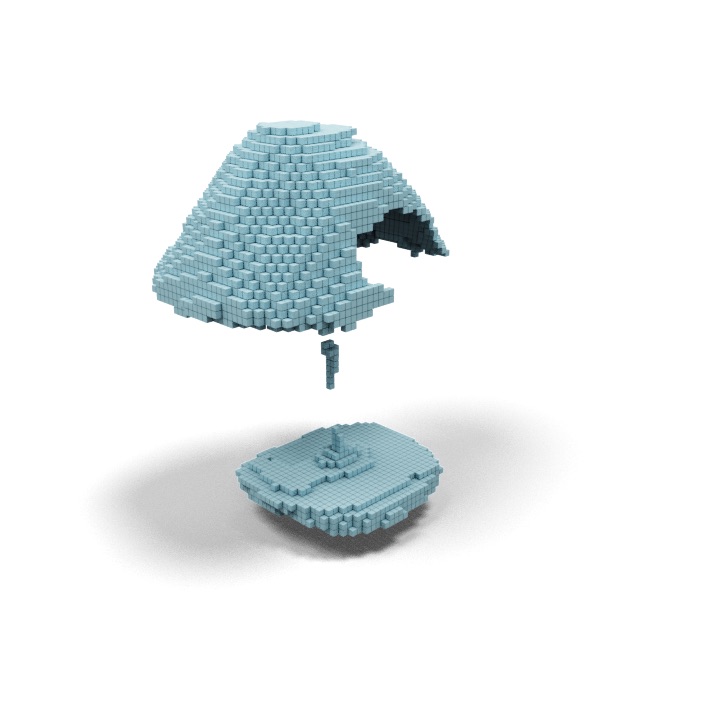} &
\includegraphics[width=0.15\linewidth]{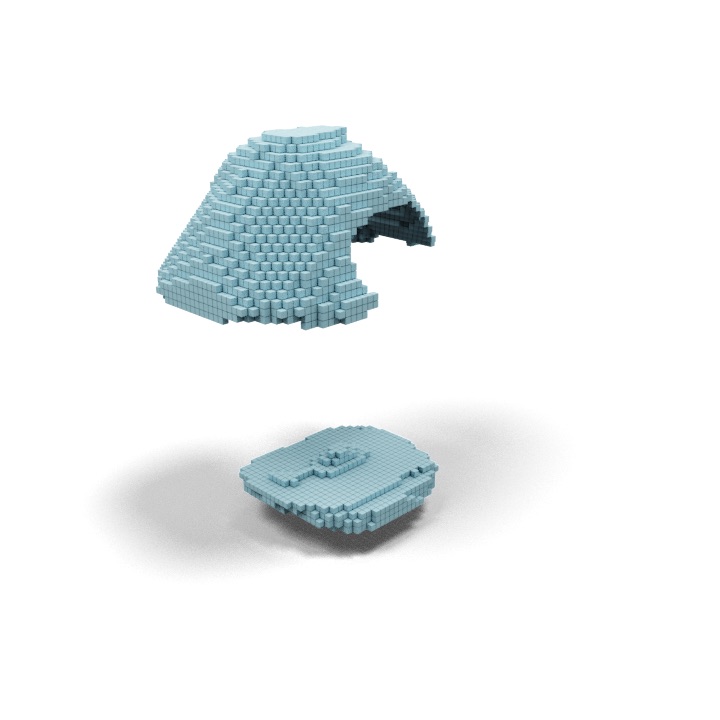} &
\includegraphics[width=0.15\linewidth]{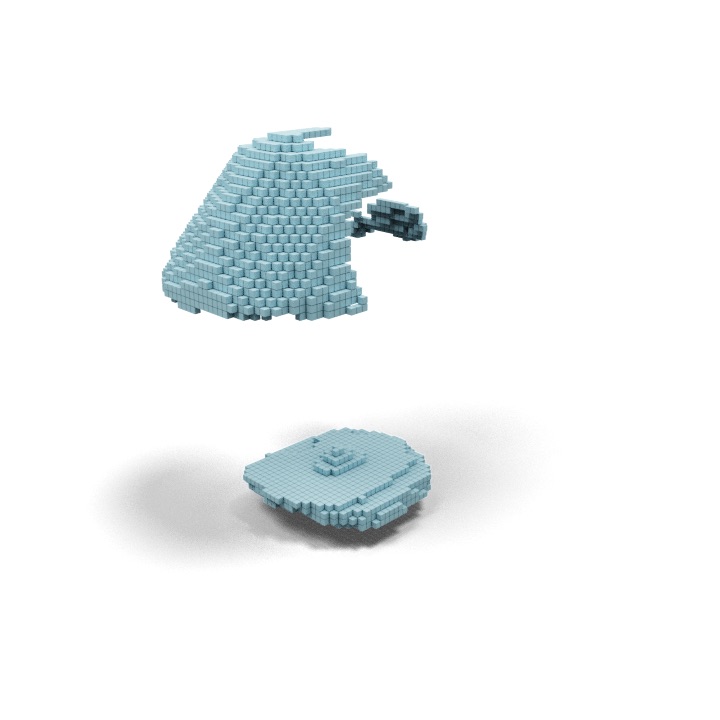}\\

\includegraphics[width=0.15\linewidth]{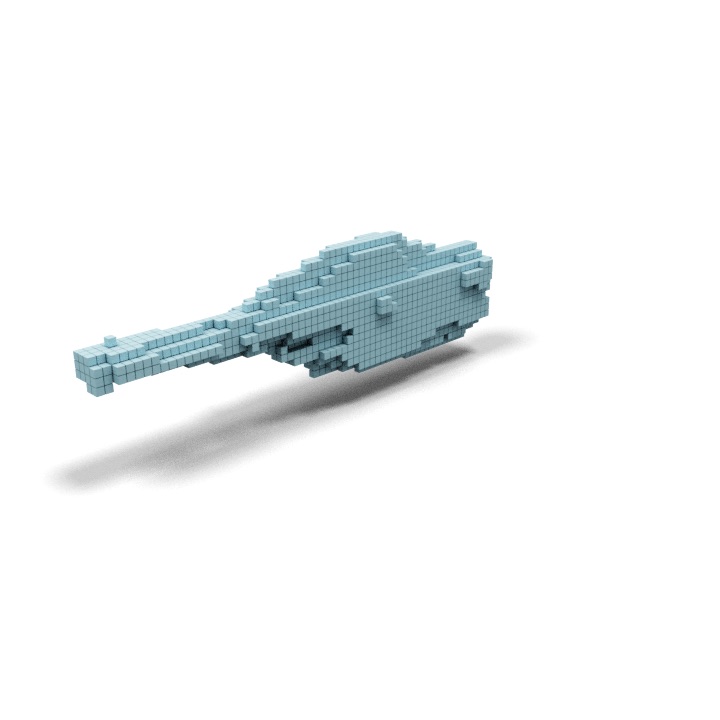} &
\includegraphics[width=0.15\linewidth]{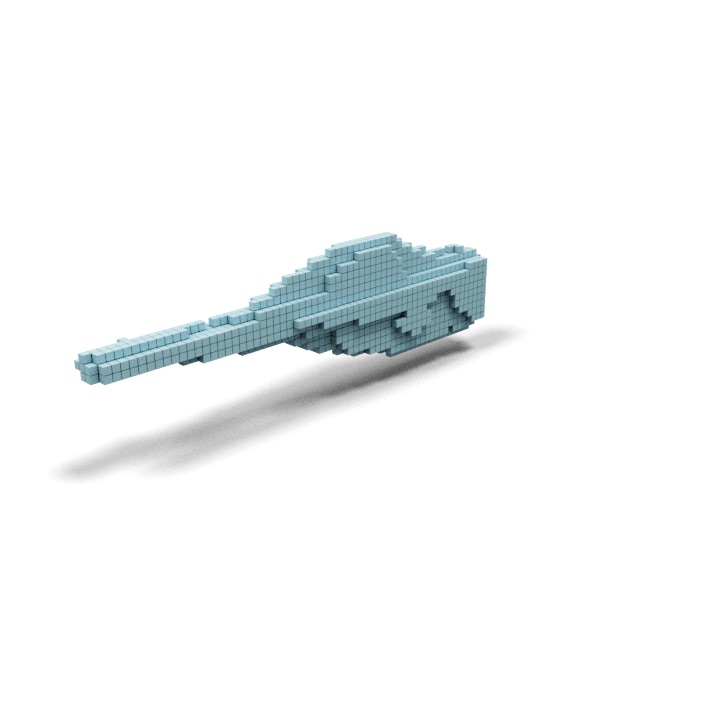} &
\includegraphics[width=0.15\linewidth]{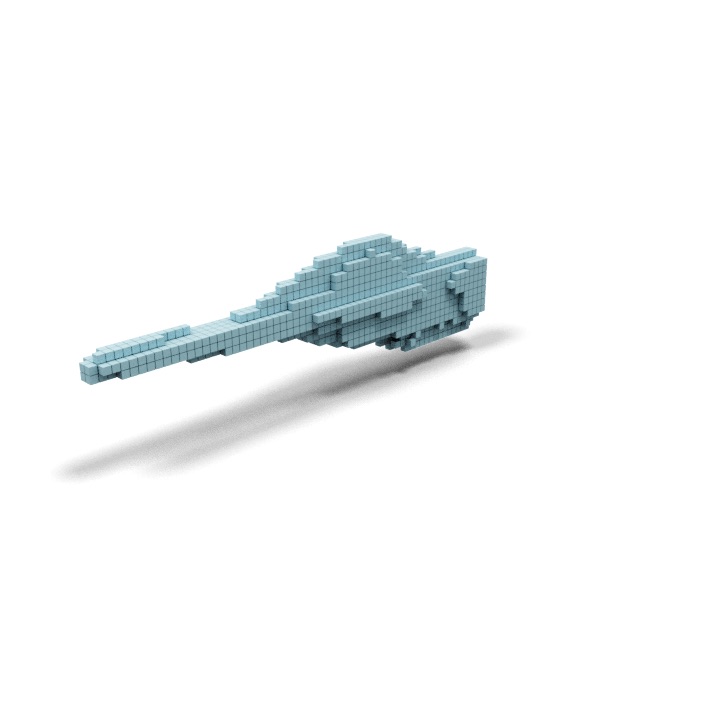} &
\includegraphics[width=0.15\linewidth]{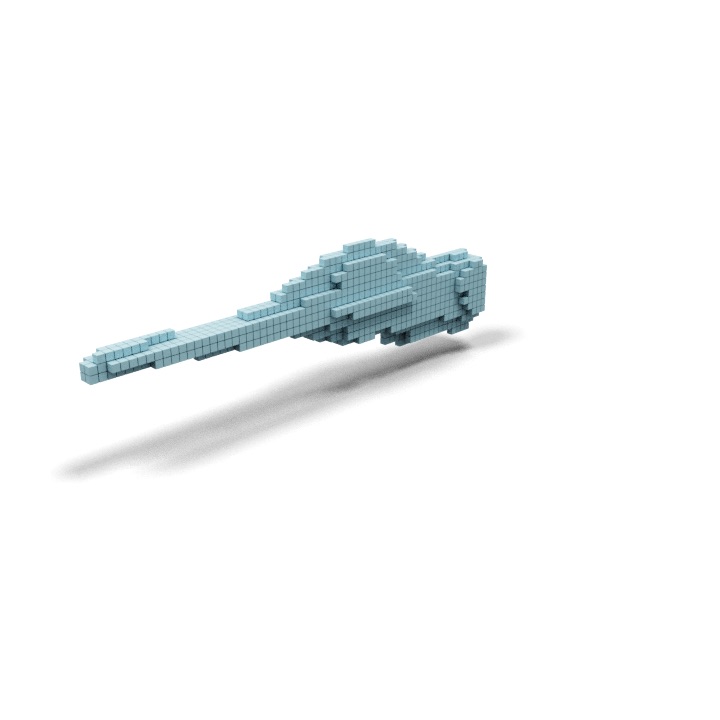} &
\includegraphics[width=0.15\linewidth]{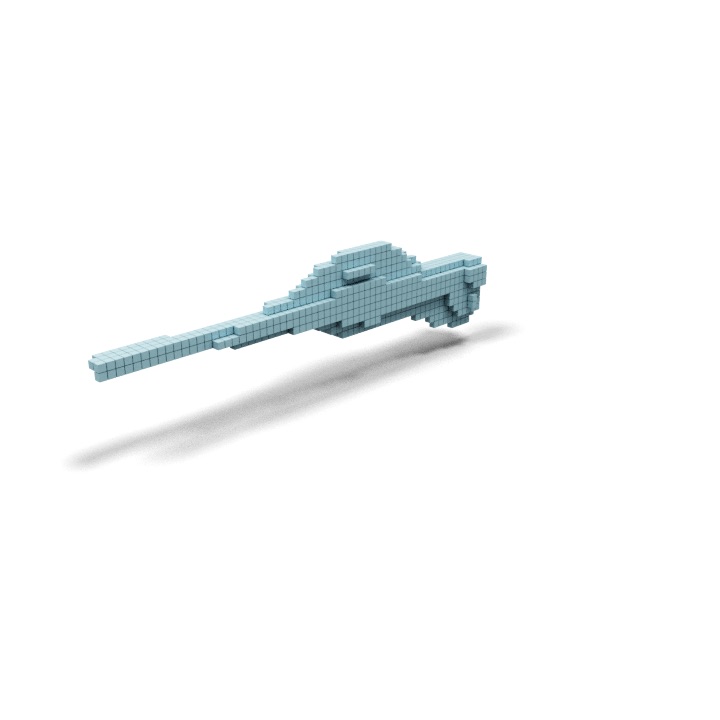} &
\includegraphics[width=0.15\linewidth]{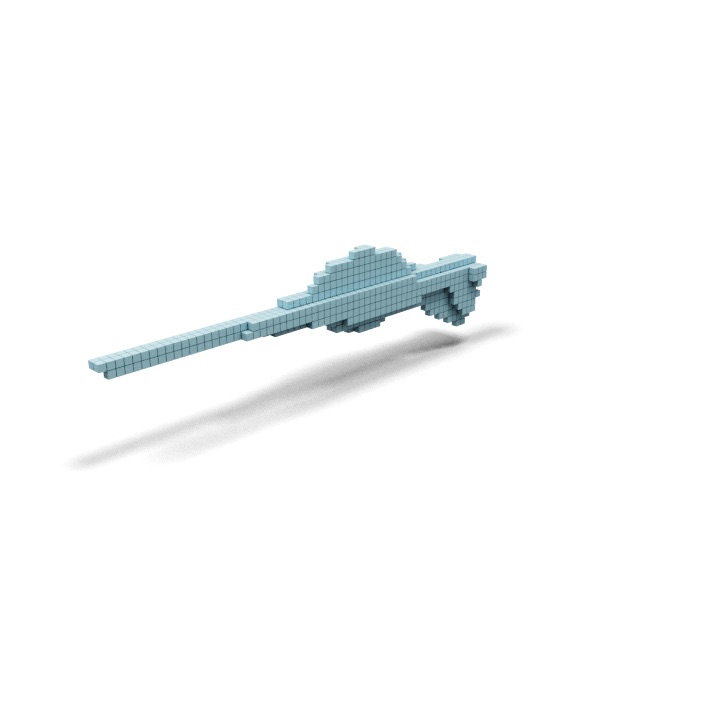}\\

\includegraphics[width=0.15\linewidth]{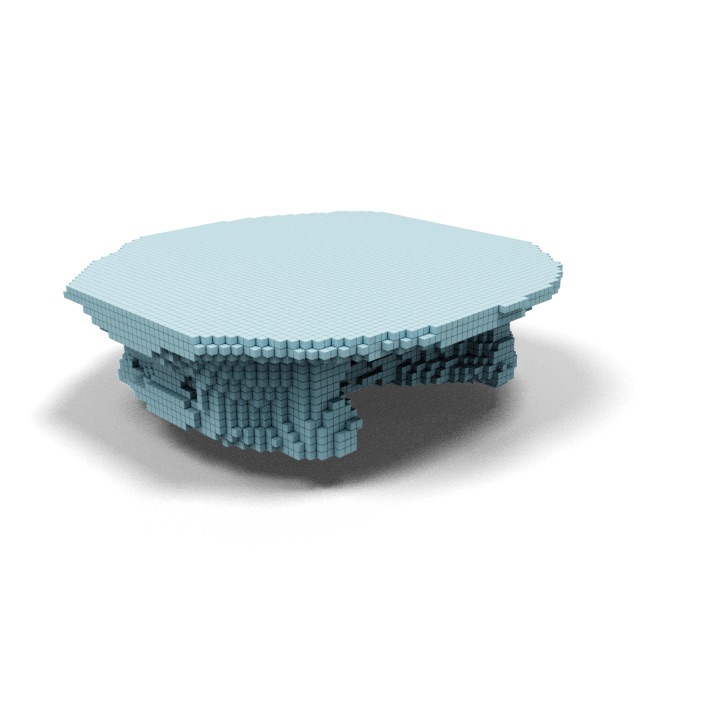} &
\includegraphics[width=0.15\linewidth]{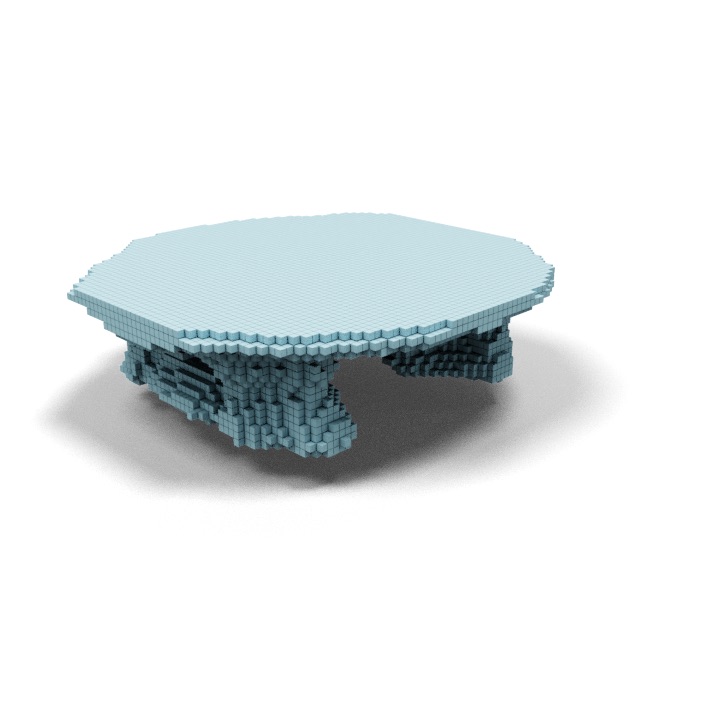} &
\includegraphics[width=0.15\linewidth]{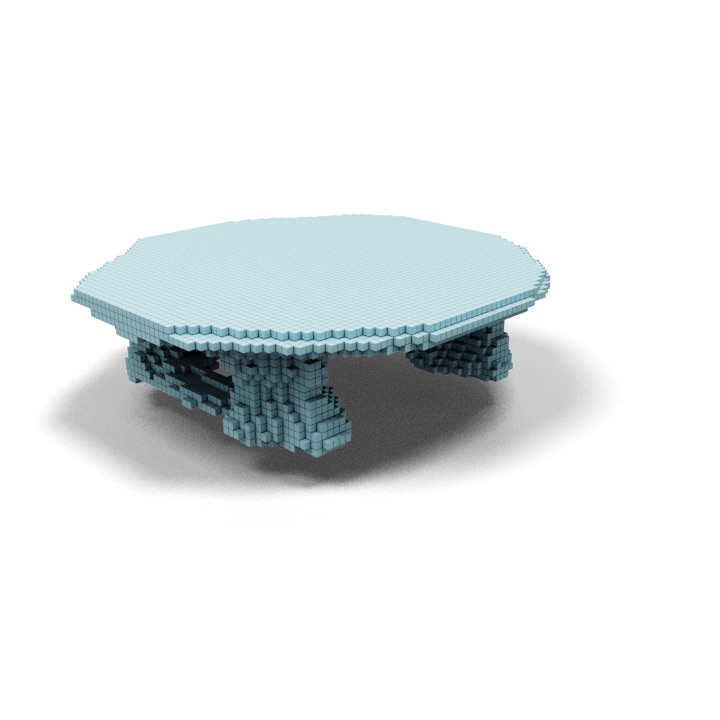} &
\includegraphics[width=0.15\linewidth]{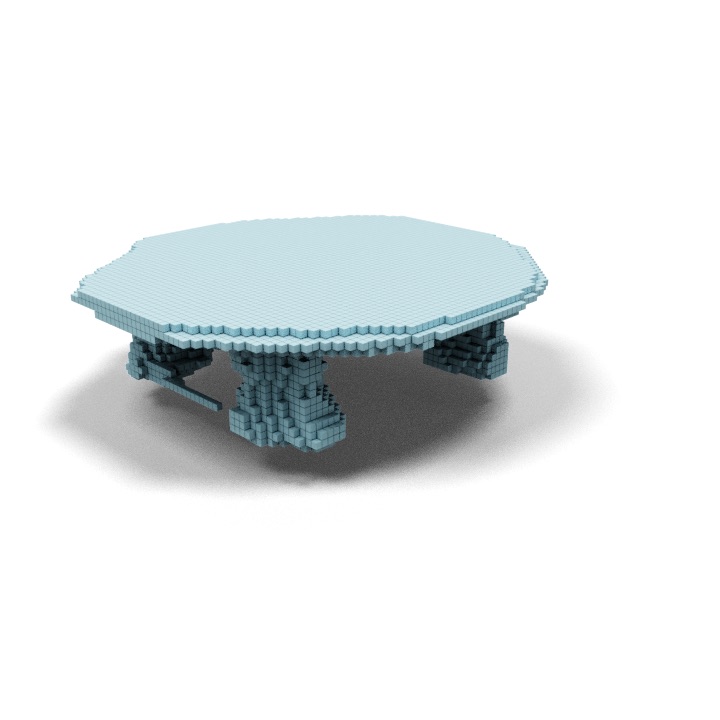} &
\includegraphics[width=0.15\linewidth]{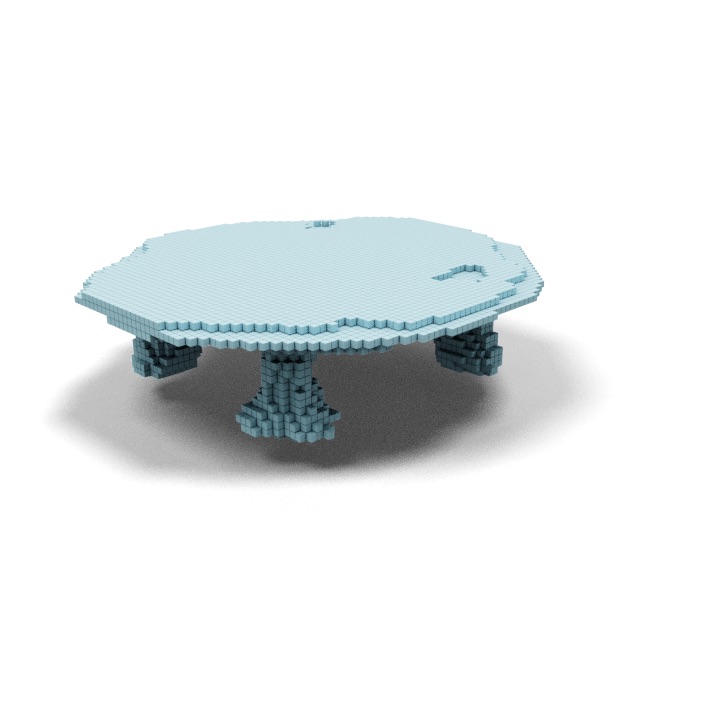} &
\includegraphics[width=0.15\linewidth]{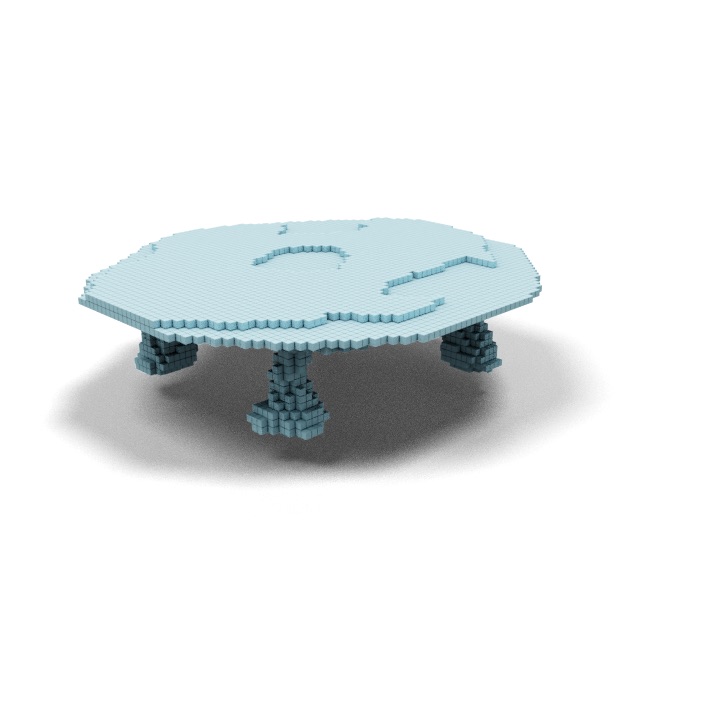}\\

\includegraphics[width=0.15\linewidth]{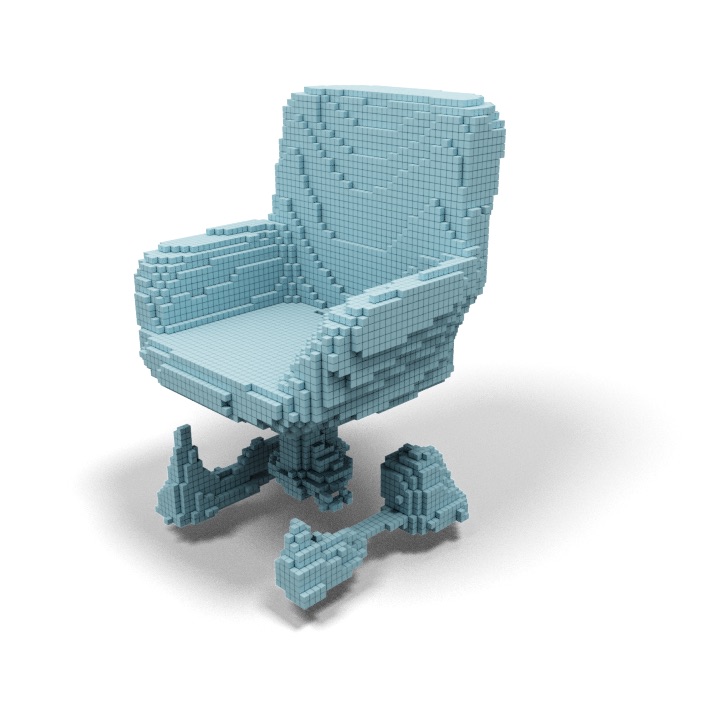} &
\includegraphics[width=0.15\linewidth]{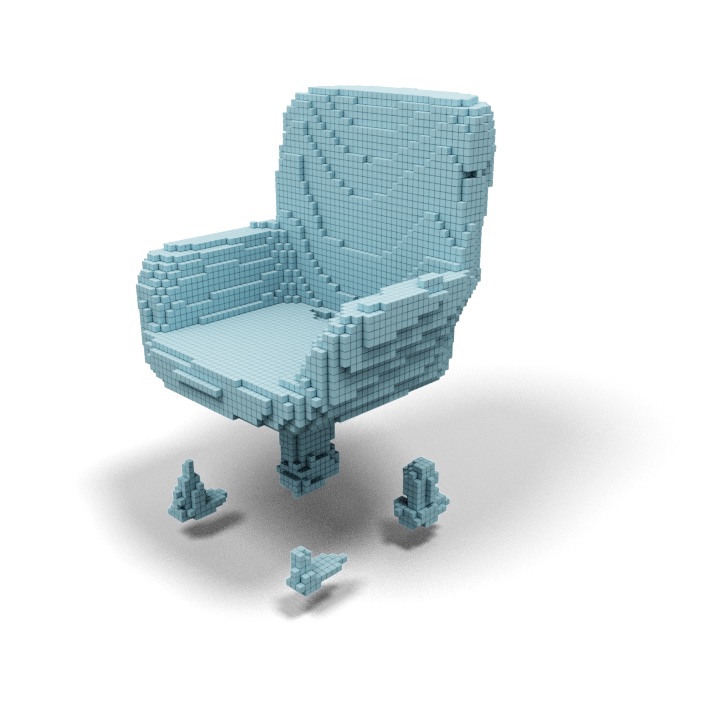} &
\includegraphics[width=0.15\linewidth]{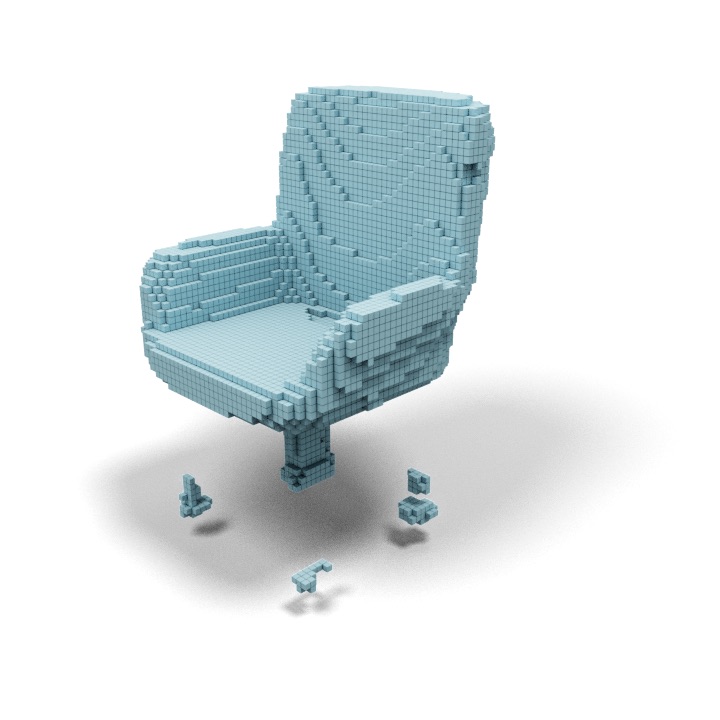} &
\includegraphics[width=0.15\linewidth]{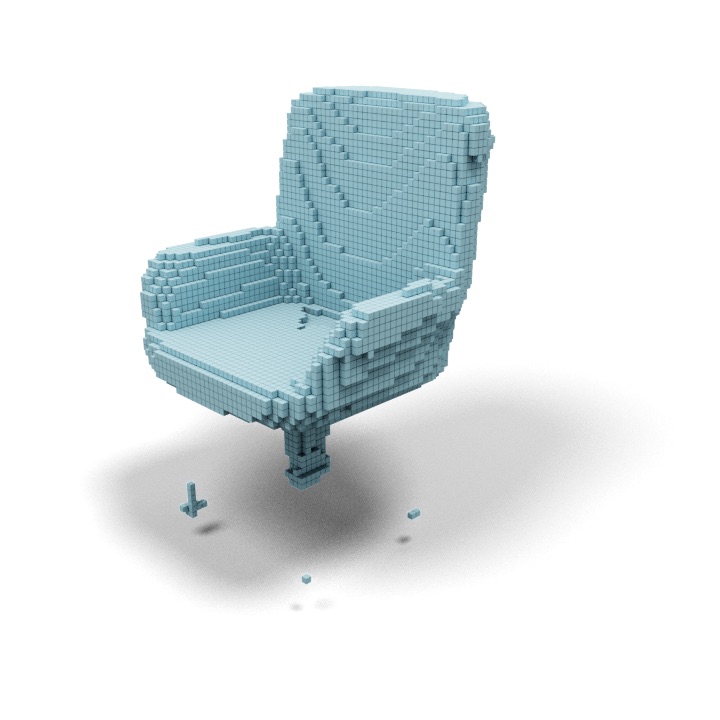} &
\includegraphics[width=0.15\linewidth]{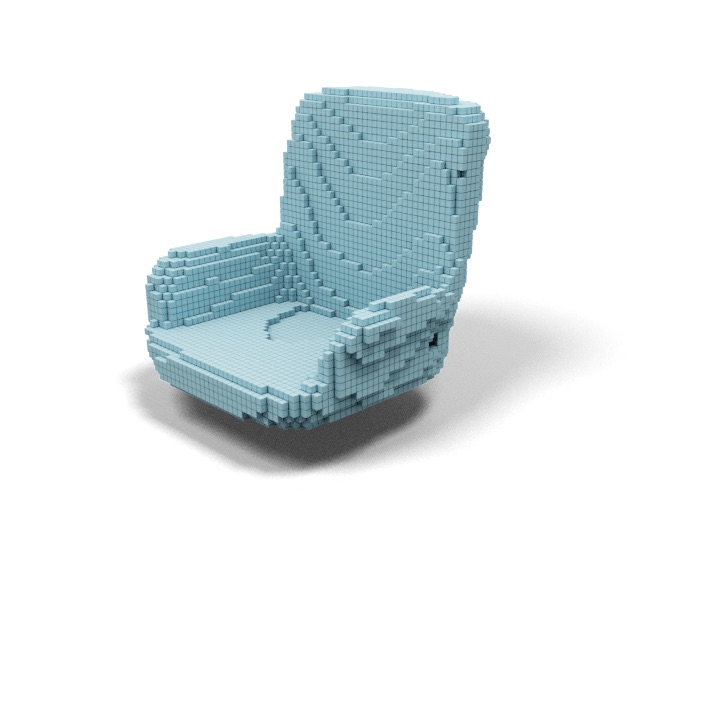} &
\includegraphics[width=0.15\linewidth]{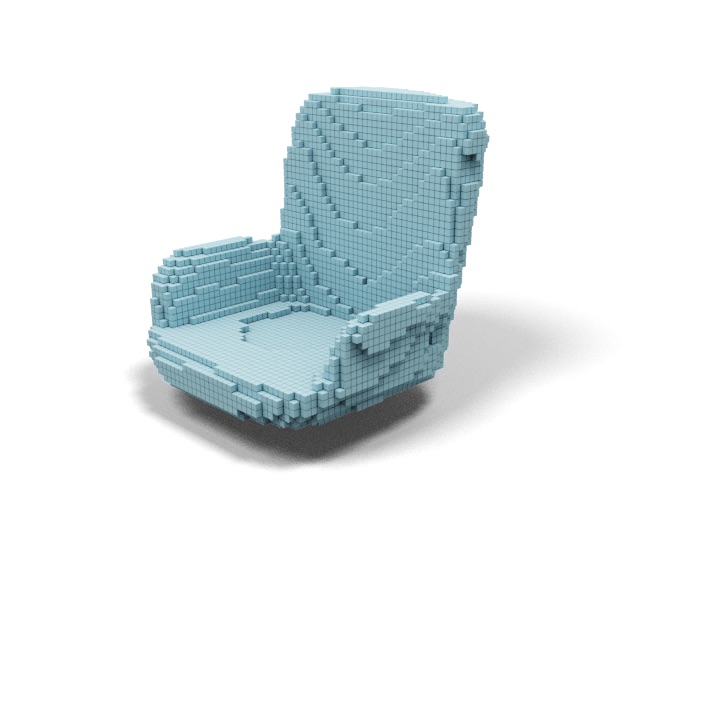}\\

Threshold = 0.025  & Threshold = 0.050 & Threshold = 0.075 & Threshold = 0.1 & Threshold = 0.2 & Threshold = 0.3 \\

\end{tabular}
}
\end{center}
  \caption{Effect of different thresholds for text: "a lamp", "a sniper rifle", "a round table" and "a swivel chair".}
\label{fig:threshold_para}
\end{figure*}

Our results are strongly affected by the threshold used to create the occupancy value. We use a constant threshold value of $0.05$ for our metrics (Acc., FID and MMD) and human perceptual evaluations. However, for our visual results we do a grid search and choose the best threshold value.  Figure \ref{fig:threshold_para}, shows the visual results of different thresholds. It can be seen that different shapes require different threshold which depends on the category and local details of the shape. We believe that our metrics and human evaluation results can be further improved if a better technique is discovered for threshold tuning. 

\section{Out of Distribution Generation}
\begin{figure*}[t!]
\begin{center}
\setlength{\tabcolsep}{2pt}
\small{
\begin{tabular}{cccccc}
\includegraphics[width=0.15\linewidth]{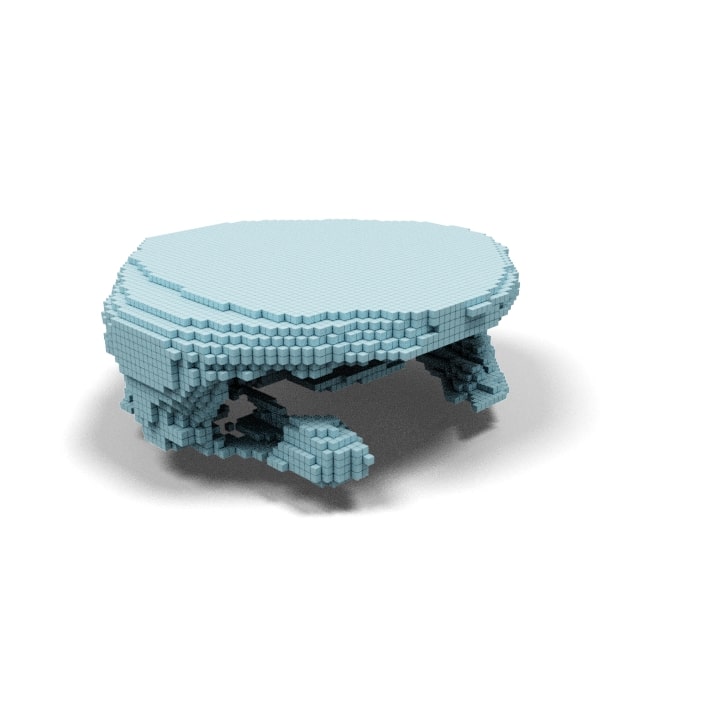} &
\includegraphics[width=0.15\linewidth]{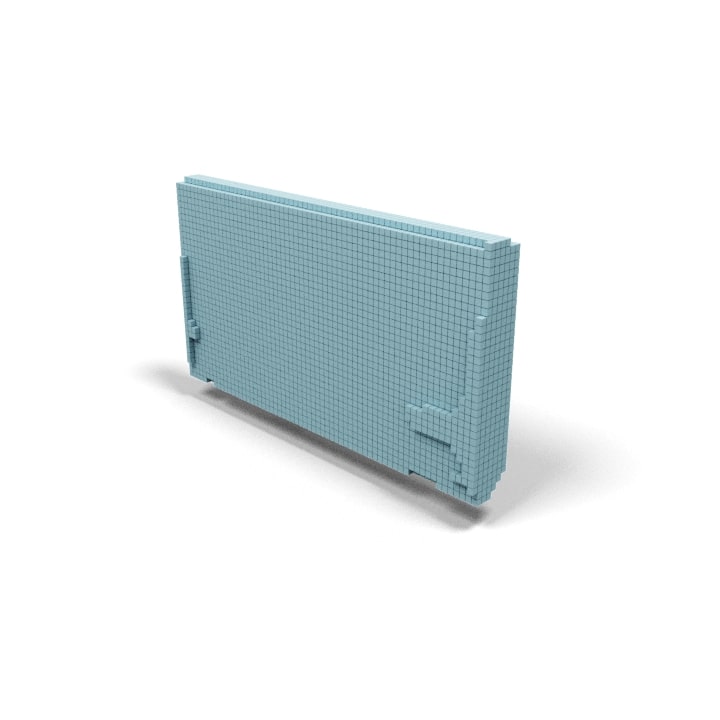} &
\includegraphics[width=0.15\linewidth]{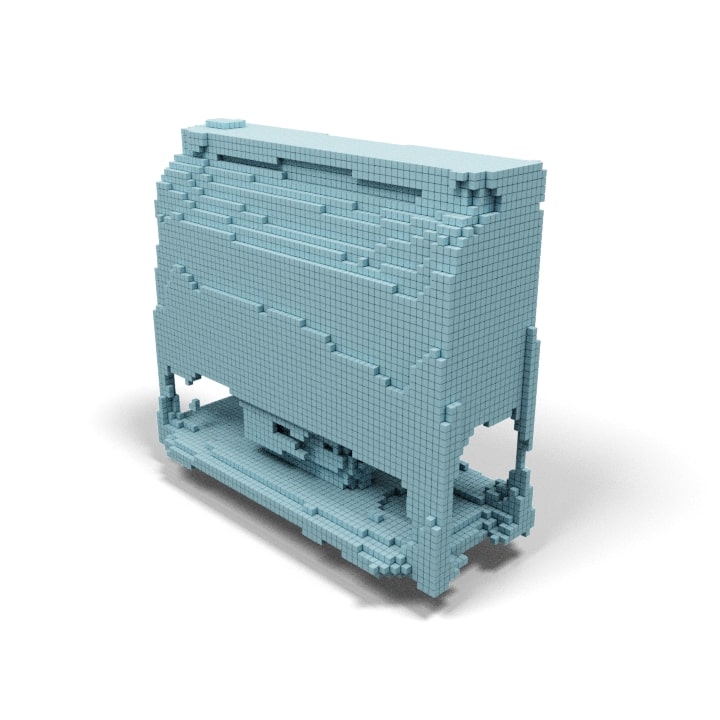} &
\includegraphics[width=0.15\linewidth]{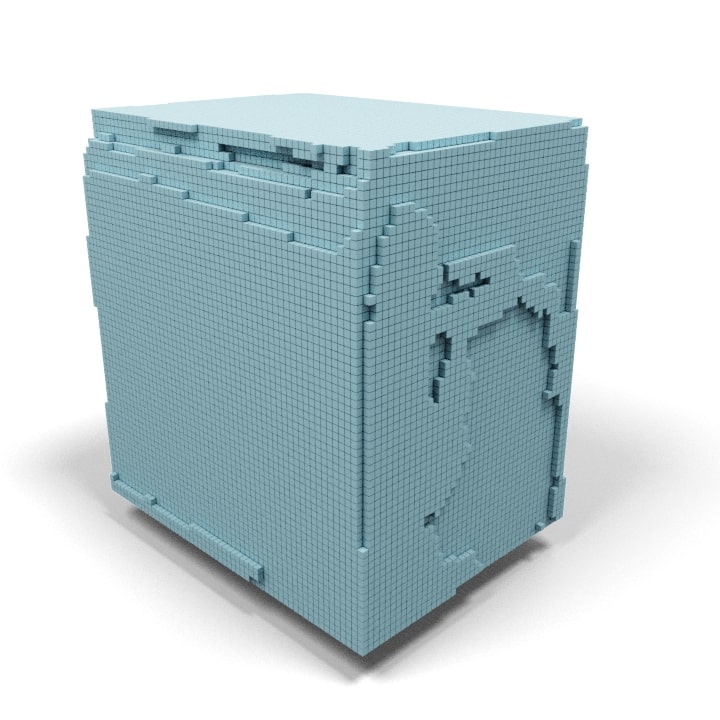} &
\includegraphics[width=0.15\linewidth]{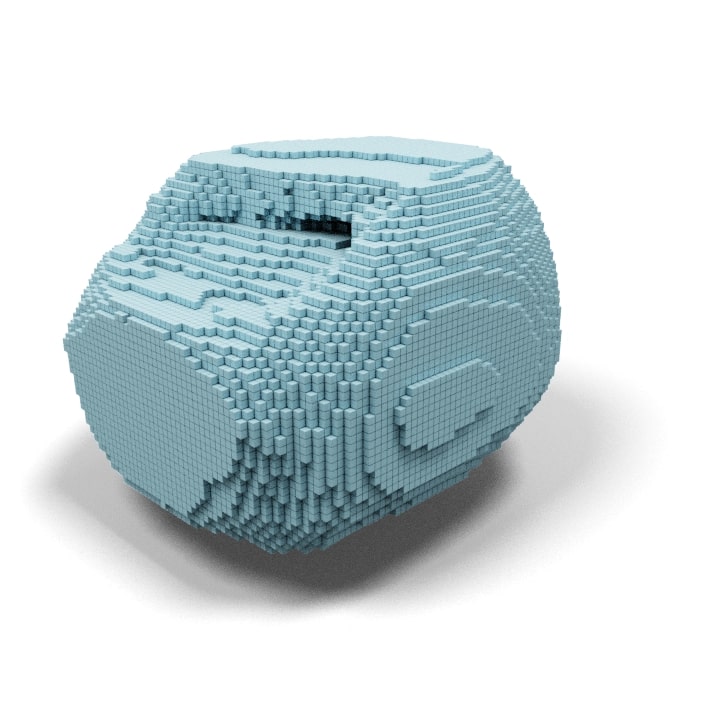} &
\includegraphics[width=0.15\linewidth]{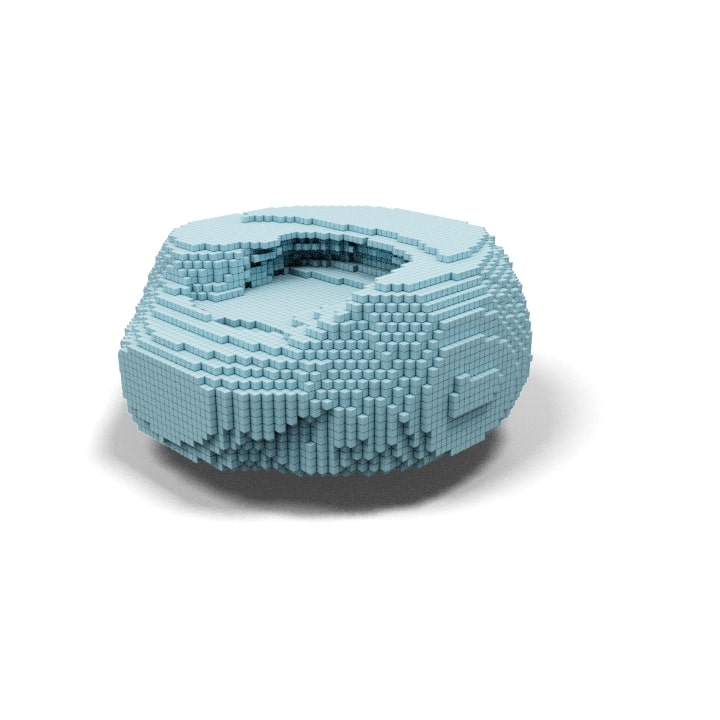} \\
``a circle'' & ``a rectangle'' & ``a square'' & ``a cuboid'' & ``a sphere'' & ``an ellipsoid''\\
\includegraphics[width=0.15\linewidth]{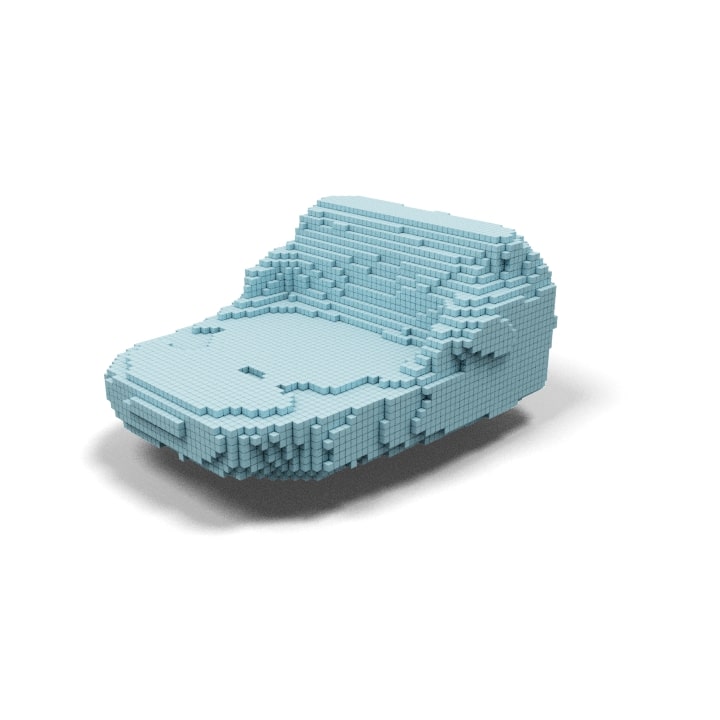} &
\includegraphics[width=0.15\linewidth]{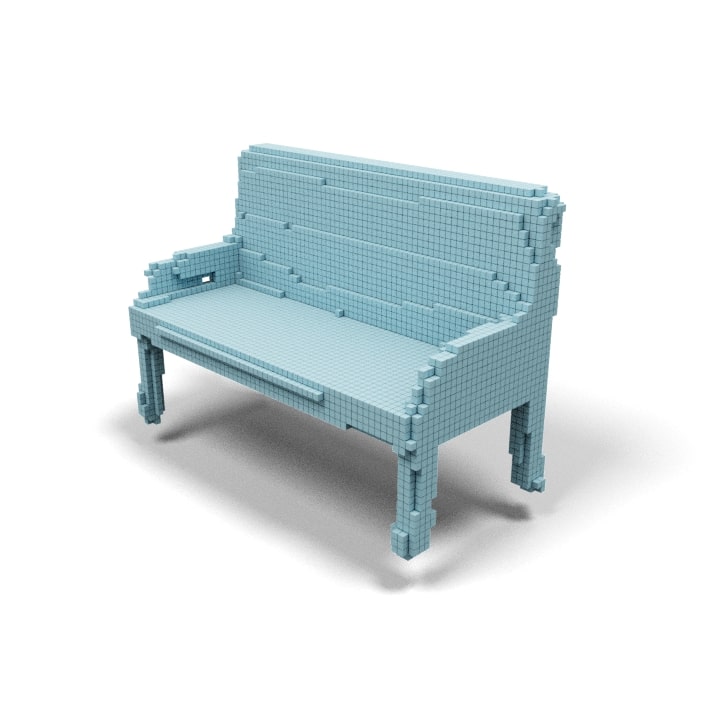} &
\includegraphics[width=0.15\linewidth]{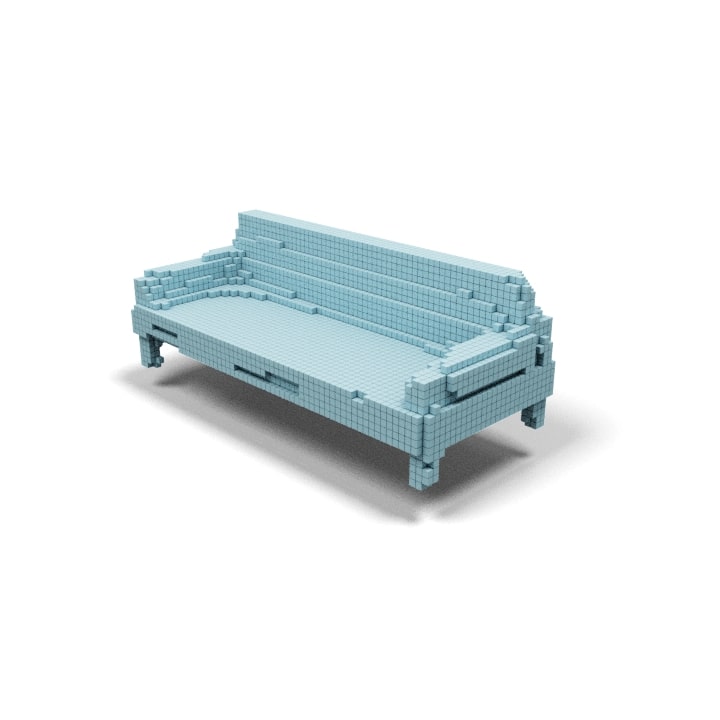} &
\includegraphics[width=0.15\linewidth]{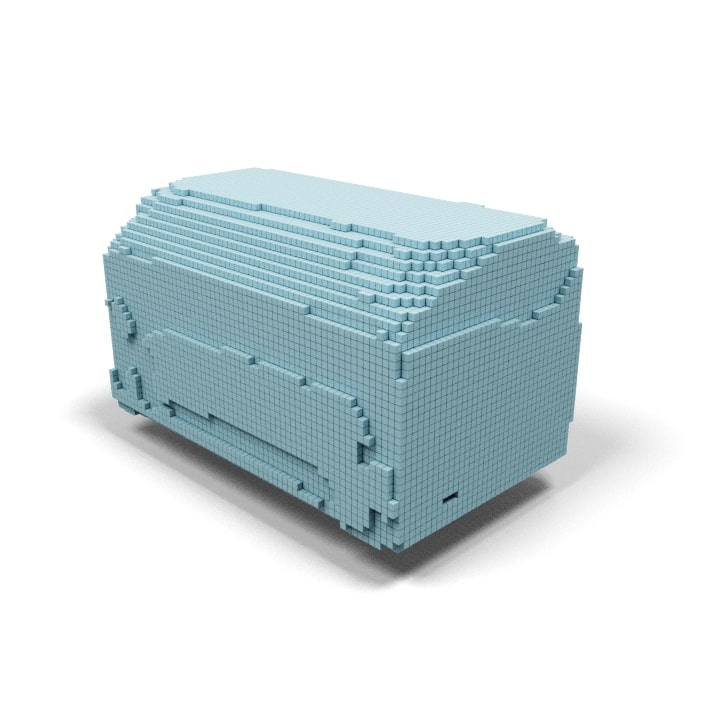} &
\includegraphics[width=0.15\linewidth]{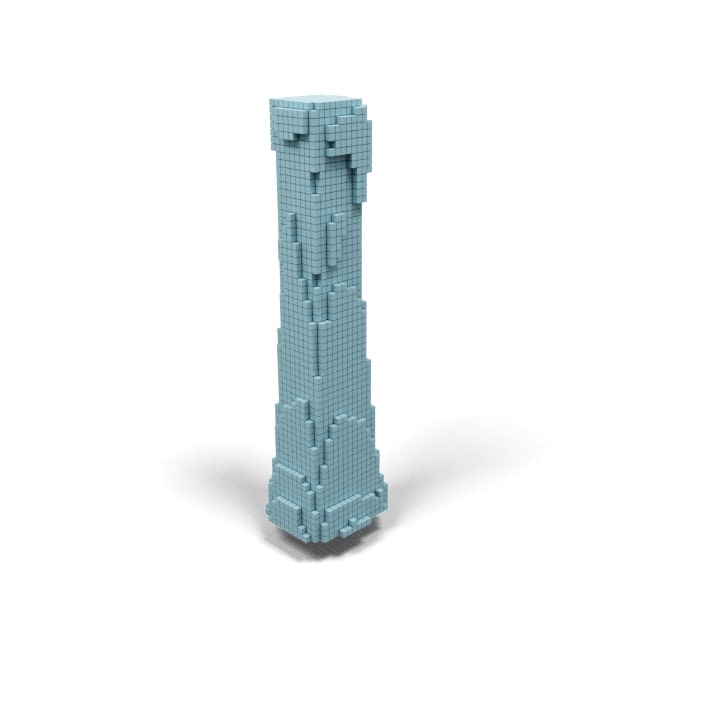} &
\includegraphics[width=0.15\linewidth]{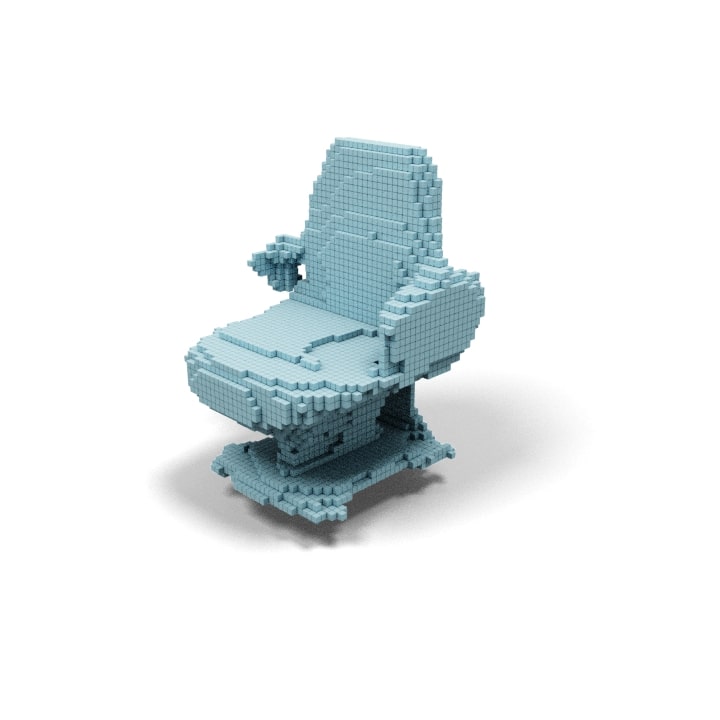} \\
``a shoe'' & ``a piano'' & ``a bed'' & ``an xbox'' & ``a tower'' & ``a human''\\
\end{tabular}
}
\end{center}
  \caption{Results using text queries that are semantically outside the dataset.}
\label{fig:gen_pics}
\end{figure*}

We also conduct experiments to see if the network can generate shapes based on text queries which are out of distribution from its training data. The results are shown in Figure~\ref{fig:gen_pics}. It can be seen from the results that the method tries to generate the desired shape  based on its training dataset. We believe extending our method to generalize on out of distribution samples might be interesting avenue to explore for future work.

\section{Visual Results for Different Prefixes}
\begin{figure*}[t!]
\begin{center}
\setlength{\tabcolsep}{2pt}
\small{
\begin{tabular}{ccccc}
\includegraphics[width=0.15\linewidth]{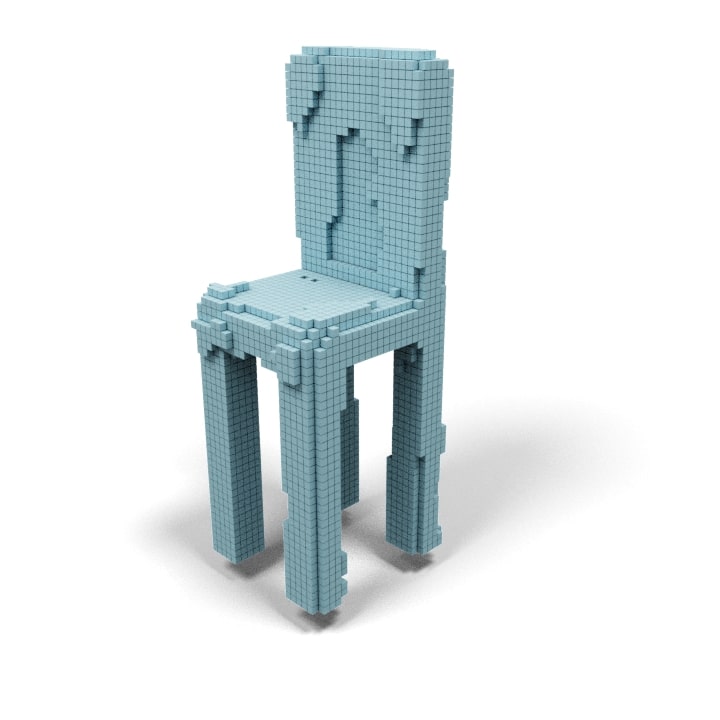} &
\includegraphics[width=0.15\linewidth]{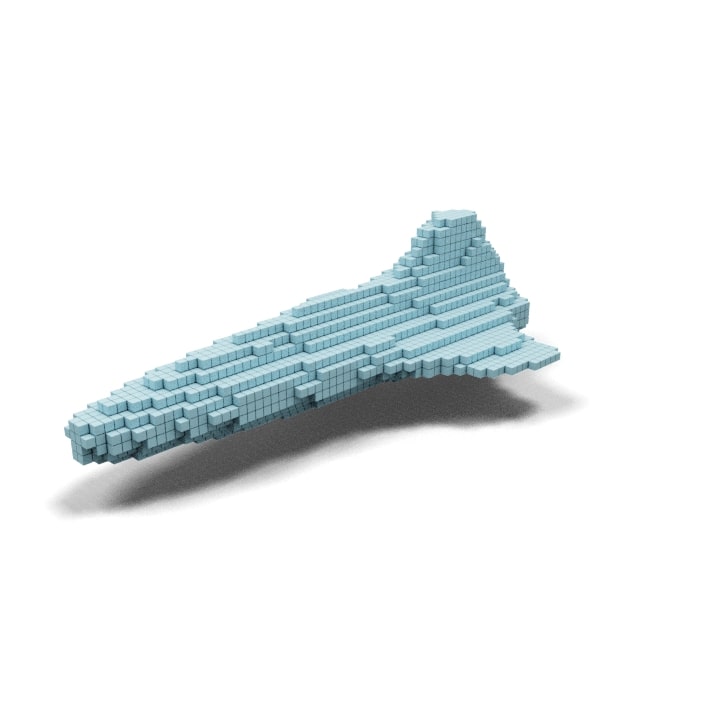} &
\includegraphics[width=0.15\linewidth]{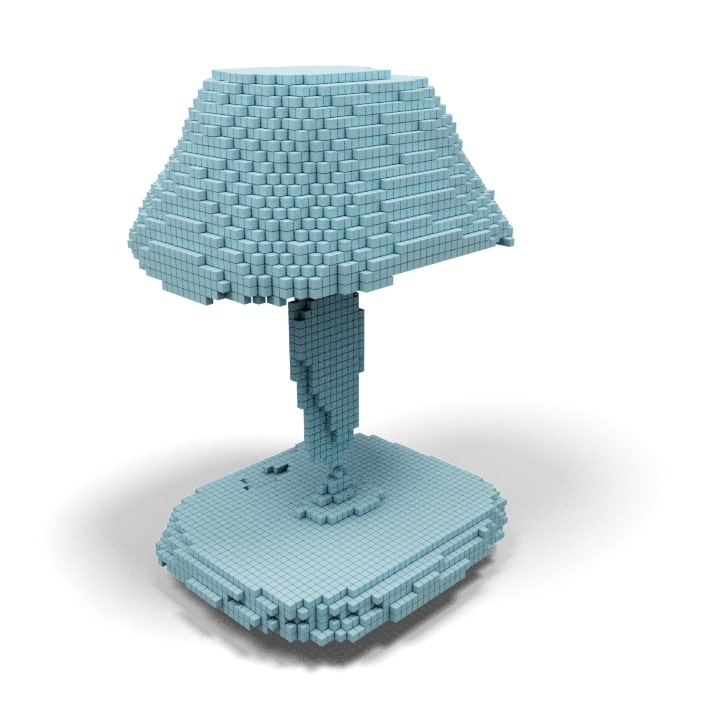} &
\includegraphics[width=0.15\linewidth]{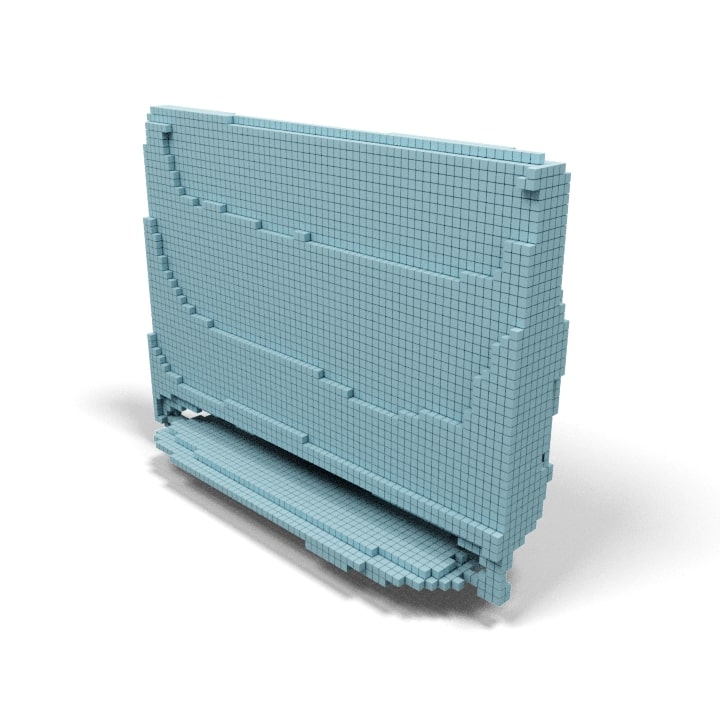} &
\includegraphics[width=0.15\linewidth]{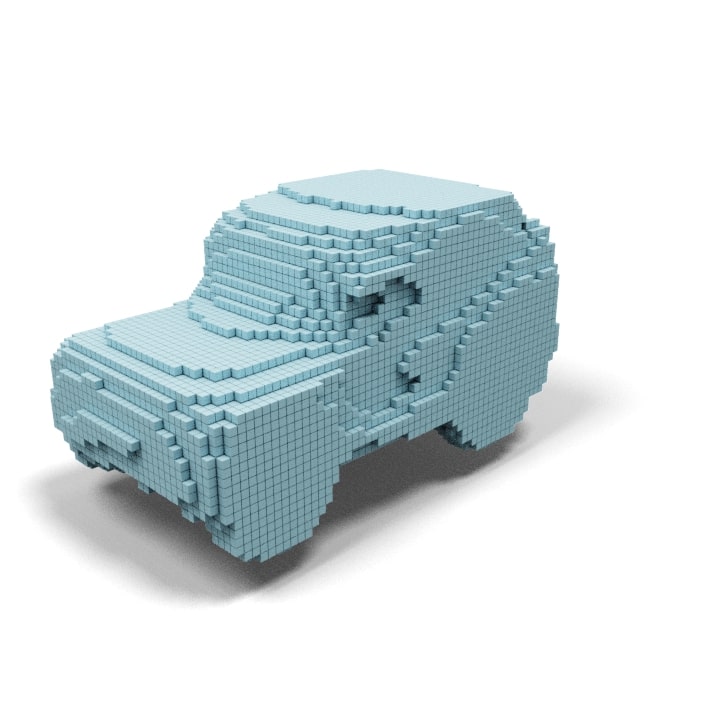} \\
``bar stool'' & ``space shuttle'' & ``table lamp'' & ``television'' &  ``beach wagon''\\
\includegraphics[width=0.15\linewidth]{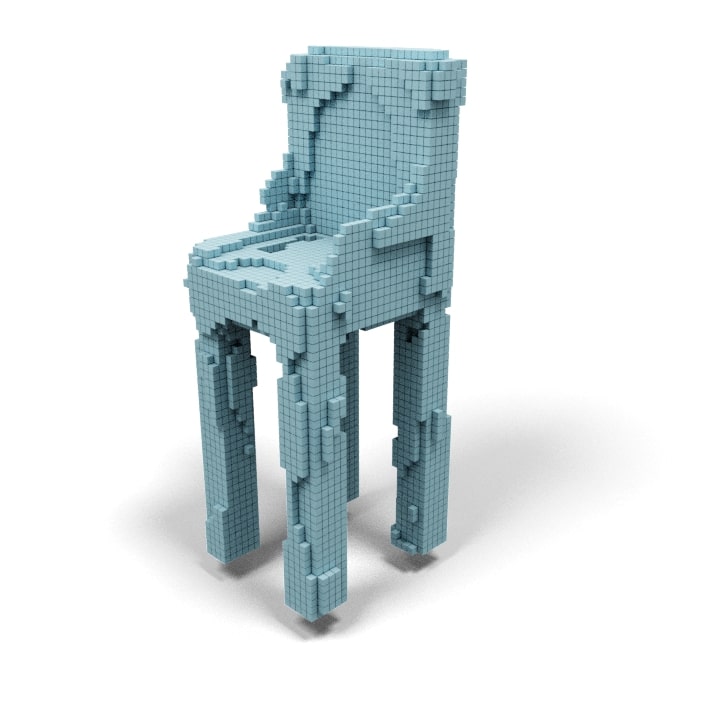} &
\includegraphics[width=0.15\linewidth]{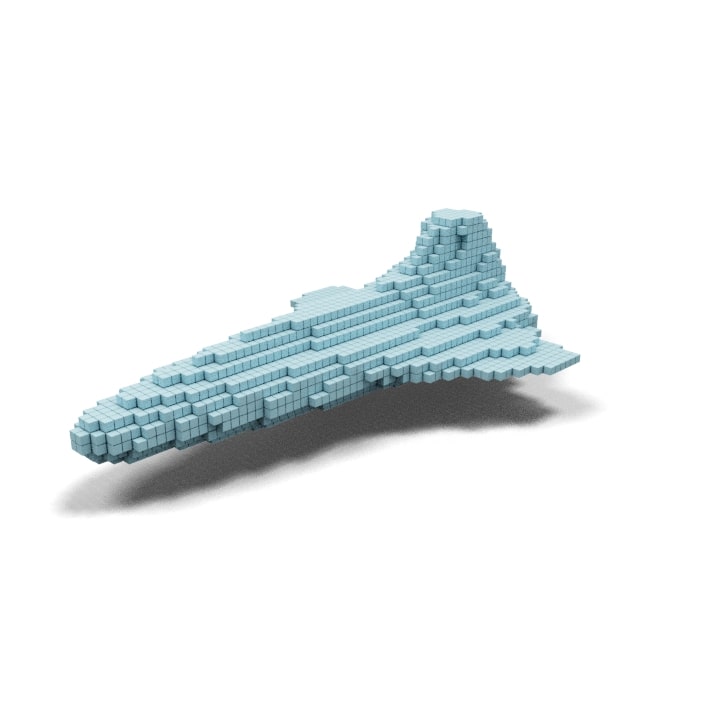} &
\includegraphics[width=0.15\linewidth]{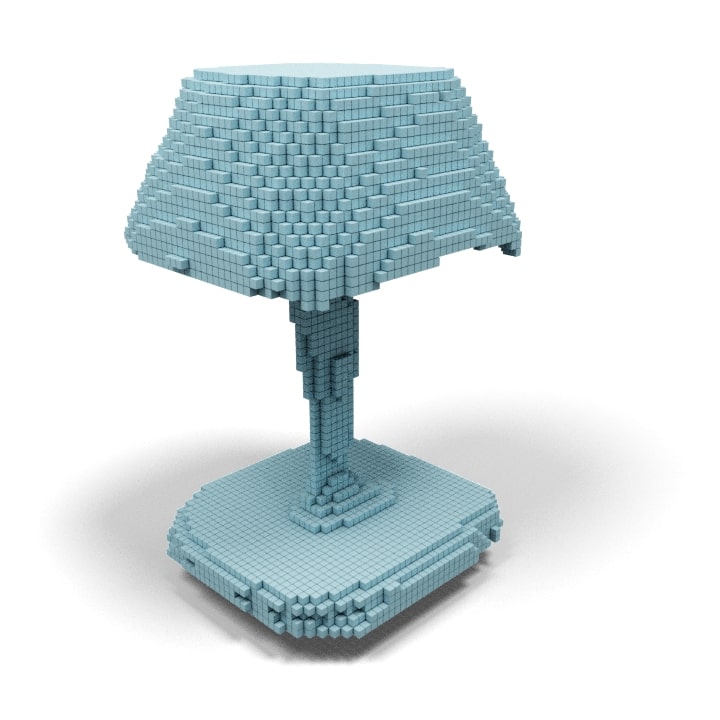} &
\includegraphics[width=0.15\linewidth]{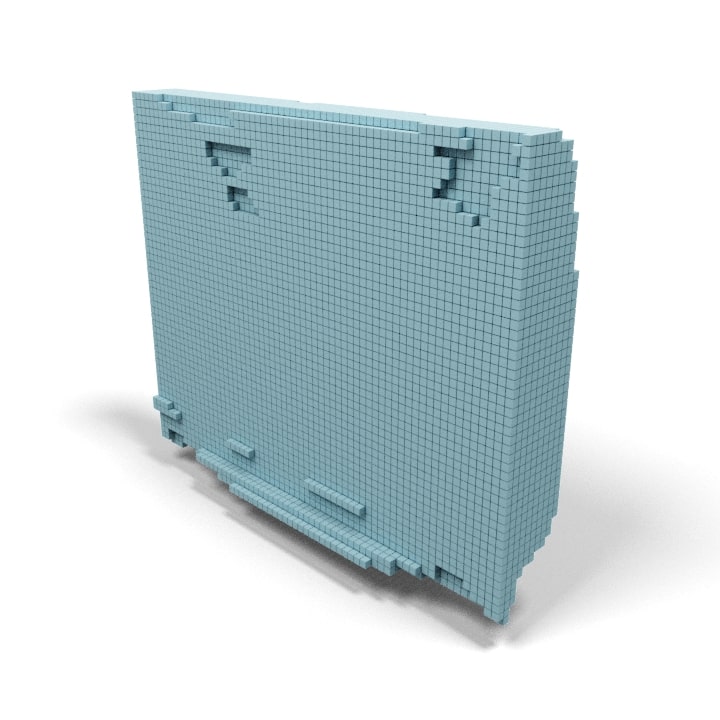} &
\includegraphics[width=0.15\linewidth]{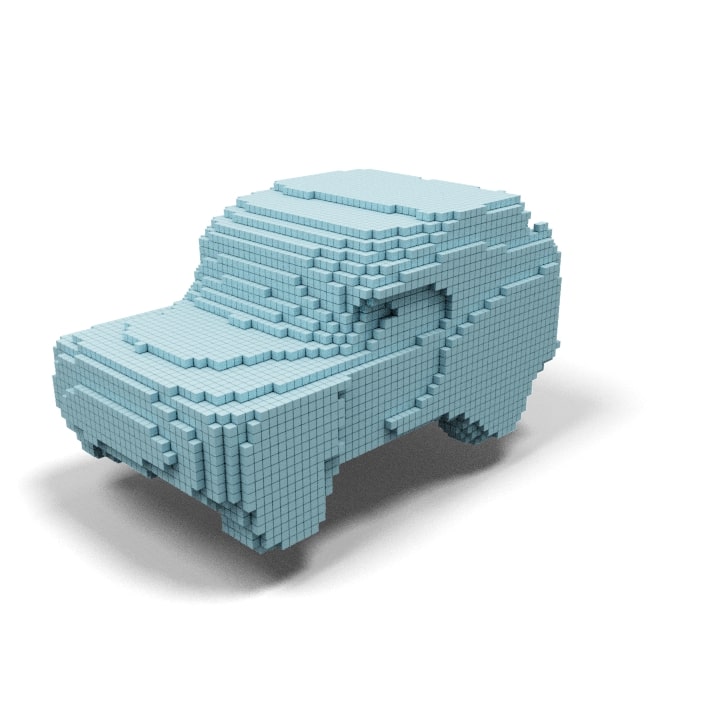} \\
``a picture of a bar stool'' & ``a picture of a space shuttle'' & ``a picture of a table lamp'' & ``a picture of a television'' &  ``a picture of a beach wagon''\\
\includegraphics[width=0.15\linewidth]{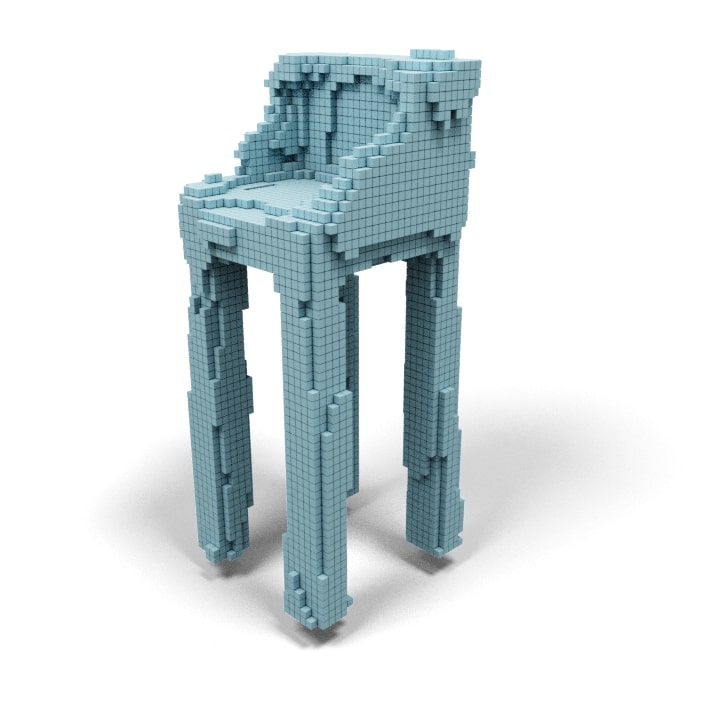} &
\includegraphics[width=0.15\linewidth]{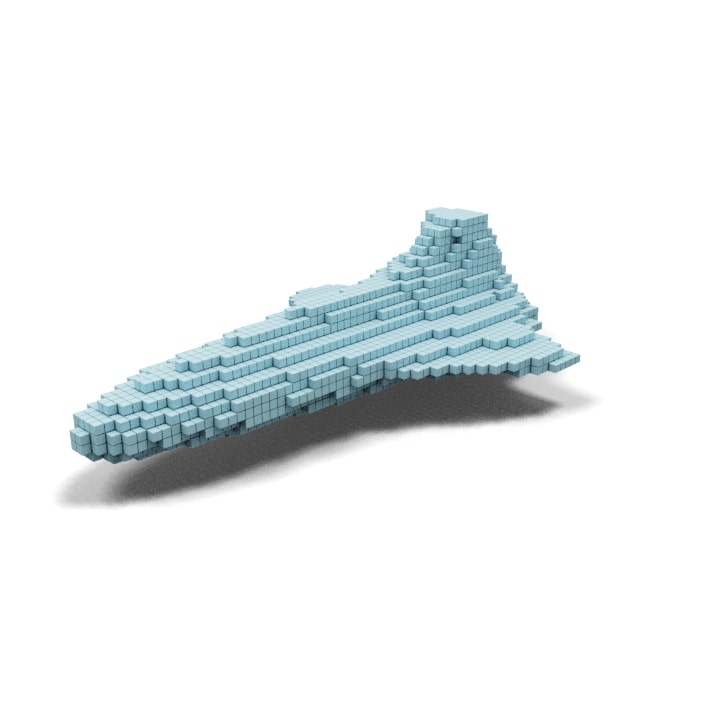} &
\includegraphics[width=0.15\linewidth]{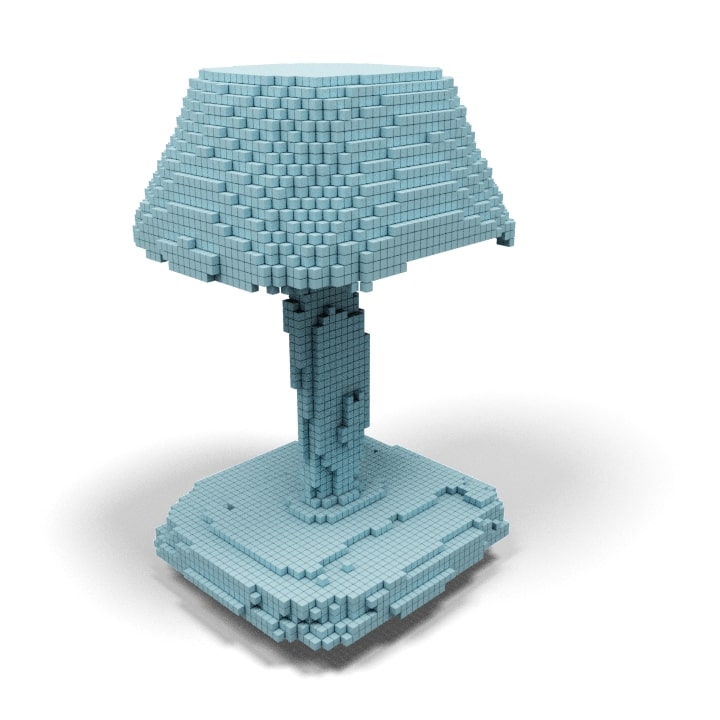} &
\includegraphics[width=0.15\linewidth]{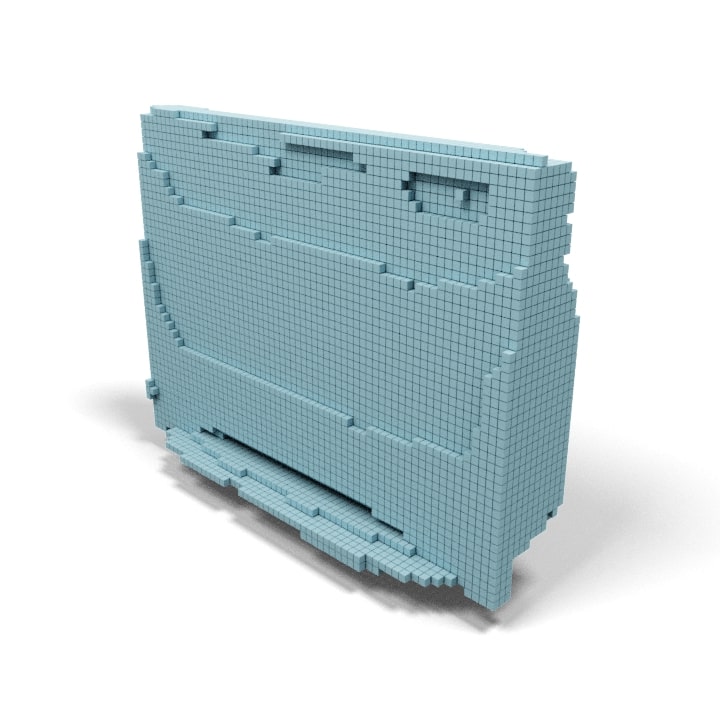} &
\includegraphics[width=0.15\linewidth]{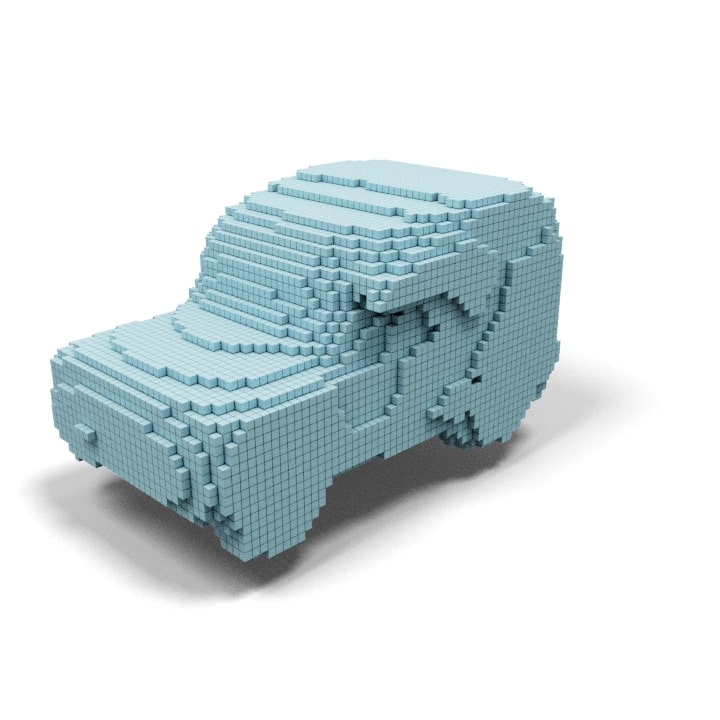} \\
``a photo of bar stool'' & ``a photo of space shuttle'' & ``a photo of table lamp'' & ``a photo of television'' &  ``a photo of beach wagon''\\
\includegraphics[width=0.15\linewidth]{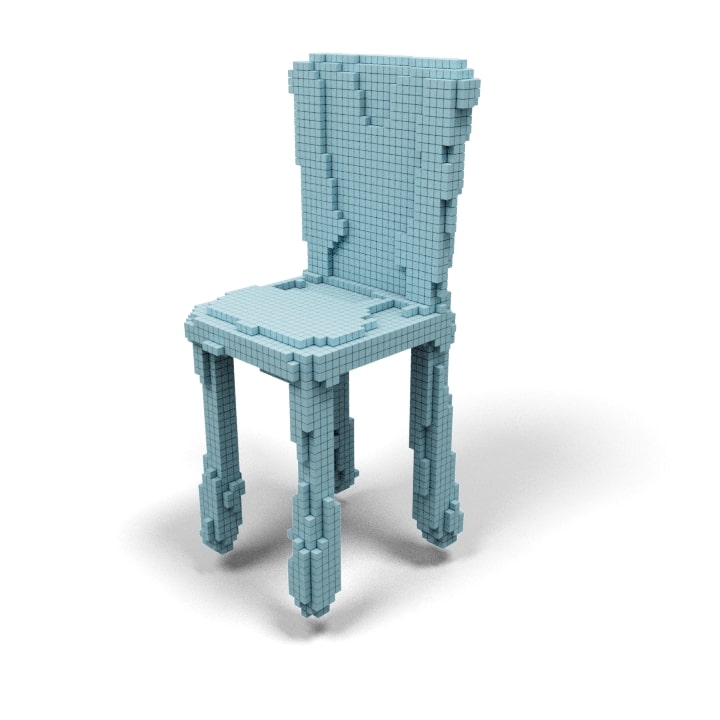} &
\includegraphics[width=0.15\linewidth]{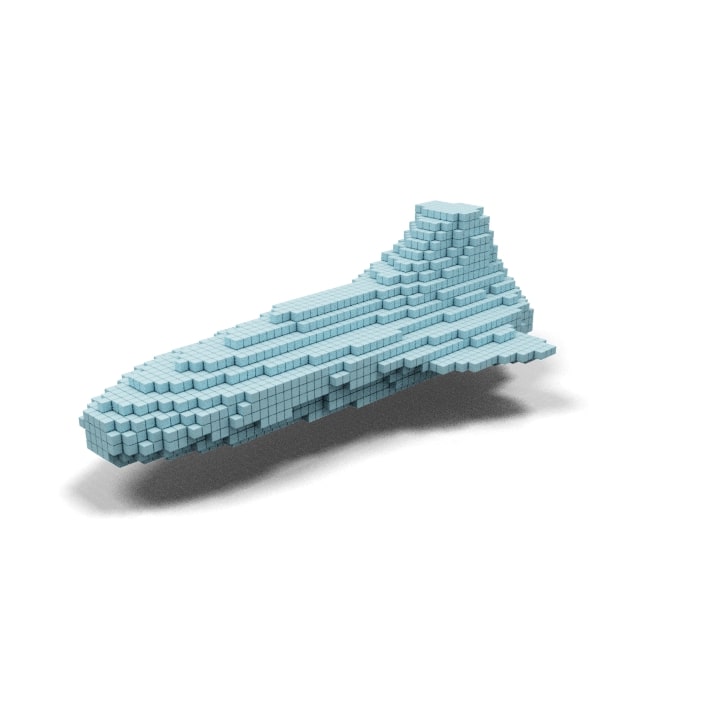} &
\includegraphics[width=0.15\linewidth]{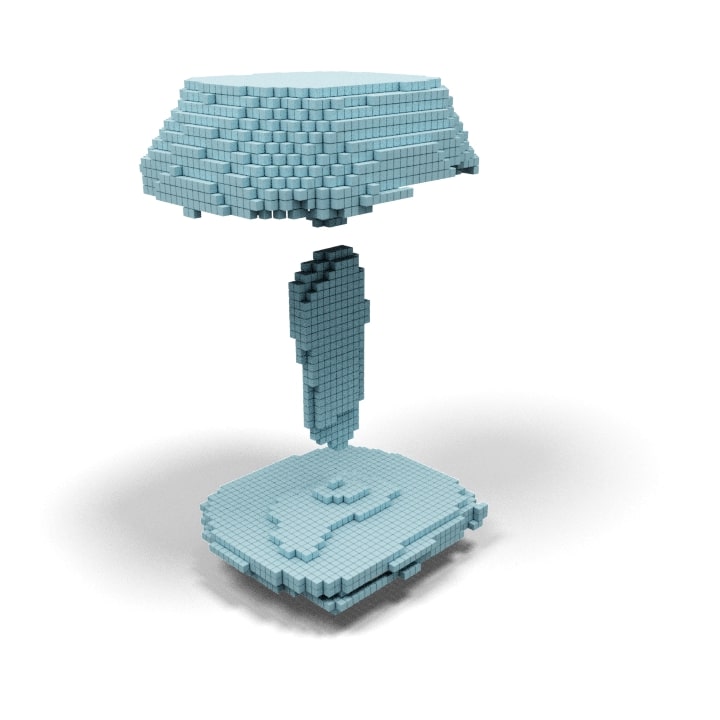} &
\includegraphics[width=0.15\linewidth]{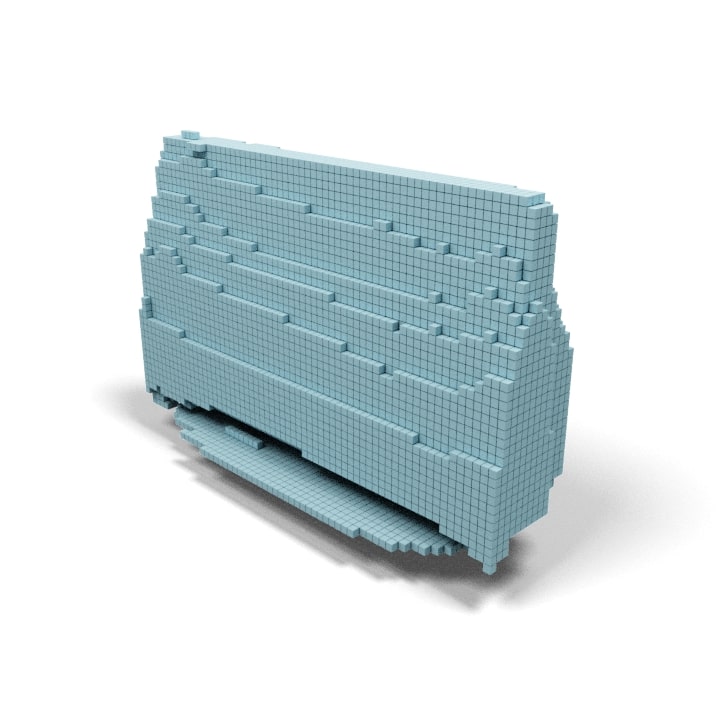} &
\includegraphics[width=0.15\linewidth]{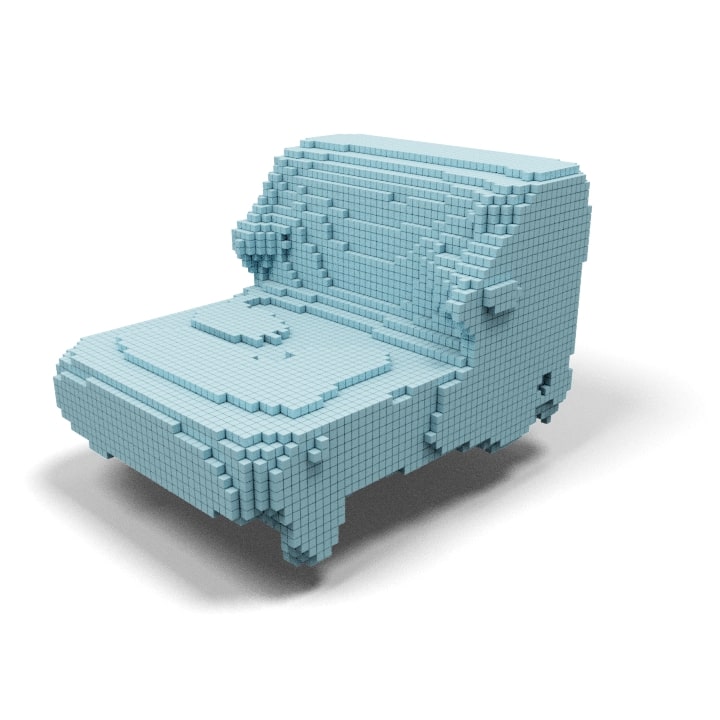} \\
``a rendering of bar stool'' & ``a rendering of space shuttle'' & ``a rendering of table lamp'' & `` a rendering of television'' &  `` a rendering of beach wagon''\\
\end{tabular}
}
\end{center}
  \caption{Results of varying the prefix for a given text query.}
\label{fig:other_prefix}
\end{figure*}

In Figure~\ref{fig:other_prefix}, we show results for different prefixes.  They indicate that for different prefixes there are small variations in generated shape. Moreover, in some prefixes such as ``a rendering of'', the visual results are worse. It would be interesting to investigate other prompts or do prompt tuning as future work. 

\section{Visual Results for more Descriptive Texts}

\begin{figure*}
\begin{center}
\setlength{\tabcolsep}{8pt}
\small{
\begin{tabular}{cccc}
\includegraphics[width=0.15\linewidth]{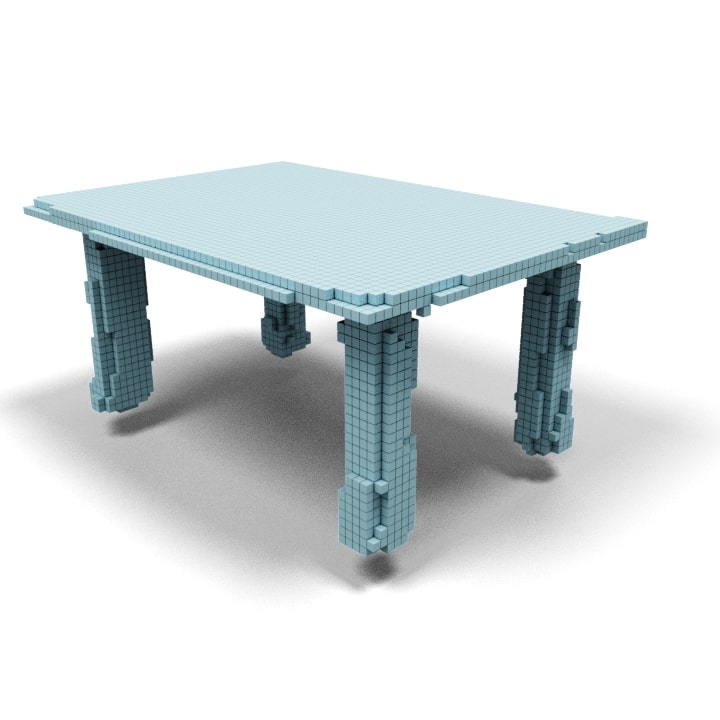} &
\includegraphics[width=0.15\linewidth]{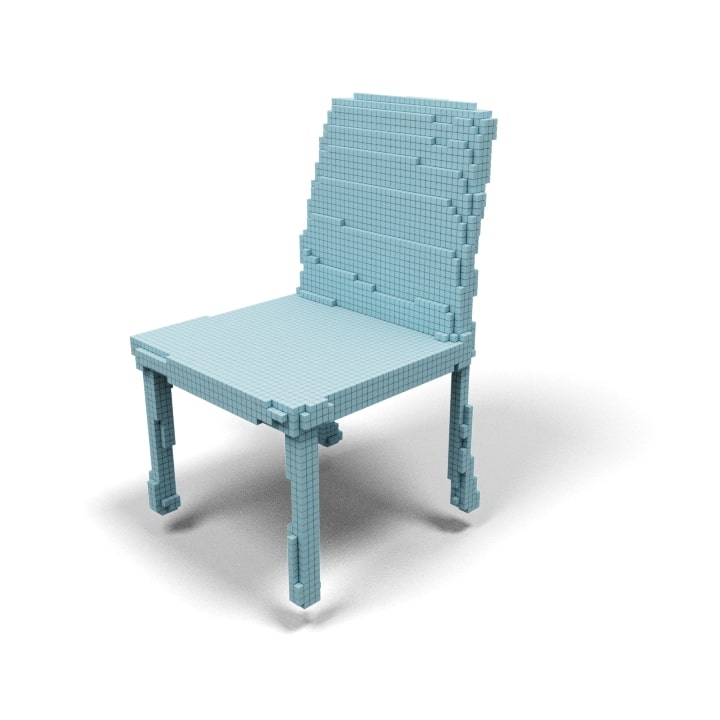} &
\includegraphics[width=0.15\linewidth]{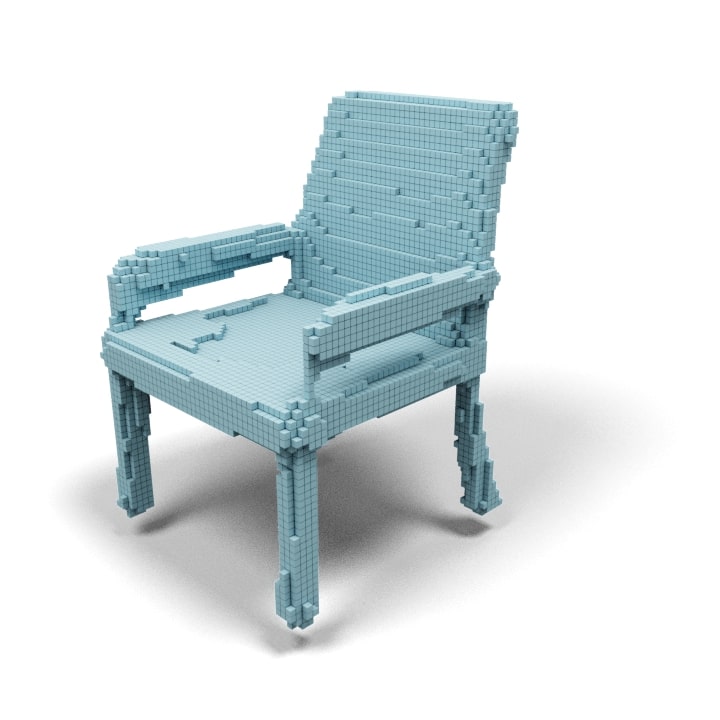} &
\includegraphics[width=0.15\linewidth]{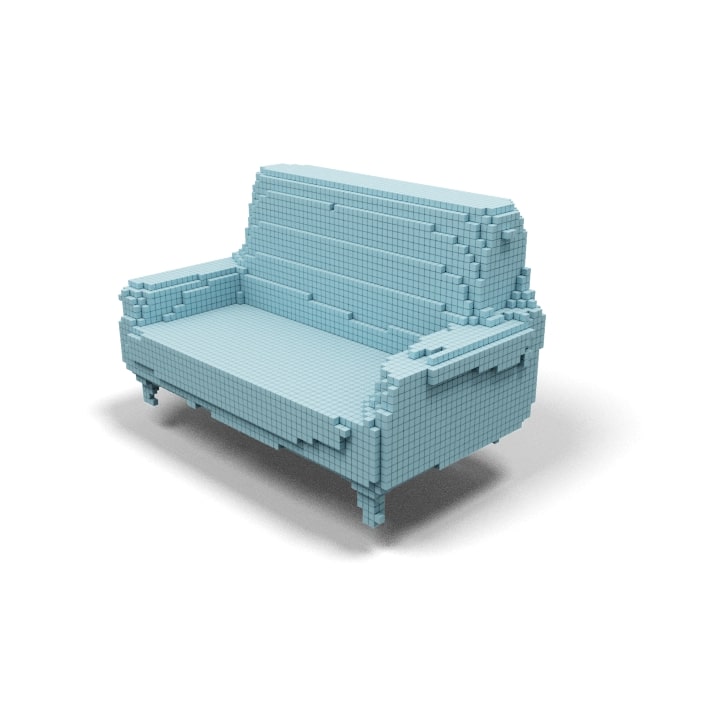}\\
\end{tabular}
}
\end{center}
\caption{Descriptive CLIP-Forge results: ``a brown table with four legs'', ``an armless chair with curved rectangular back'', ``an armed chair with curved rectangular back'', ``big sofa having two legs of black color, backrest, armrest, and sitting of black color''.}
\label{fig:long_sentence}
\end{figure*}

We show additional results using text queries that are longer and more descriptive in Figure~\ref{fig:long_sentence}. It can be seen that CLIP-Forge is able to capture certain shape-related attributes. Non-shape related descriptions such as color is not captured but could potentially bias the generation. We believe that combining our method with semi-supervised learning can enable more fine control of shape generation using text.

\section{Additional Qualitative Results}
\begin{figure*}[t!]
\begin{center}
\setlength{\tabcolsep}{2pt}
\small{
\begin{tabular}{cccccc}

\includegraphics[width=0.15\linewidth]{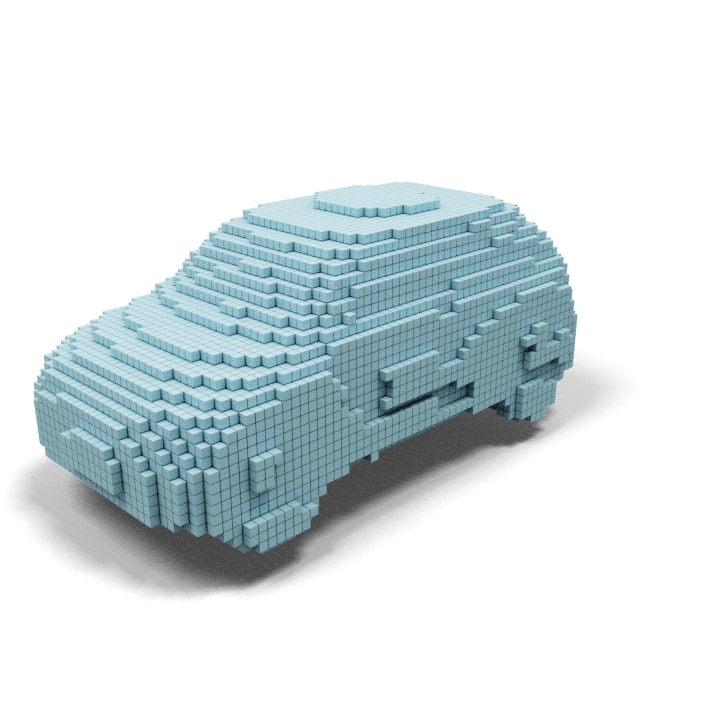} &
\includegraphics[width=0.15\linewidth]{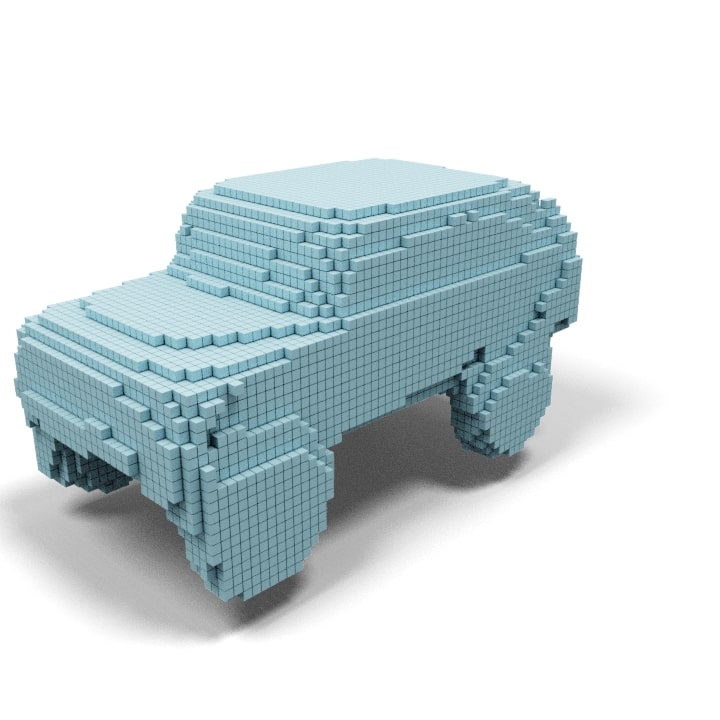} &
\includegraphics[width=0.15\linewidth]{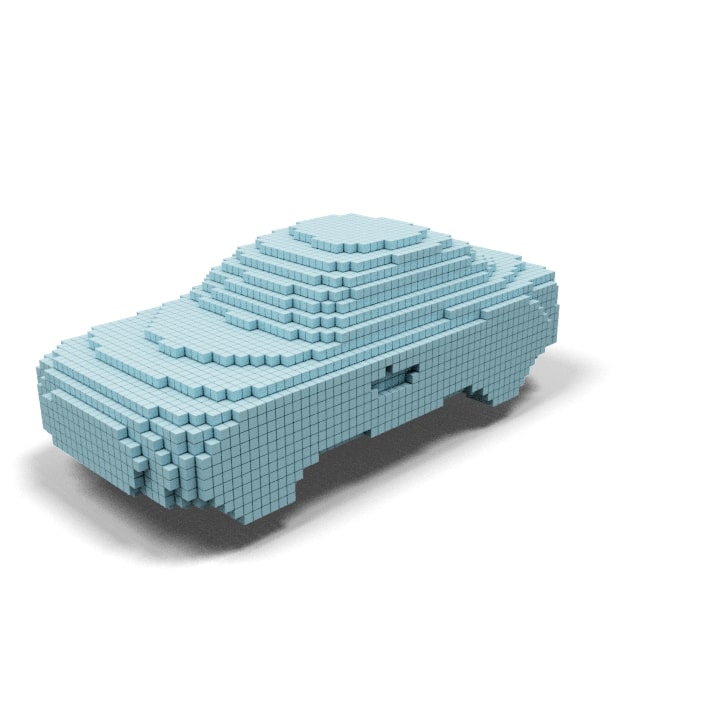} &
\includegraphics[width=0.15\linewidth]{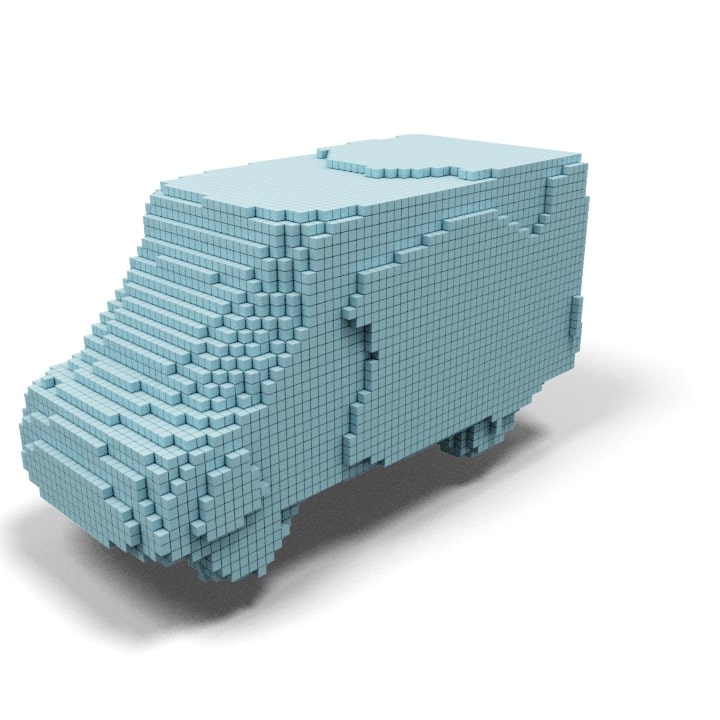} &
\includegraphics[width=0.15\linewidth]{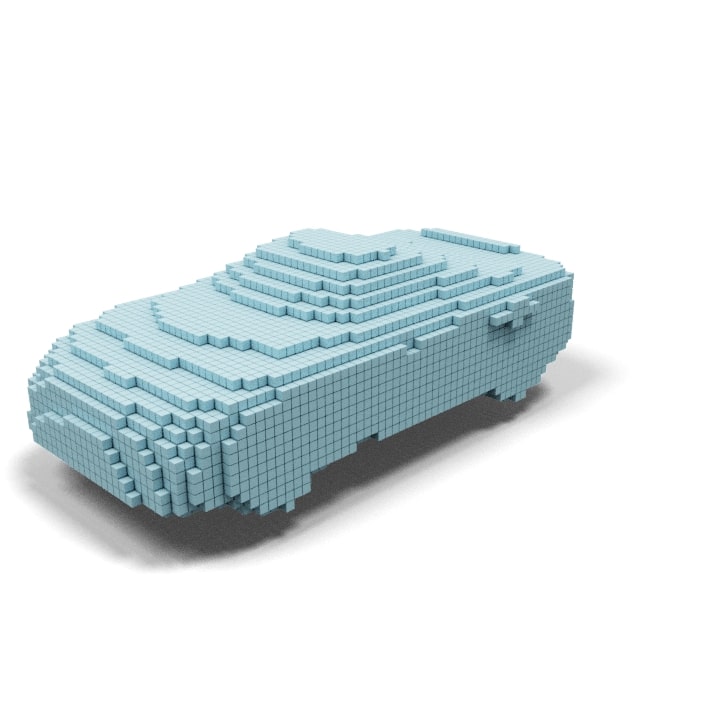} &
\includegraphics[width=0.15\linewidth]{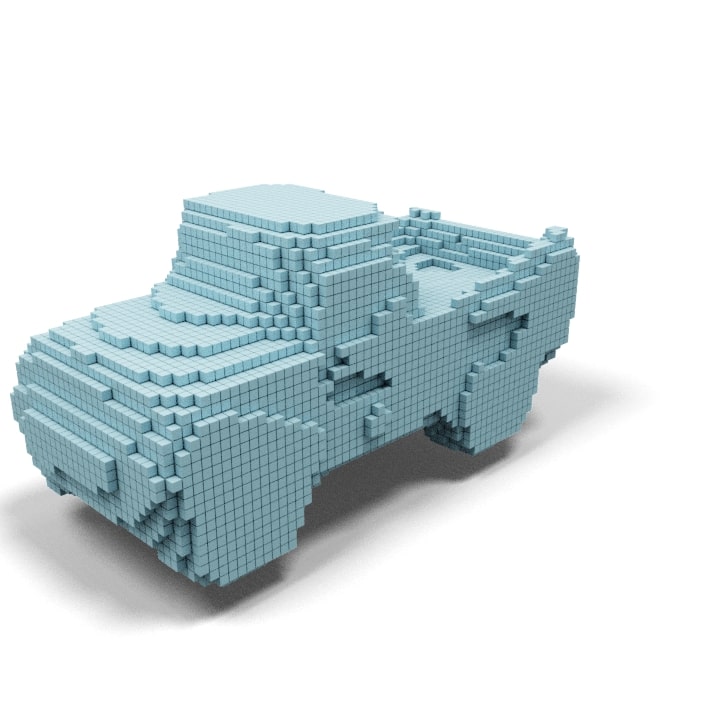}\\
``a hatchback'' & ``a suv'' & ``a sedan'' & ``a  van'' & ``a sports car'' & ``a truck''\\
\includegraphics[width=0.15\linewidth]{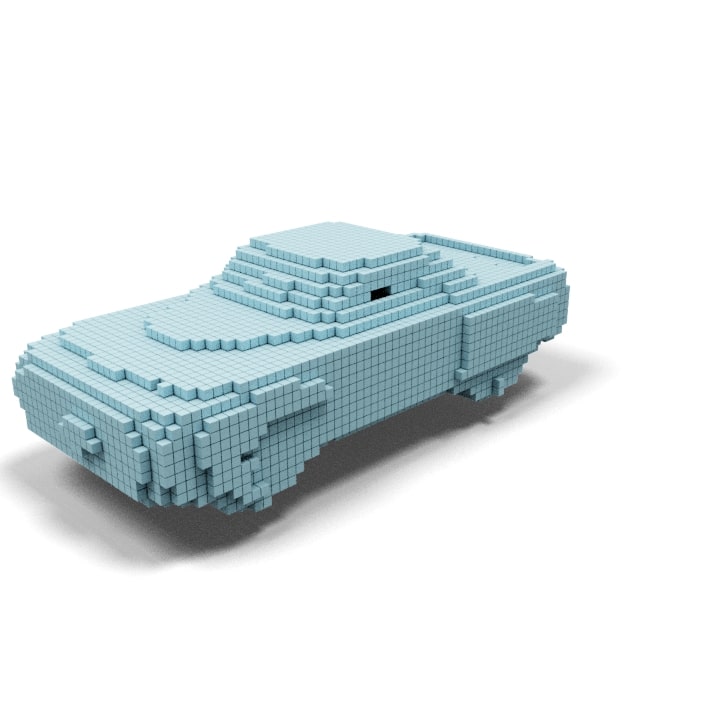} &
\includegraphics[width=0.15\linewidth]{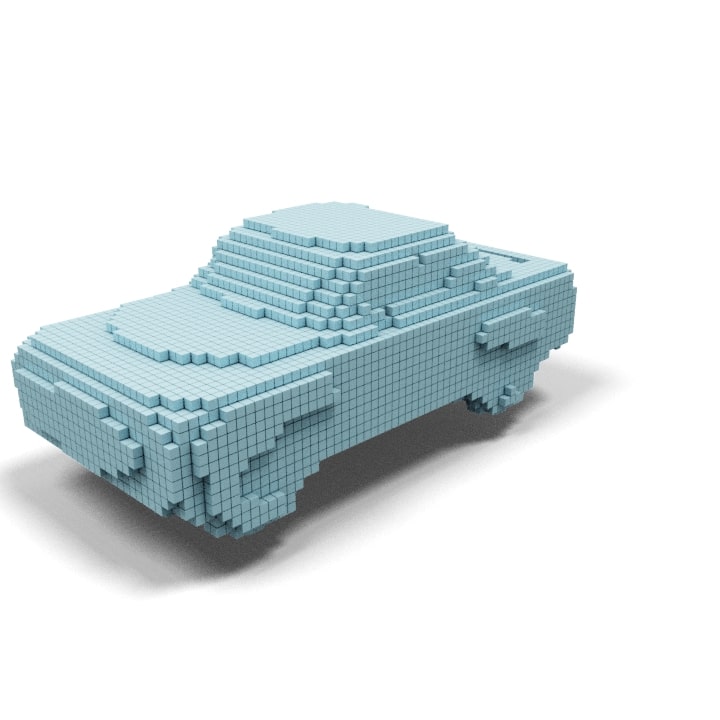} &
\includegraphics[width=0.15\linewidth]{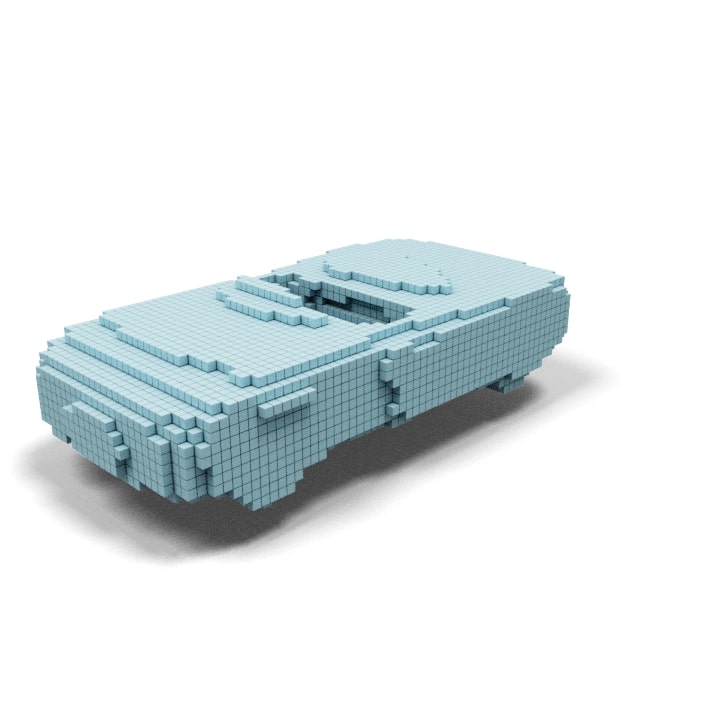} &
\includegraphics[width=0.15\linewidth]{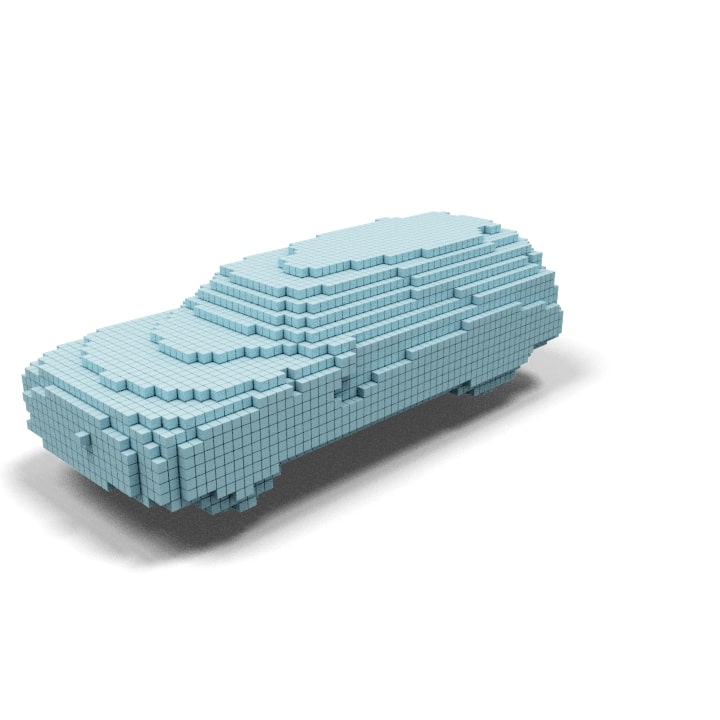} &
\includegraphics[width=0.15\linewidth]{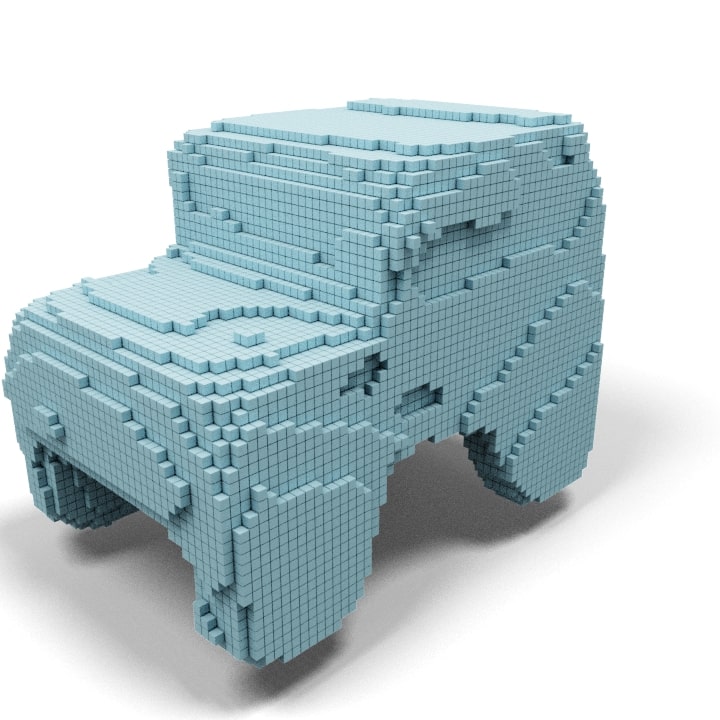} &
\includegraphics[width=0.15\linewidth]{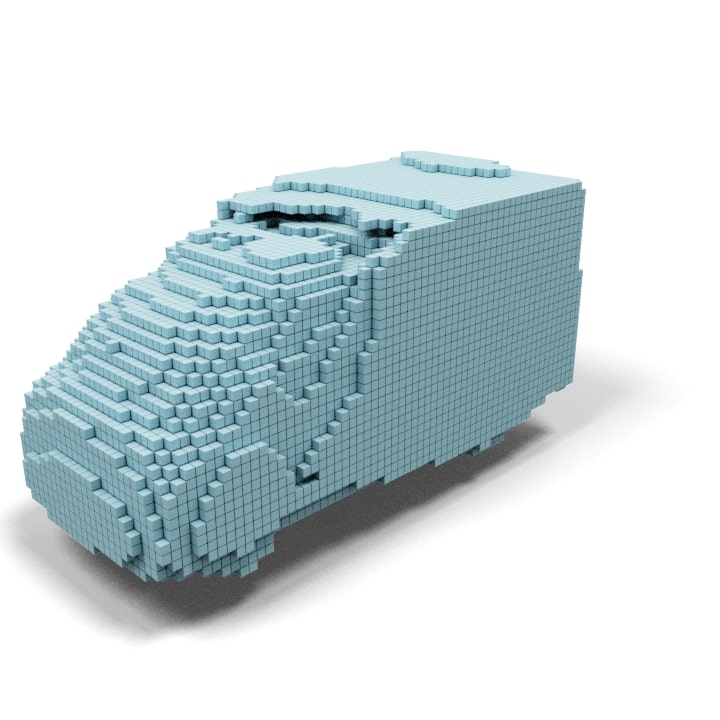}\\
``a muscle car'' & ``a retro car'' & ``a roadster car'' & ``a limo'' & ``a jeep'' & ``a bus''\\
\includegraphics[width=0.15\linewidth]{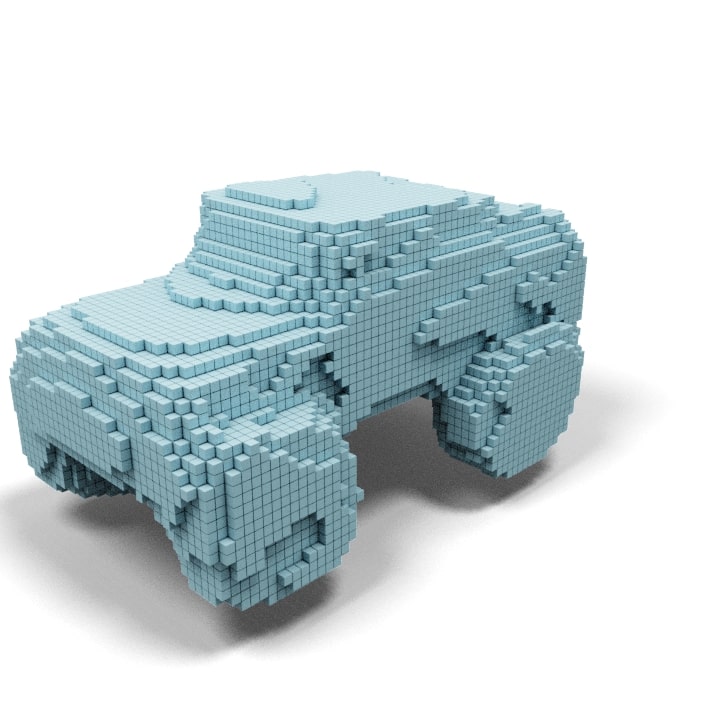} &
\includegraphics[width=0.15\linewidth]{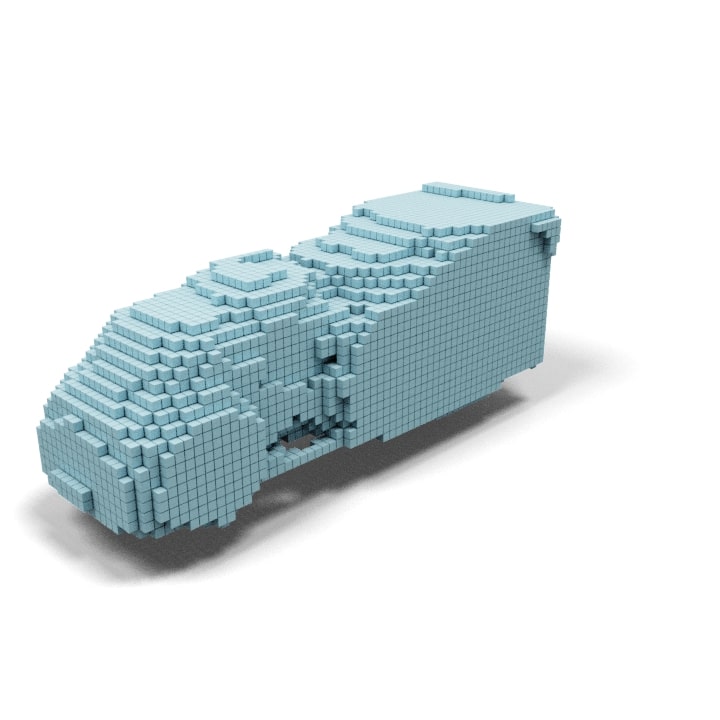} &
\includegraphics[width=0.15\linewidth]{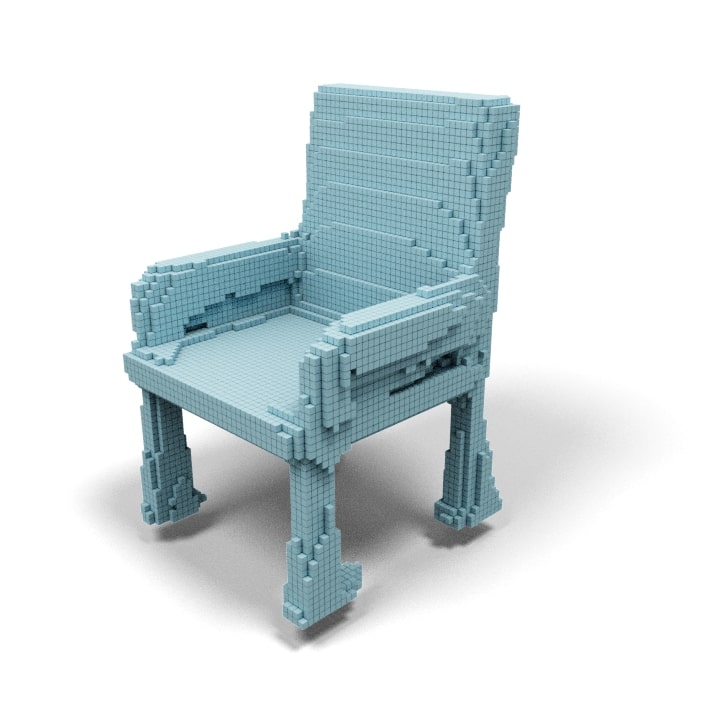} &
\includegraphics[width=0.15\linewidth]{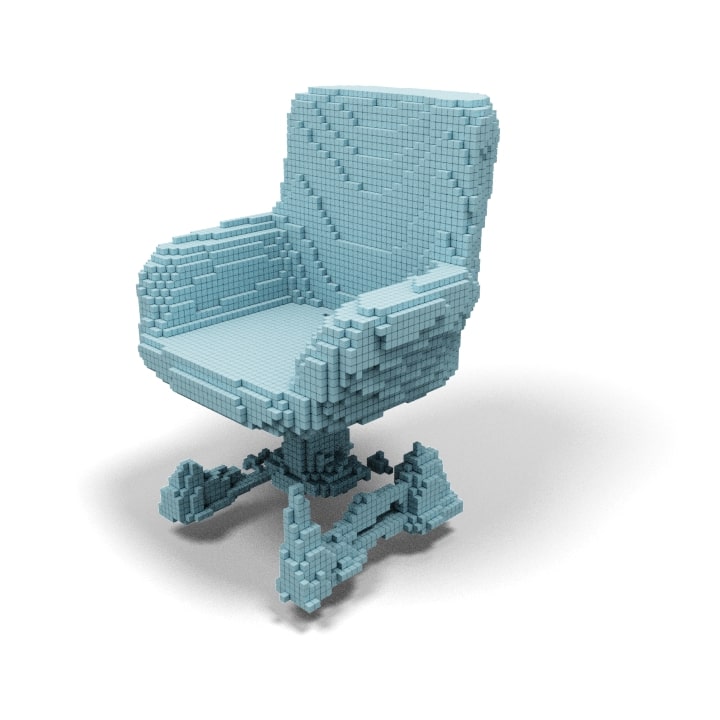} &
\includegraphics[width=0.15\linewidth]{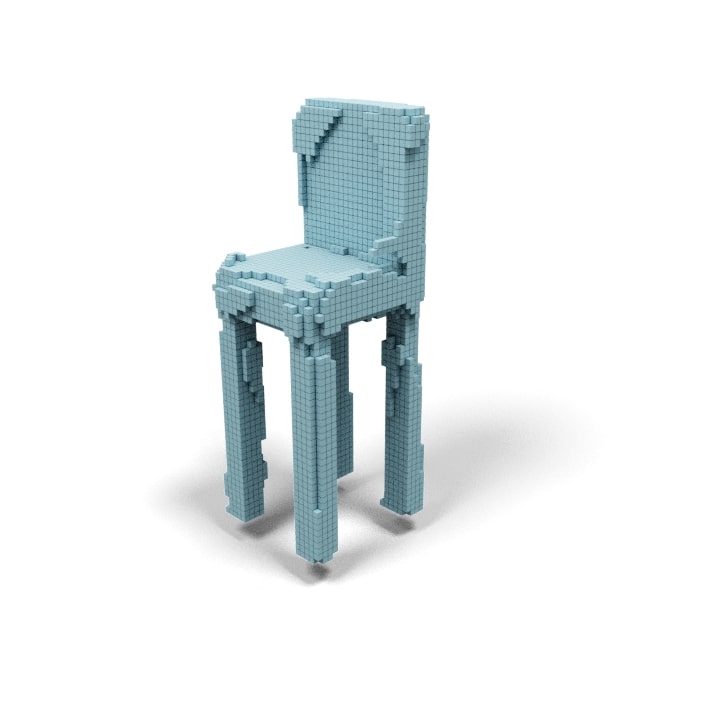} &
\includegraphics[width=0.15\linewidth]{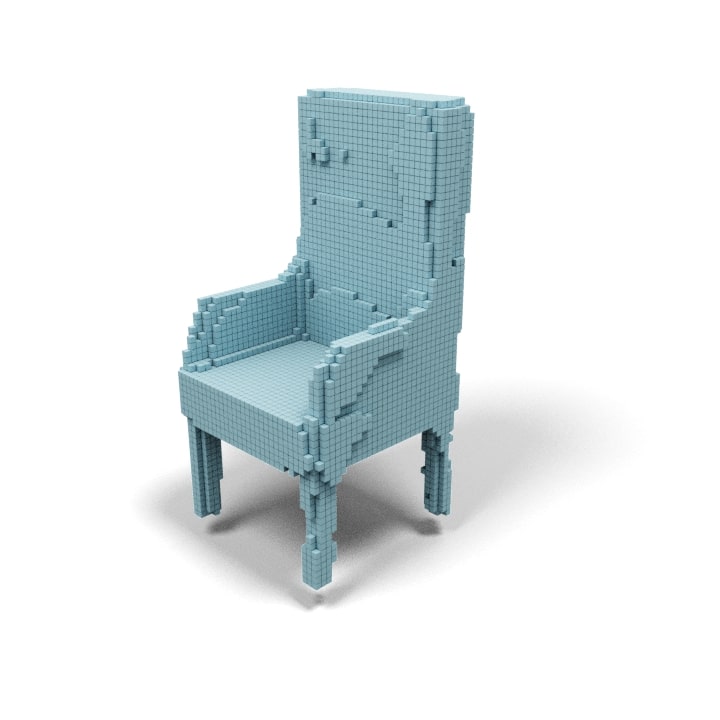}\\
``a monster truck'' & ``a shuttle-bus'' & ``an arm chair'' & ``a swivel chair'' & ``a bar stool'' & ``a wing chair''\\
\includegraphics[width=0.15\linewidth]{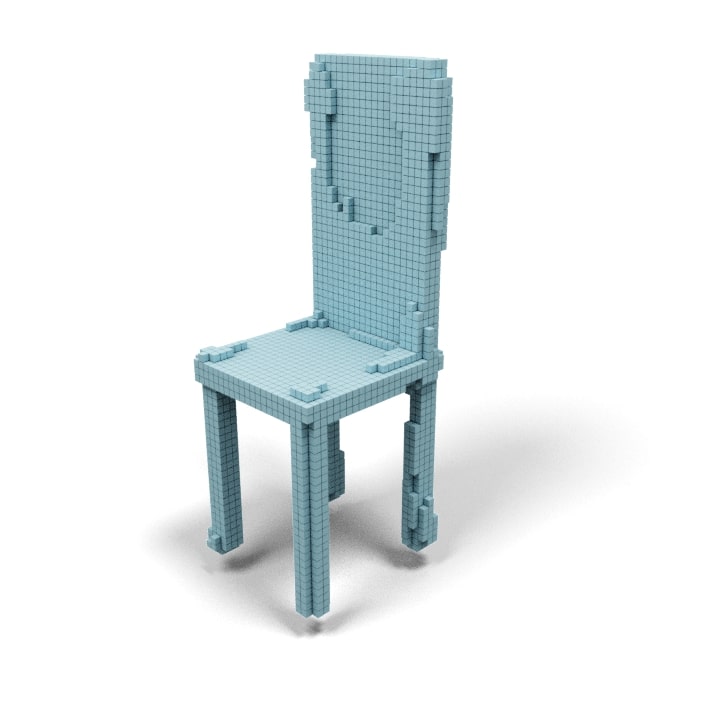} &
\includegraphics[width=0.15\linewidth]{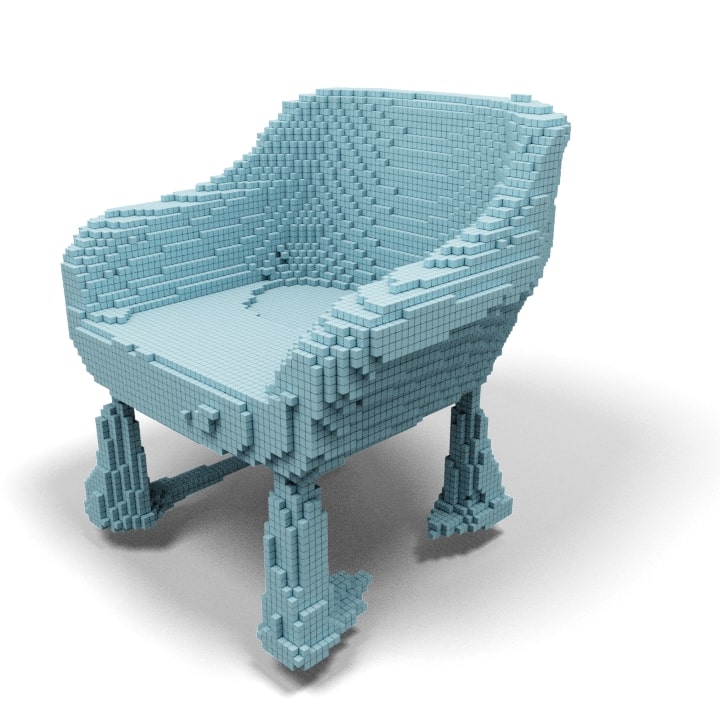} &
\includegraphics[width=0.15\linewidth]{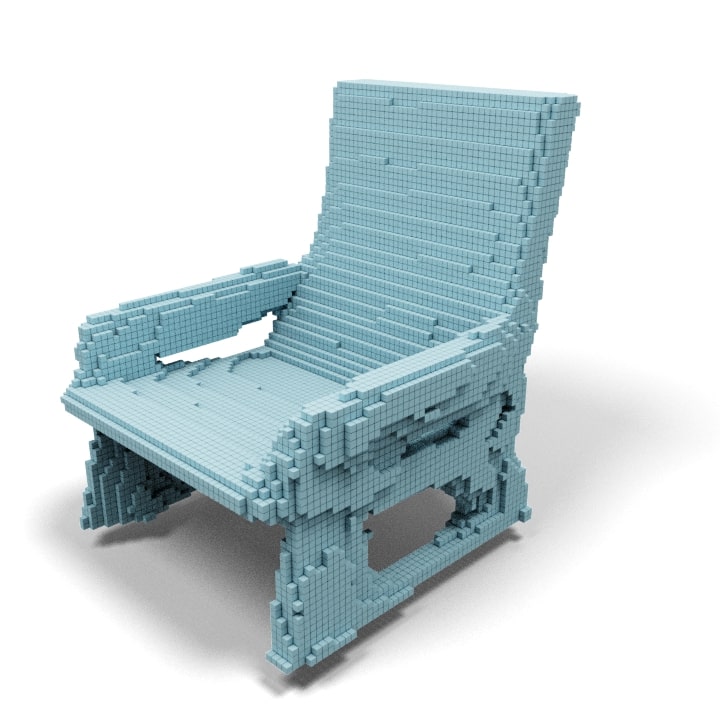} &
\includegraphics[width=0.15\linewidth]{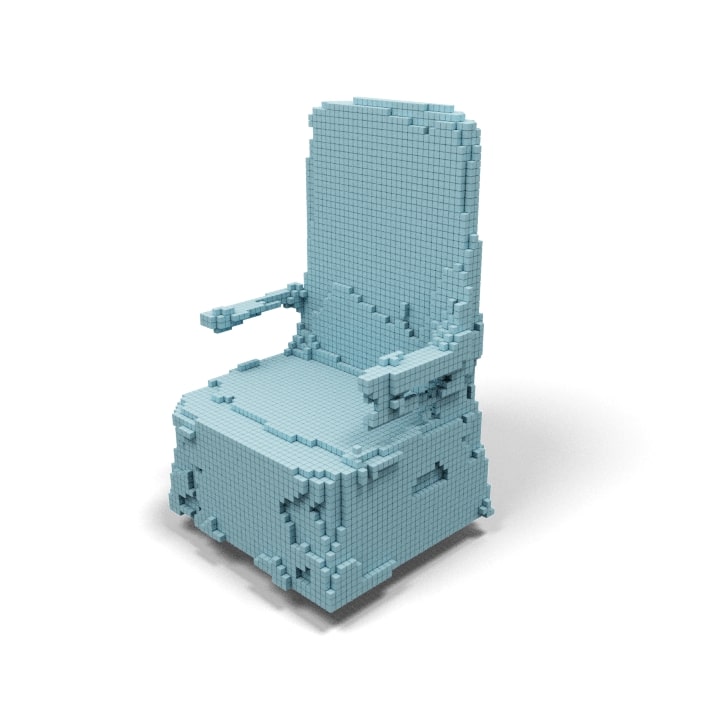} &
\includegraphics[width=0.15\linewidth]{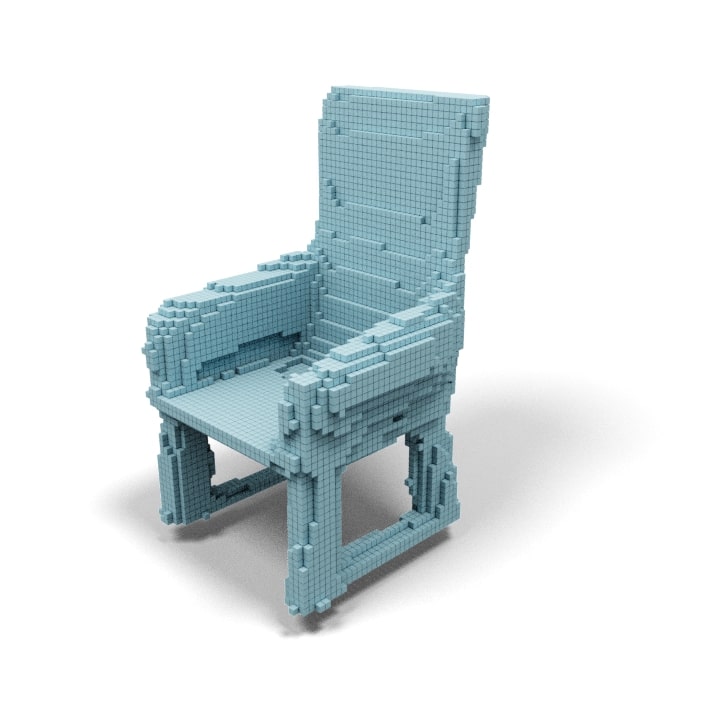} &
\includegraphics[width=0.15\linewidth]{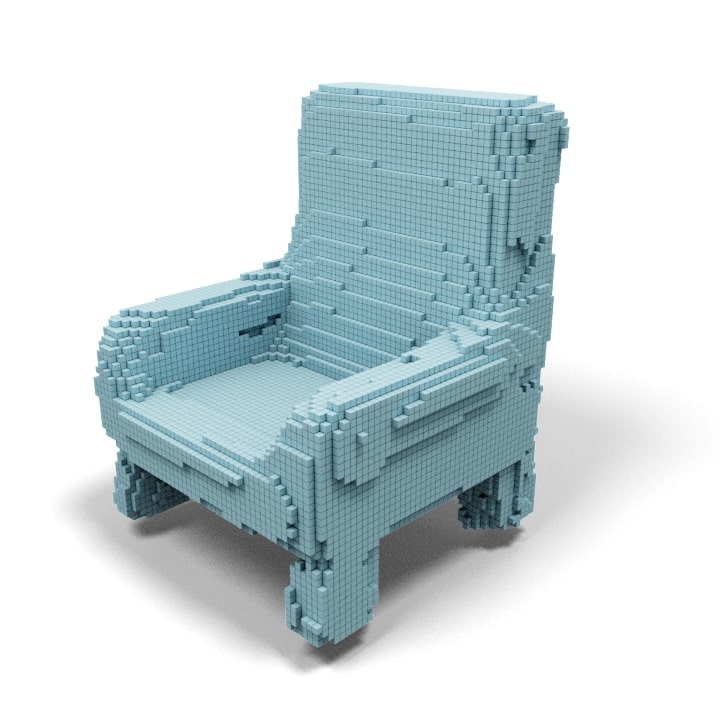}\\
``a  vertical back chair'' & ``a bowl chair'' & ``a lounge chair'' & ``a cathedra' & ``a rocking chair'' & ``a recliner''\\
\includegraphics[width=0.15\linewidth]{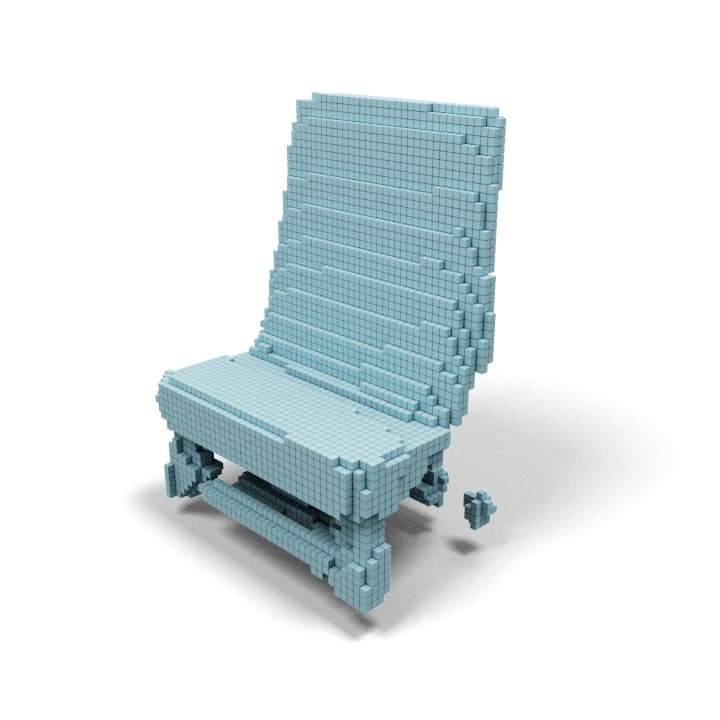} &
\includegraphics[width=0.15\linewidth]{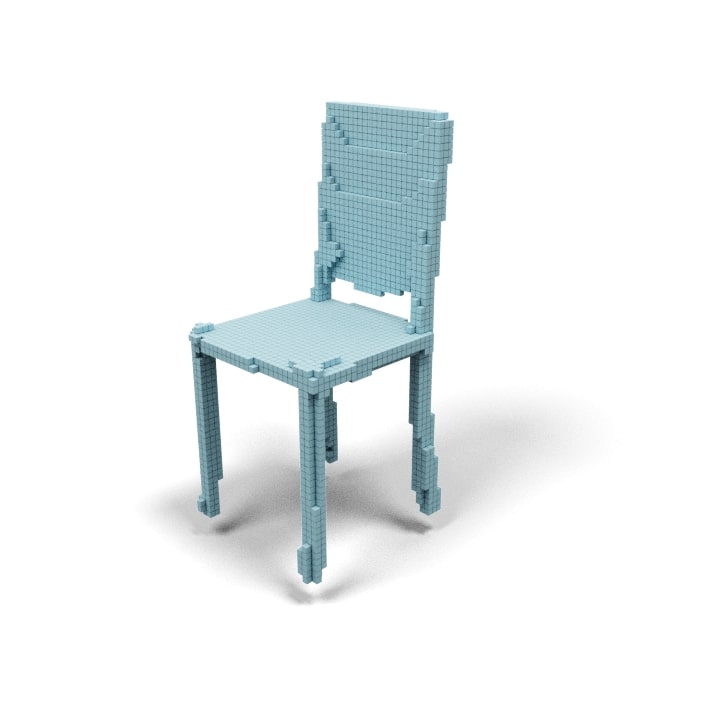} &
\includegraphics[width=0.15\linewidth]{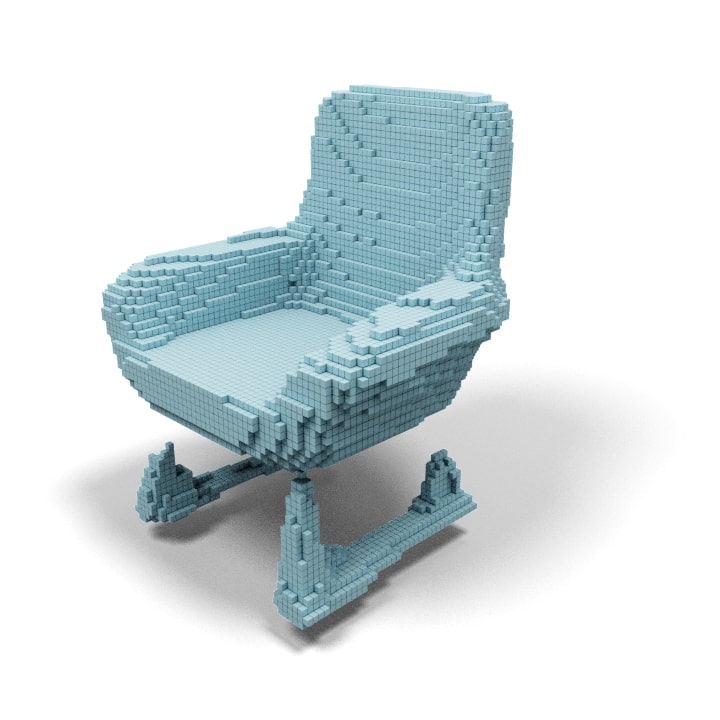} &
\includegraphics[width=0.15\linewidth]{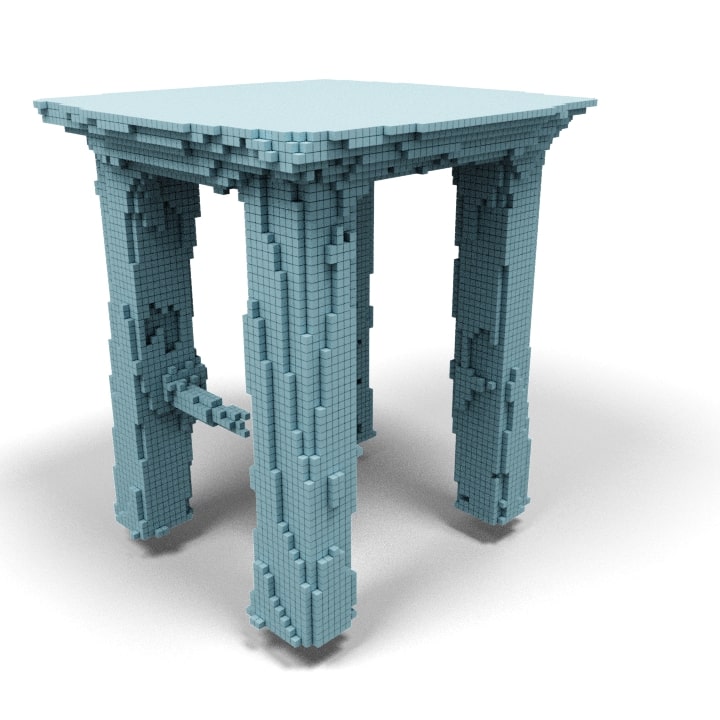} &
\includegraphics[width=0.15\linewidth]{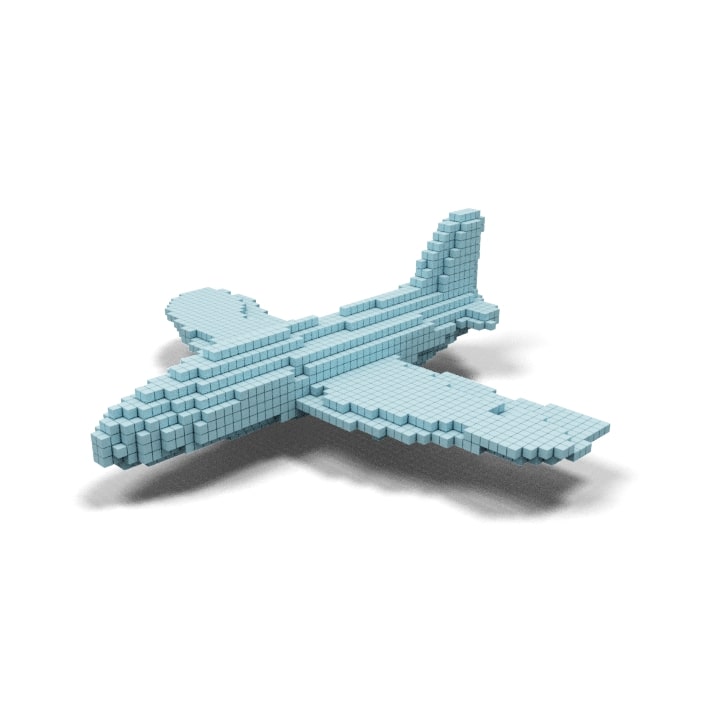} &
\includegraphics[width=0.15\linewidth]{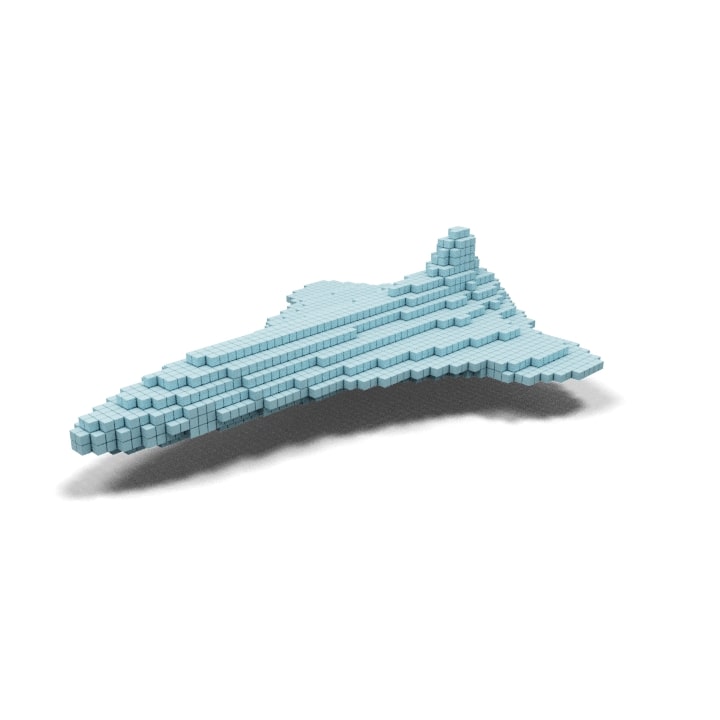}\\
``a  sling'' & ``n ladder back chair'' & ``an egg chair'' & ``a stool' & ``a commercial plane'' & ``a jet''\\
\includegraphics[width=0.15\linewidth]{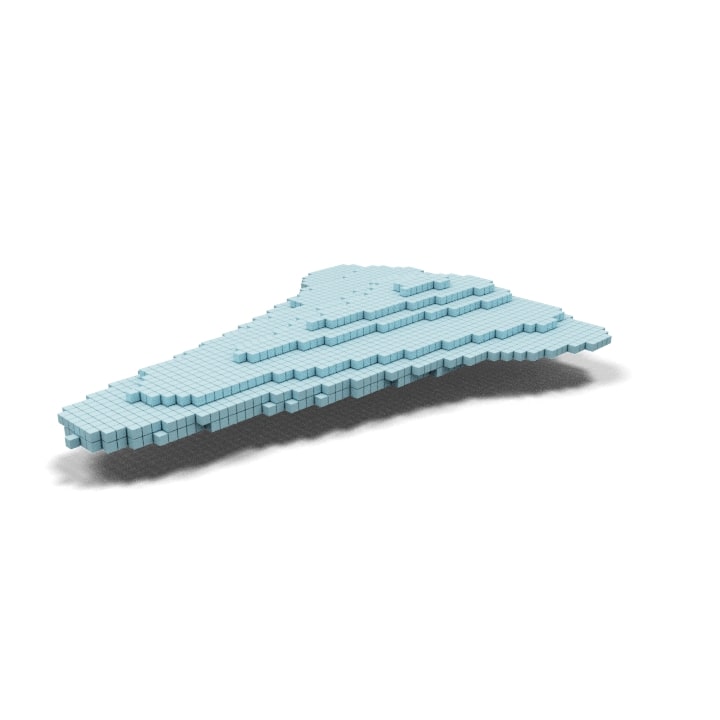} &
\includegraphics[width=0.15\linewidth]{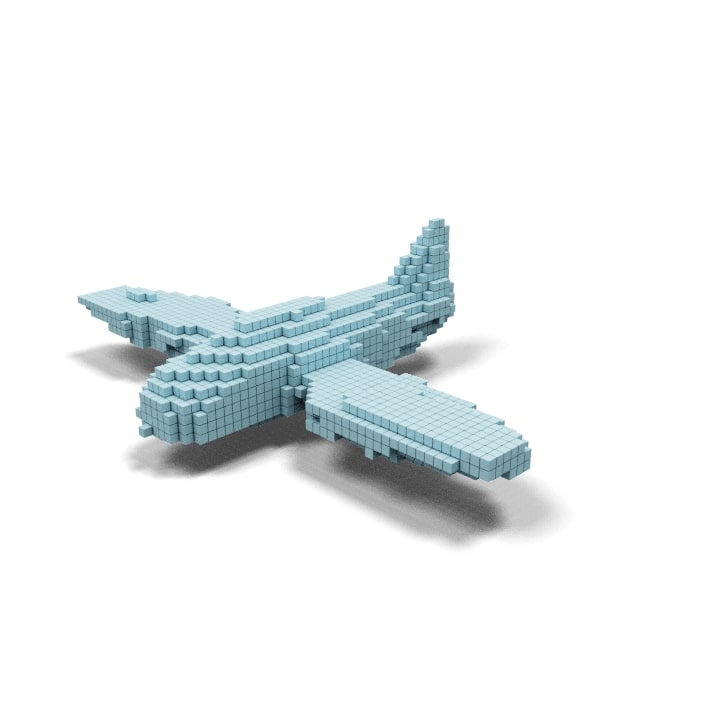} &
\includegraphics[width=0.15\linewidth]{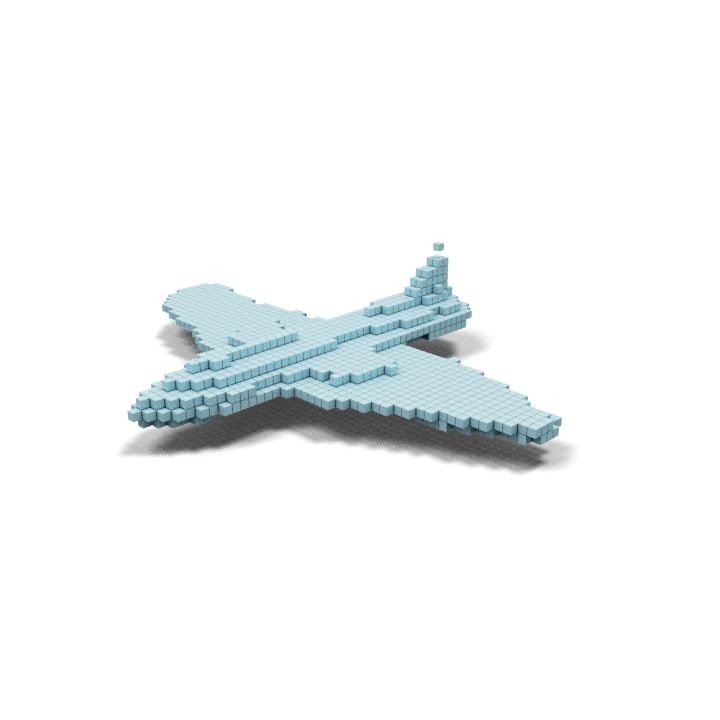} &
\includegraphics[width=0.15\linewidth]{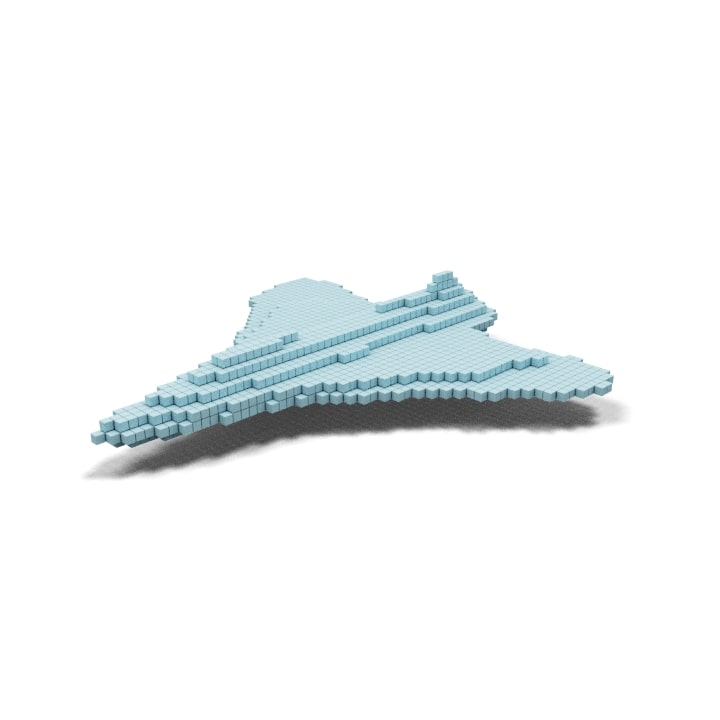} &
\includegraphics[width=0.15\linewidth]{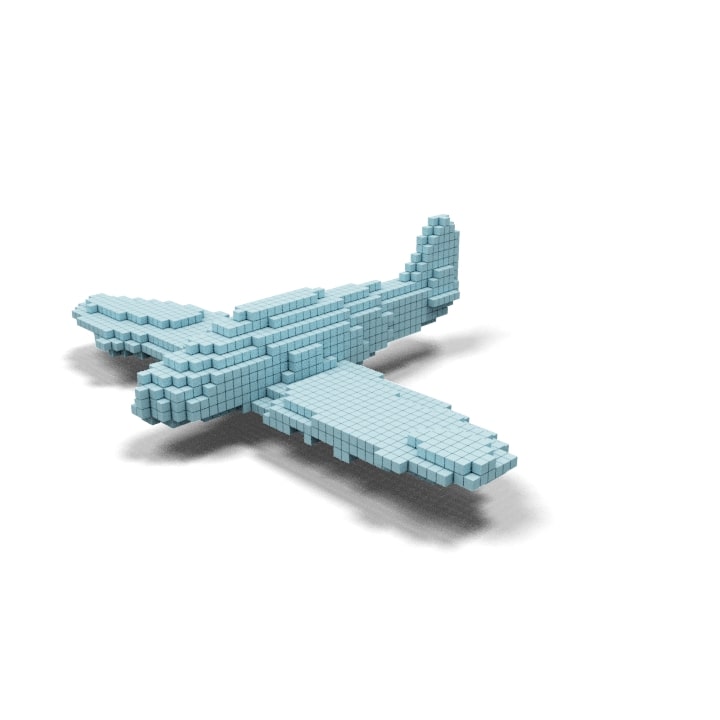} &
\includegraphics[width=0.15\linewidth]{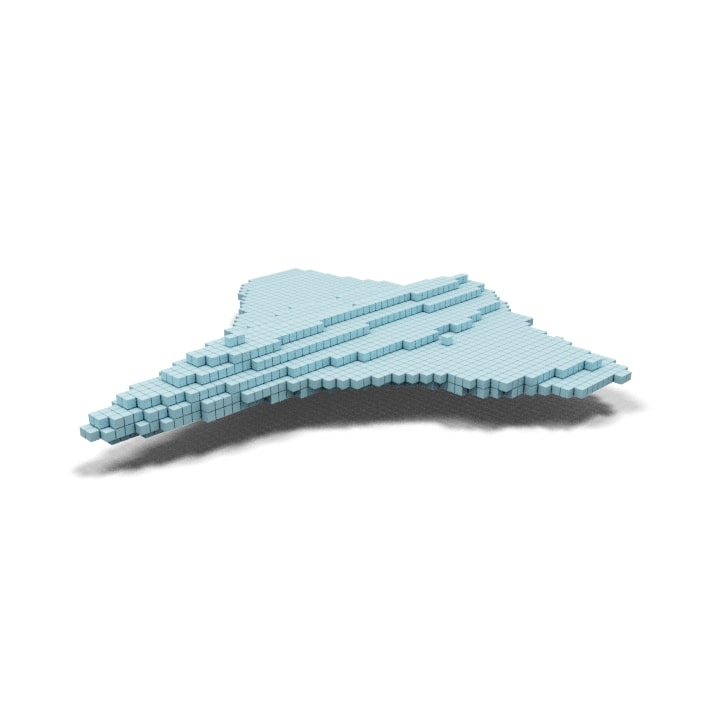}\\
``a delta wing'' & ``a seaplane'' & ``a straight wing plane'' & ``a swept wing plane' & ``a biplane'' & ``a fighter plane''\\
\includegraphics[width=0.15\linewidth]{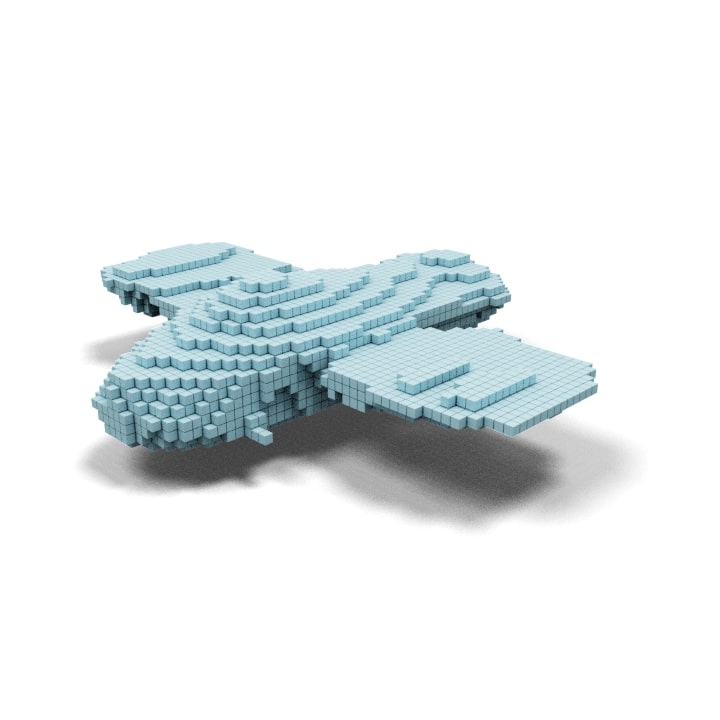} &
\includegraphics[width=0.15\linewidth]{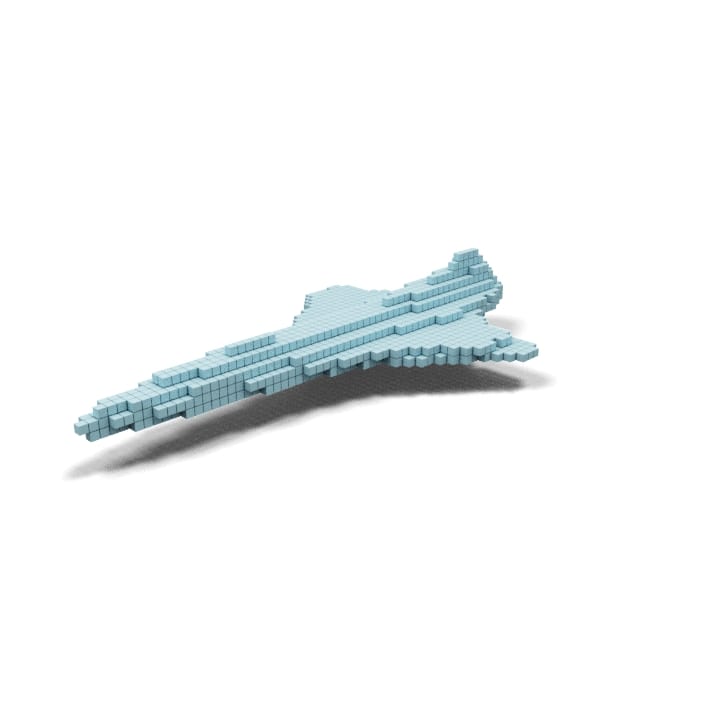} &
\includegraphics[width=0.15\linewidth]{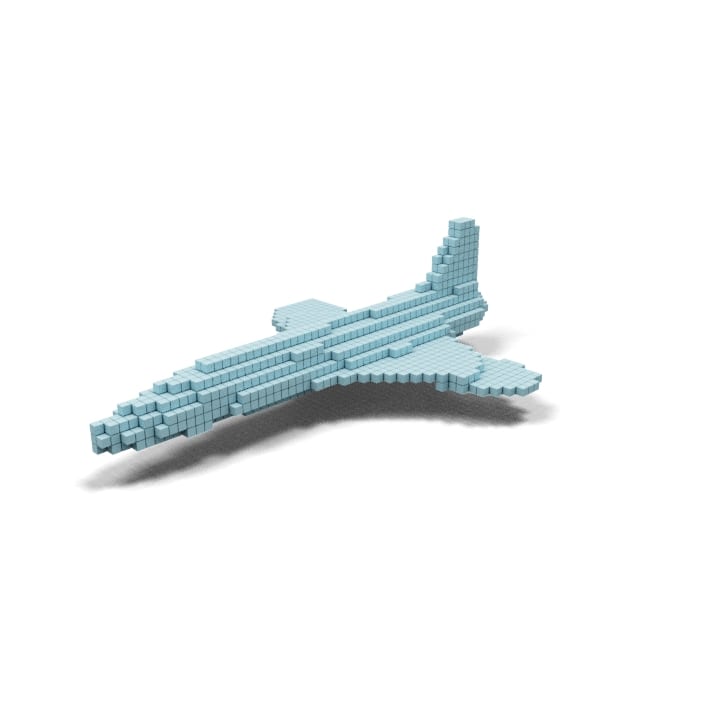} &
\includegraphics[width=0.15\linewidth]{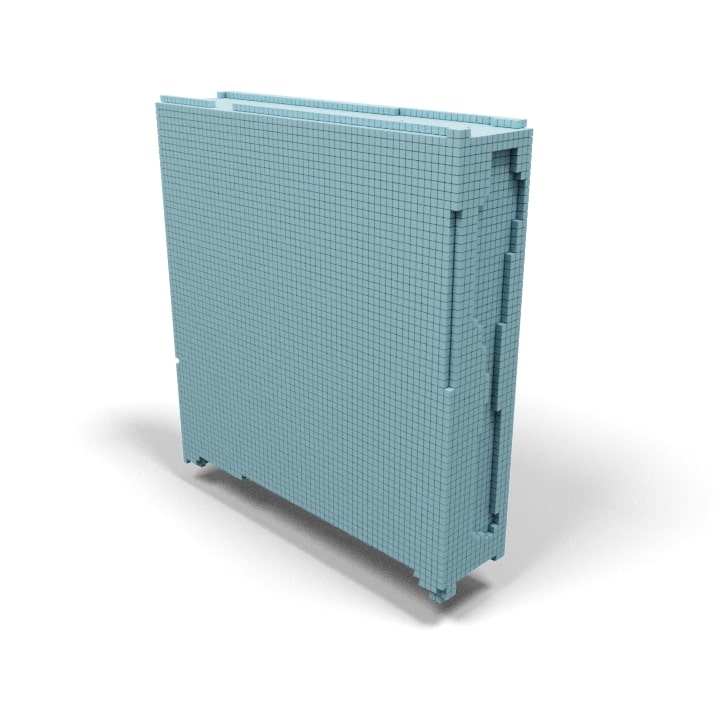} &
\includegraphics[width=0.15\linewidth]{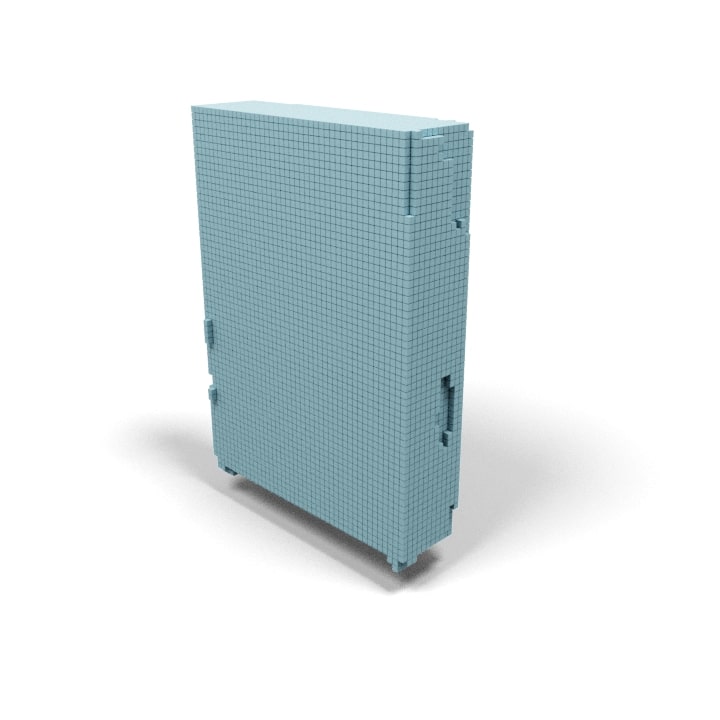} &
\includegraphics[width=0.15\linewidth]{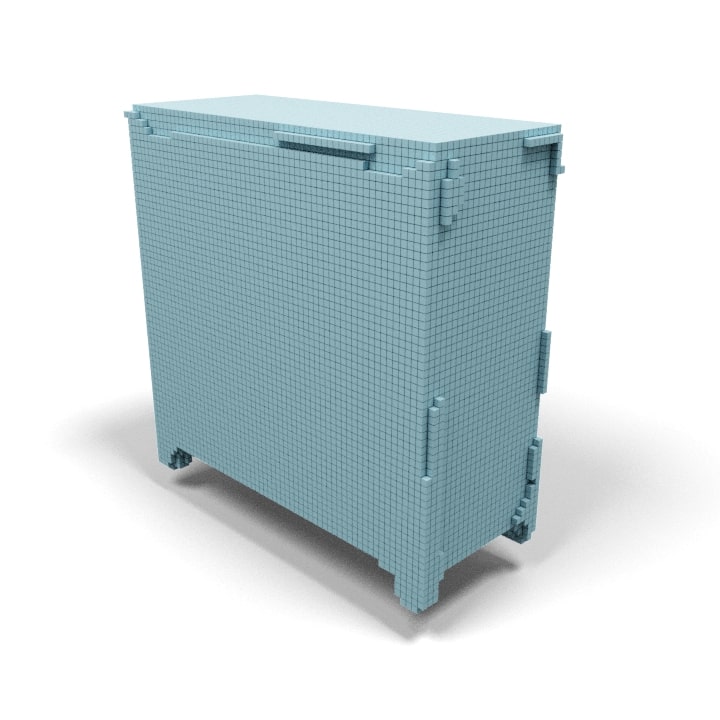}\\
``a  military drone'' & ``a supersonic plane'' & ``a rocket plane'' & ``a cabinet' & ``a garage cabinet'' & ``a desk cabinet''\\

\end{tabular}
}
\end{center}
  \caption{Additional shape generation results using sub-category text queries of CLIP-Forge.}
\label{fig:sub_category_pics}
\end{figure*}

\begin{figure*}[t!]
\begin{center}
\setlength{\tabcolsep}{2pt}
\small{
\begin{tabular}{cccccc}

\includegraphics[width=0.15\linewidth]{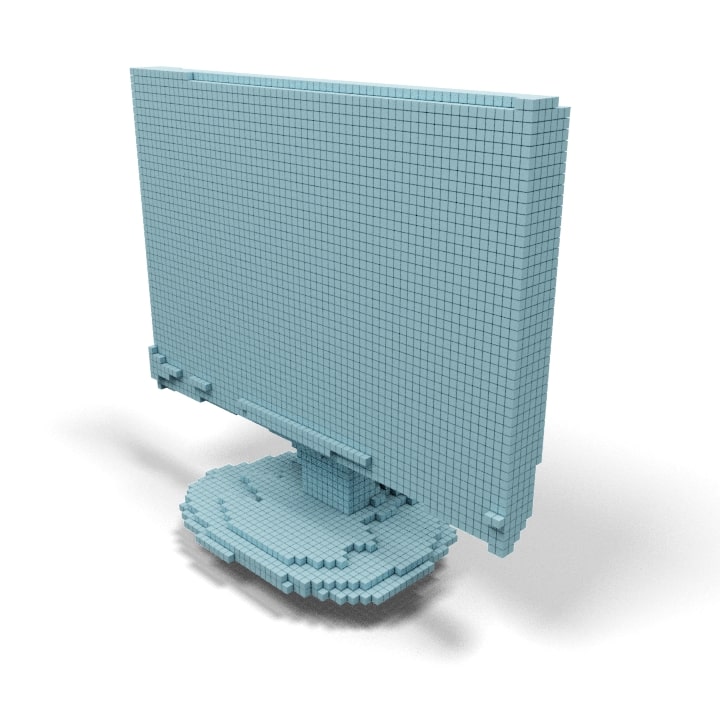} &
\includegraphics[width=0.15\linewidth]{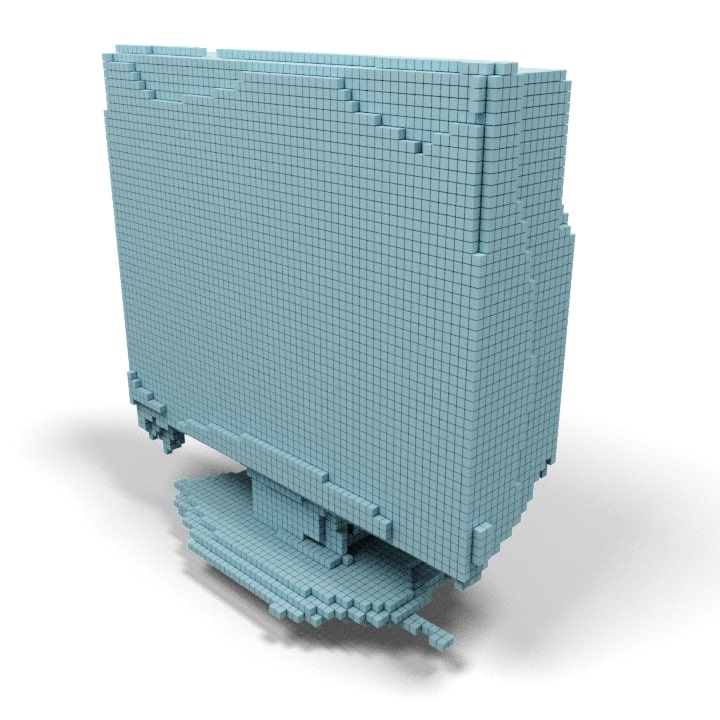} &
\includegraphics[width=0.15\linewidth]{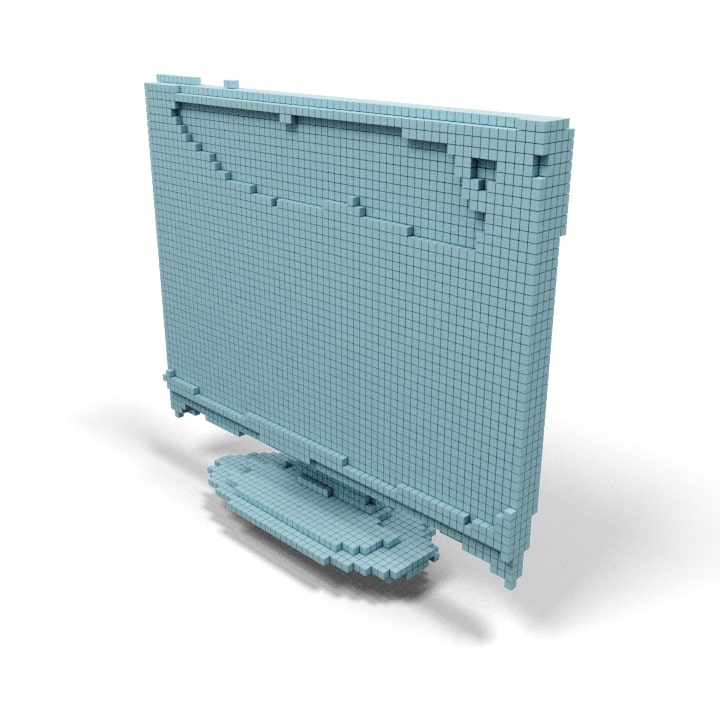} &
\includegraphics[width=0.15\linewidth]{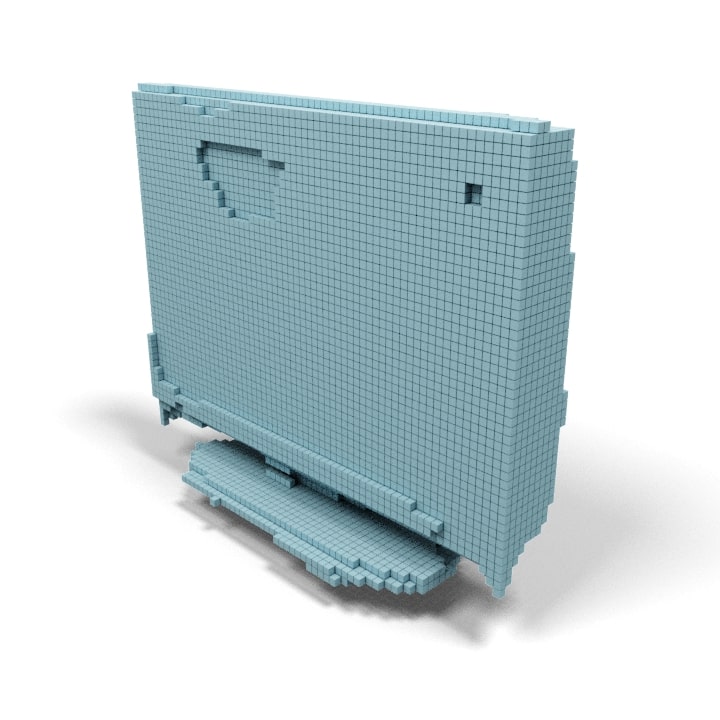} &
\includegraphics[width=0.15\linewidth]{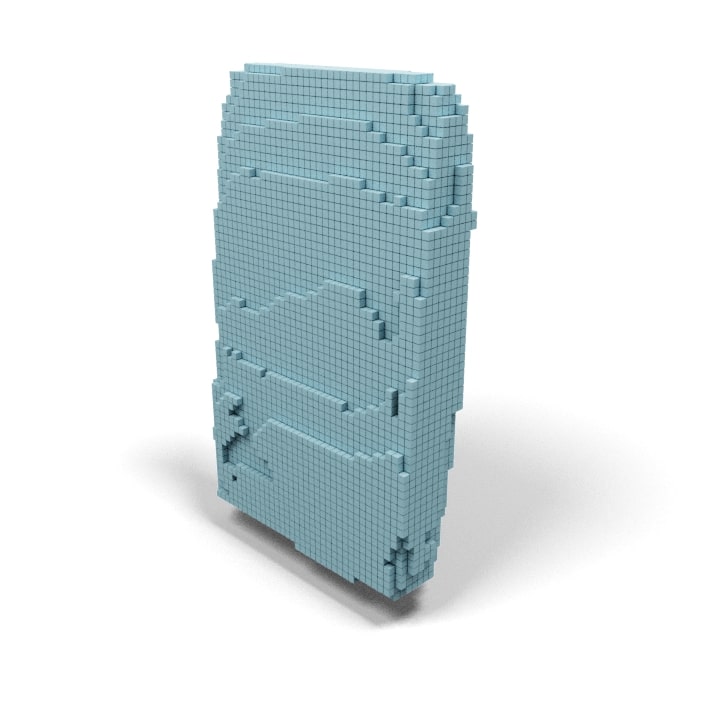} &
\includegraphics[width=0.15\linewidth]{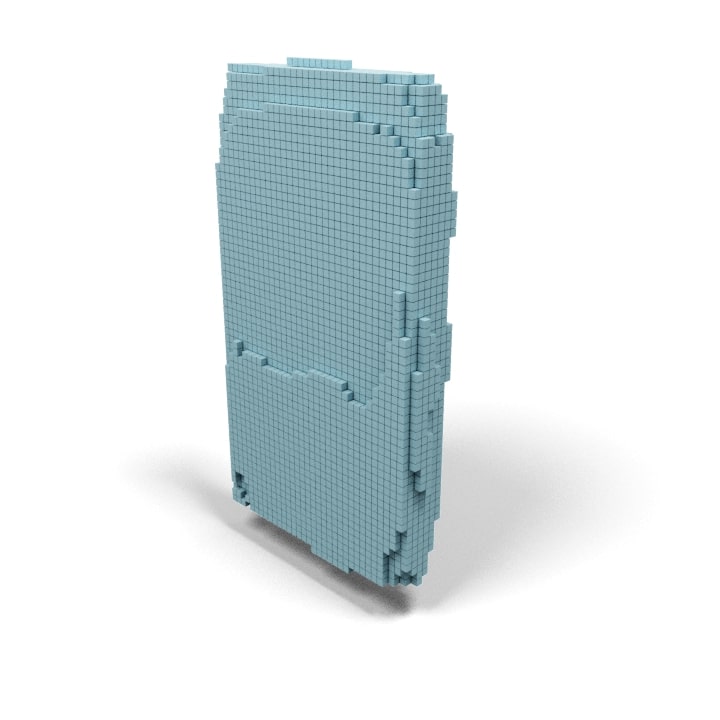}\\
``a  monitor'' & ``a crt monitor'' & ``a flat panel display'' & ``a television' & ``a mobile phone'' & ``a flip-phone''\\
\includegraphics[width=0.15\linewidth]{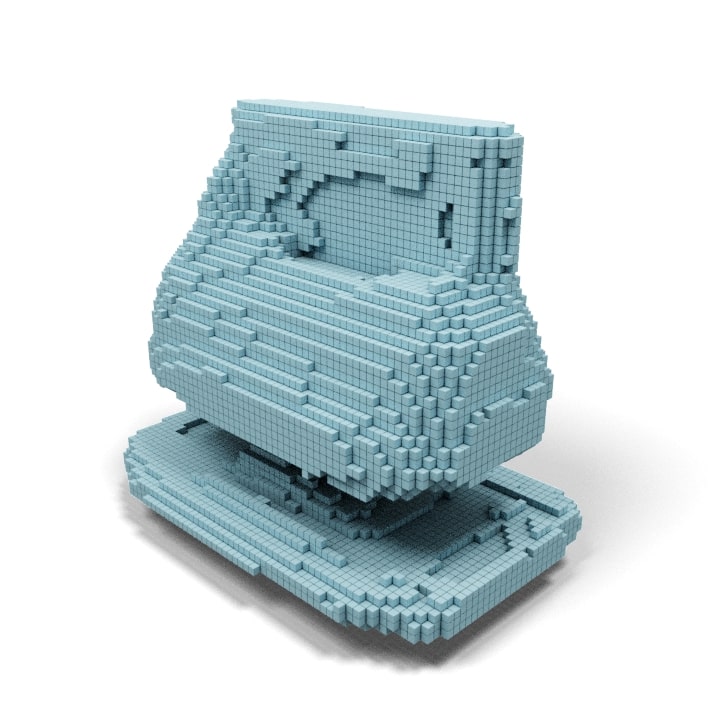} &
\includegraphics[width=0.15\linewidth]{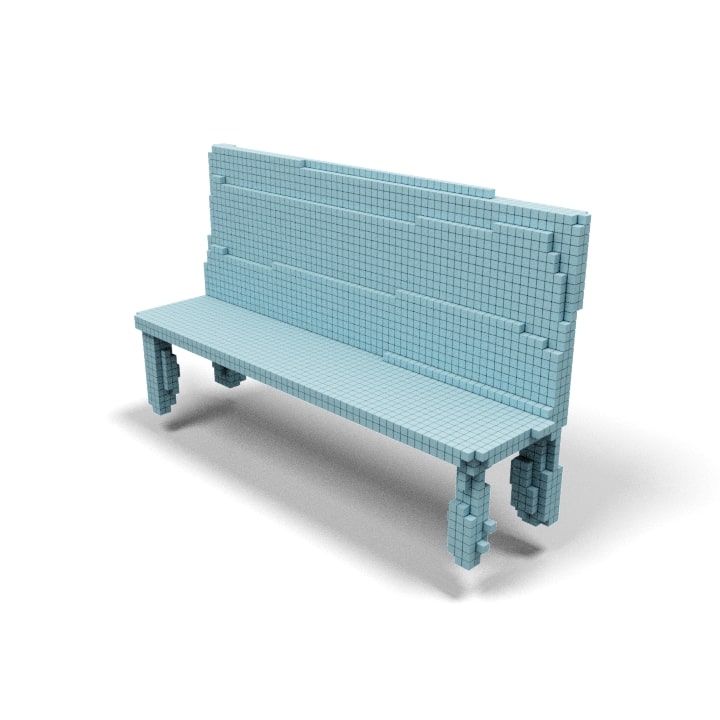} &
\includegraphics[width=0.15\linewidth]{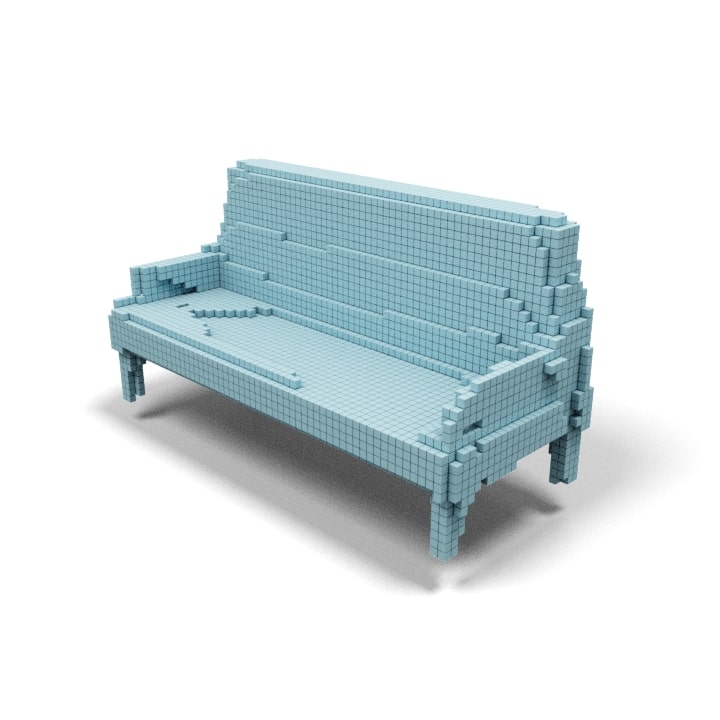} &
\includegraphics[width=0.15\linewidth]{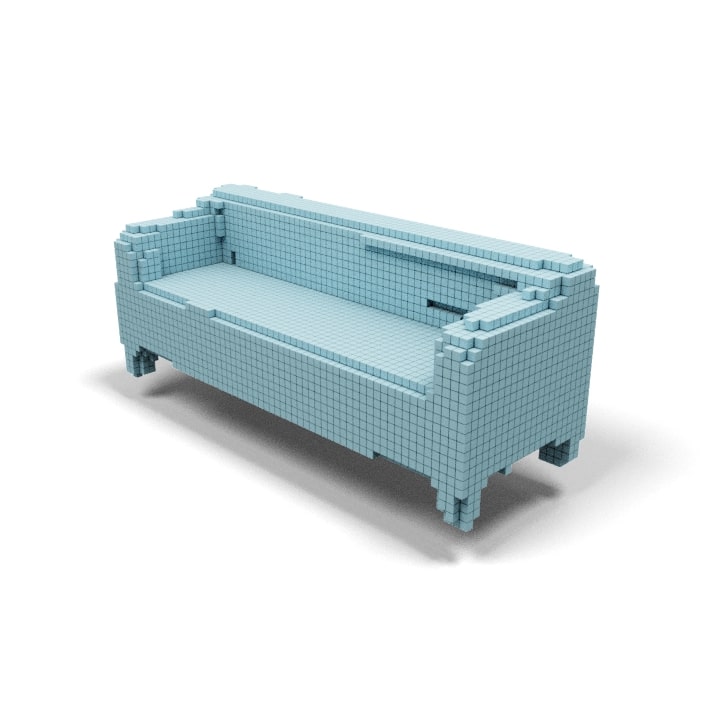} &
\includegraphics[width=0.15\linewidth]{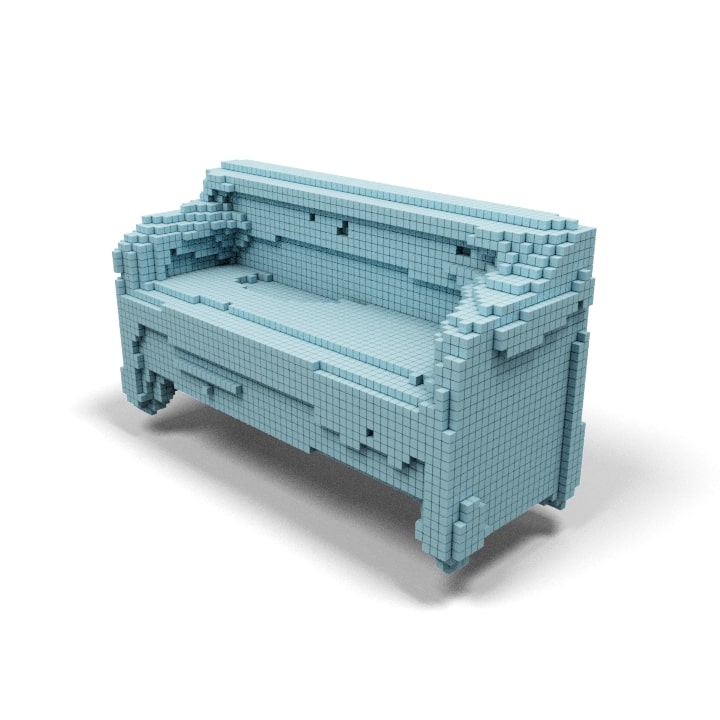} &
\includegraphics[width=0.15\linewidth]{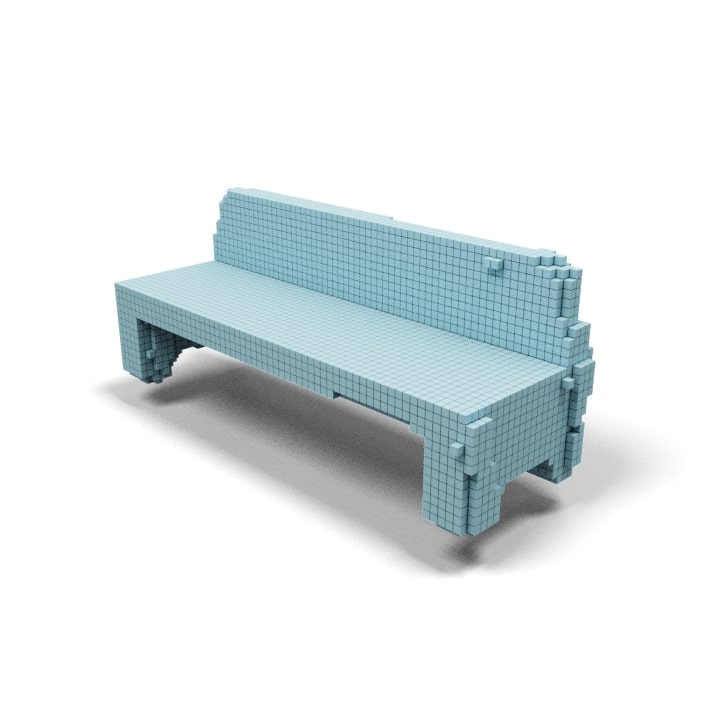}\\
``a desk phone'' & ``a back bench'' & ``a pew'' & ``a storage bench' & ``a laboratory bench'' & ``a flat bench''\\
\includegraphics[width=0.15\linewidth]{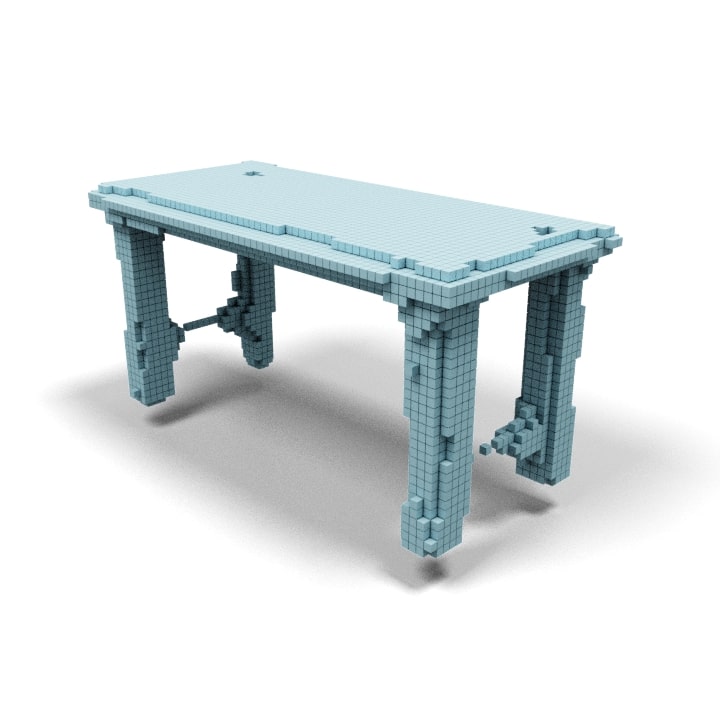} &
\includegraphics[width=0.15\linewidth]{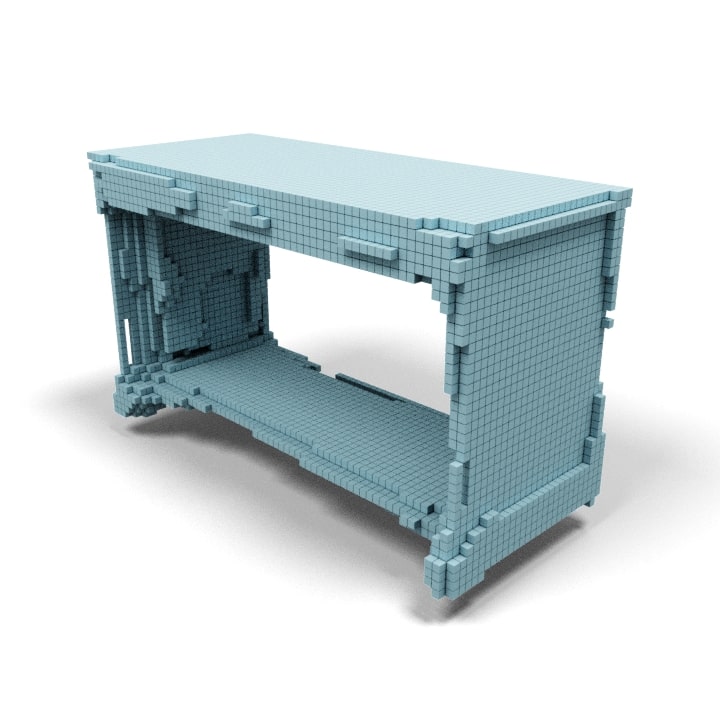} &
\includegraphics[width=0.15\linewidth]{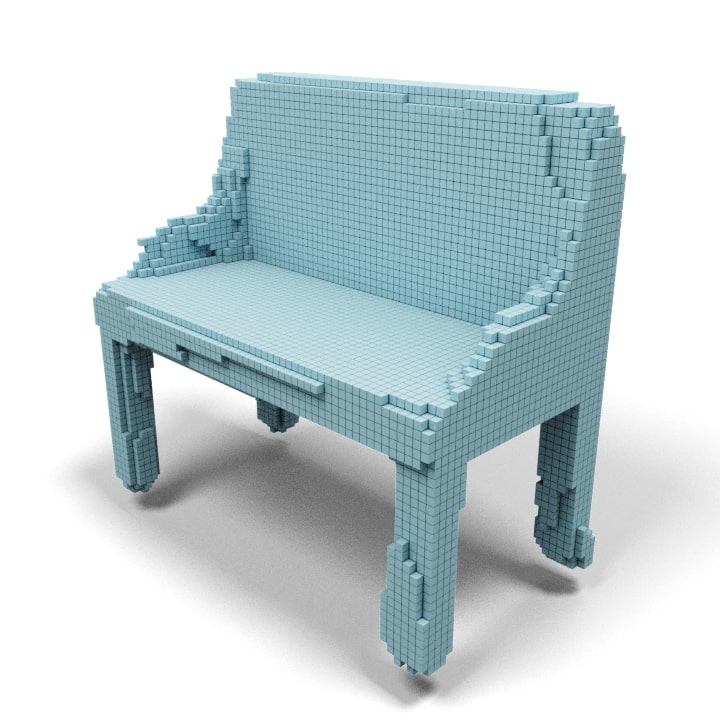} &
\includegraphics[width=0.15\linewidth]{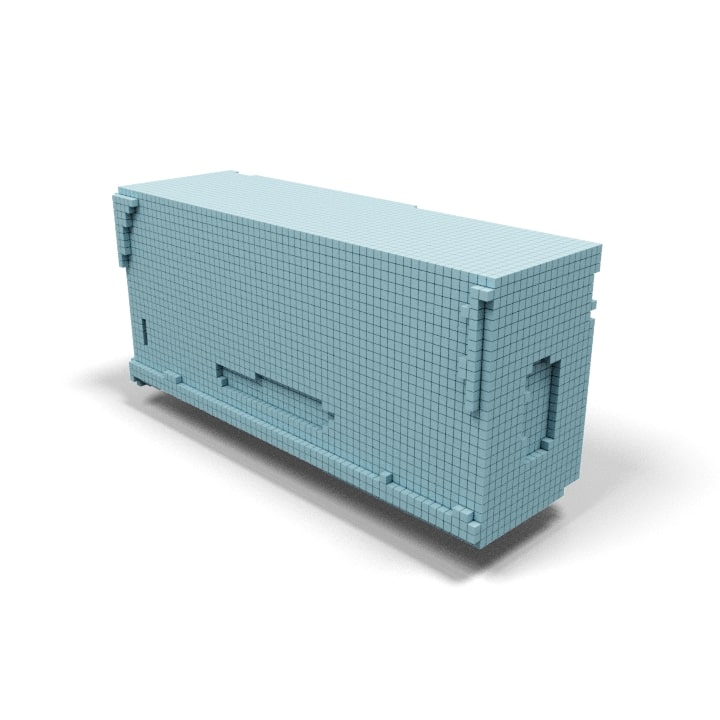} &
\includegraphics[width=0.15\linewidth]{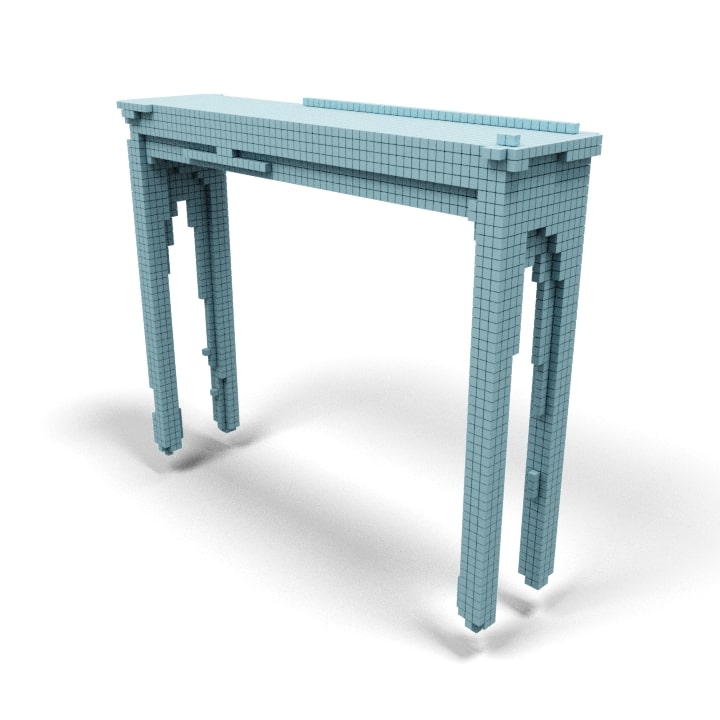} &
\includegraphics[width=0.15\linewidth]{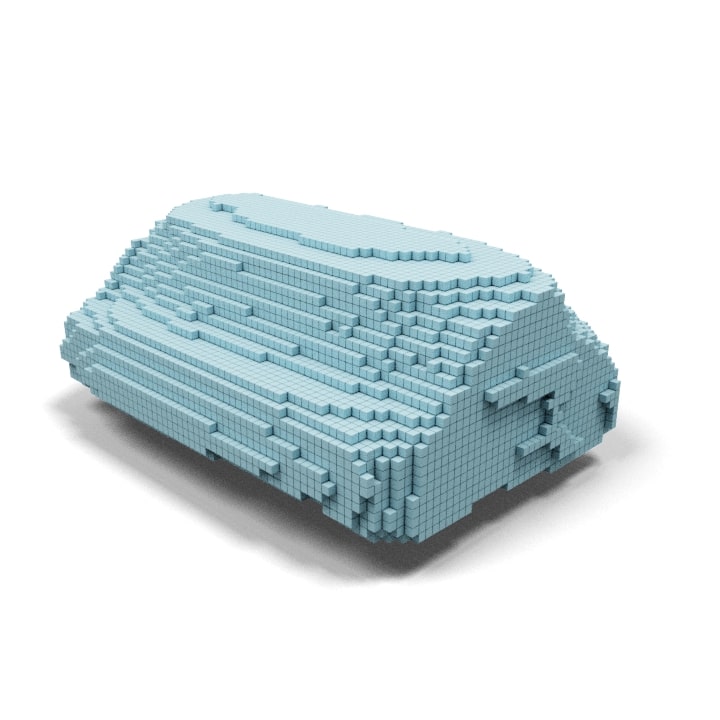}\\
``a refectory table'' & ``a desk'' & ``a dressing table'' & ``a counter' & ``a console table'' & ``an operating table''\\
\includegraphics[width=0.15\linewidth]{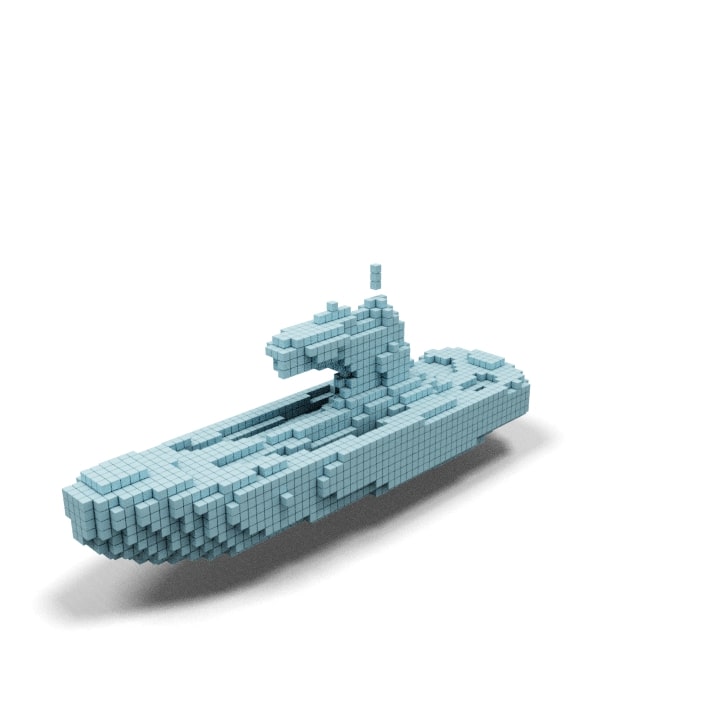} &
\includegraphics[width=0.15\linewidth]{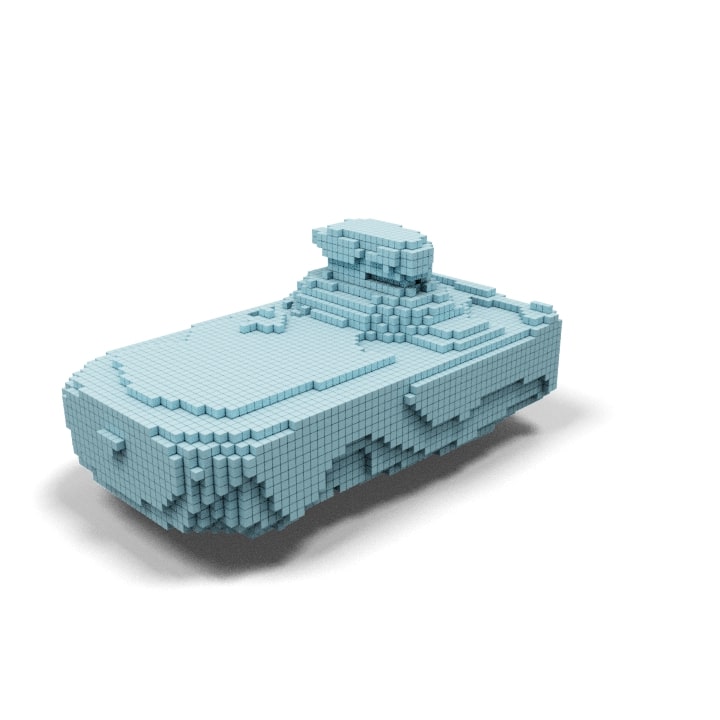} &
\includegraphics[width=0.15\linewidth]{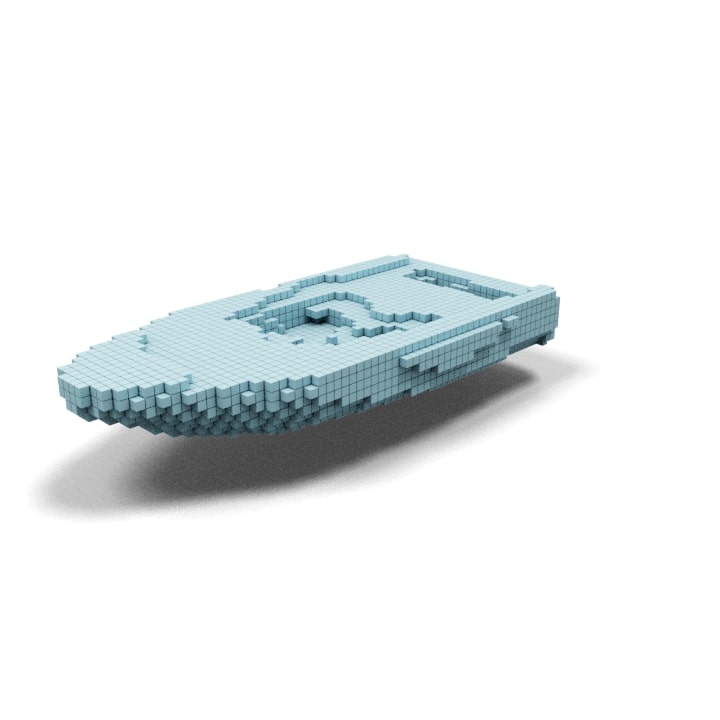} &
\includegraphics[width=0.15\linewidth]{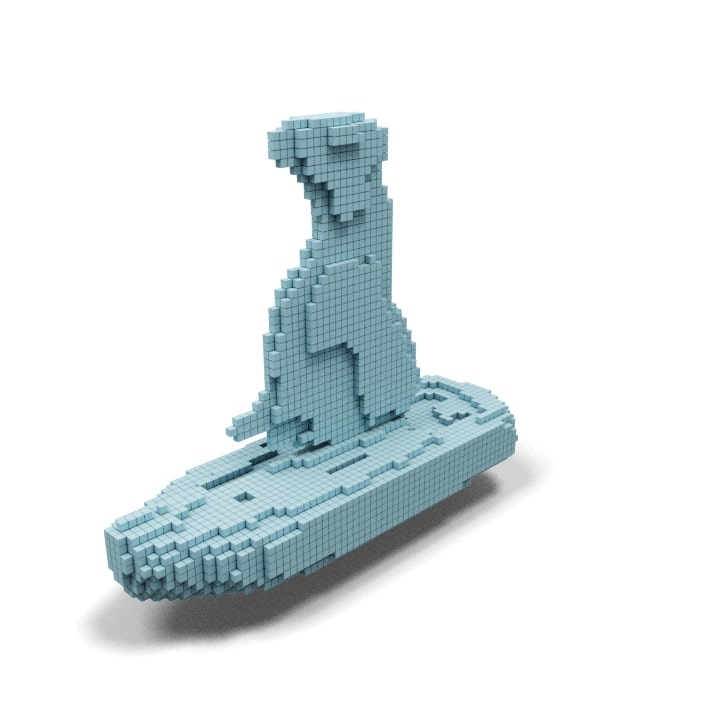} &
\includegraphics[width=0.15\linewidth]{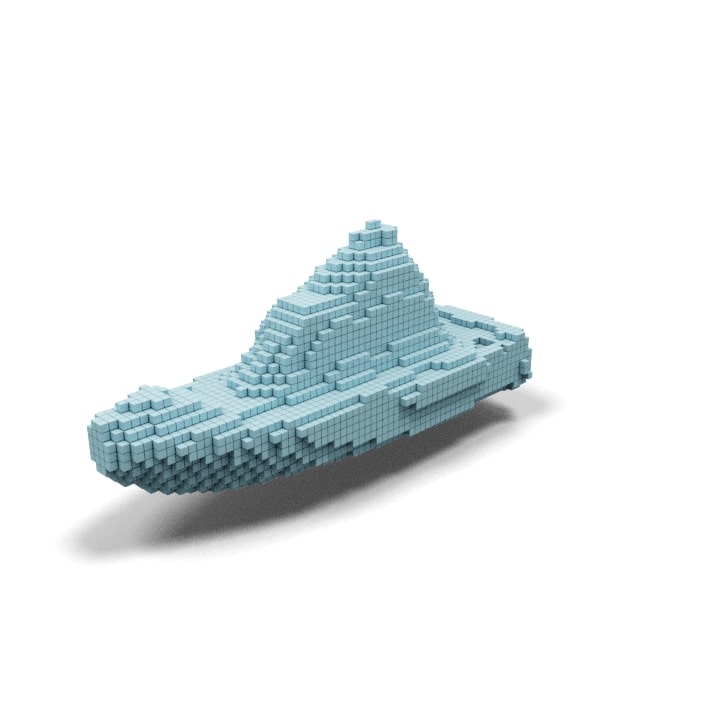} &
\includegraphics[width=0.15\linewidth]{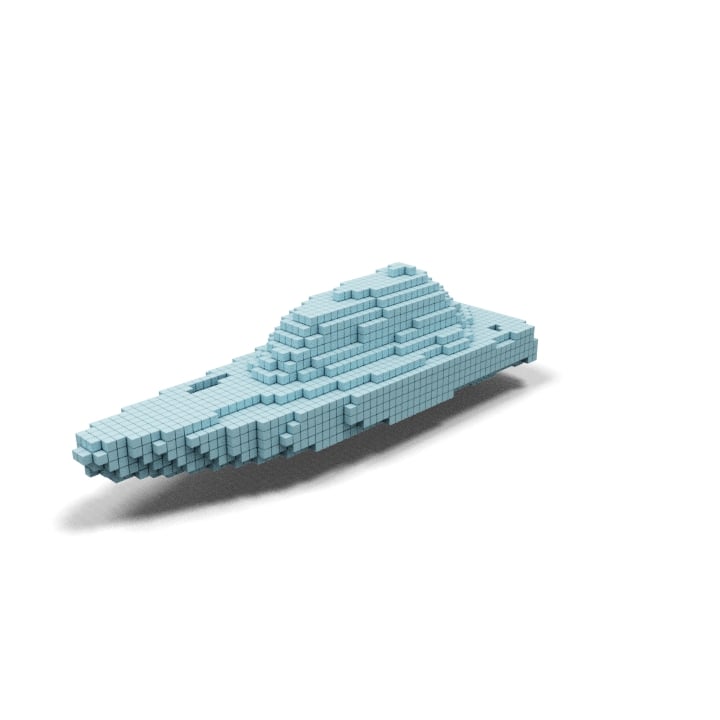}\\
``a war ship'' & ``a cabin cruiser'' & ``a speedboat'' & ``a sail boat' & ``a yacht'' & ``a mega yacht''\\
\includegraphics[width=0.15\linewidth]{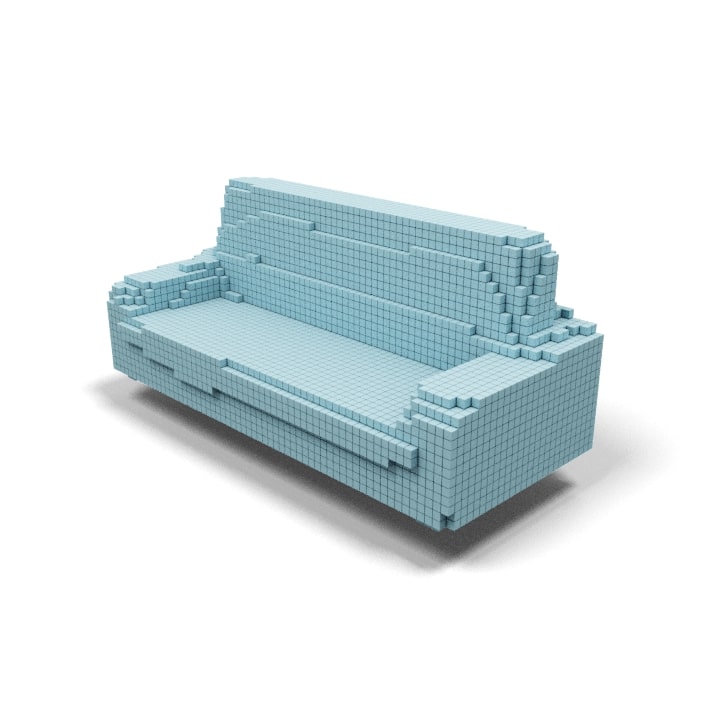} &
\includegraphics[width=0.15\linewidth]{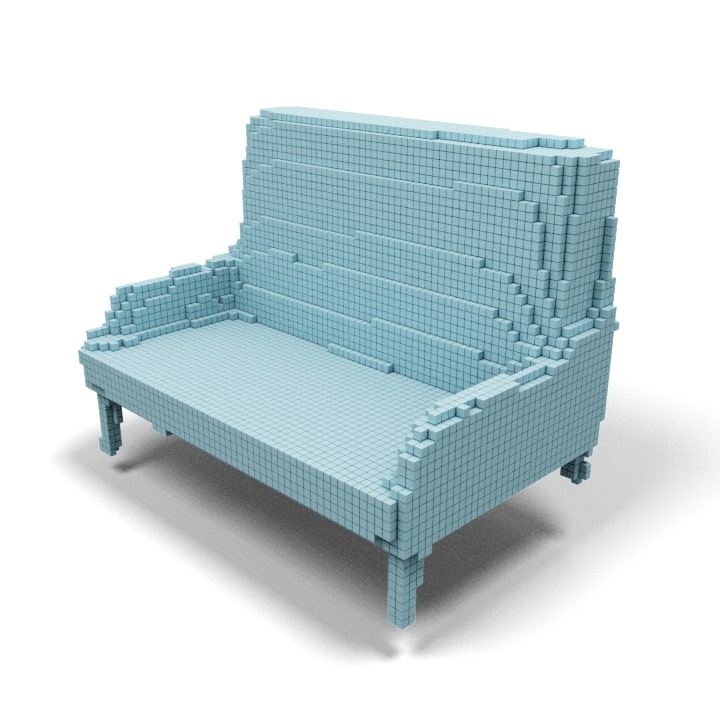} &
\includegraphics[width=0.15\linewidth]{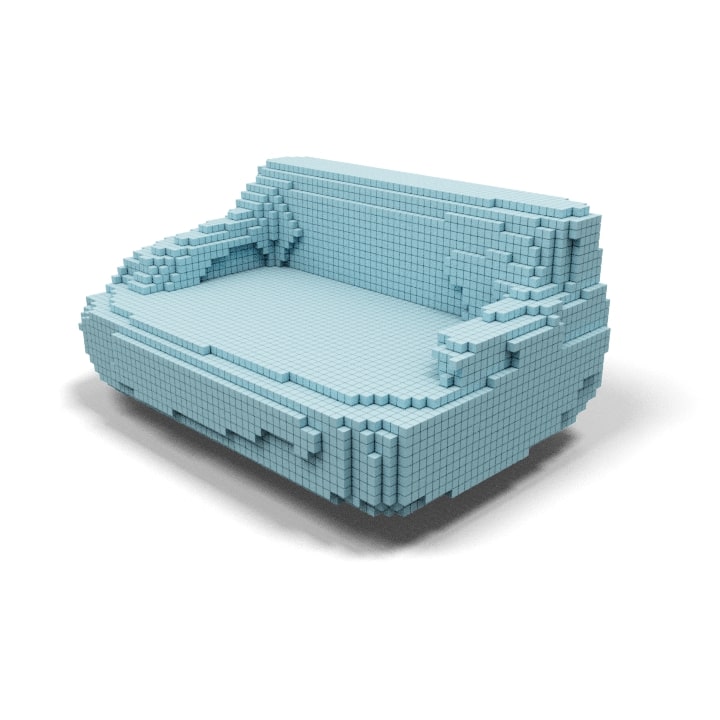} &
\includegraphics[width=0.15\linewidth]{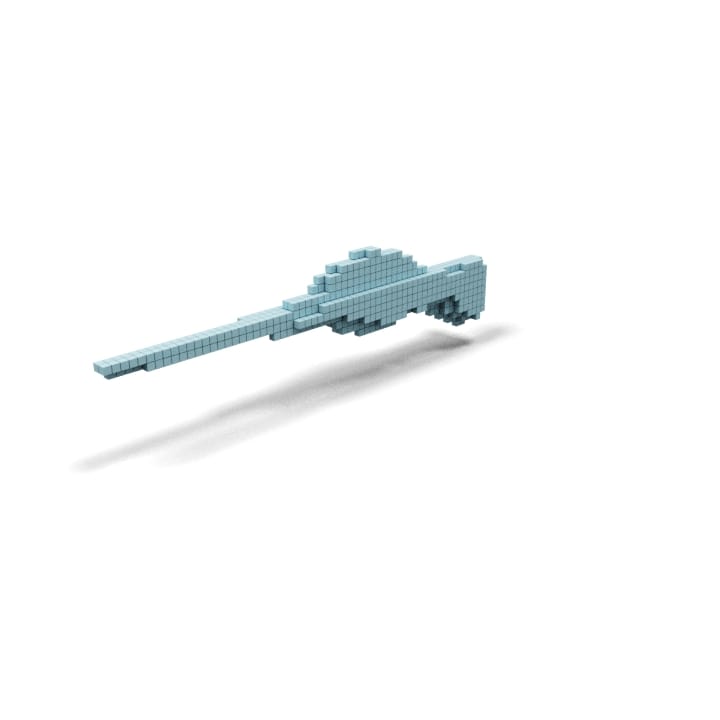} &
\includegraphics[width=0.15\linewidth]{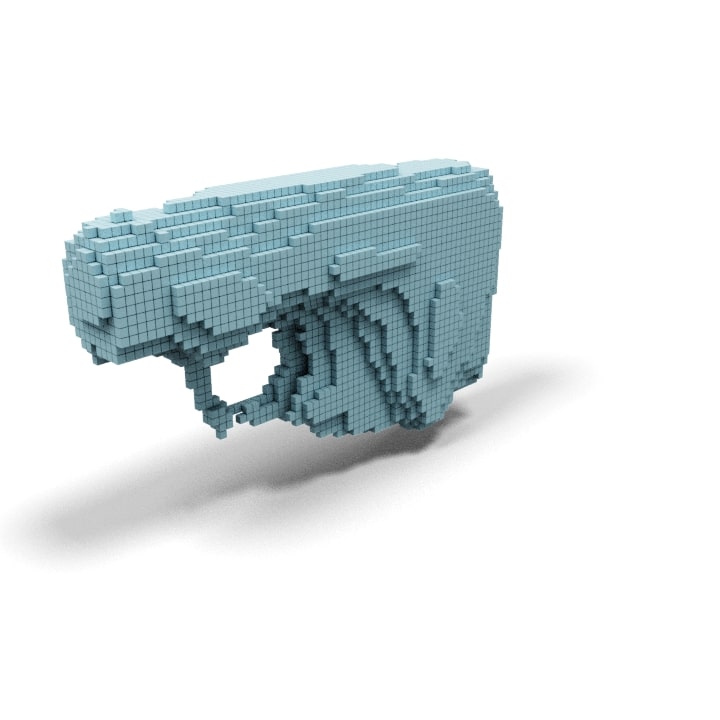} &
\includegraphics[width=0.15\linewidth]{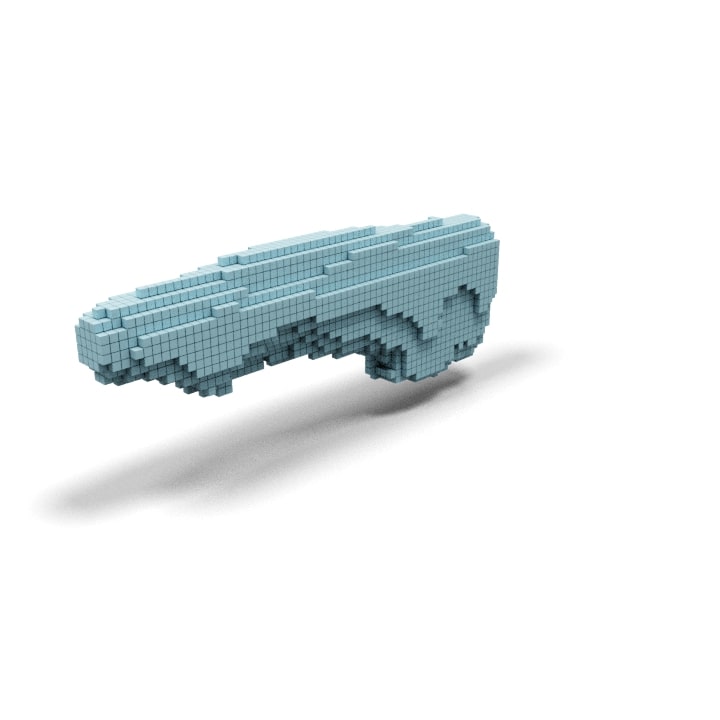}\\
``a chesterfield'' & ``a love seat'' & ``an ottoman'' & ``a sniper rifle' & ``a pistol'' & ``a shotgun''\\
\includegraphics[width=0.15\linewidth]{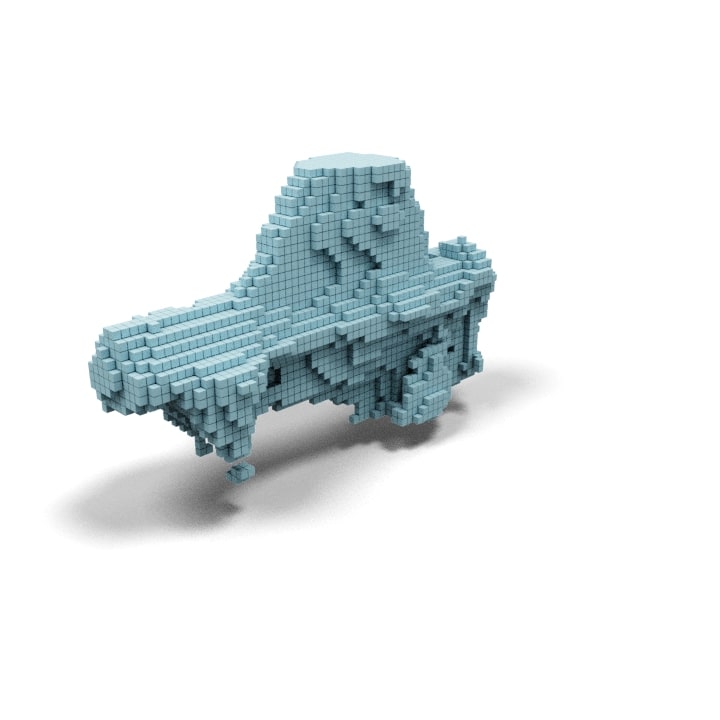} &
\includegraphics[width=0.15\linewidth]{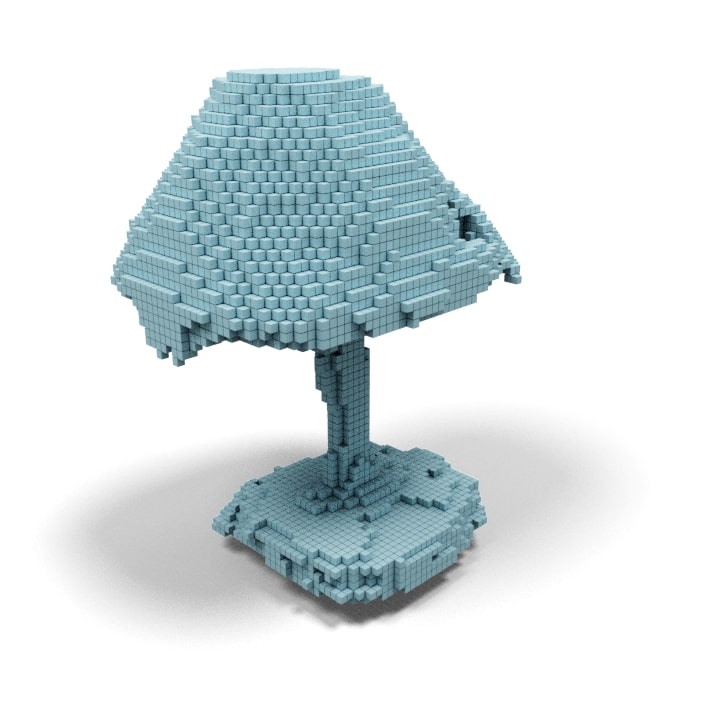} &
\includegraphics[width=0.15\linewidth]{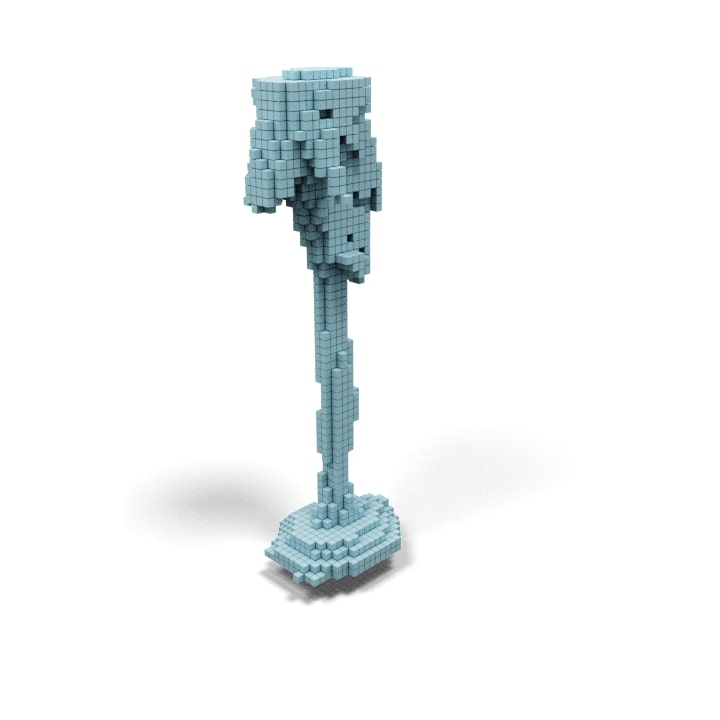} &
\includegraphics[width=0.15\linewidth]{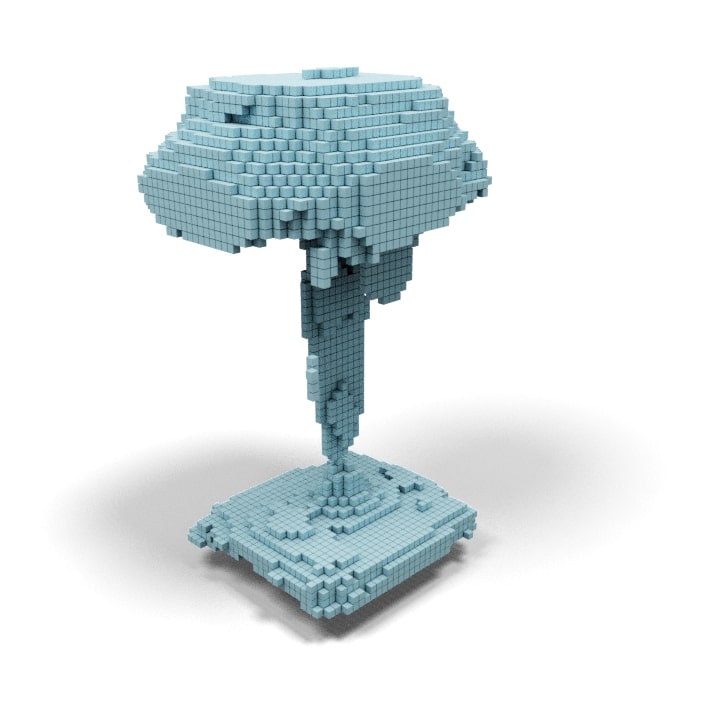} &
\includegraphics[width=0.15\linewidth]{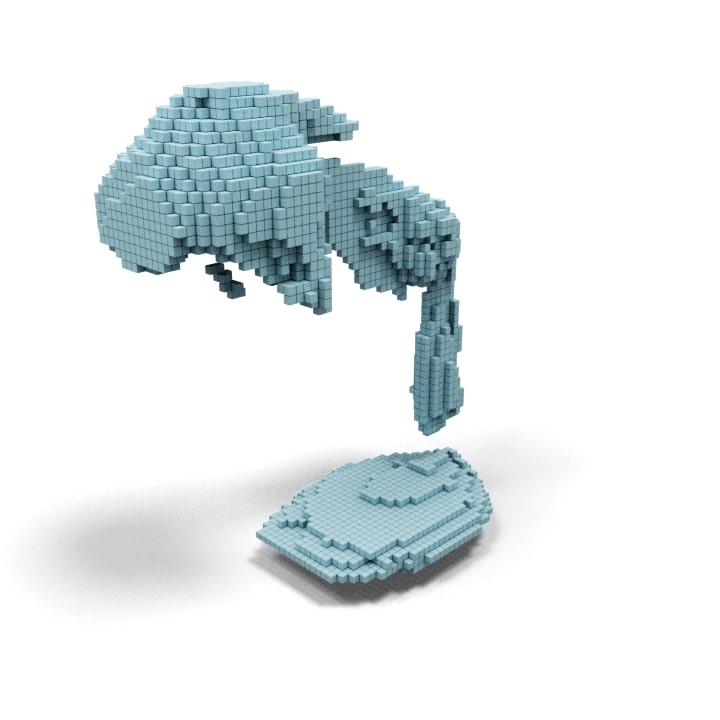} &
\includegraphics[width=0.15\linewidth]{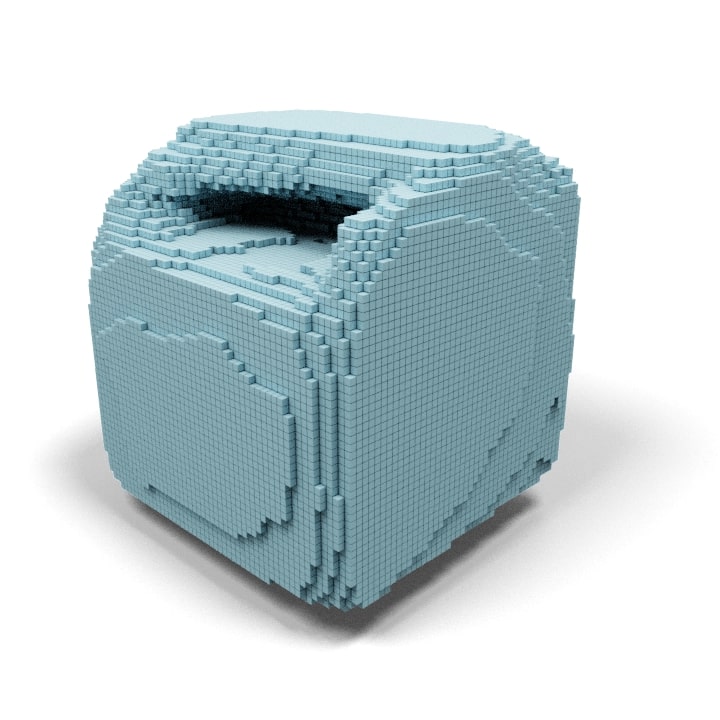}\\
``a machine gun'' & ``a lamp'' & ``a street lamp'' & ``a LED table lamp' & ``a swing arm lamp'' & ``a subwoofer speaker''\\
\end{tabular}
}
\end{center}
  \caption{Additional shape generation results using sub-category text queries of CLIP-Forge (continued).}
\label{fig:sub_category_pics_cont}
\end{figure*}

\begin{figure*}[t!]
\begin{center}
\setlength{\tabcolsep}{2pt}
\small{
\begin{tabular}{cccccc}

\includegraphics[width=0.15\linewidth]{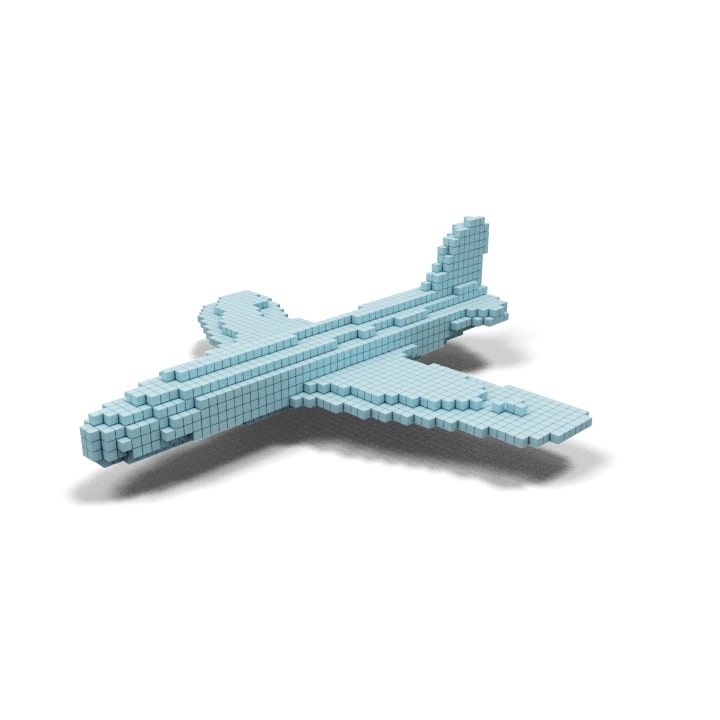} &
\includegraphics[width=0.15\linewidth]{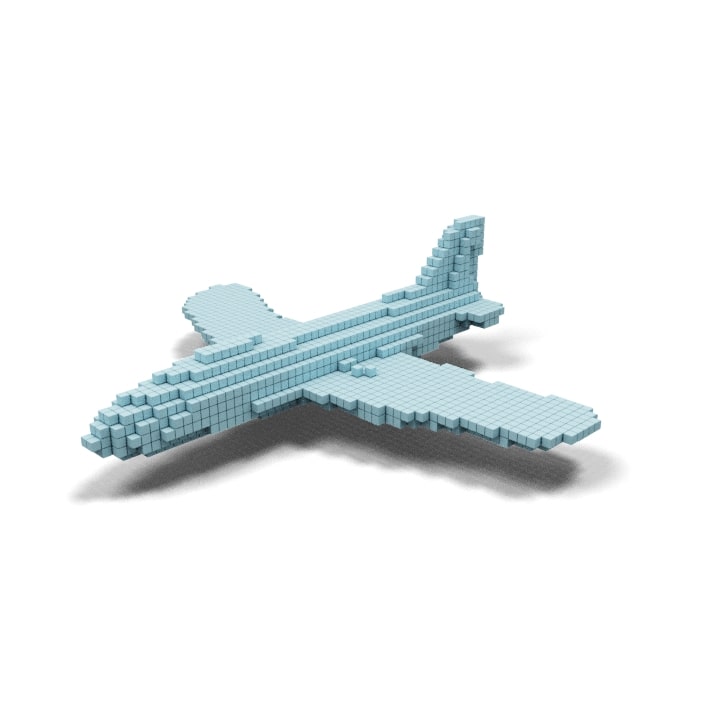} &
\includegraphics[width=0.15\linewidth]{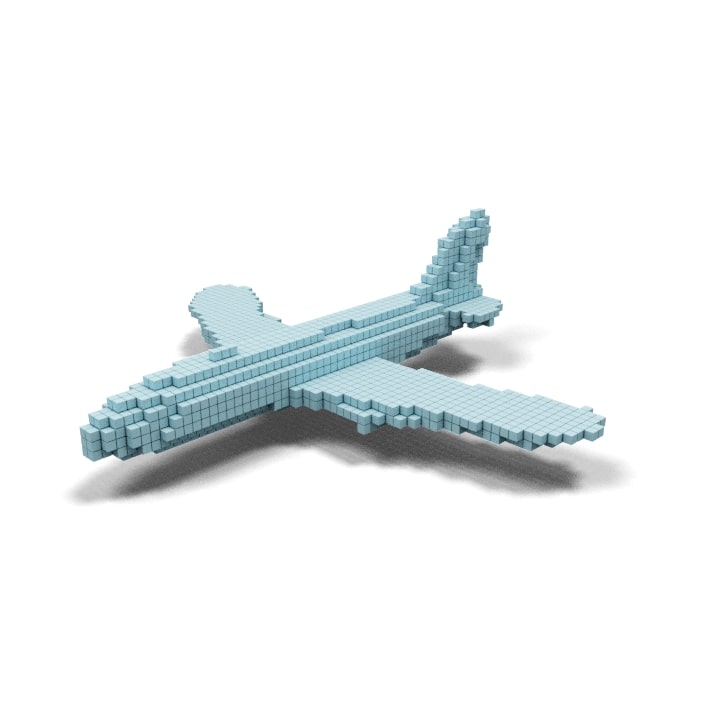} &
\includegraphics[width=0.15\linewidth]{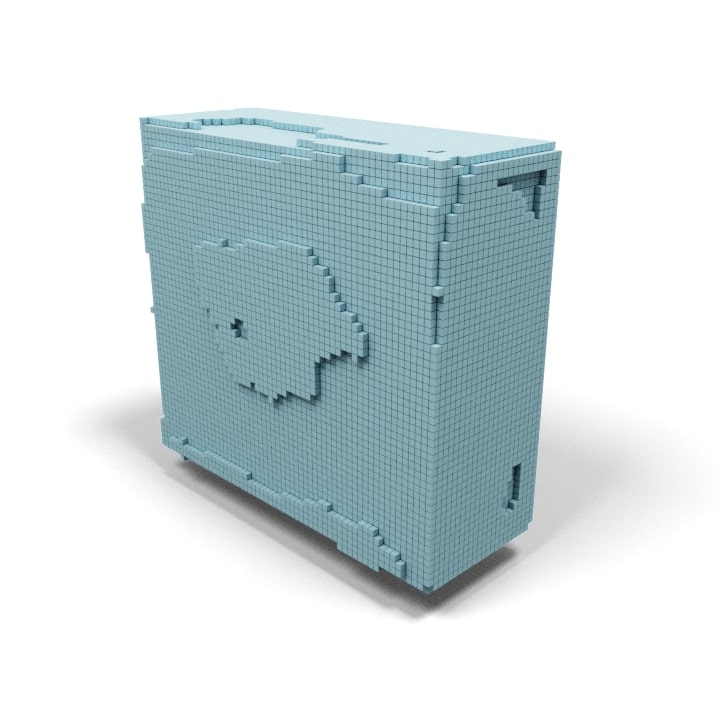} &
\includegraphics[width=0.15\linewidth]{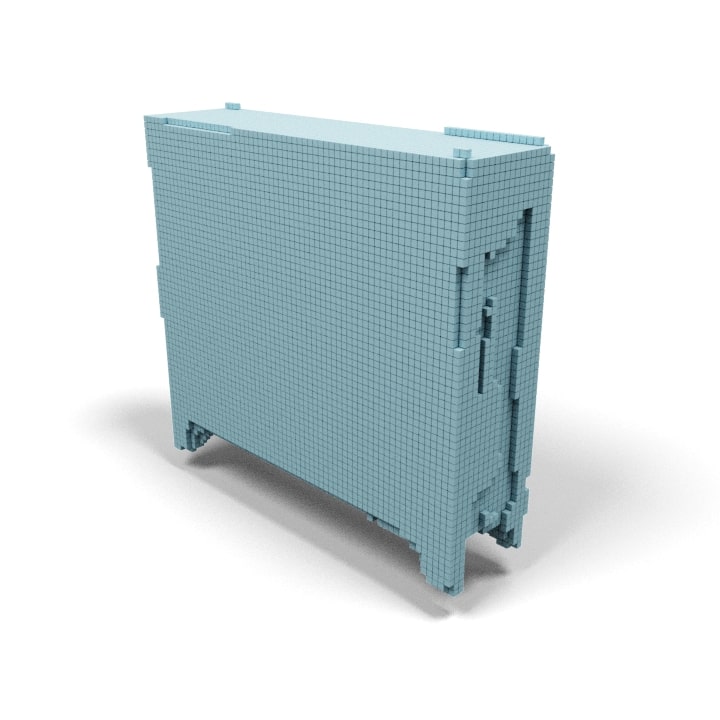} &
\includegraphics[width=0.15\linewidth]{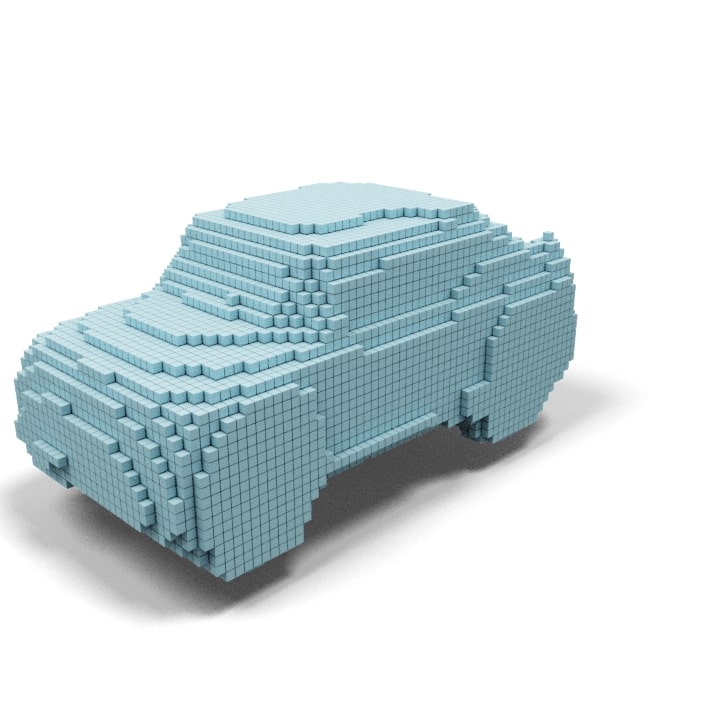}\\
``an aeroplane'' & ``a plane'' & ``an airplane'' & ``a container'' & ``a dresser'' & ``an auto''\\
\includegraphics[width=0.15\linewidth]{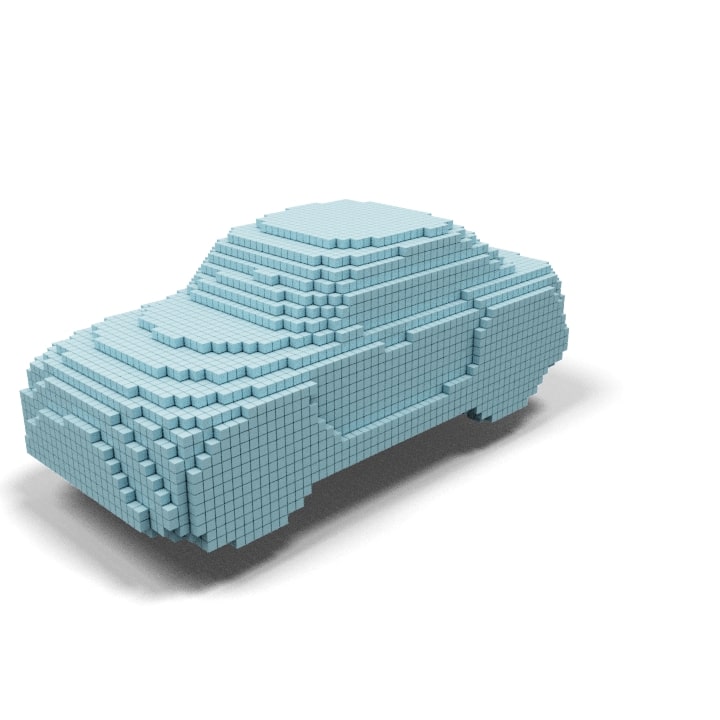} &
\includegraphics[width=0.15\linewidth]{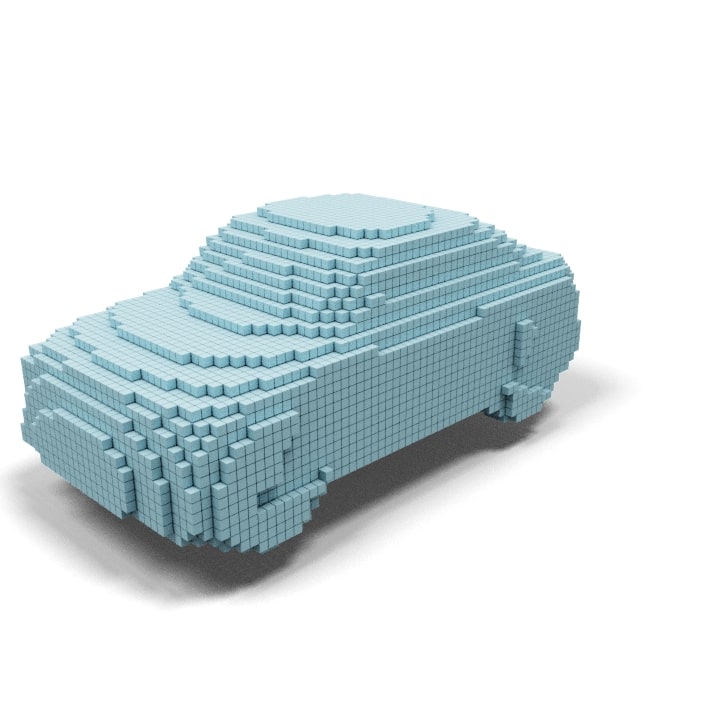} &
\includegraphics[width=0.15\linewidth]{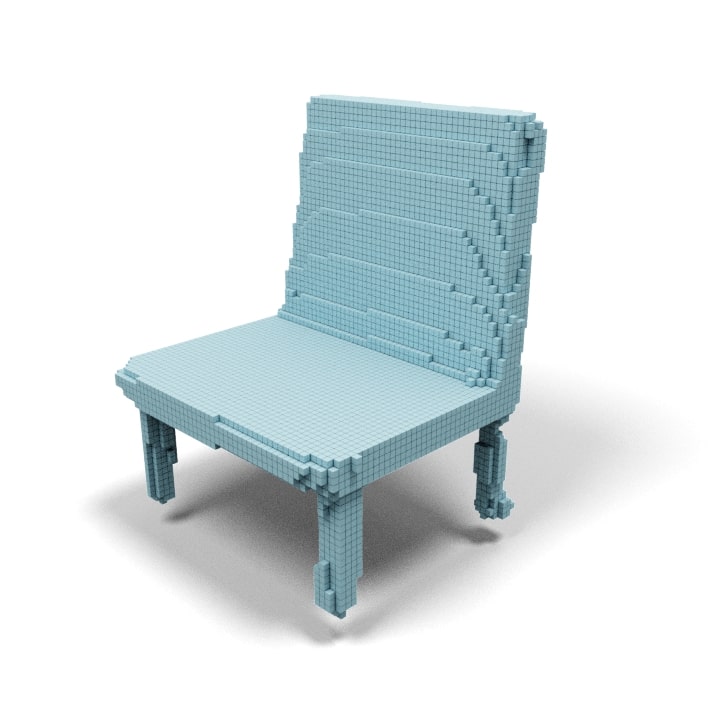} &
\includegraphics[width=0.15\linewidth]{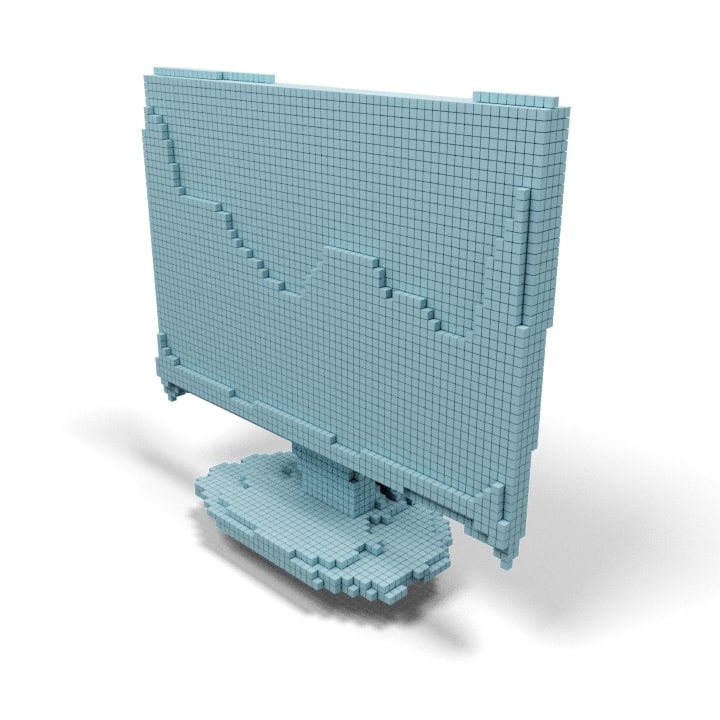} &
\includegraphics[width=0.15\linewidth]{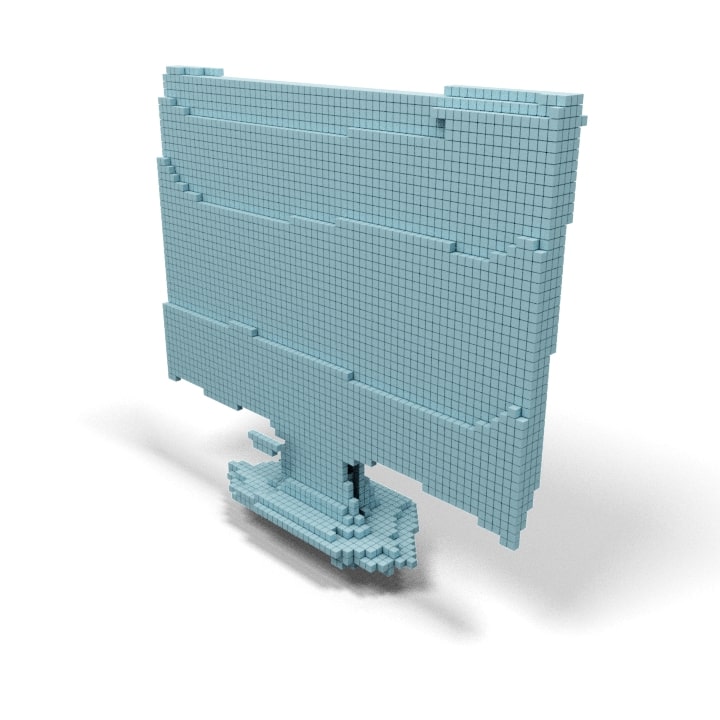} &
\includegraphics[width=0.15\linewidth]{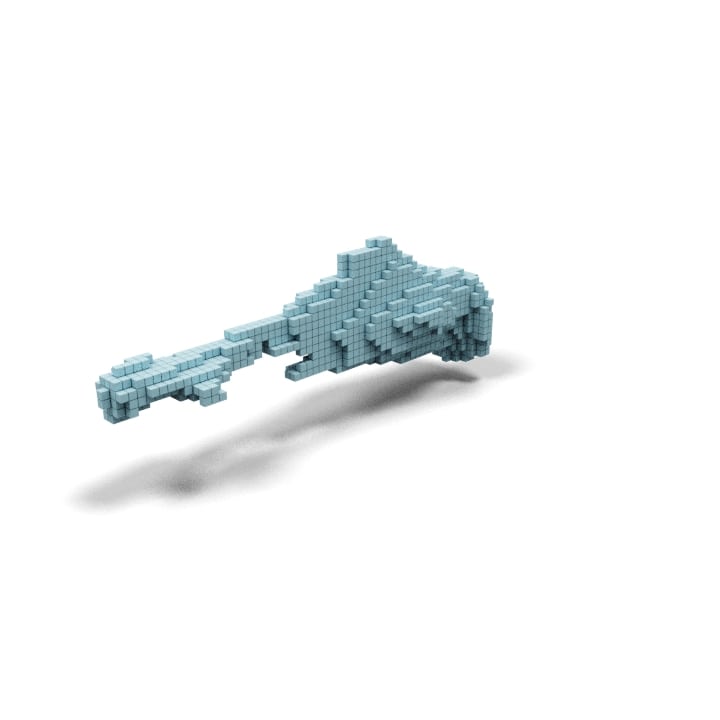}\\
``an automobile'' & ``a motor car'' & ``a seat'' & ``a digital display'' & ``a screen'' & ``a weapon''\\
\includegraphics[width=0.15\linewidth]{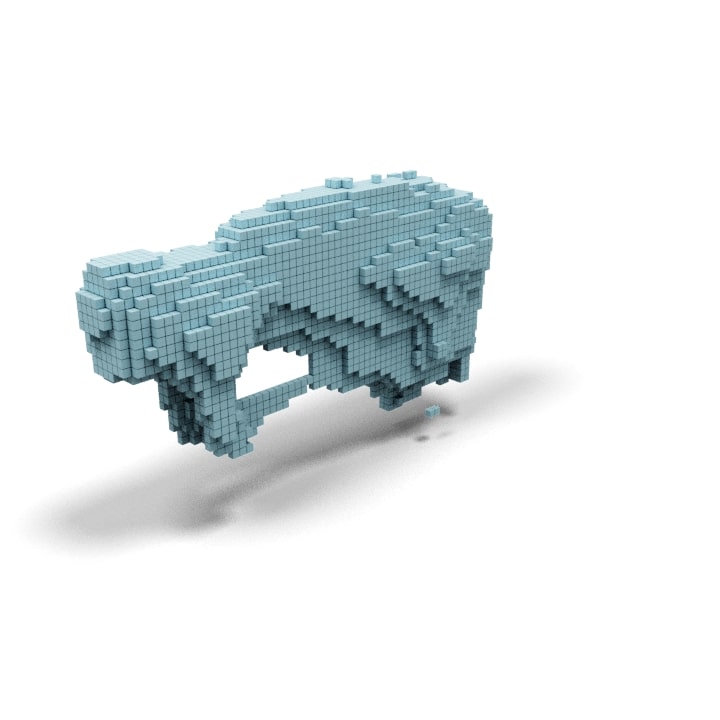} &
\includegraphics[width=0.15\linewidth]{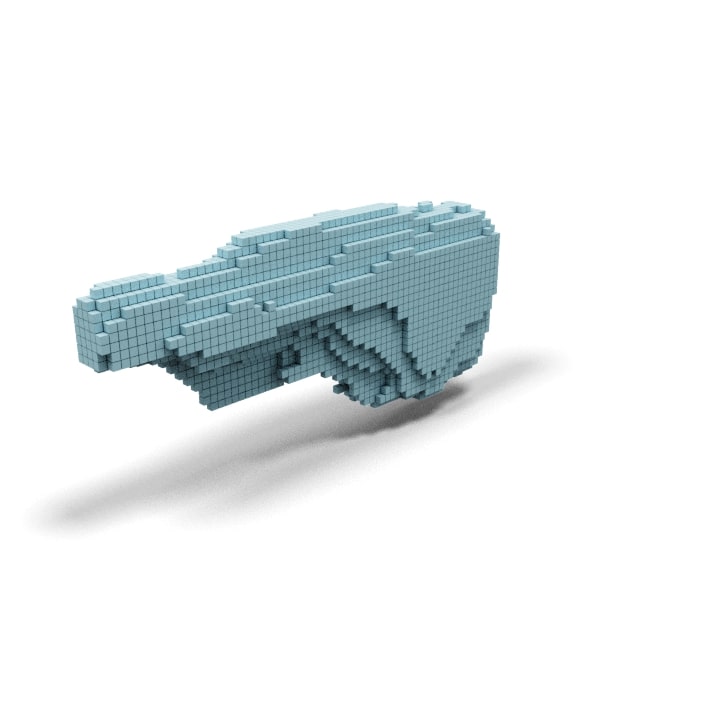} &
\includegraphics[width=0.15\linewidth]{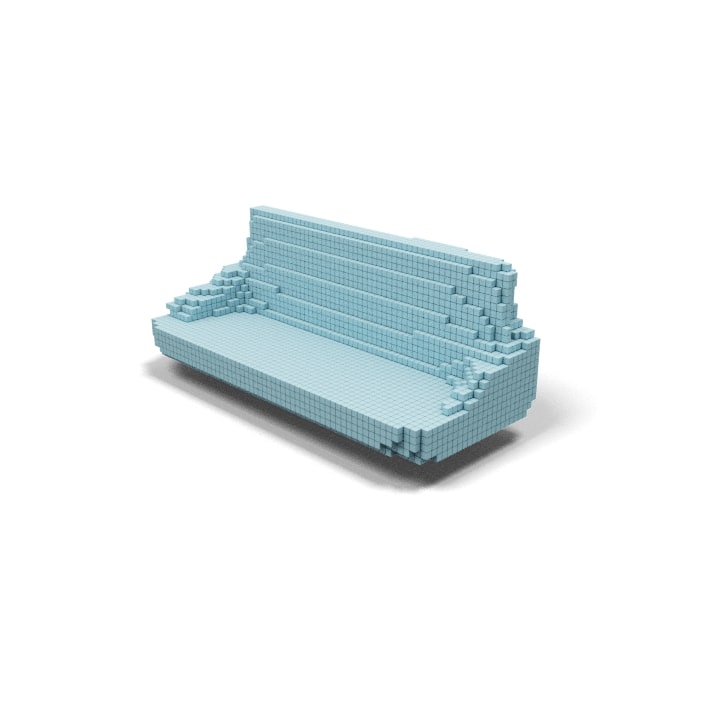} &
\includegraphics[width=0.15\linewidth]{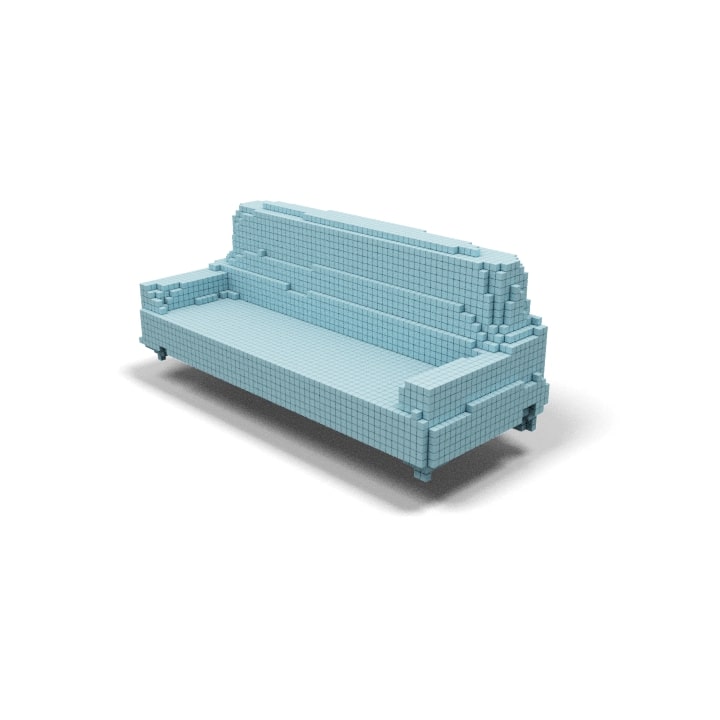} &
\includegraphics[width=0.15\linewidth]{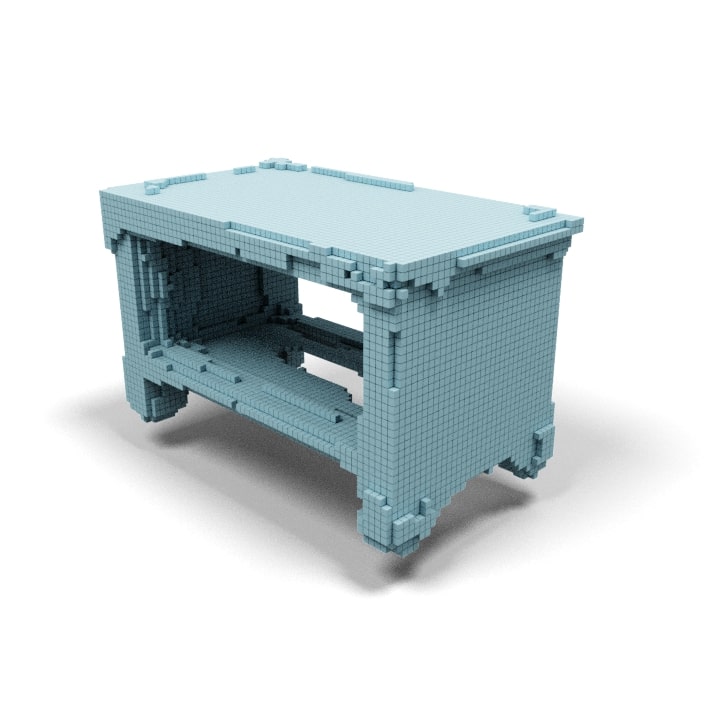} &
\includegraphics[width=0.15\linewidth]{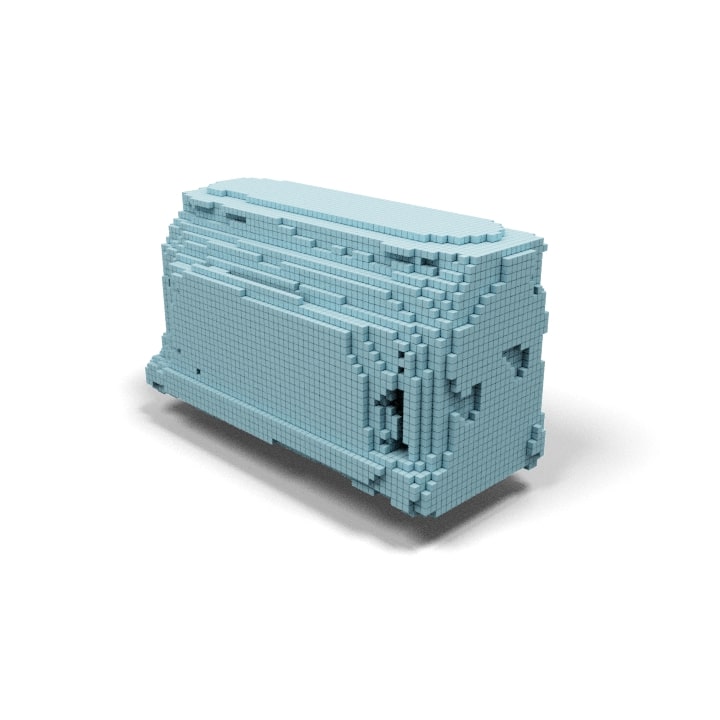}\\
``a shooter'' & ``a firearm'' & ``a lounge'' & ``a couch'' & ``an altar table'' & ``a workbench''\\
\includegraphics[width=0.15\linewidth]{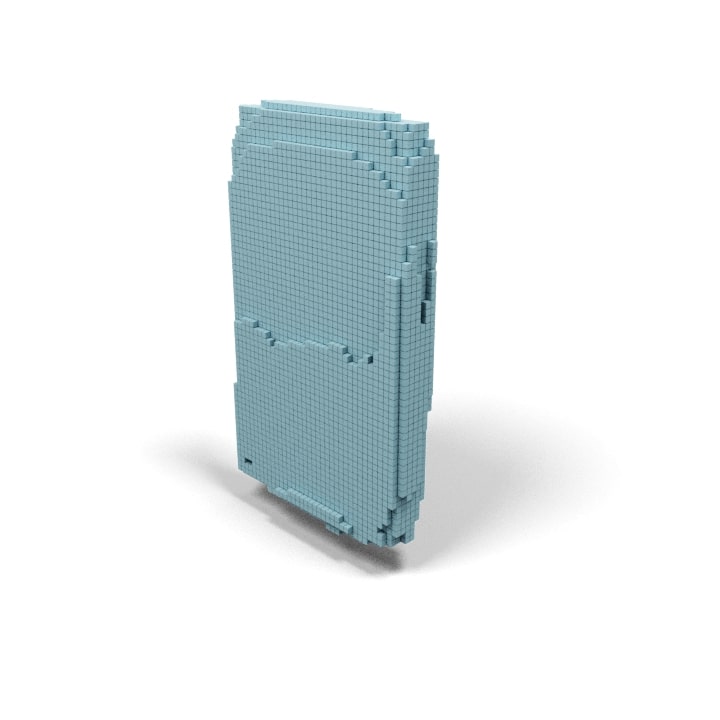} &
\includegraphics[width=0.15\linewidth]{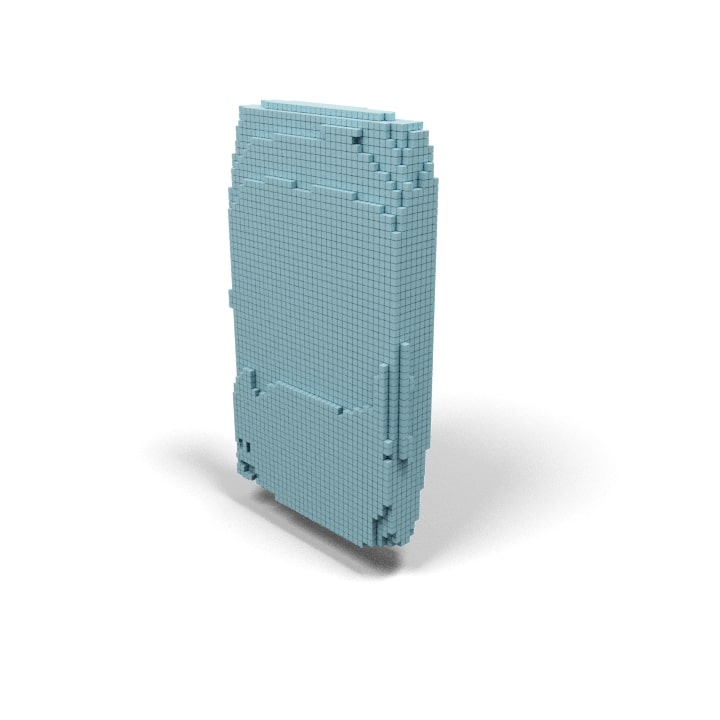} &
\includegraphics[width=0.15\linewidth]{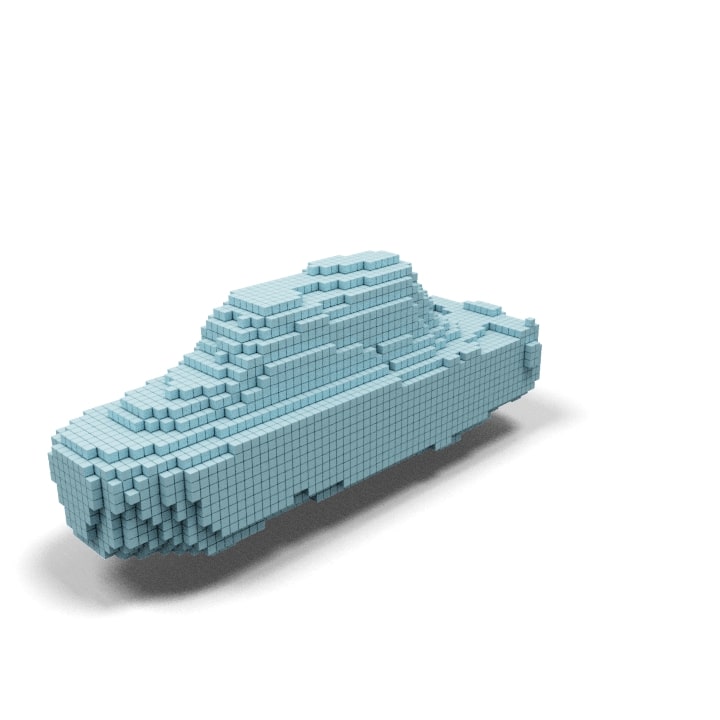} &
\includegraphics[width=0.15\linewidth]{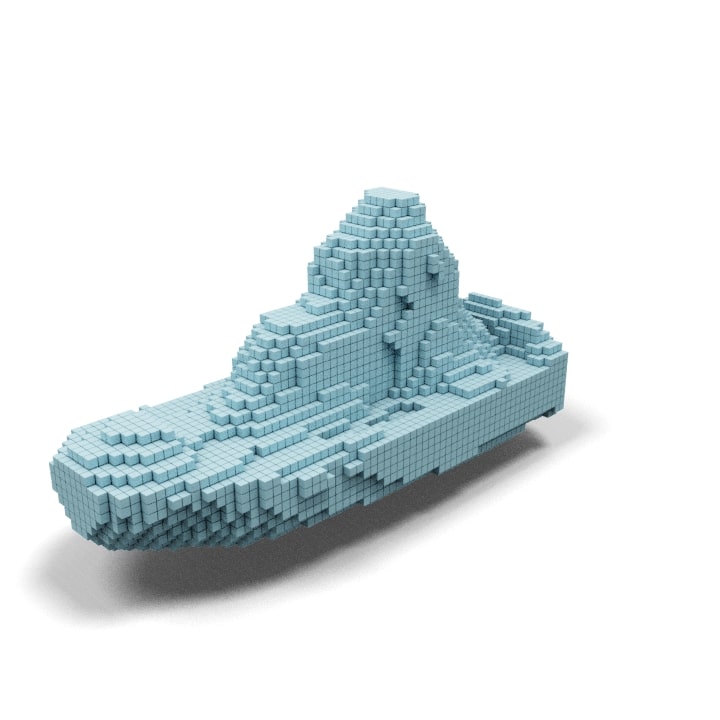} &
\includegraphics[width=0.15\linewidth]{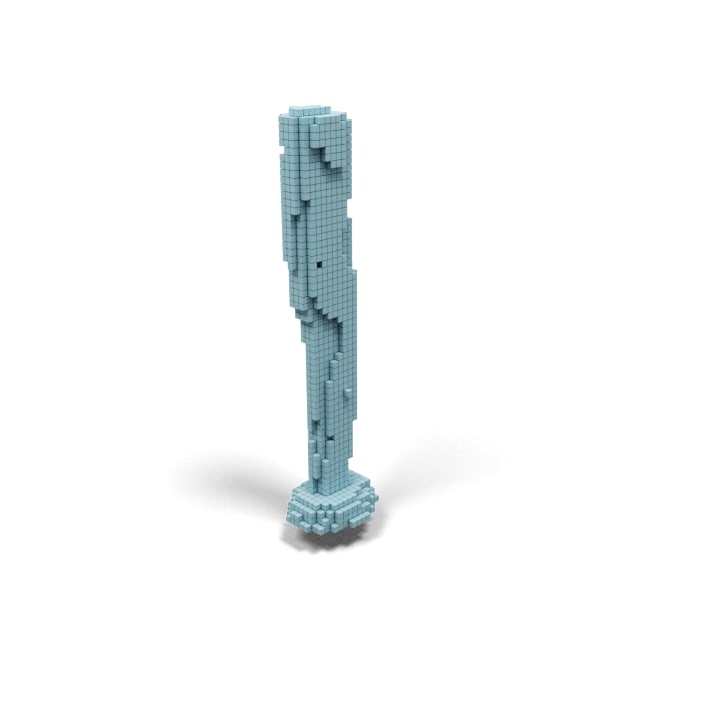} &
\includegraphics[width=0.15\linewidth]{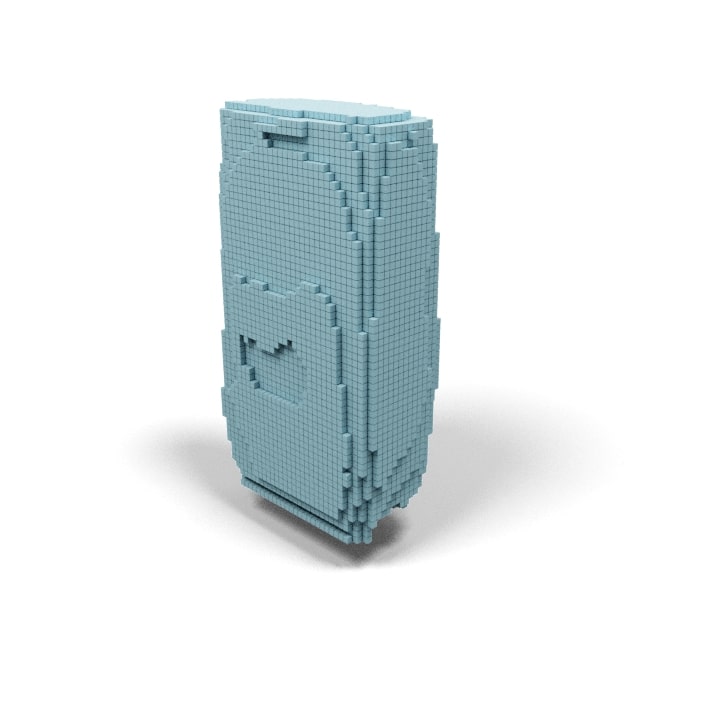}\\
``a cellphone'' & ``a cellular phone'' & ``a ferry'' & ``a ship'' & ``a torch'' & `a tannoy''\\
\end{tabular}
}
\end{center}
  \caption{Additional shape generation results using category and synonyms based text queries of CLIP-Forge.}
\label{fig:category_pics}
\end{figure*}

\begin{figure*}[t!]
\begin{center}
\setlength{\tabcolsep}{2pt}
\small{
\begin{tabular}{cccccc}
\includegraphics[width=0.15\linewidth]{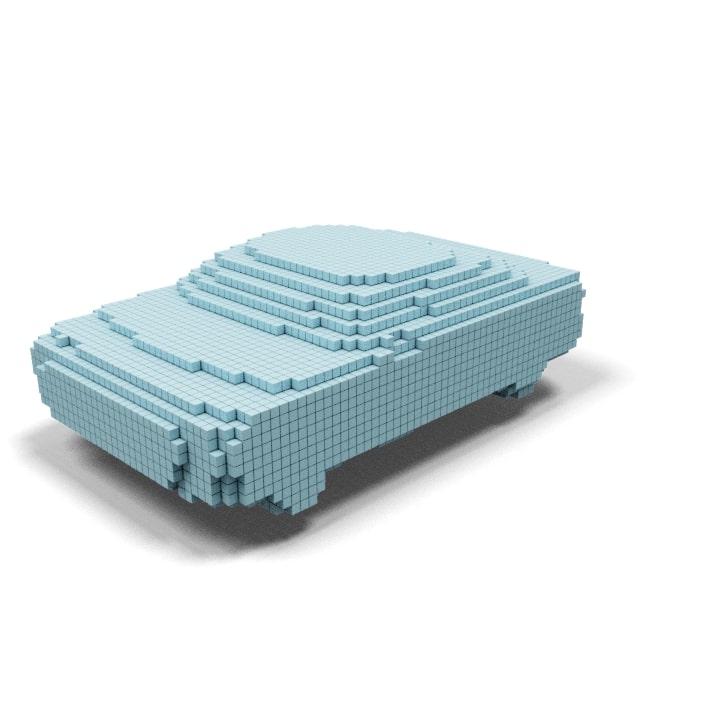} &
\includegraphics[width=0.15\linewidth]{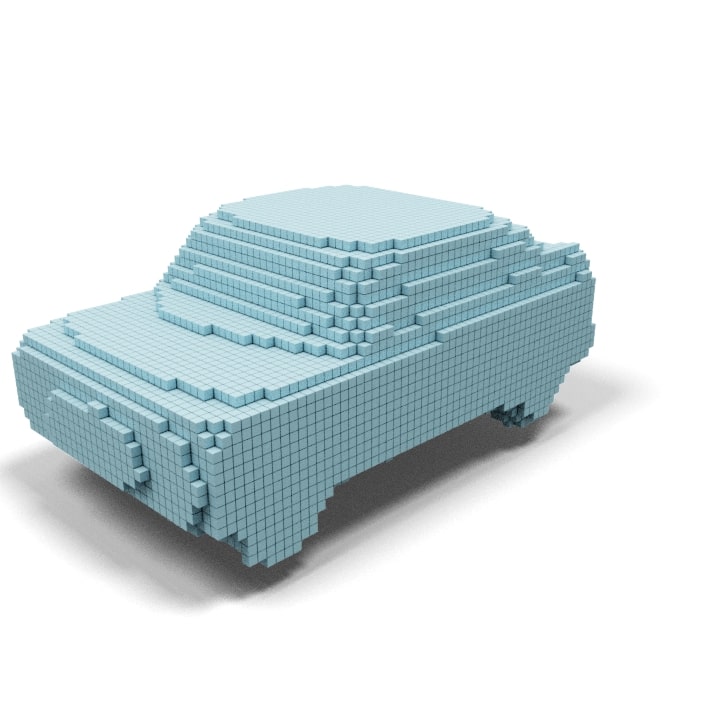} &
\includegraphics[width=0.15\linewidth]{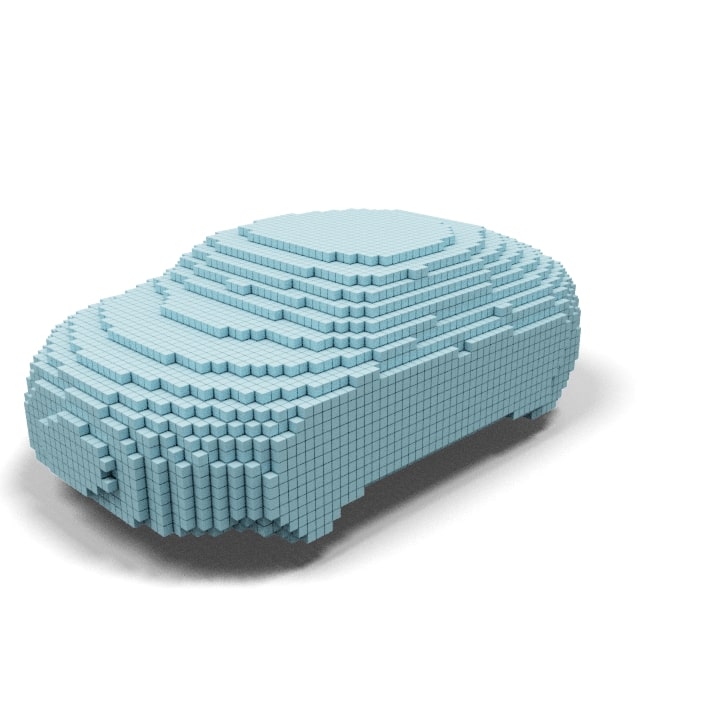} &
\includegraphics[width=0.15\linewidth]{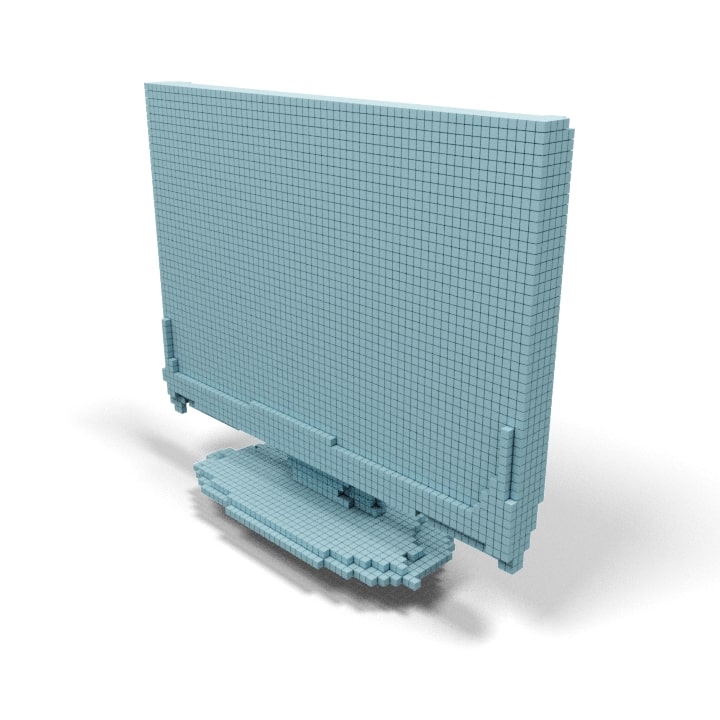} &
\includegraphics[width=0.15\linewidth]{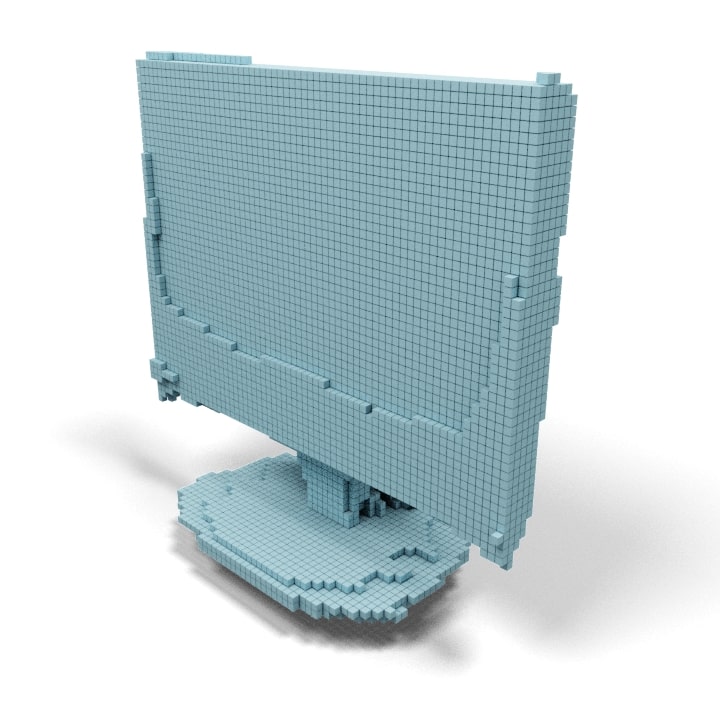} &
\includegraphics[width=0.15\linewidth]{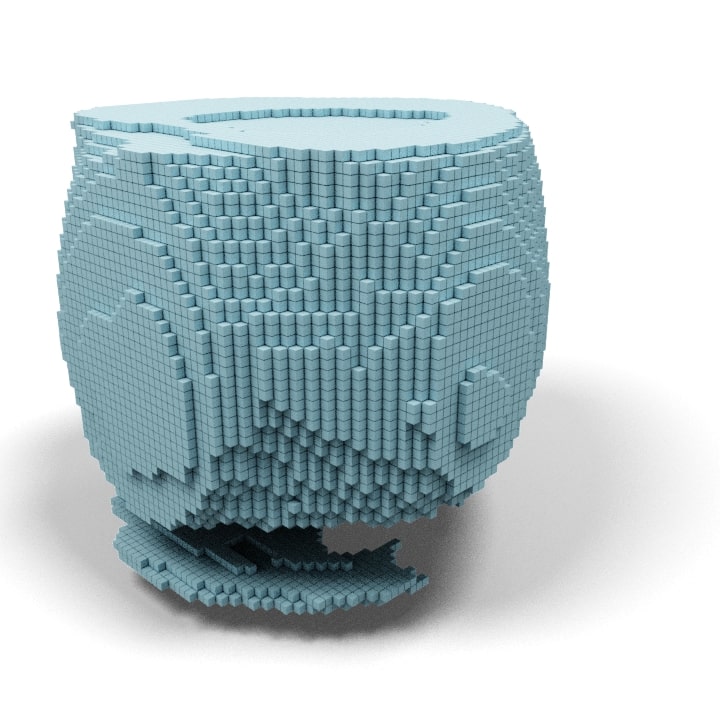} \\
``a  rectangular car'' & ``a square car'' & ``a round car'' & ``a rectangular monitor'' & ``a square monitor'' & ``a round monitor''\\
\includegraphics[width=0.15\linewidth]{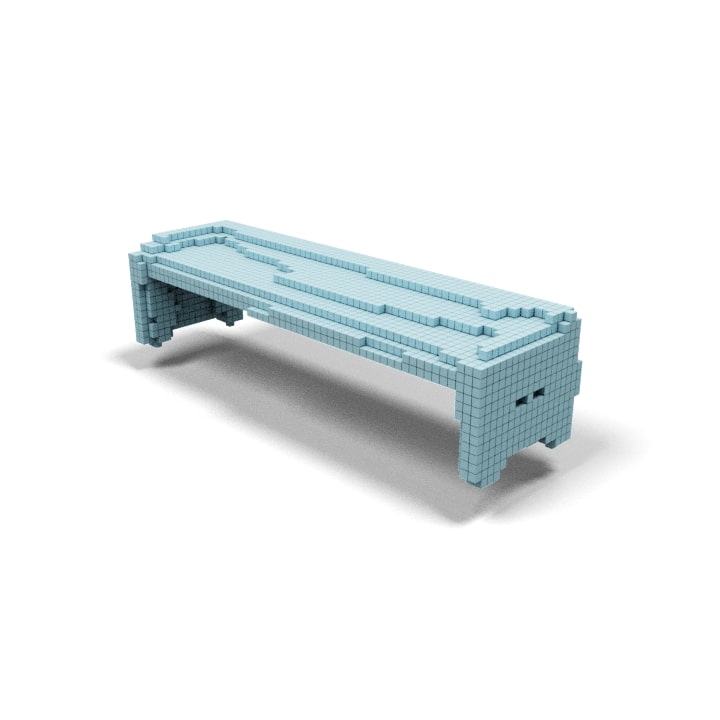} &
\includegraphics[width=0.15\linewidth]{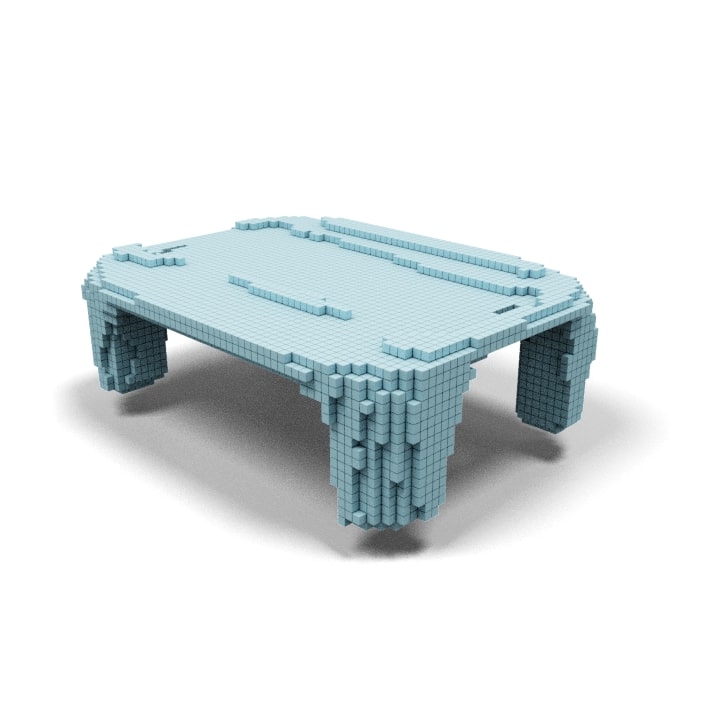} &
\includegraphics[width=0.15\linewidth]{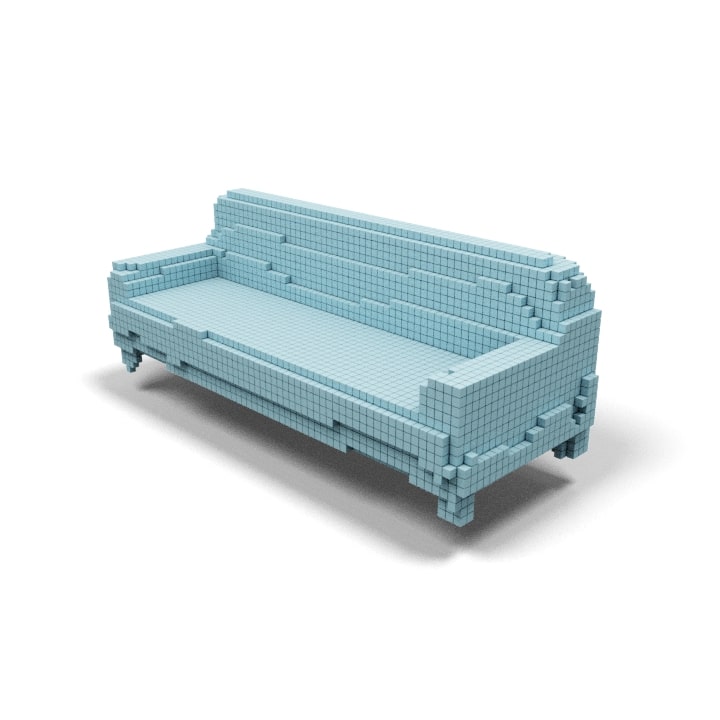} &
\includegraphics[width=0.15\linewidth]{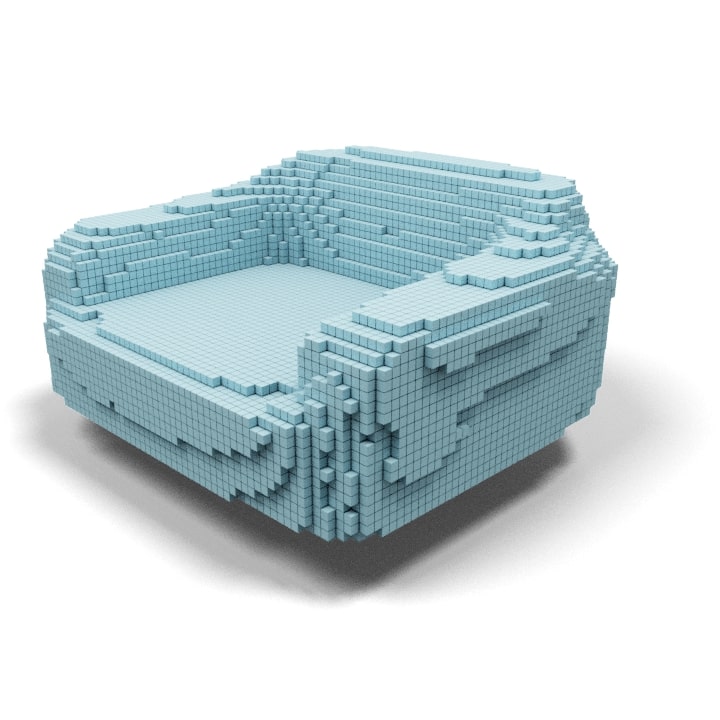} &
\includegraphics[width=0.15\linewidth]{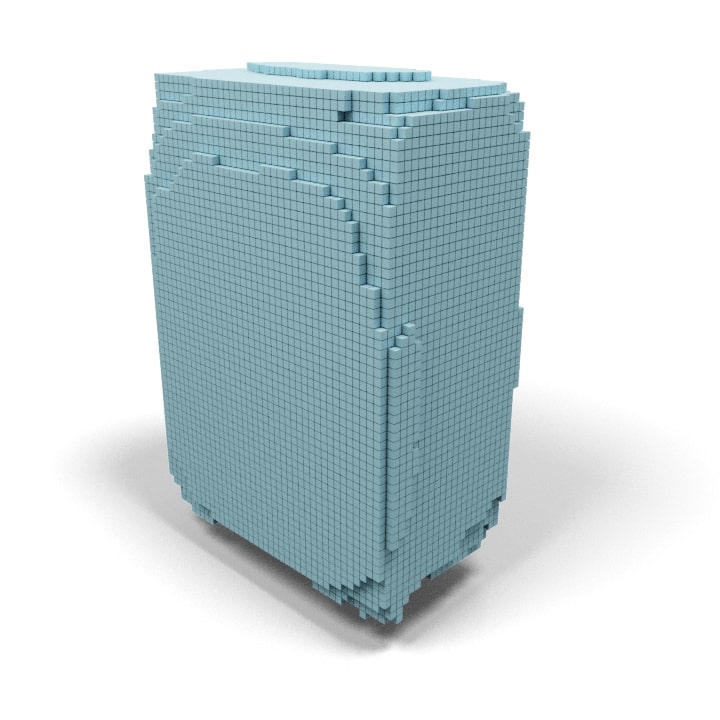} &
\includegraphics[width=0.15\linewidth]{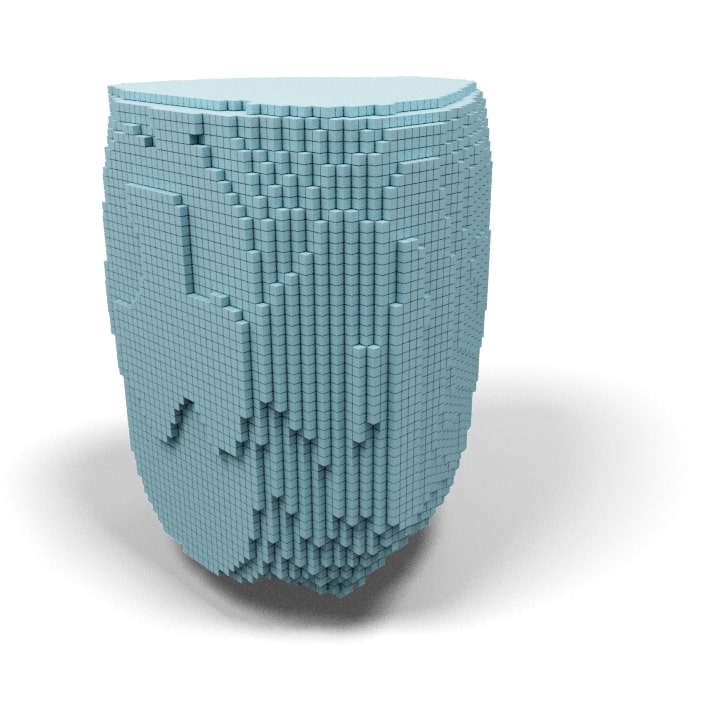}\\
``a rectangular bench'' & ``a circular bench'' & ``a rectangular sofa'' & ``a round sofa'' & ``a square loudspeaker'' & ``a round loudspeaker''\\
\end{tabular}
}
\end{center}
  \caption{Additional shape generation results using attribute-based text queries of CLIP-Forge.}
\label{fig:attributes}
\end{figure*}

\begin{figure*}[t!]
\begin{center}
\setlength{\tabcolsep}{2pt}
\small{
\begin{tabular}{cccccc}
\includegraphics[width=0.15\linewidth]{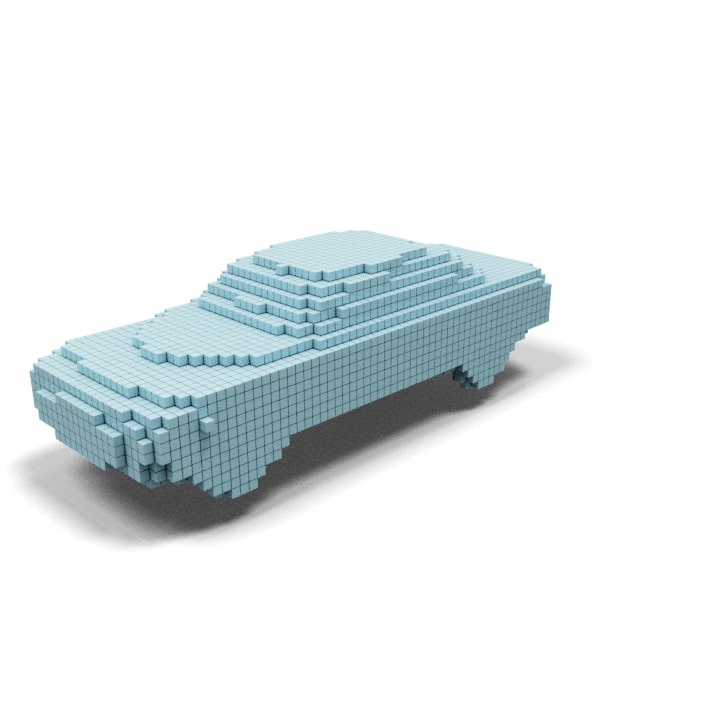} &
\includegraphics[width=0.15\linewidth]{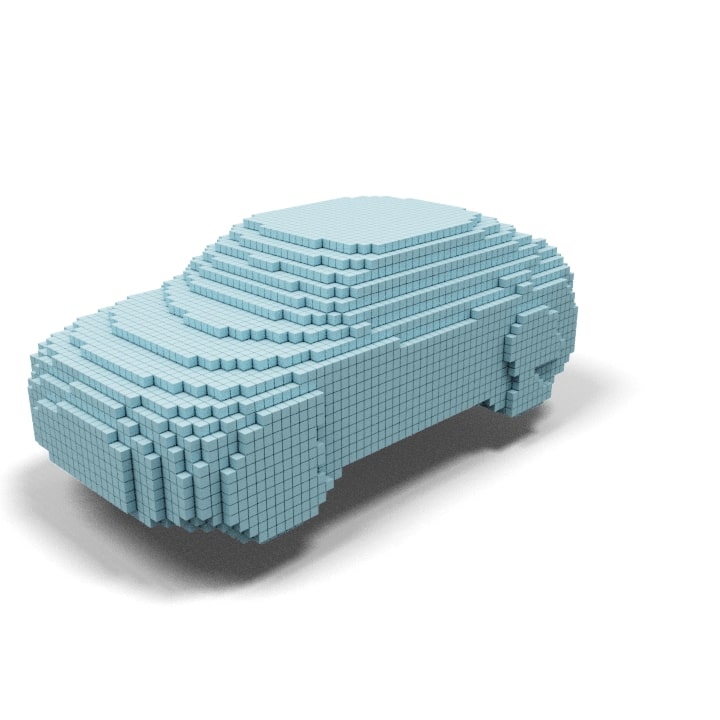} &
\includegraphics[width=0.15\linewidth]{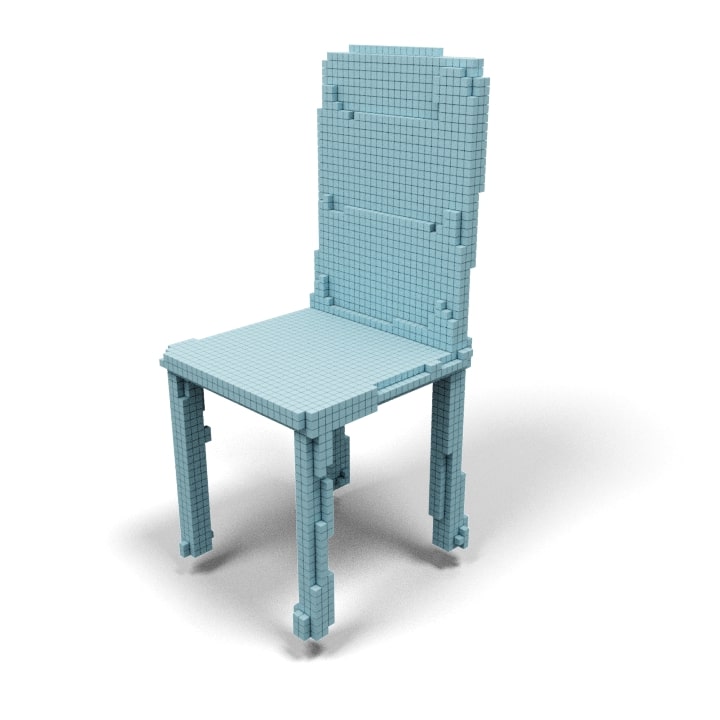} &
\includegraphics[width=0.15\linewidth]{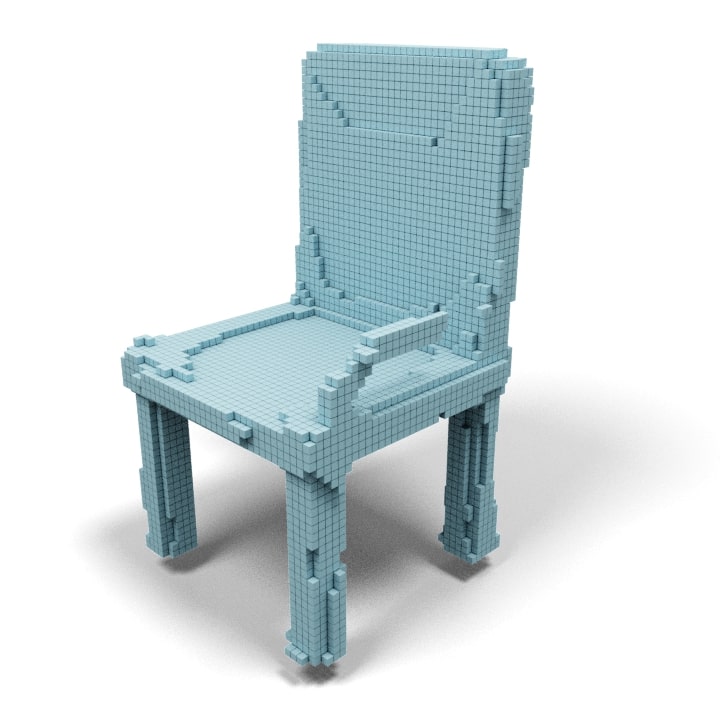} &
\includegraphics[width=0.15\linewidth]{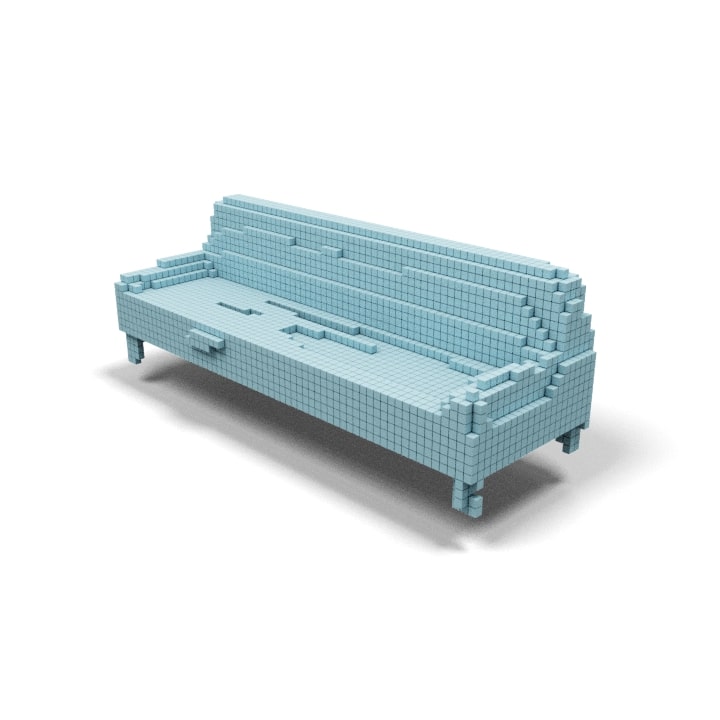} &
\includegraphics[width=0.15\linewidth]{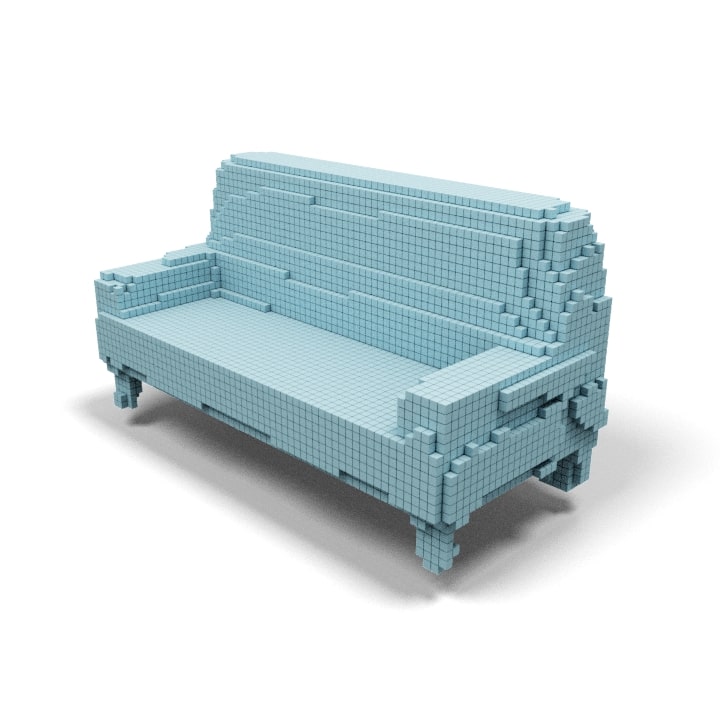}\\
``a thin car'' & ``a thick car'' & ``a thin chair'' & ``a  thick chair'' & ``a thin sofa'' & ``a thick sofa''\\
\includegraphics[width=0.15\linewidth]{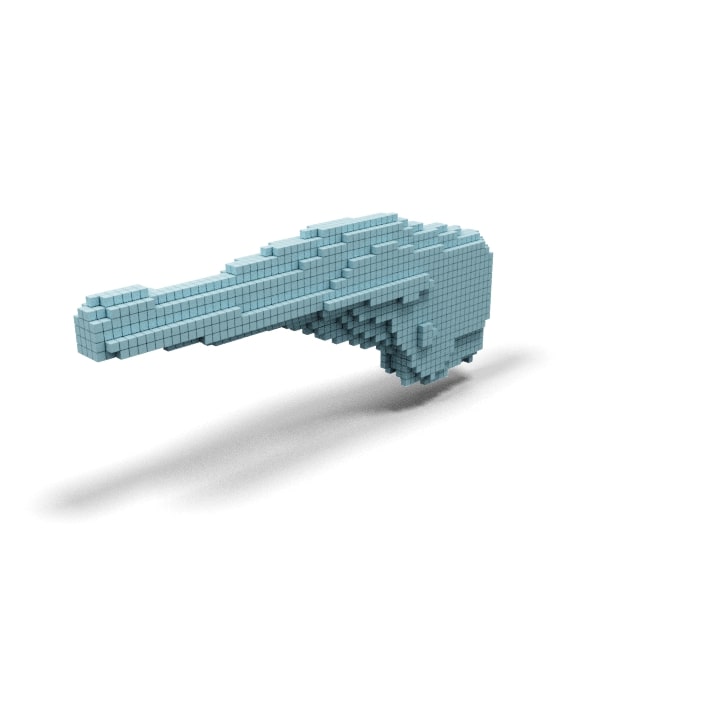} &
\includegraphics[width=0.15\linewidth]{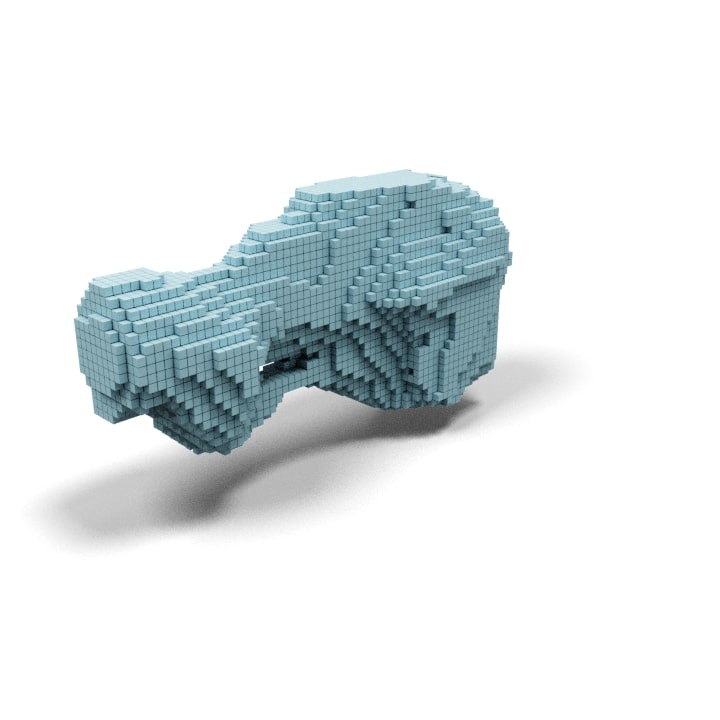} &
\includegraphics[width=0.15\linewidth]{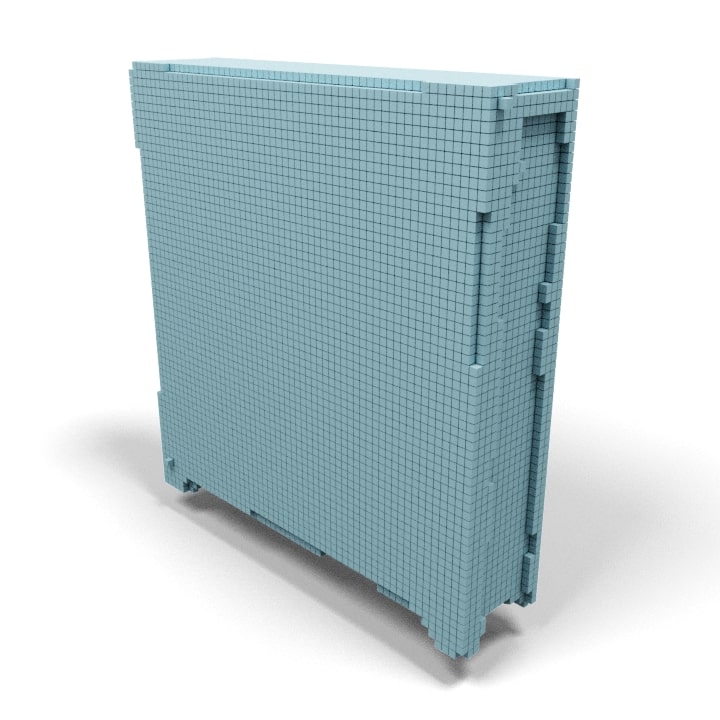} &
\includegraphics[width=0.15\linewidth]{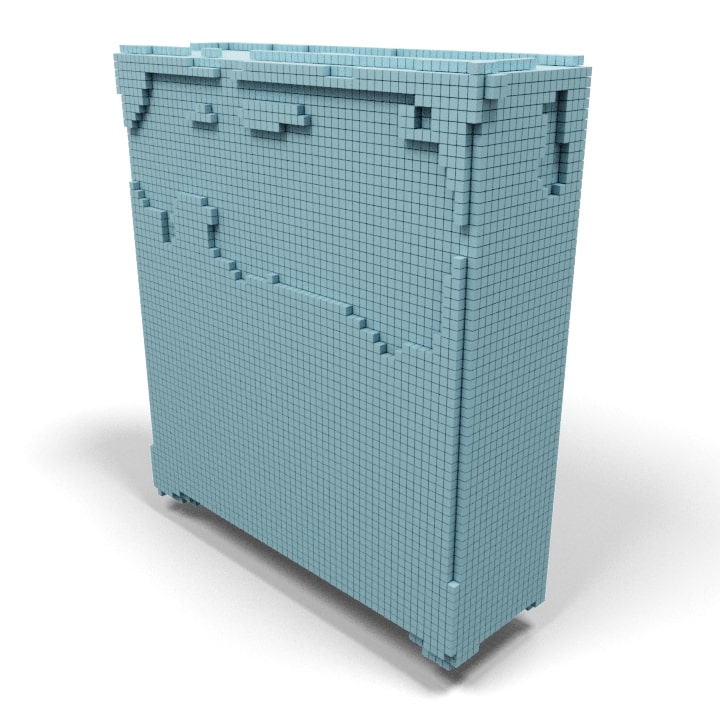} &
\includegraphics[width=0.15\linewidth]{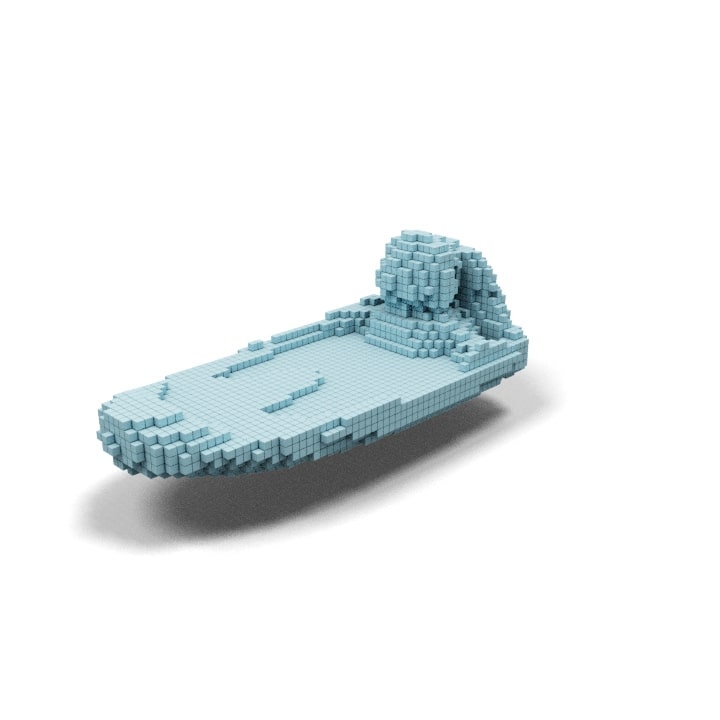} &
\includegraphics[width=0.15\linewidth]{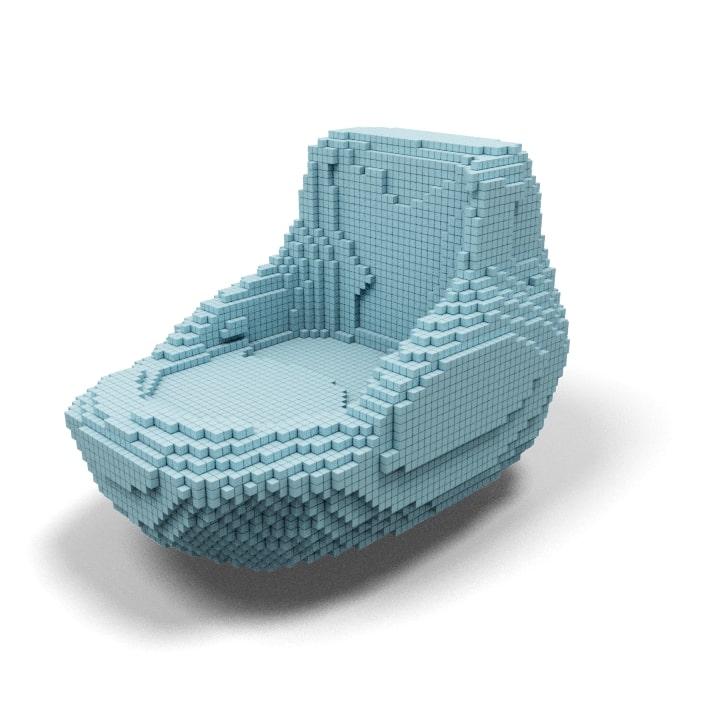}\\
``a thin gun'' & ``a thick gun'' & ``a thin cabinet'' & ``a thick cabinet'' & ``a thin boat'' & ``a thick boat''\\
\end{tabular}
}
\end{center}
  \caption{Additional shape generation results using attribute-based text queries of CLIP-Forge (continued).}
\label{fig:attributes_cont}
\end{figure*}

\begin{figure*}[t!]
\begin{center}
\setlength{\tabcolsep}{2pt}
\small{
\begin{tabular}{cccccc}
\includegraphics[width=0.15\linewidth]{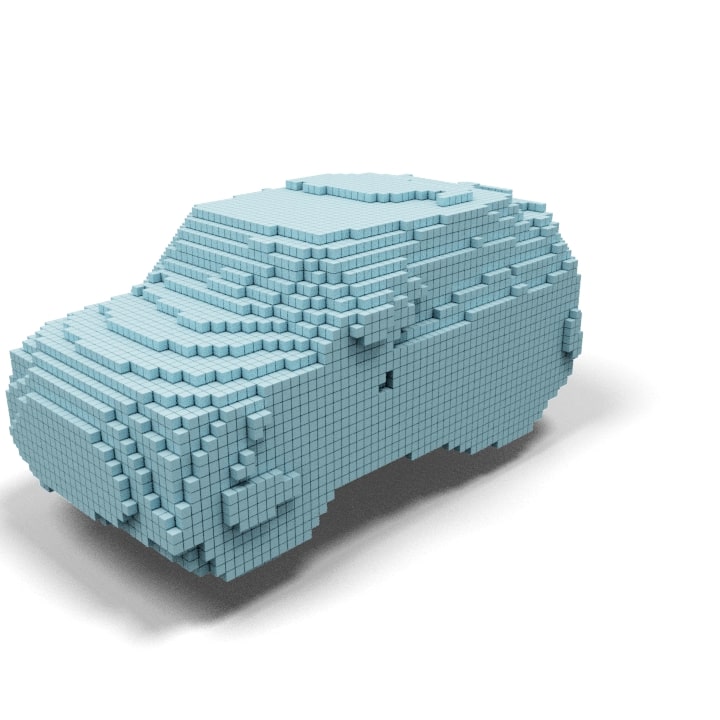} &
\includegraphics[width=0.15\linewidth]{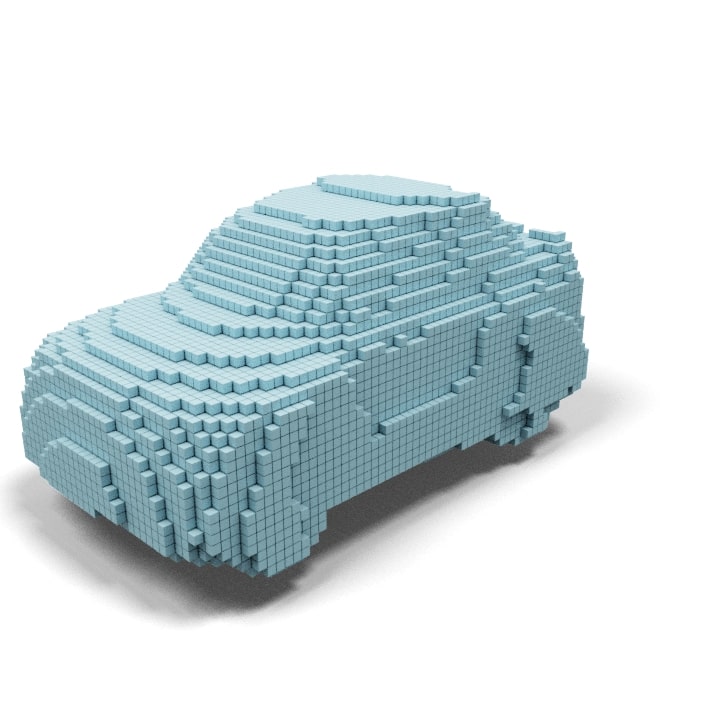} &
\includegraphics[width=0.15\linewidth]{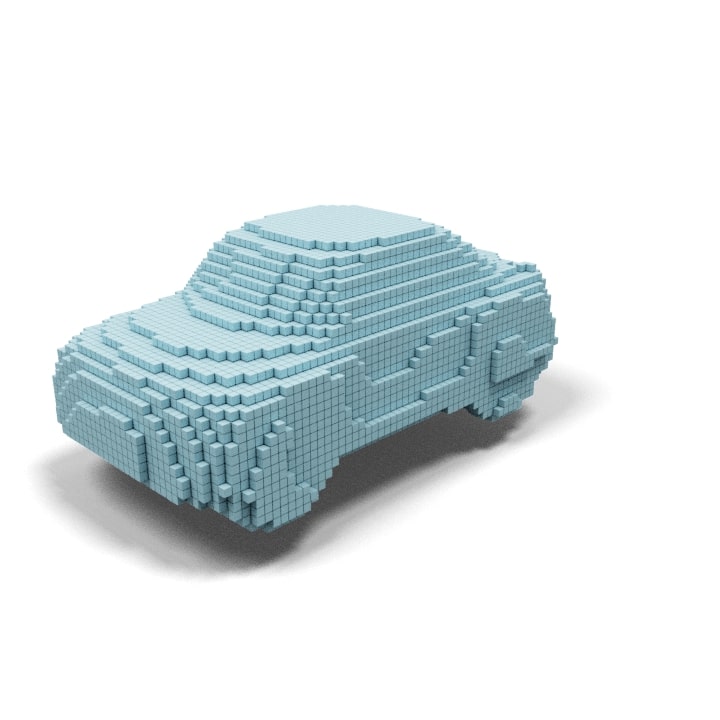} &
\includegraphics[width=0.15\linewidth]{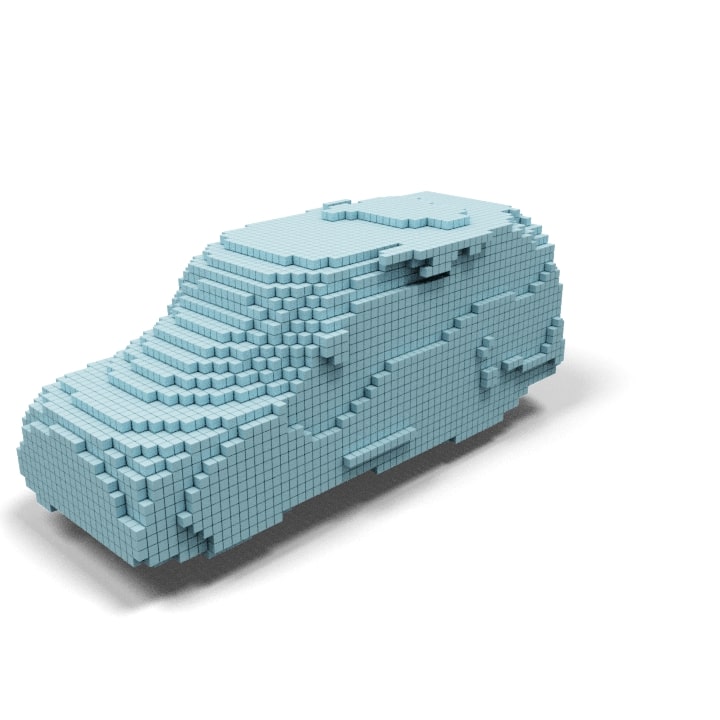} &
\includegraphics[width=0.15\linewidth]{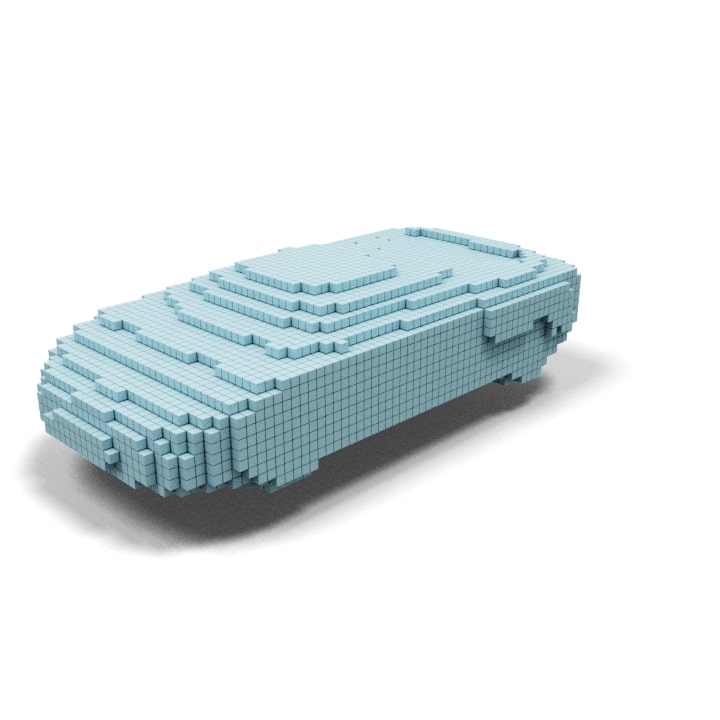} &
\includegraphics[width=0.15\linewidth]{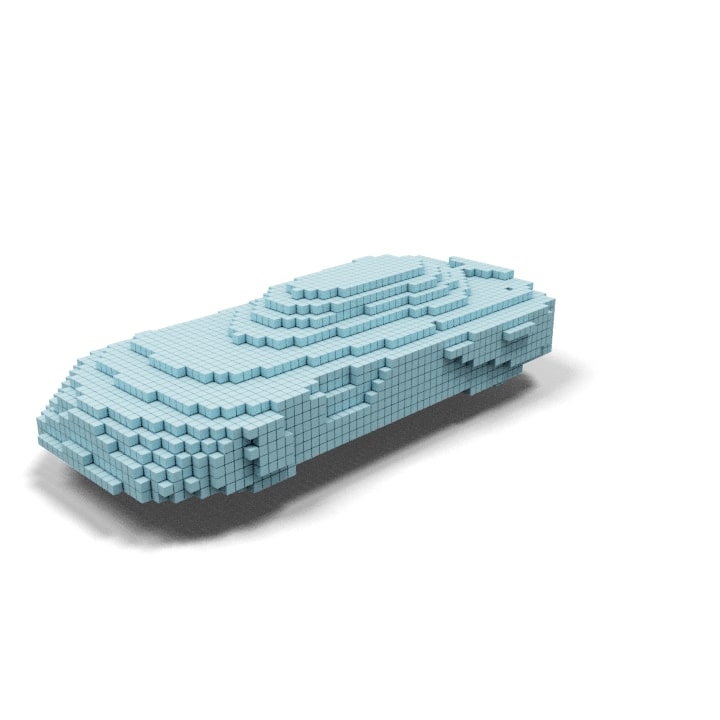} \\
``an auto'' & ``a honda'' & ``a toyota'' & ``an ambulance'' & ``a lamborghini'' & ``a tesla''\\
\includegraphics[width=0.15\linewidth]{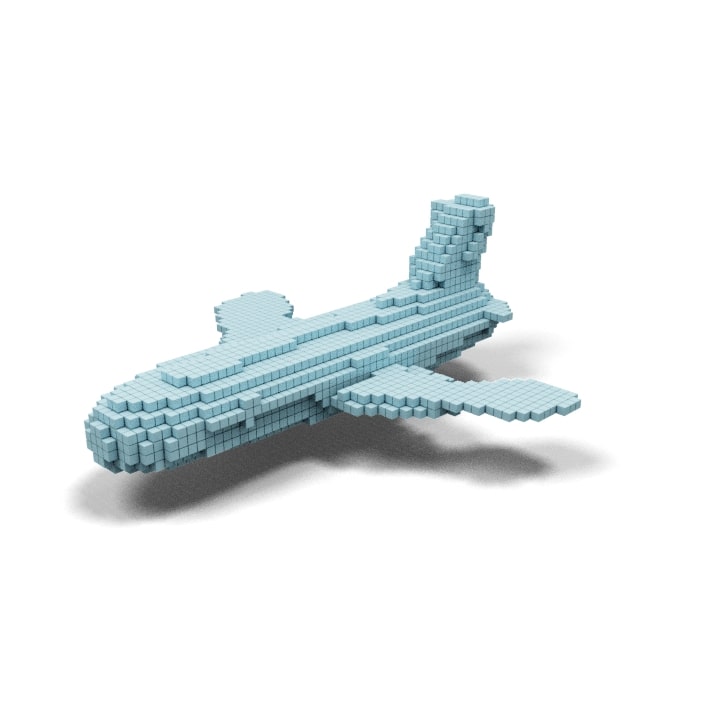} &
\includegraphics[width=0.15\linewidth]{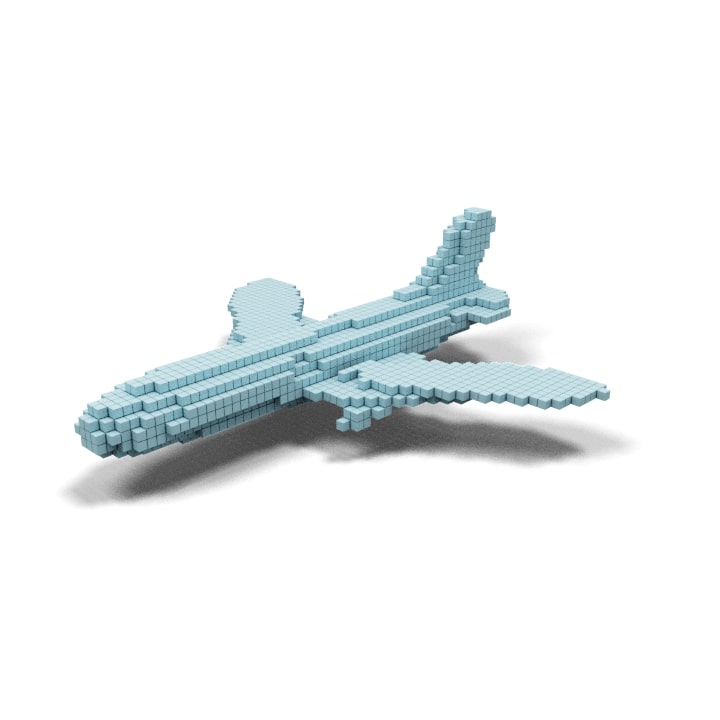} &
\includegraphics[width=0.15\linewidth]{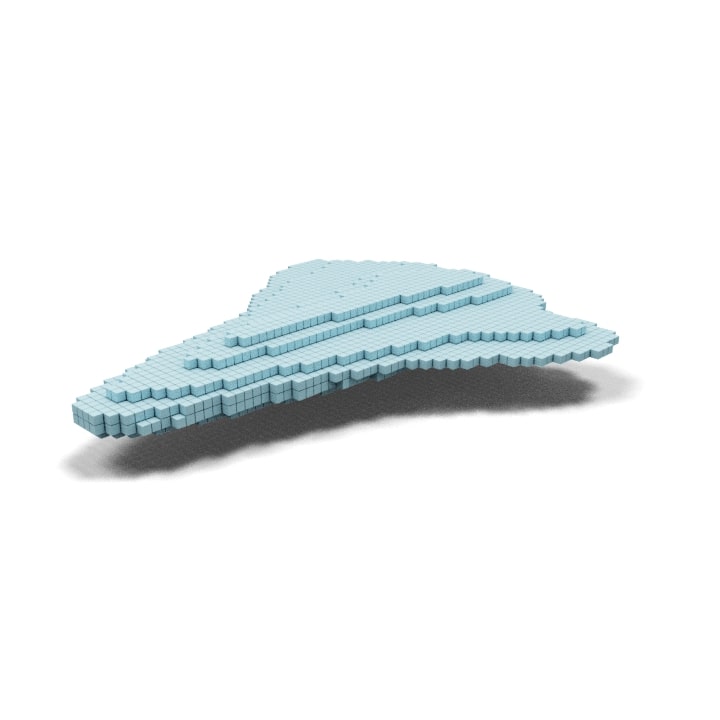} &
\includegraphics[width=0.15\linewidth]{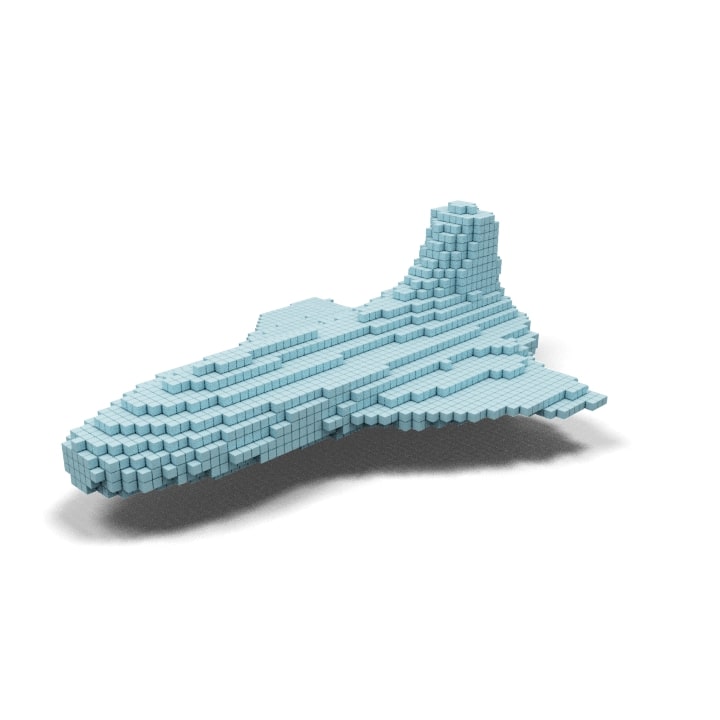} &
\includegraphics[width=0.15\linewidth]{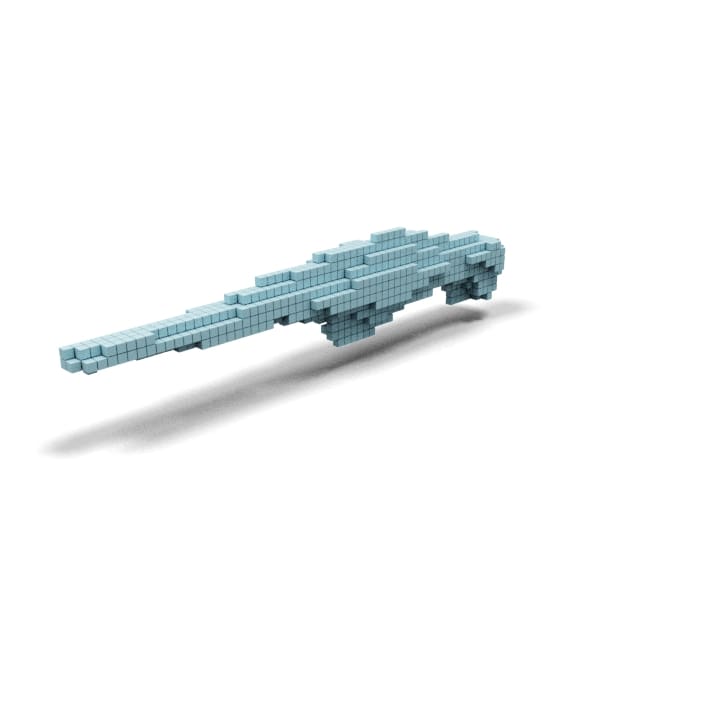} &
\includegraphics[width=0.15\linewidth]{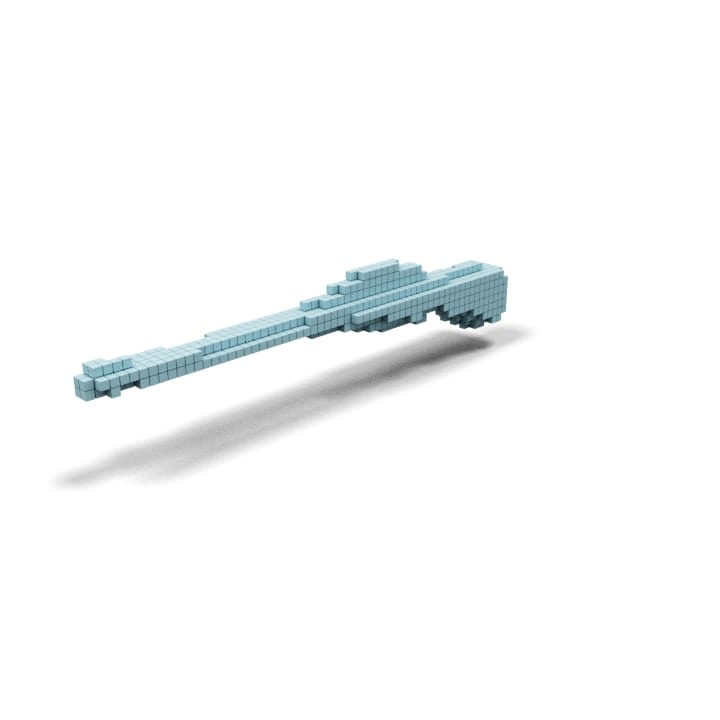} \\
``an Airbus A380plus'' & ``a boeing 747'' & ``a Lockheed SR-71 blackbird'' & ``a private jet'' & `a M-16'' & ``a M1 Garand''\\
\includegraphics[width=0.15\linewidth]{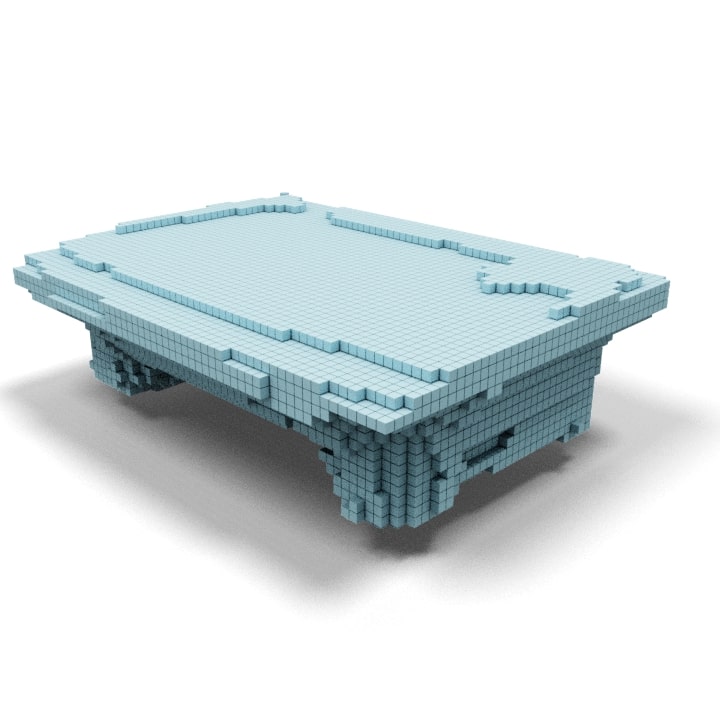} &
\includegraphics[width=0.15\linewidth]{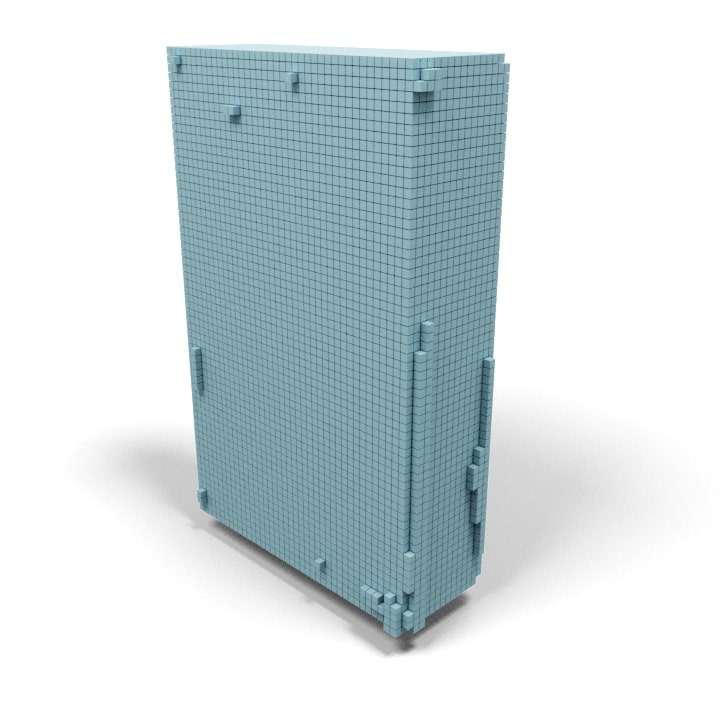} &
\includegraphics[width=0.15\linewidth]{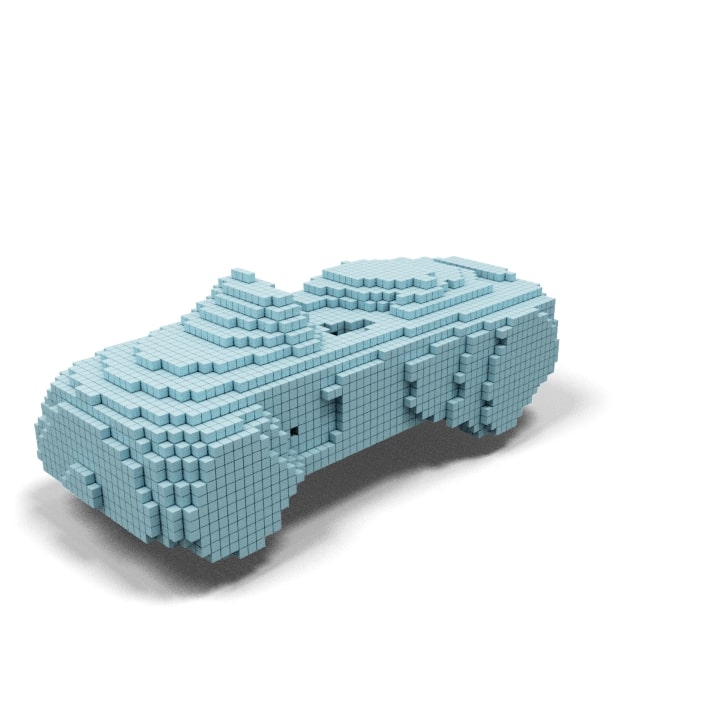} &
\includegraphics[width=0.15\linewidth]{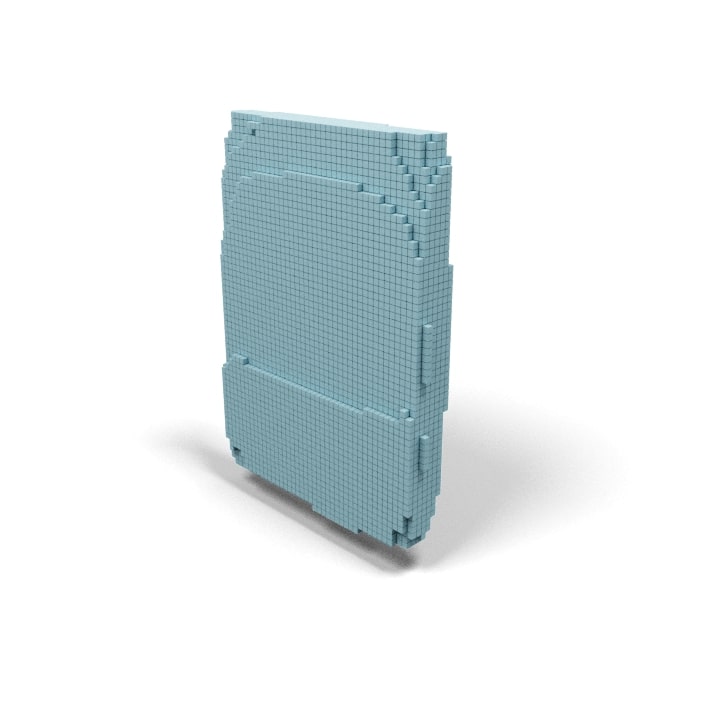} &
\includegraphics[width=0.15\linewidth]{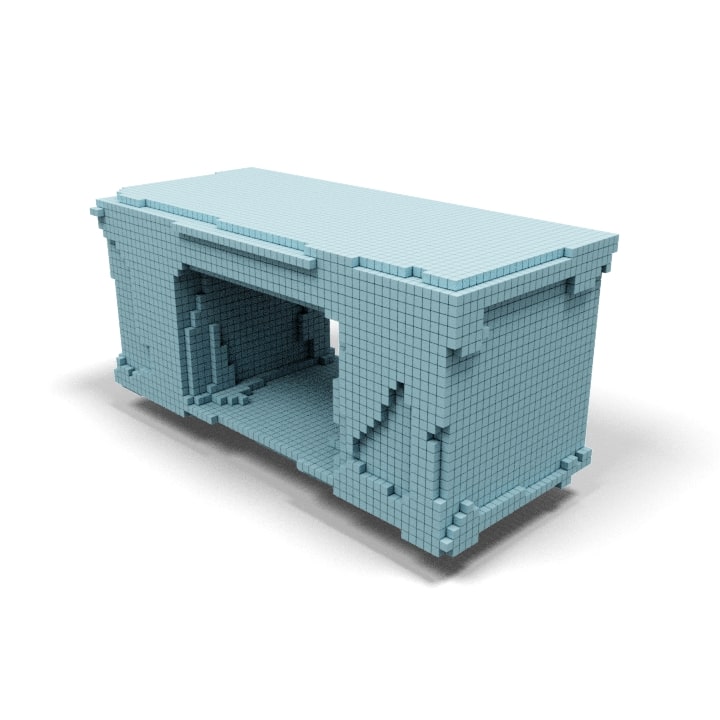} &
\includegraphics[width=0.15\linewidth]{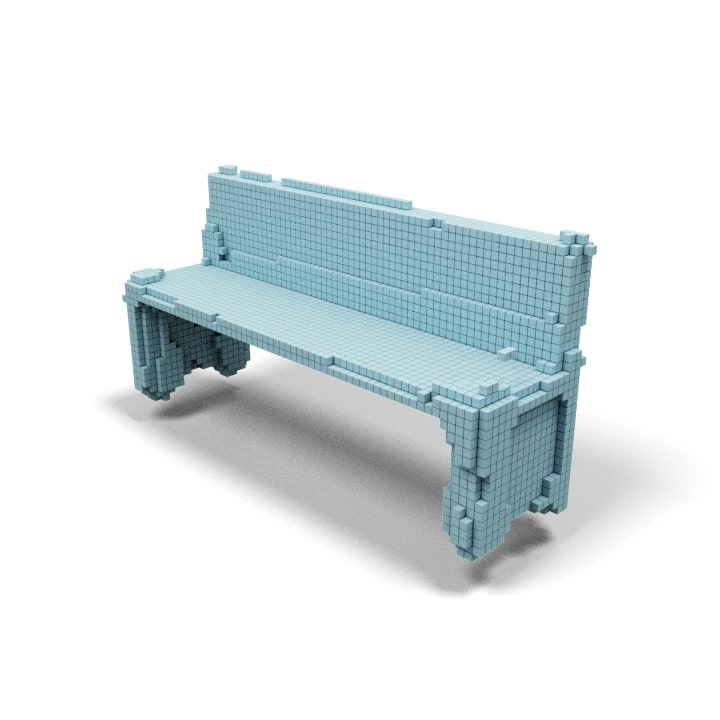}\\
``a billiard table'' & ``a locker'' & ``a gas guzzler''  & ``an iphone'' & ``a worktop'' &  ``a park bench''\\
\end{tabular}
}
\end{center}
  \caption{Additional shape generation results using common name text queries of CLIP-Forge.}
\label{fig:common_names}
\end{figure*}

In this part, we show more visual results for shape generation conditioned with text based on sub-category (Figure~\ref{fig:sub_category_pics} and Figure~\ref{fig:sub_category_pics_cont}), synonyms (Figure~\ref{fig:category_pics}), shape attributes (Figure~\ref{fig:attributes} and Figure~\ref{fig:attributes_cont}) and common names (Figure~\ref{fig:common_names}). Moreover, we also show more visuals for text based multiple shape generation (Figure~\ref{fig:qual_multiple}) and interpolation (Figure~\ref{fig:interp}). It can be seen from all these results that our method is good at generating 3D shapes based on text queries. However, in same cases for example ``a swivel chair'', it cannot construct all the details. Furthermore, on some sub-categories such as ``an operating table'' it cannot generate accurate shapes.

\begin{figure*}[t!]
\begin{center}
\setlength{\tabcolsep}{2pt}
\small{
\begin{tabular}{cccccc}
\includegraphics[width=0.15\linewidth]{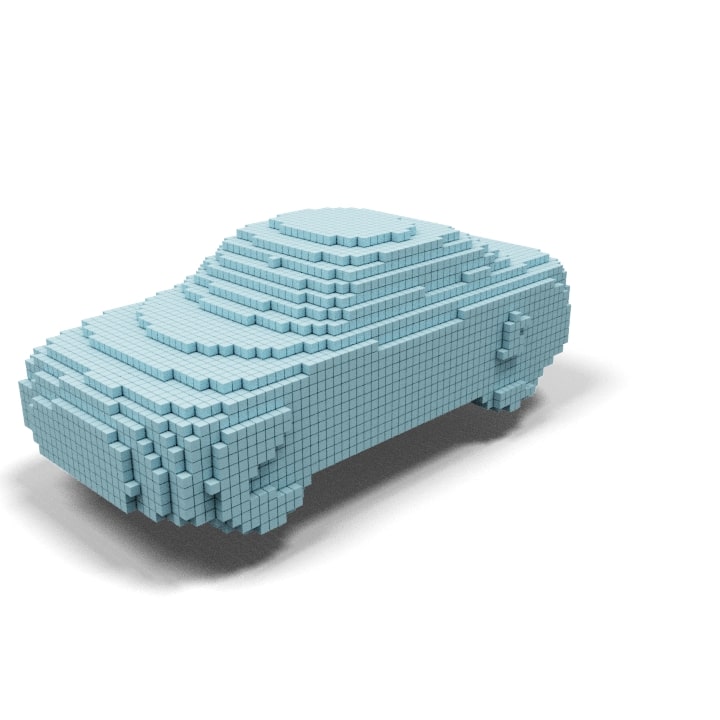} &
\includegraphics[width=0.15\linewidth]{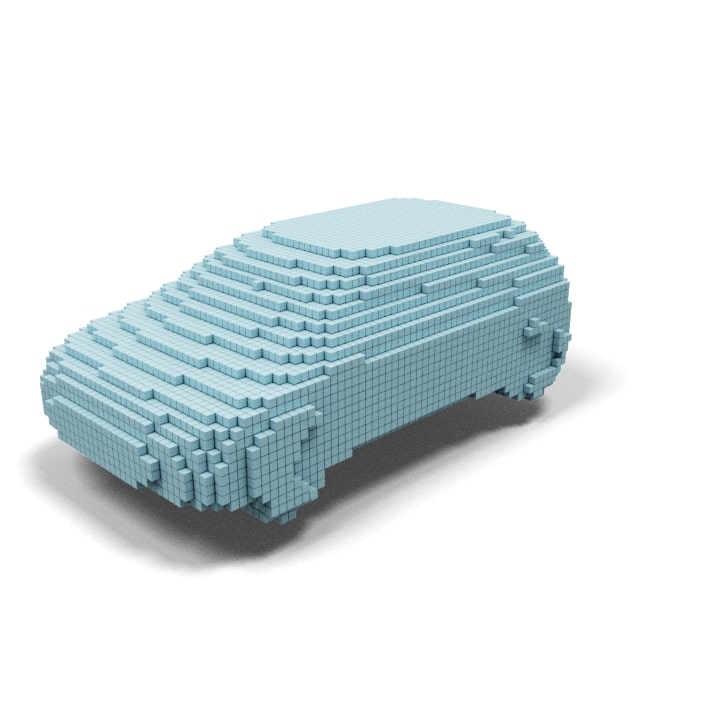} &
\includegraphics[width=0.15\linewidth]{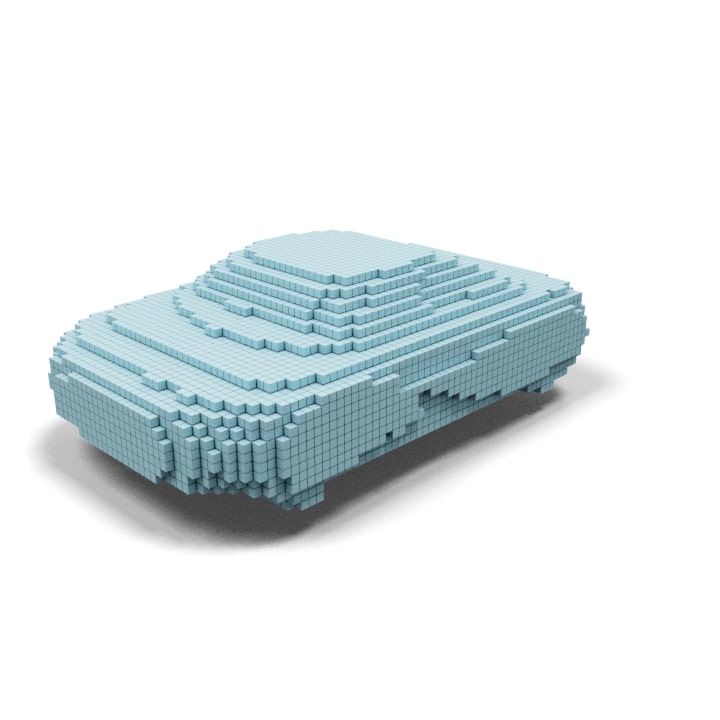} &
\includegraphics[width=0.15\linewidth]{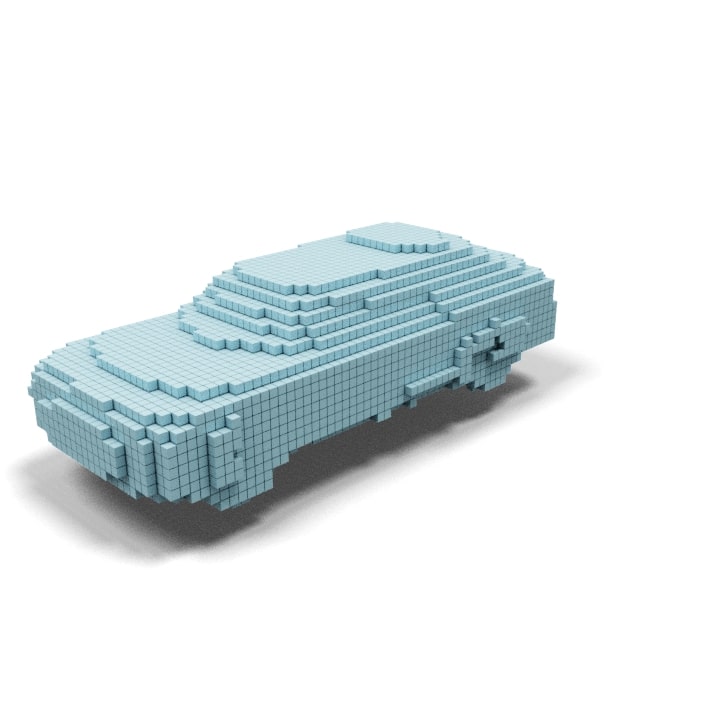} &
\includegraphics[width=0.15\linewidth]{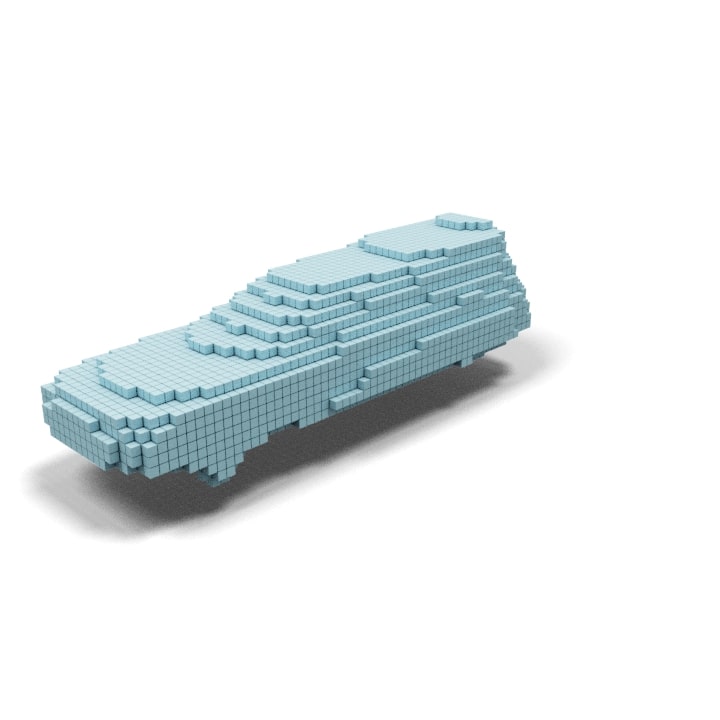} &
\includegraphics[width=0.15\linewidth]{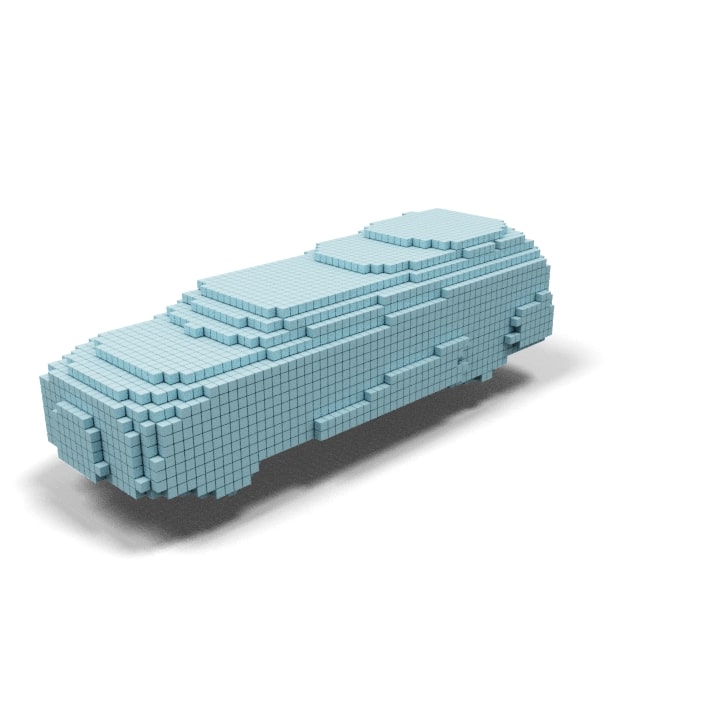}\\
\multicolumn{3}{c}{``a car''} & \multicolumn{3}{c}{``a limo''}\\
\includegraphics[width=0.15\linewidth]{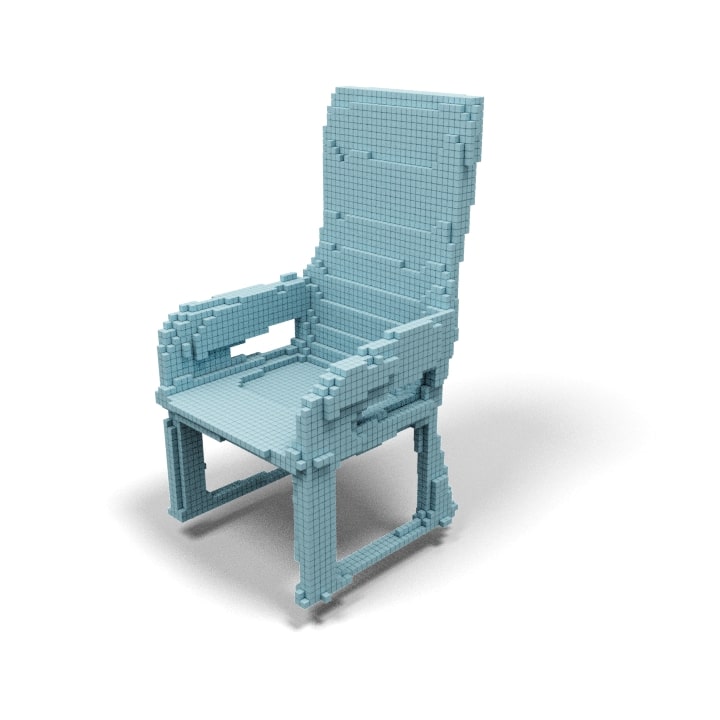} &
\includegraphics[width=0.15\linewidth]{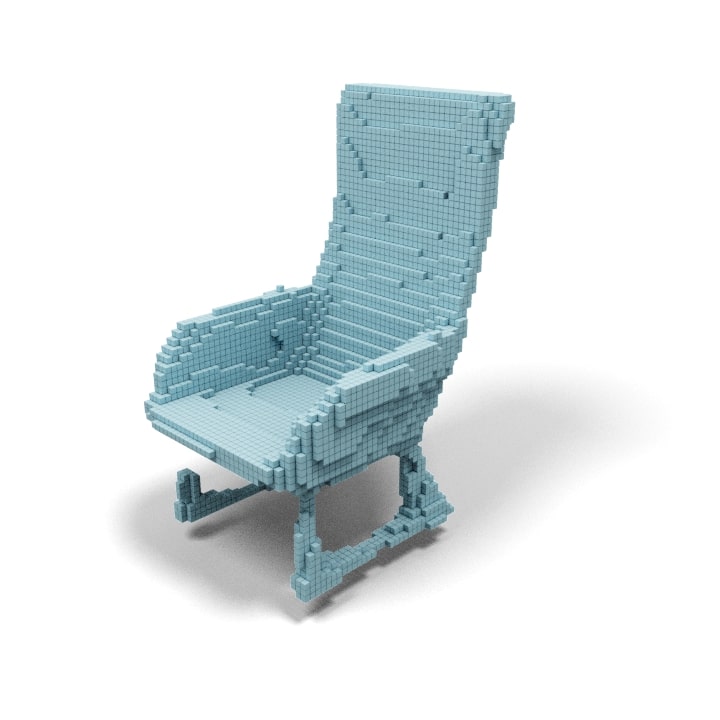} &
\includegraphics[width=0.15\linewidth]{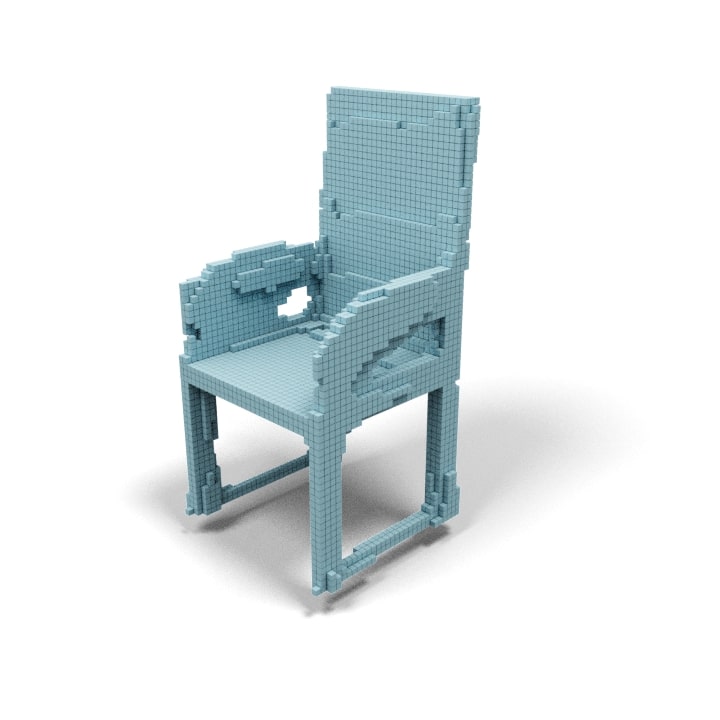} &
\includegraphics[width=0.15\linewidth]{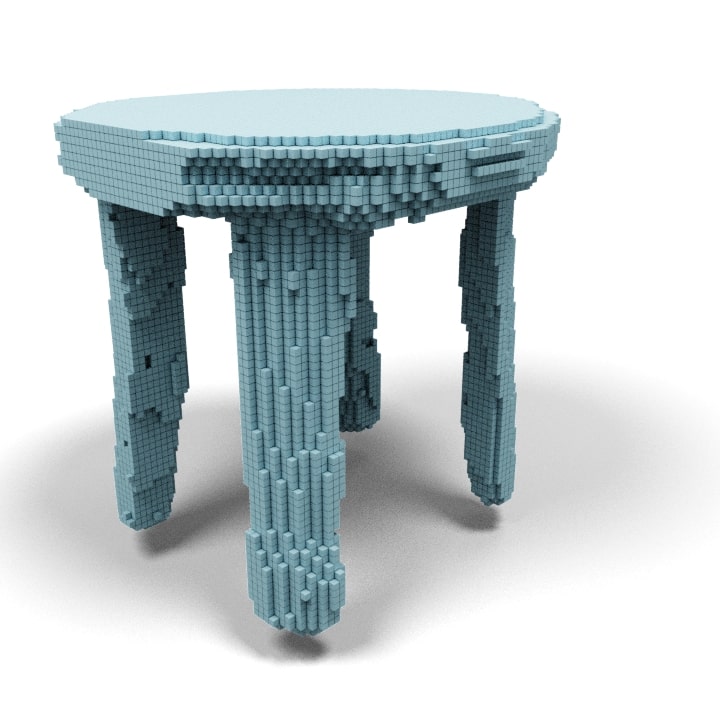} &
\includegraphics[width=0.15\linewidth]{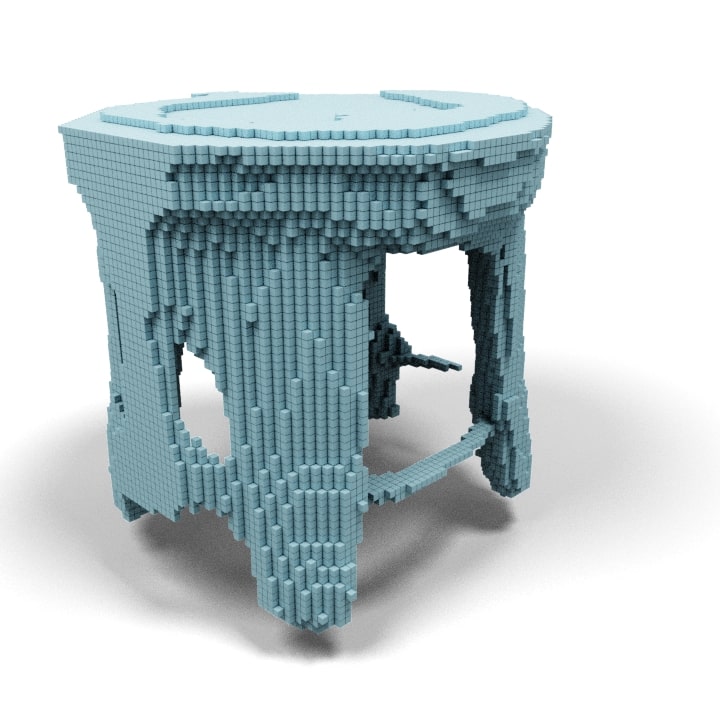} &
\includegraphics[width=0.15\linewidth]{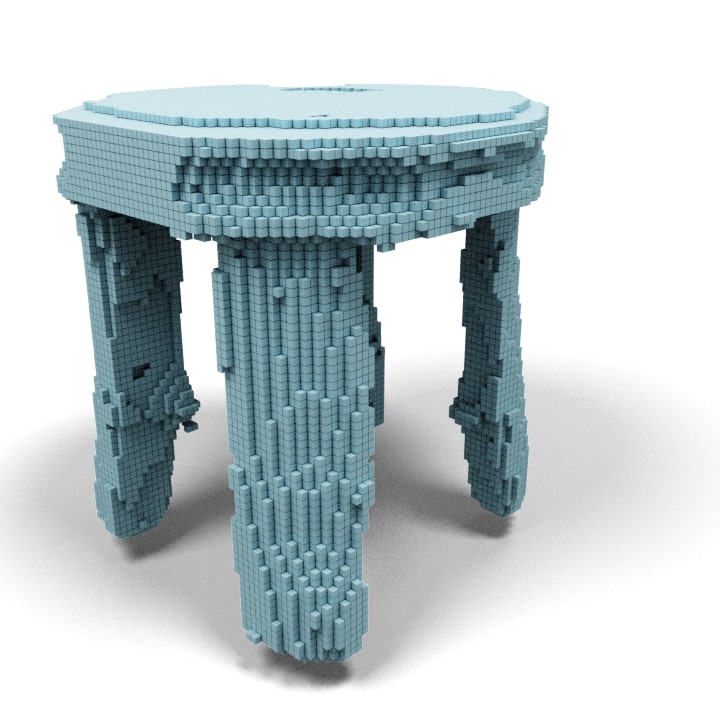}\\
\multicolumn{3}{c}{``a rocking chair''} & \multicolumn{3}{c}{``a round stool''}\\
\includegraphics[width=0.15\linewidth]{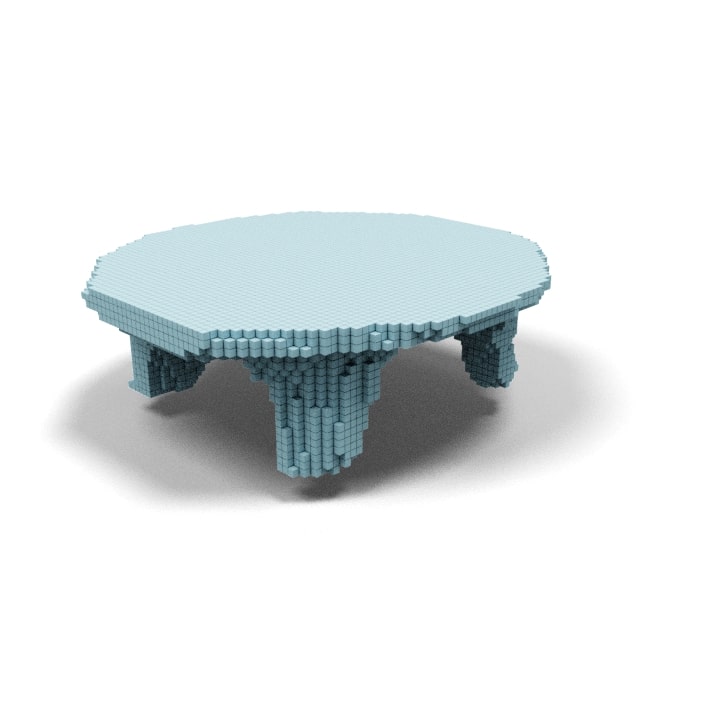} &
\includegraphics[width=0.15\linewidth]{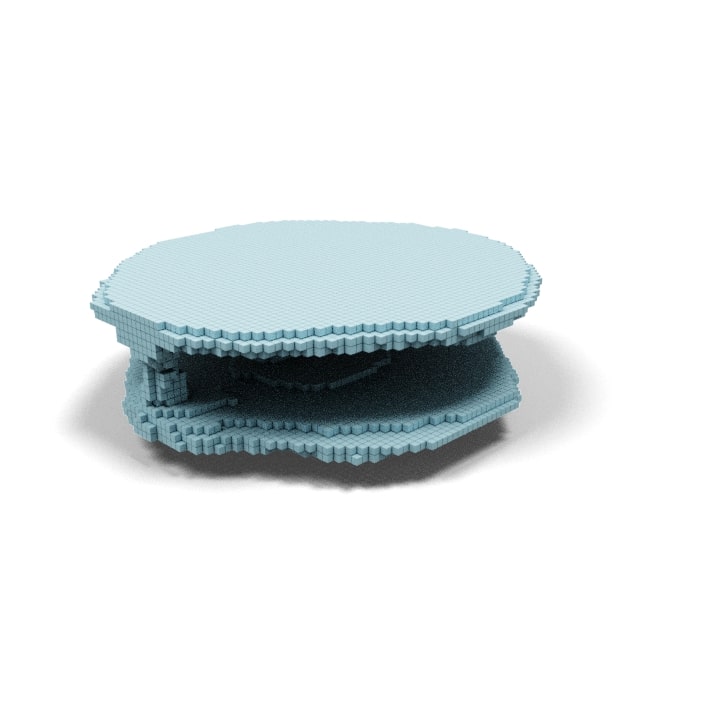} &
\includegraphics[width=0.15\linewidth]{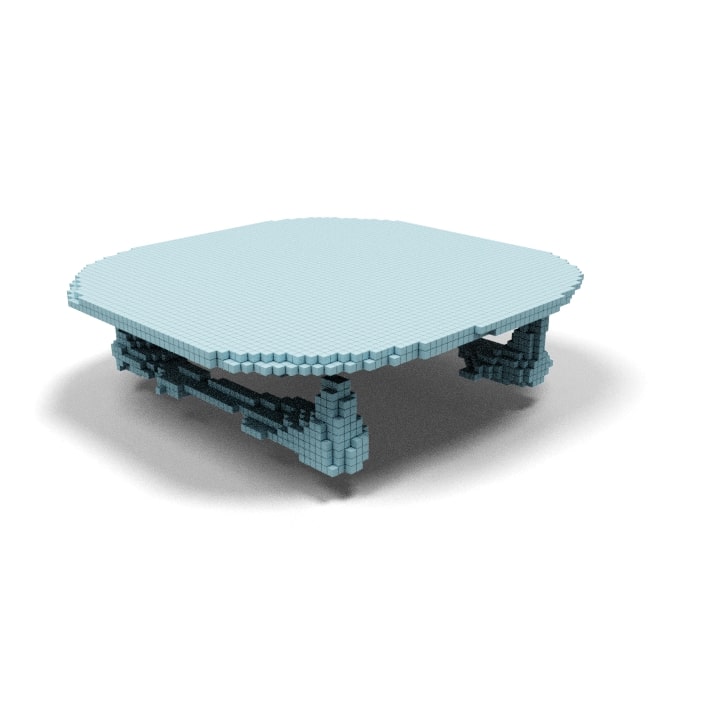} &
\includegraphics[width=0.15\linewidth]{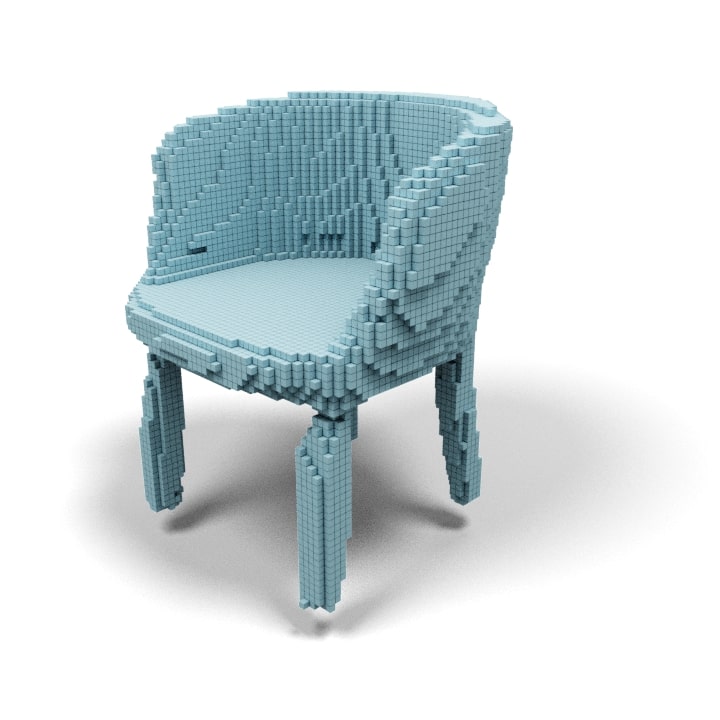} &
\includegraphics[width=0.15\linewidth]{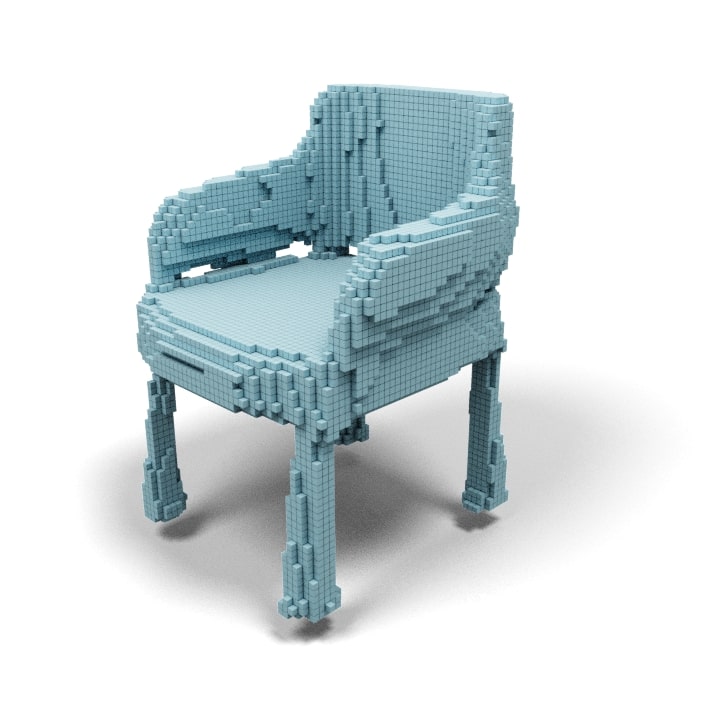} &
\includegraphics[width=0.15\linewidth]{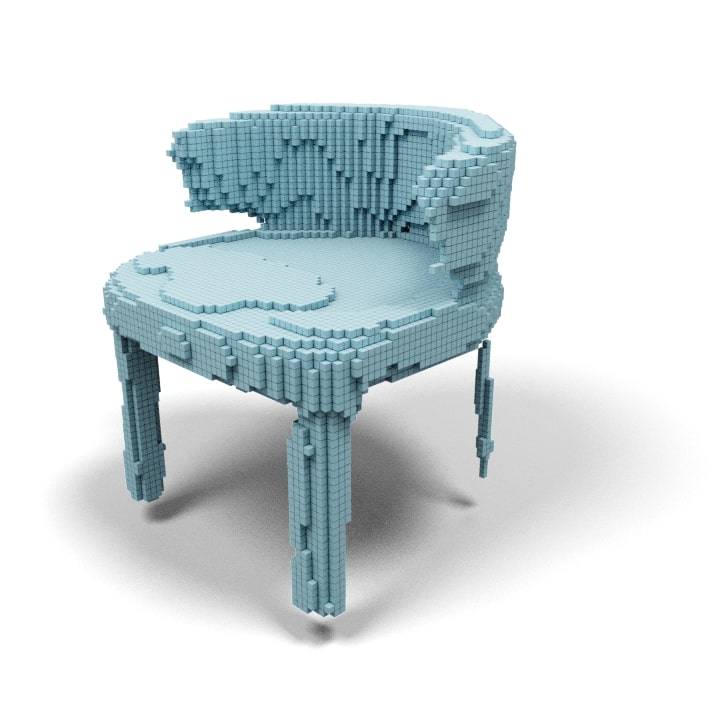}\\
\multicolumn{3}{c}{``a round table''} & \multicolumn{3}{c}{``a round chair''}\\
\end{tabular}
}
\end{center}
  \caption{Additional results for multiple shapes generation.}
\label{fig:qual_multiple}
\end{figure*}

\begin{figure*}[t!]
\begin{center}
\setlength{\tabcolsep}{5pt}
\small{
\begin{tabular}{ccccc}
\includegraphics[width=0.15\linewidth]{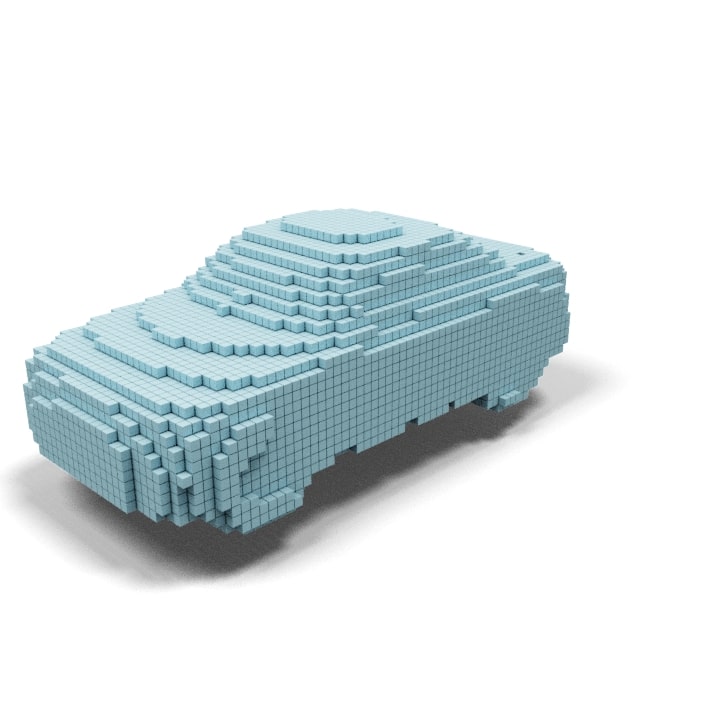} &
\includegraphics[width=0.15\linewidth]{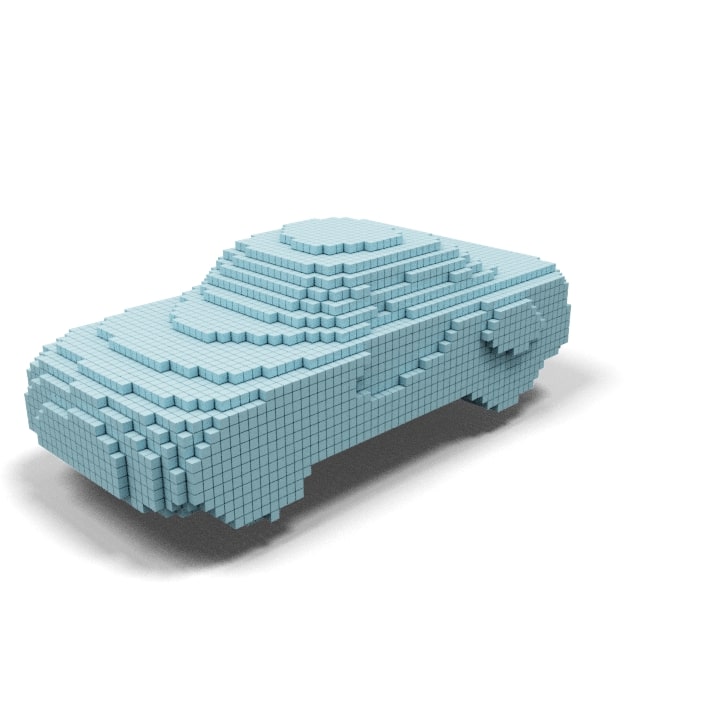} &
\includegraphics[width=0.15\linewidth]{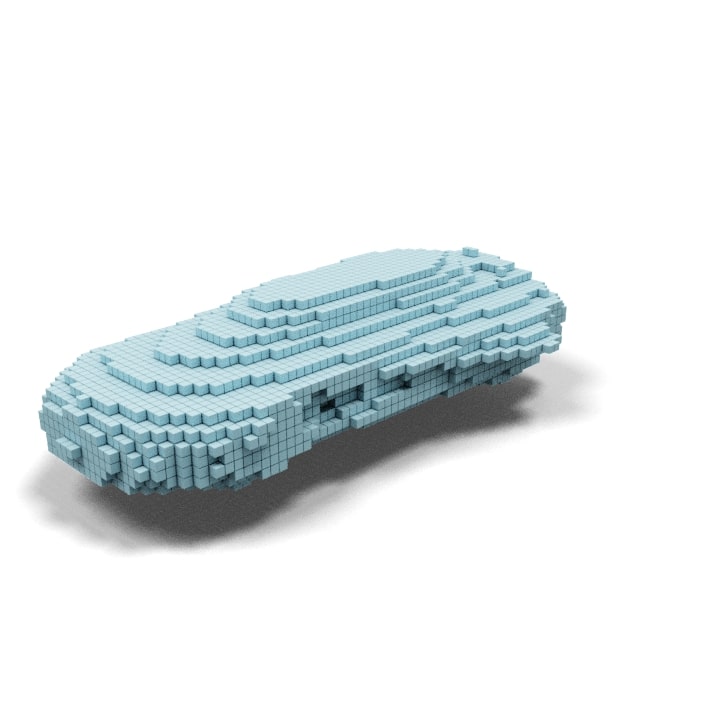} &
\includegraphics[width=0.15\linewidth]{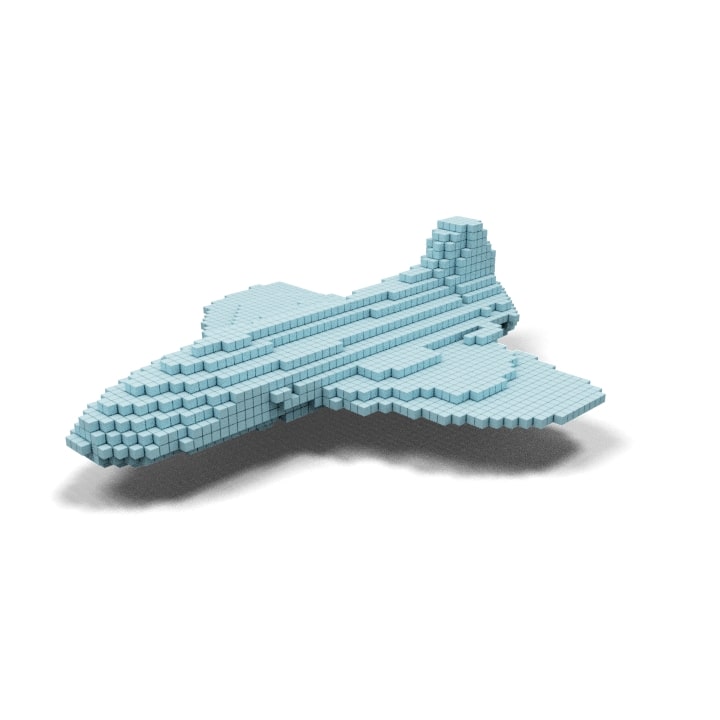} &
\includegraphics[width=0.15\linewidth]{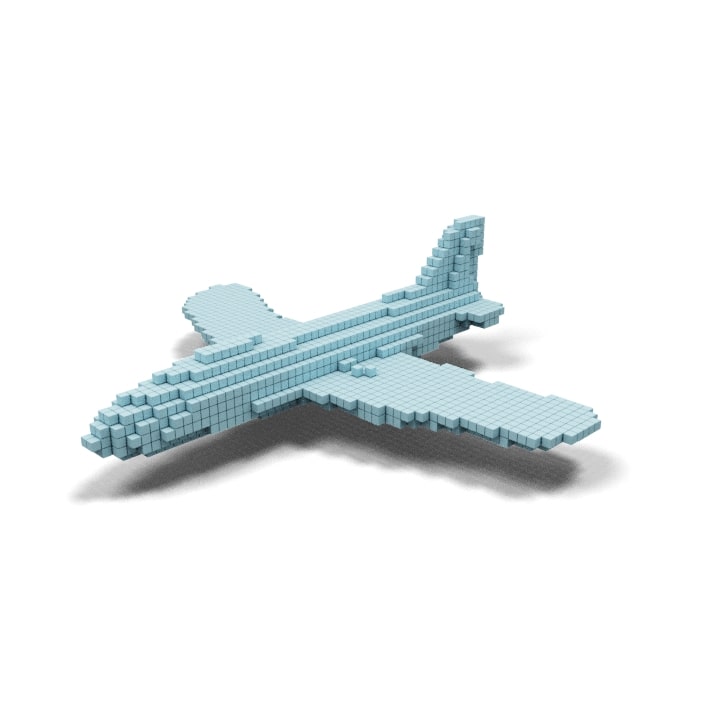}\\
\multicolumn{5}{c}{``a car'' $\rightarrow$ ``a plane''}\\

\includegraphics[width=0.15\linewidth]{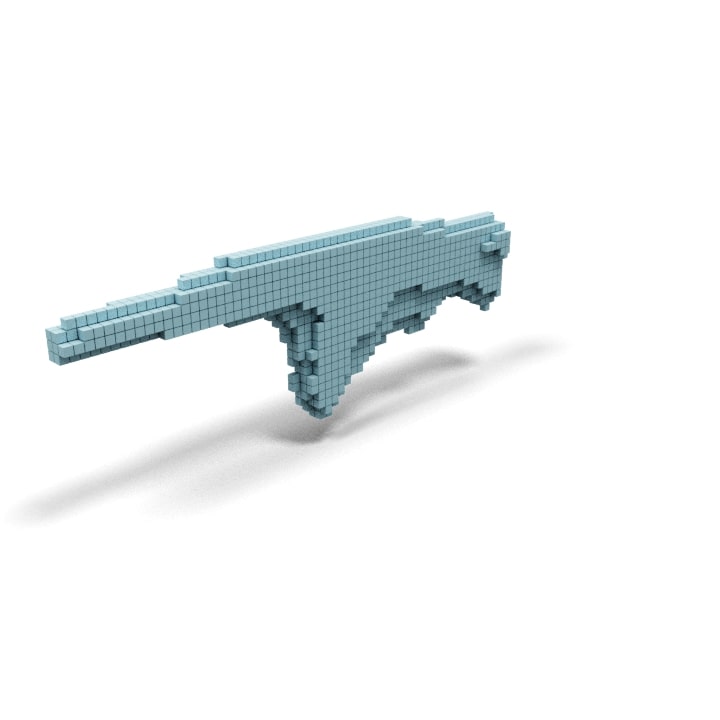} &
\includegraphics[width=0.15\linewidth]{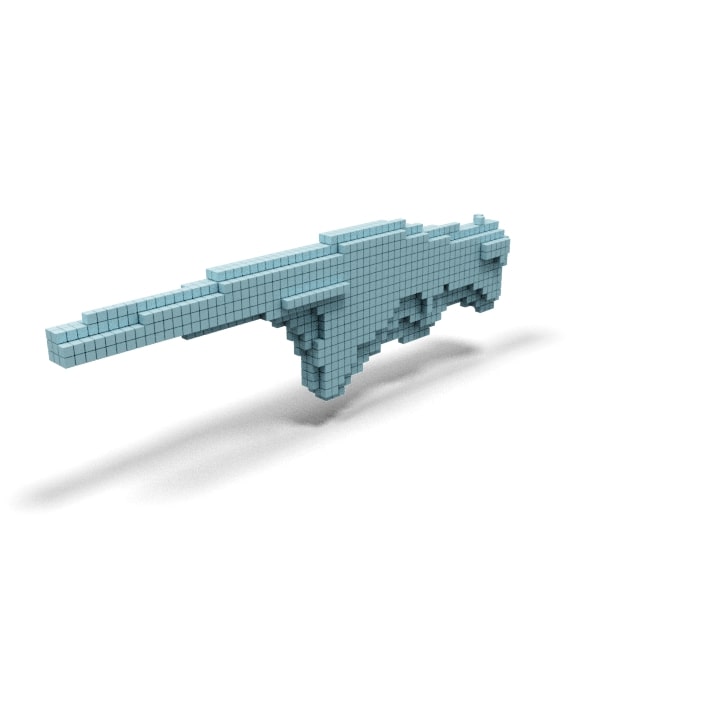} &
\includegraphics[width=0.15\linewidth]{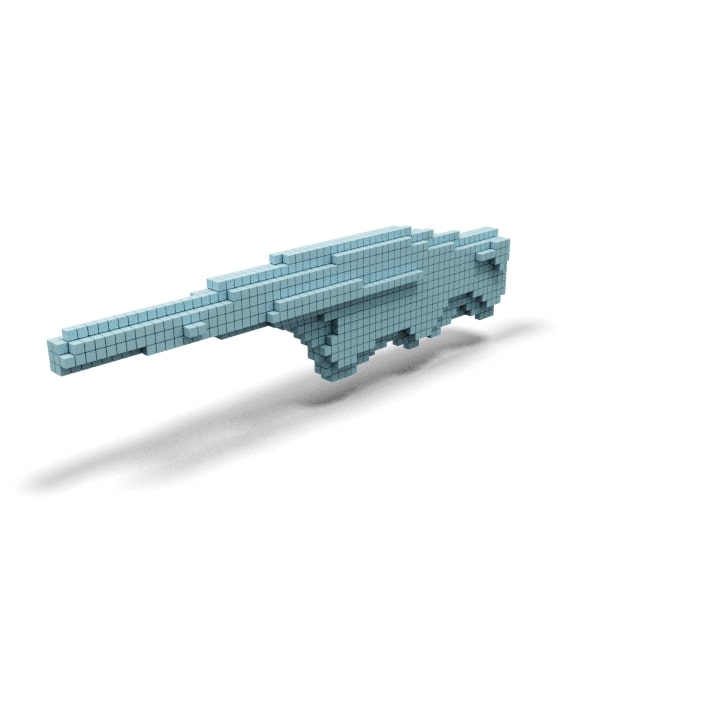} &
\includegraphics[width=0.15\linewidth]{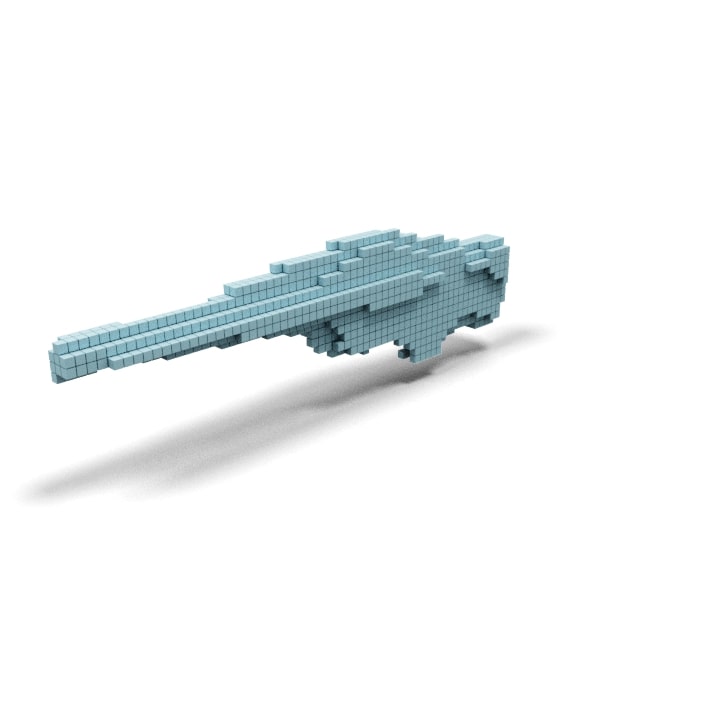} &
\includegraphics[width=0.15\linewidth]{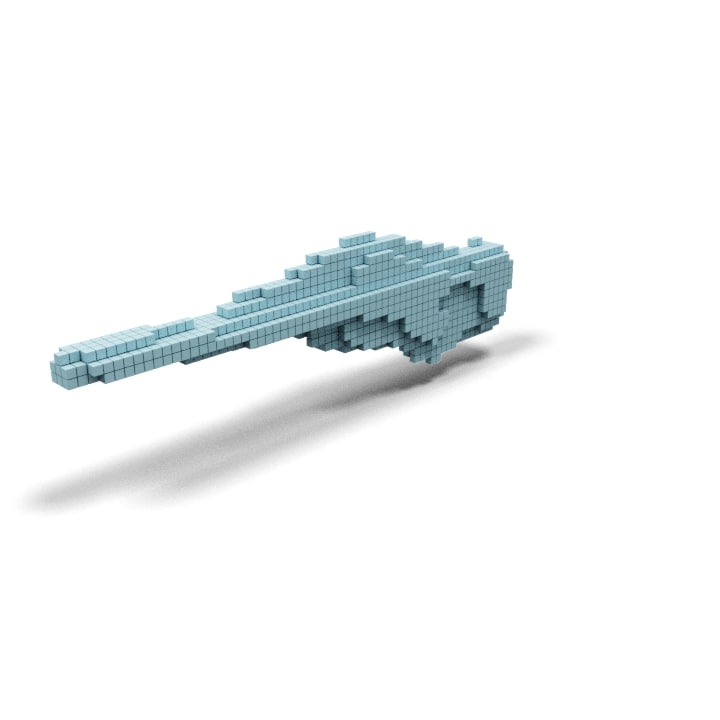}\\
\multicolumn{5}{c}{``an ak-47'' $\rightarrow$ ``a rifle''}\\
\includegraphics[width=0.15\linewidth]{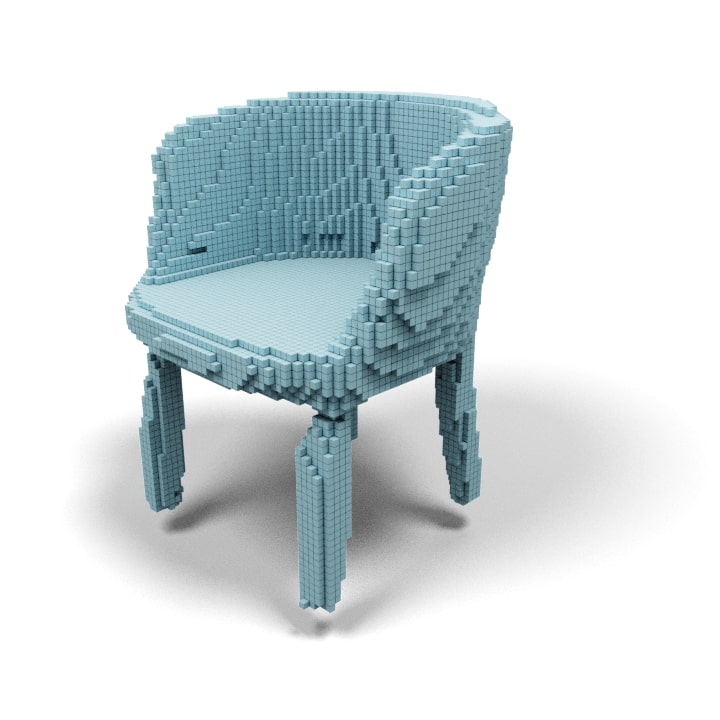} &
\includegraphics[width=0.15\linewidth]{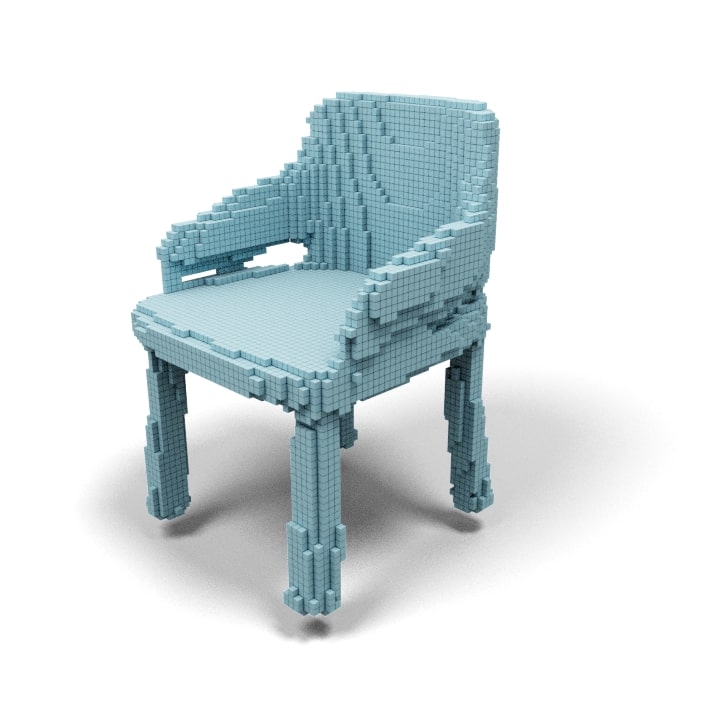} &
\includegraphics[width=0.15\linewidth]{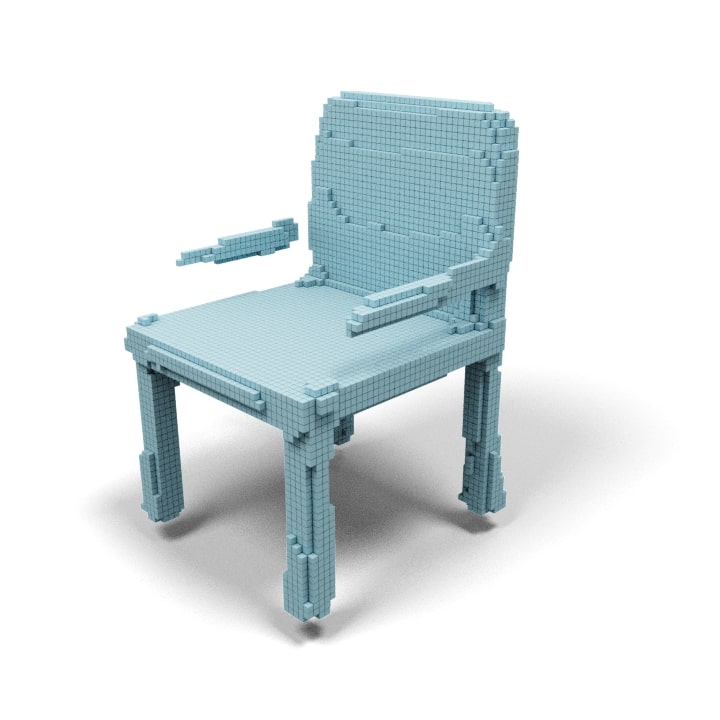} &
\includegraphics[width=0.15\linewidth]{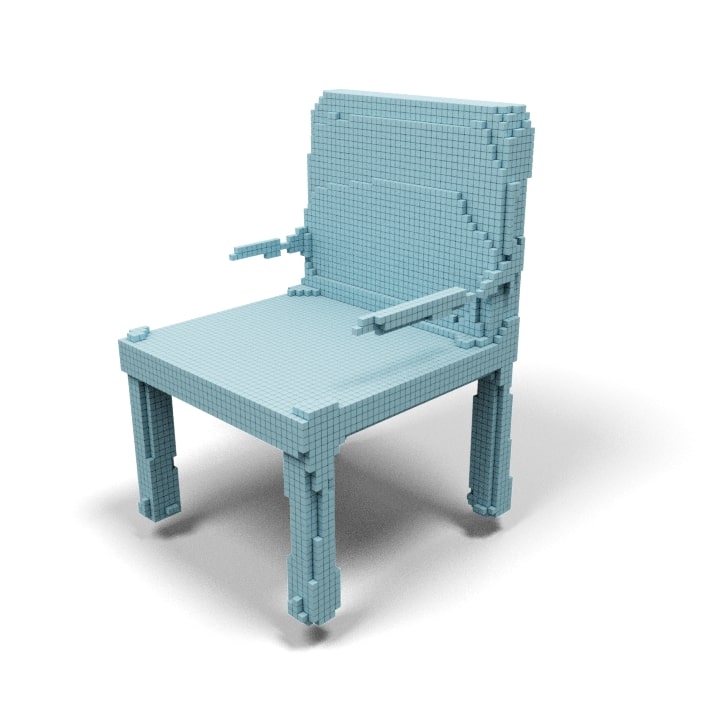} &
\includegraphics[width=0.15\linewidth]{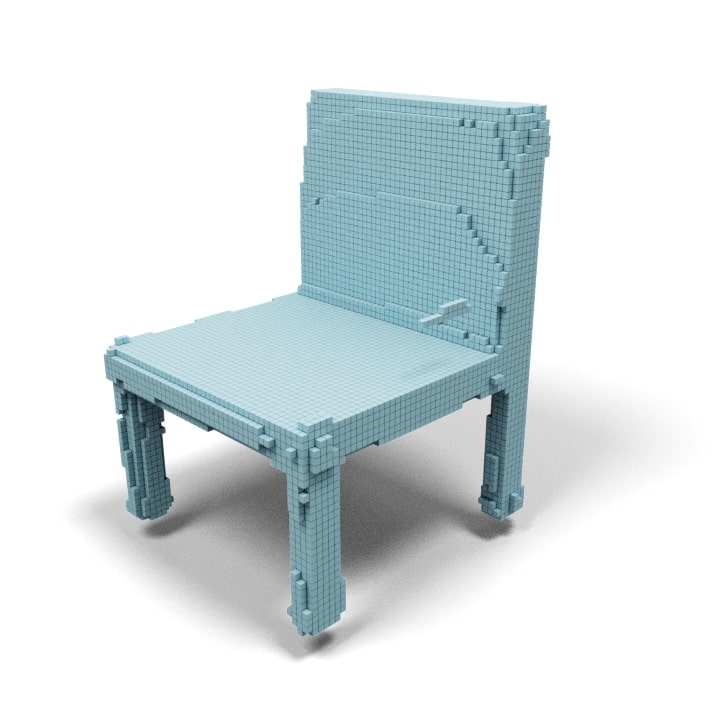}\\
\multicolumn{5}{c}{``a round chair'' $\rightarrow$ ``a rectangular chair''}\\
\end{tabular}
}
\end{center}
\caption{Additional Interpolation results between two text queries.}
\label{fig:interp}
\end{figure*}

\begin{figure*}[t!]
\begin{center}
\setlength{\tabcolsep}{13pt}
\small{
\begin{tabular}{cccccc}
\includegraphics[width=0.10\linewidth]{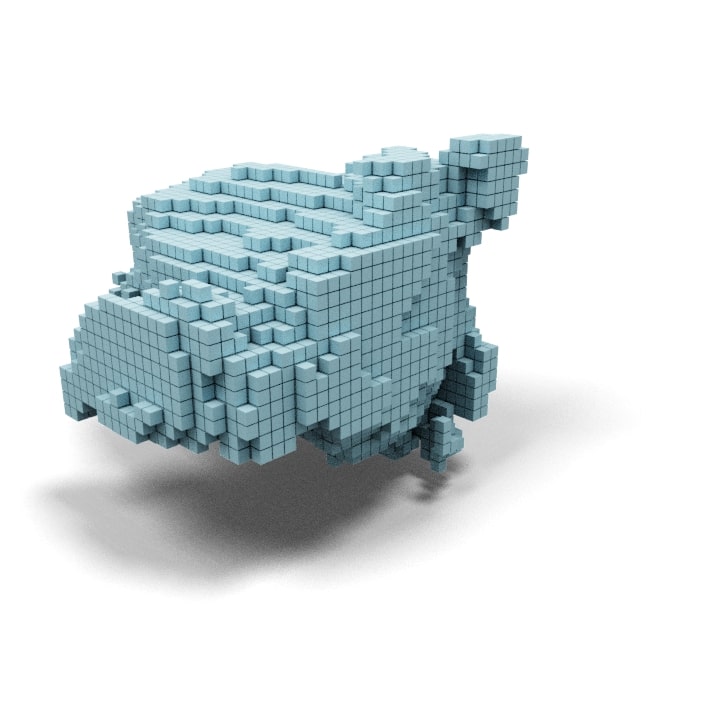} &
\includegraphics[width=0.10\linewidth]{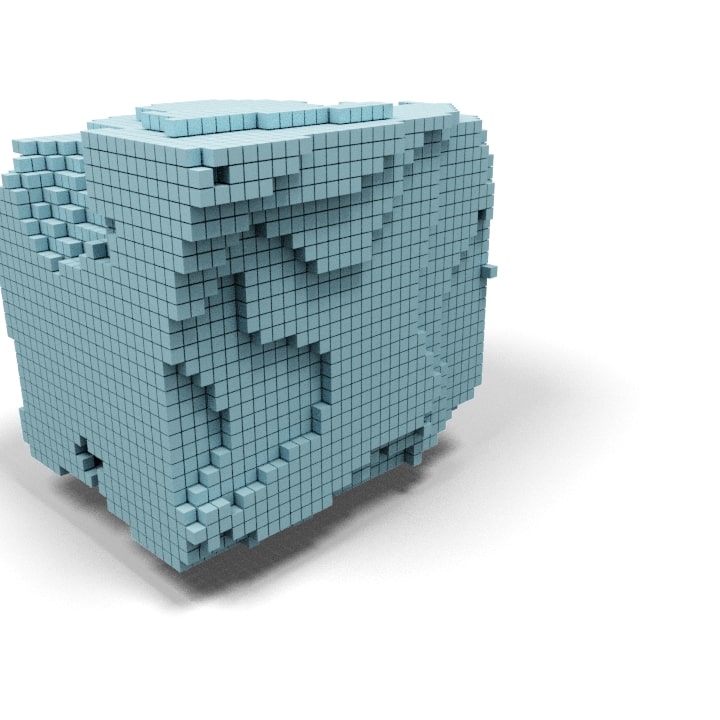} &
\includegraphics[width=0.10\linewidth]{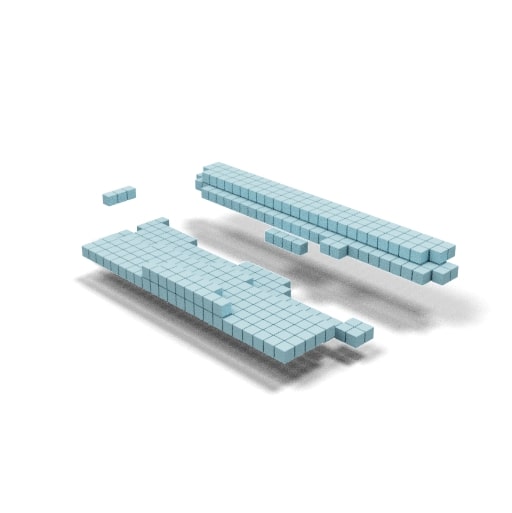} &
\includegraphics[width=0.10\linewidth]{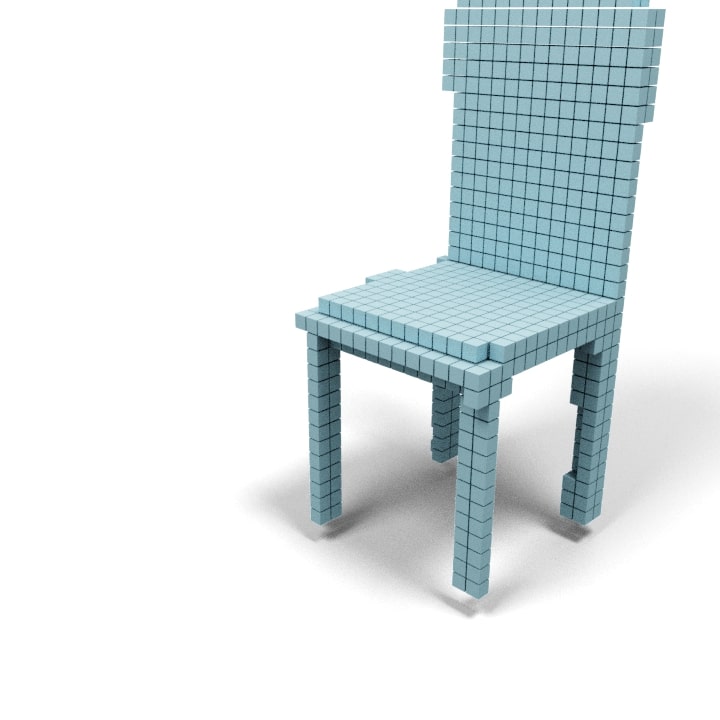} &
\includegraphics[width=0.10\linewidth]{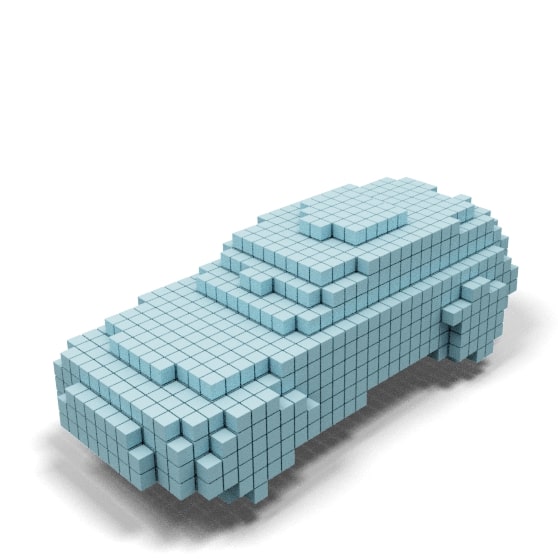} &
\includegraphics[width=0.10\linewidth]{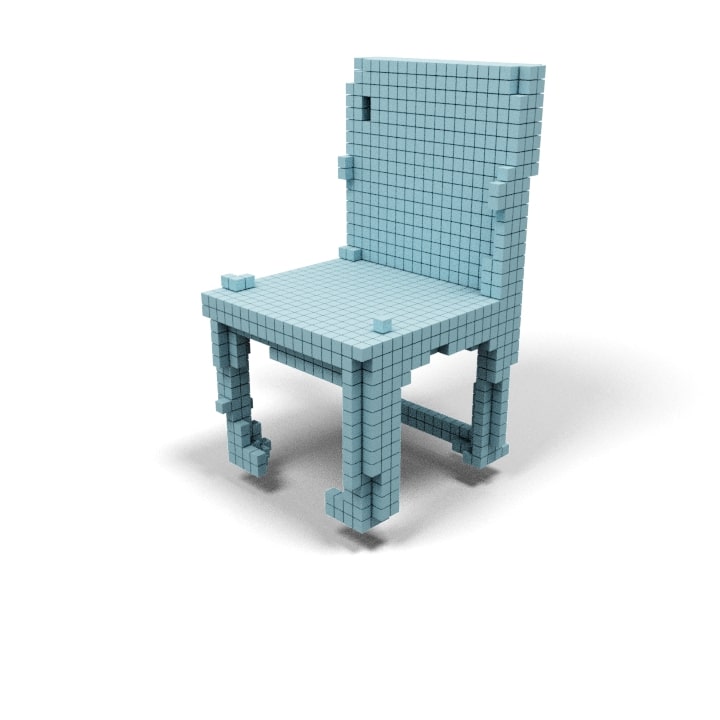}\\
\multicolumn{2}{c}{CMA-T2S} & \multicolumn{2}{c}{supervised-T2S} & \multicolumn{2}{c}{supervised-SN13}\\
\end{tabular}
}
\end{center}
  \caption{Qualitative results for supervised baselines using ``a sports car'' and ``a vertical back chair''.}
\label{fig:baselines}
\end{figure*}

\section{Human Perceptual Evaluation}
\label{sec:human_eval_supp}
\begin{figure*}[t!]
\begin{center}
\setlength{\tabcolsep}{10pt}
\small{
\begin{tabular}{ccc}
\includegraphics[width=0.2\linewidth]{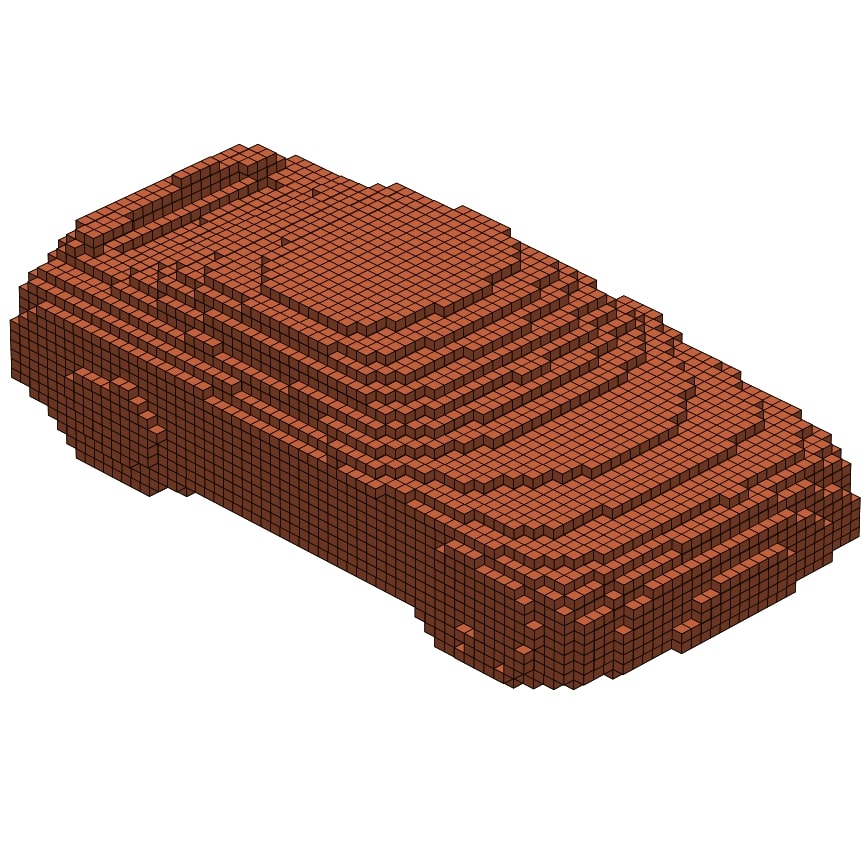} & 
\includegraphics[width=0.2\linewidth]{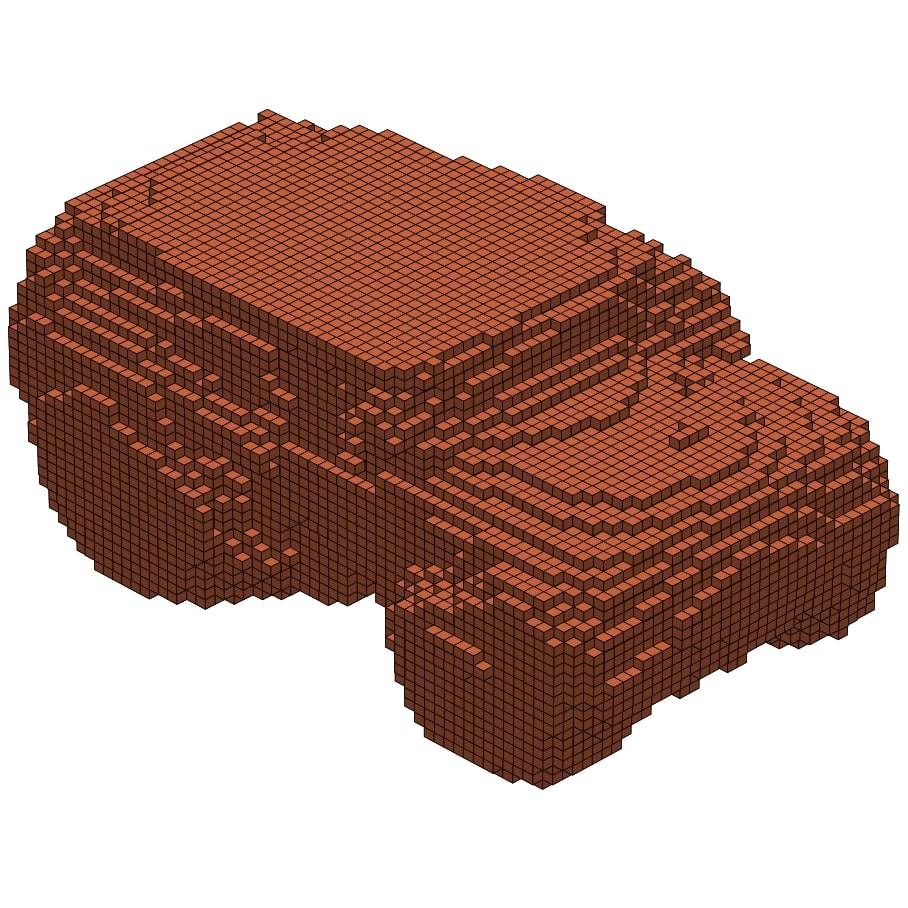} & 
\includegraphics[width=0.2\linewidth]{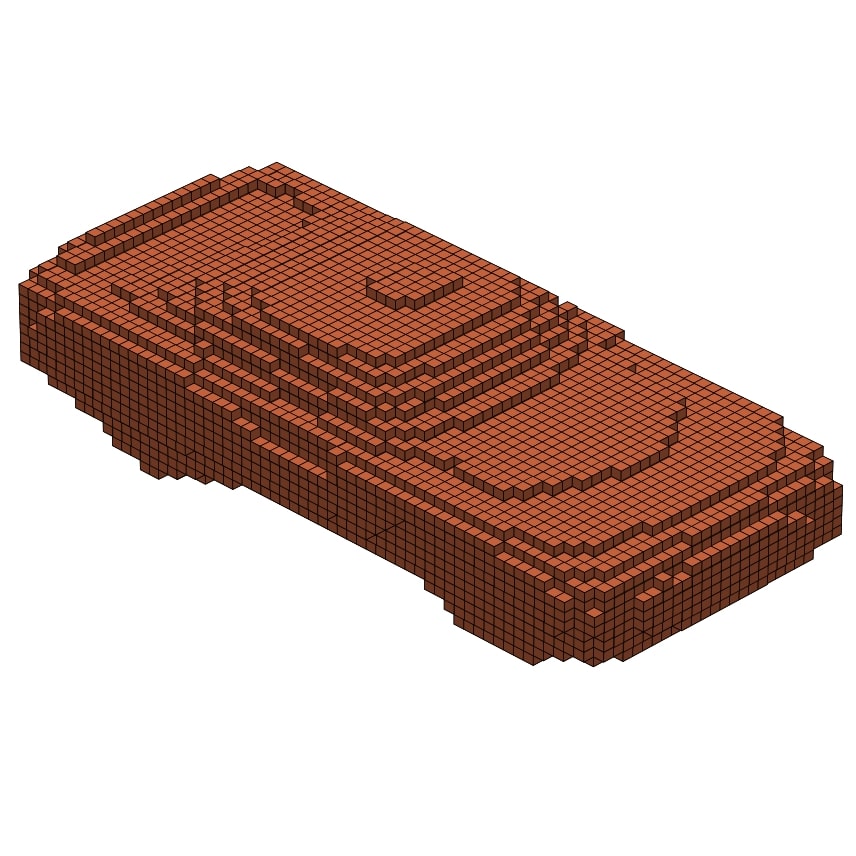} \\
``A car'' & ``A monster truck'' 8/9 & ``A muscle car'' 2/9 \\
\includegraphics[width=0.2\linewidth]{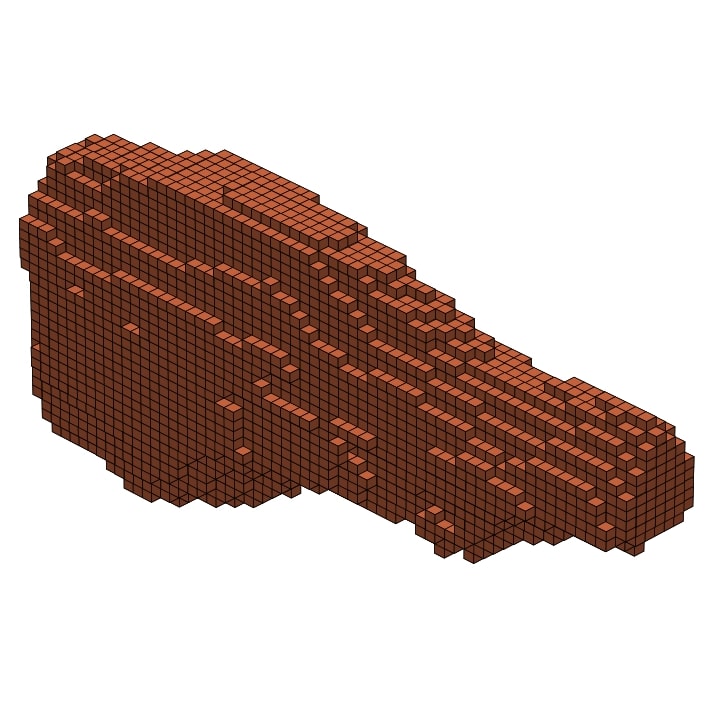} &
\includegraphics[width=0.2\linewidth]{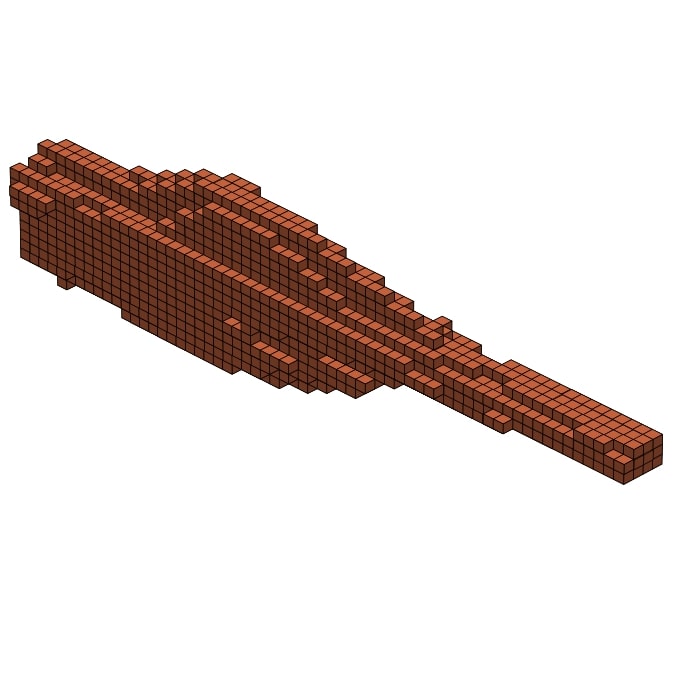} &
\includegraphics[width=0.2\linewidth]{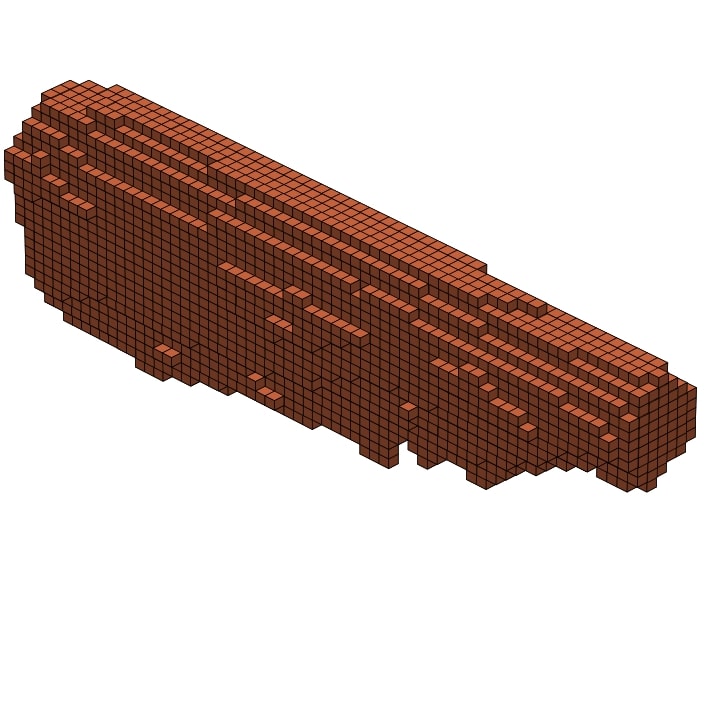} \\
``A gun'' & ``A sniper rifle'' 8/9 & ``A shotgun'' 3/9 \\  \\
\includegraphics[width=0.2\linewidth]{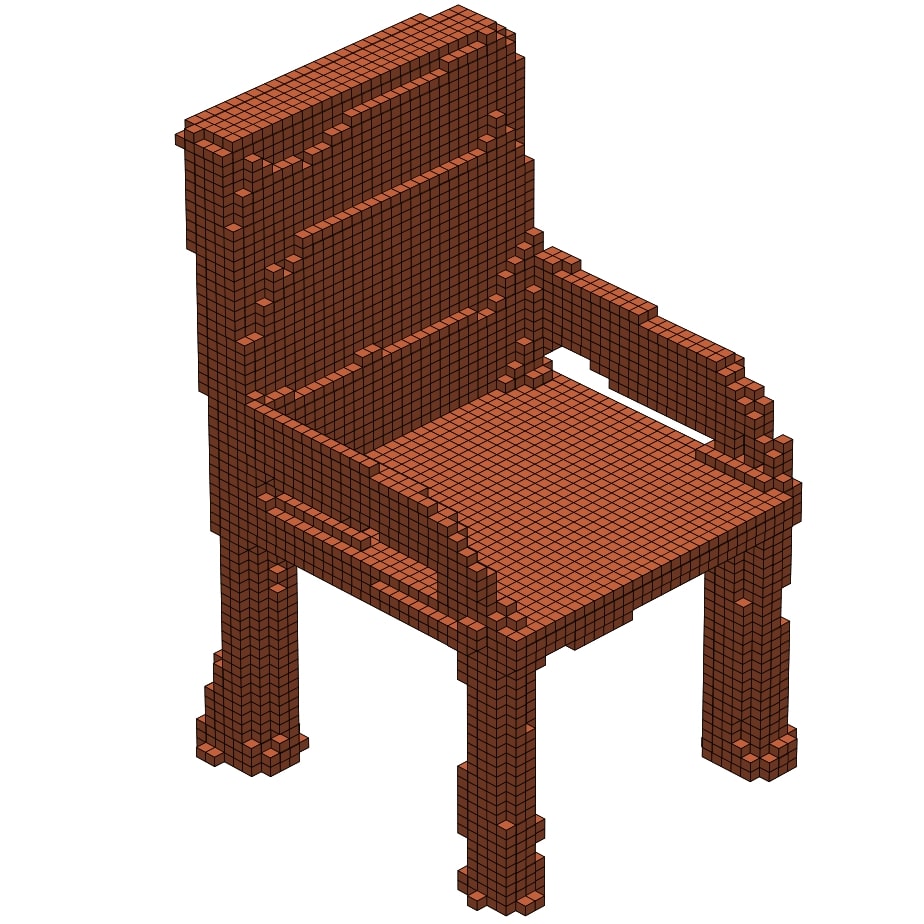} &
\includegraphics[width=0.2\linewidth]{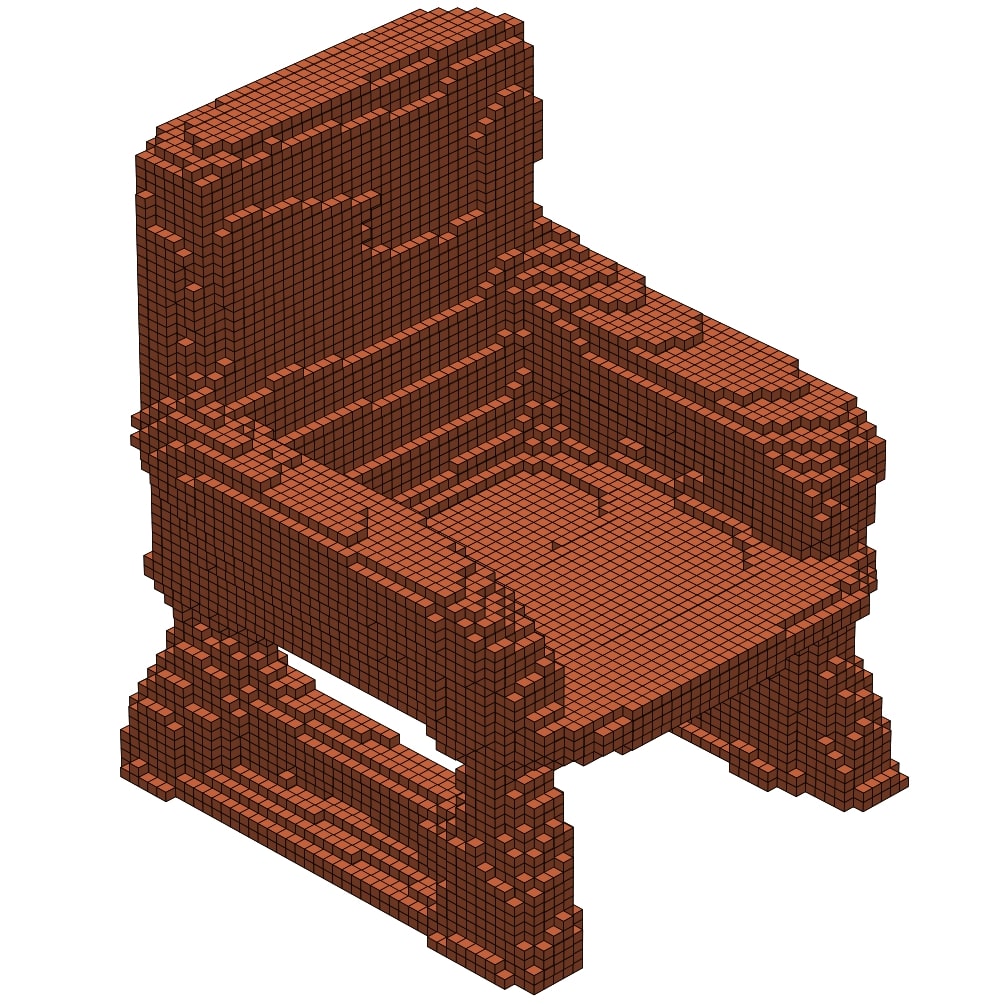} &
\includegraphics[width=0.2\linewidth]{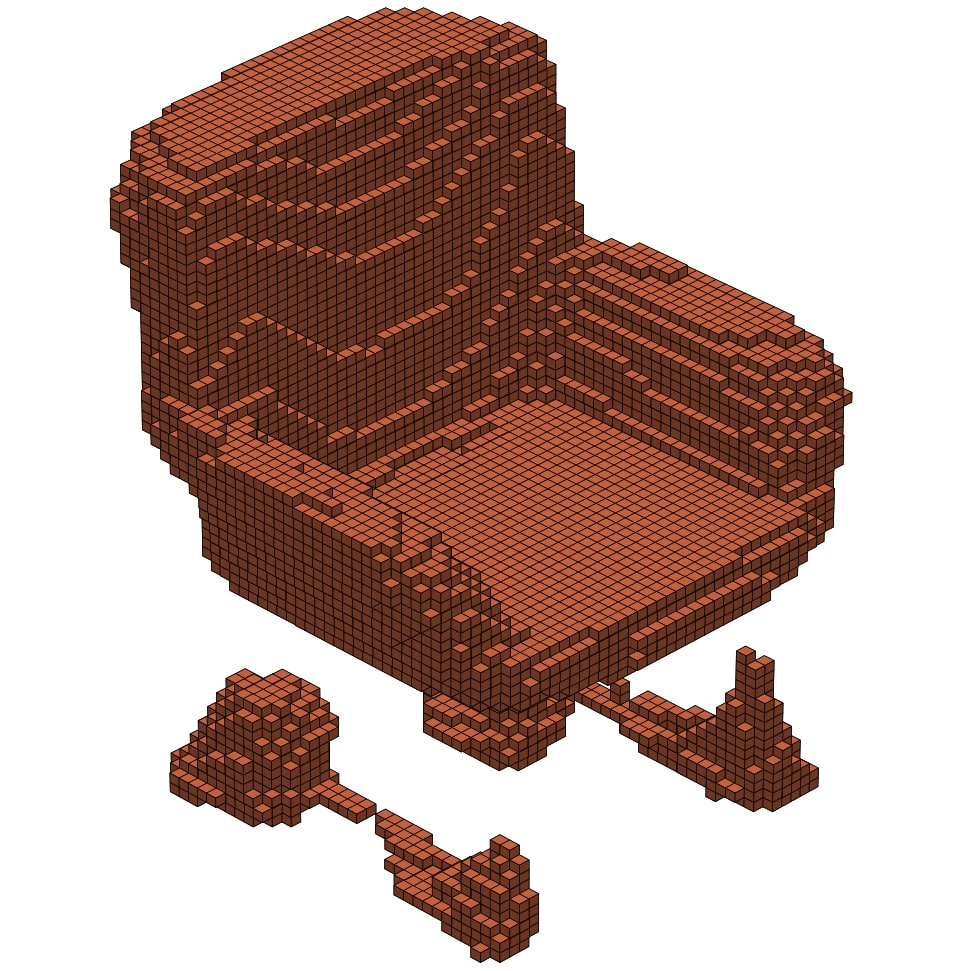} \\
``A chair'' & ``A wheelchair'' 8/9 & ``A swivel chair'' 4/9 \\
\includegraphics[width=0.2\linewidth]{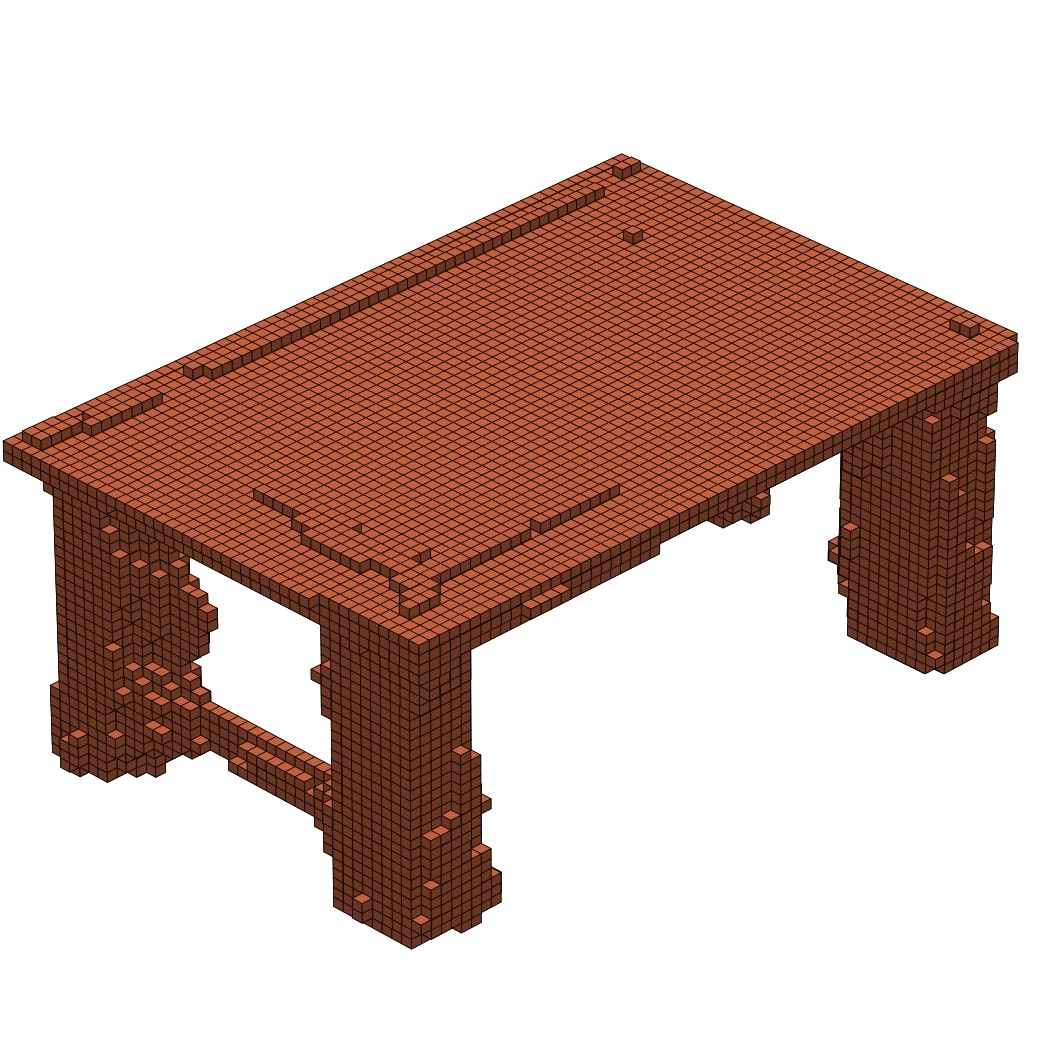} &
\includegraphics[width=0.2\linewidth]{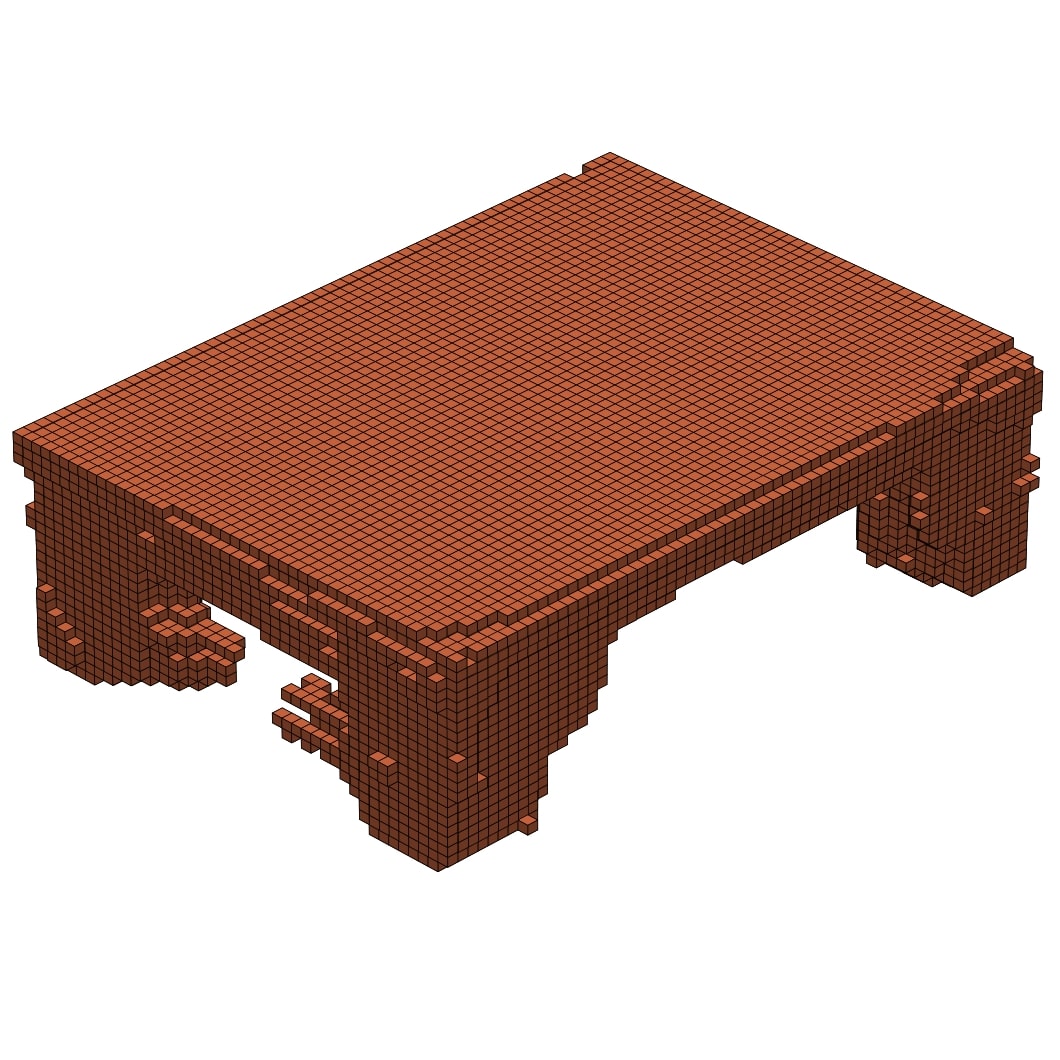} &
\includegraphics[width=0.2\linewidth]{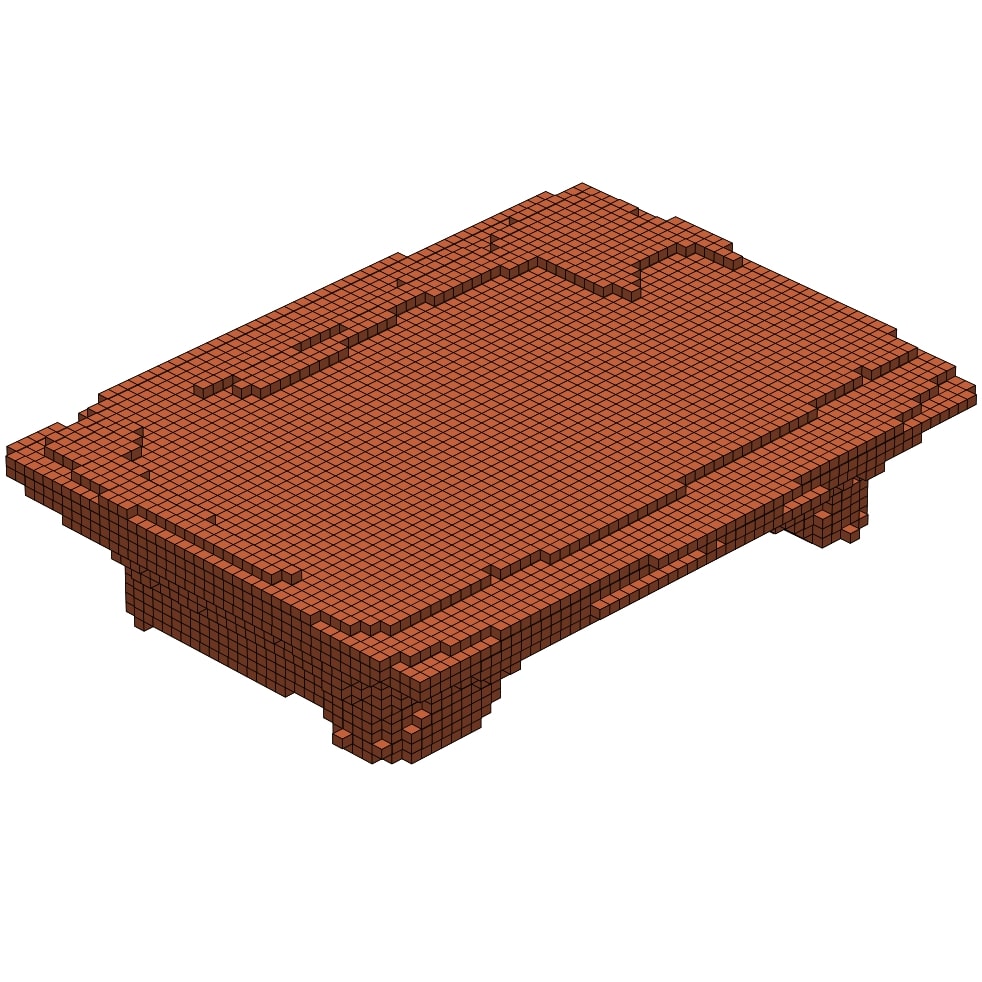} \\
``A table'' & ``A refactory table'' 5/9 & ``A billiard table'' 2/9 \\
\includegraphics[width=0.2\linewidth]{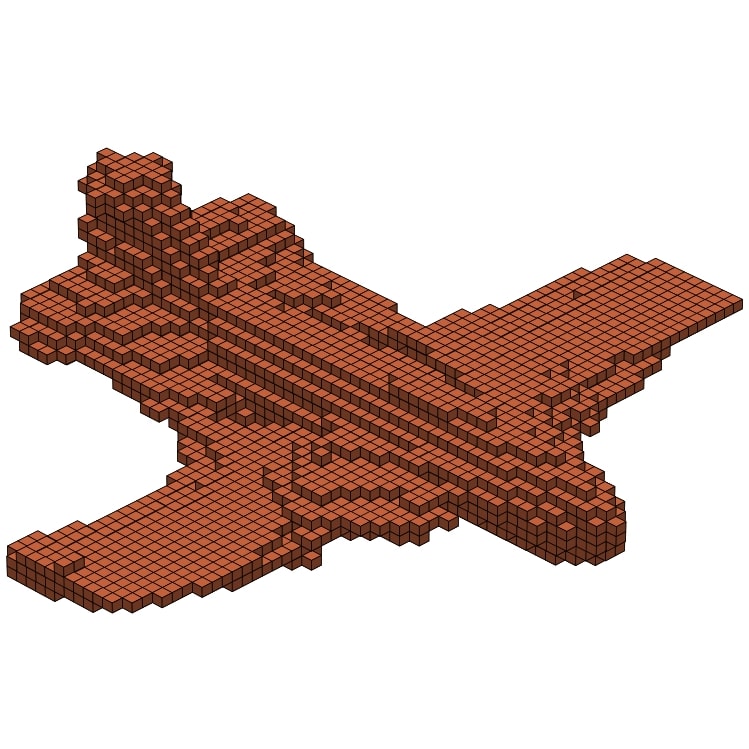} &
\includegraphics[width=0.2\linewidth]{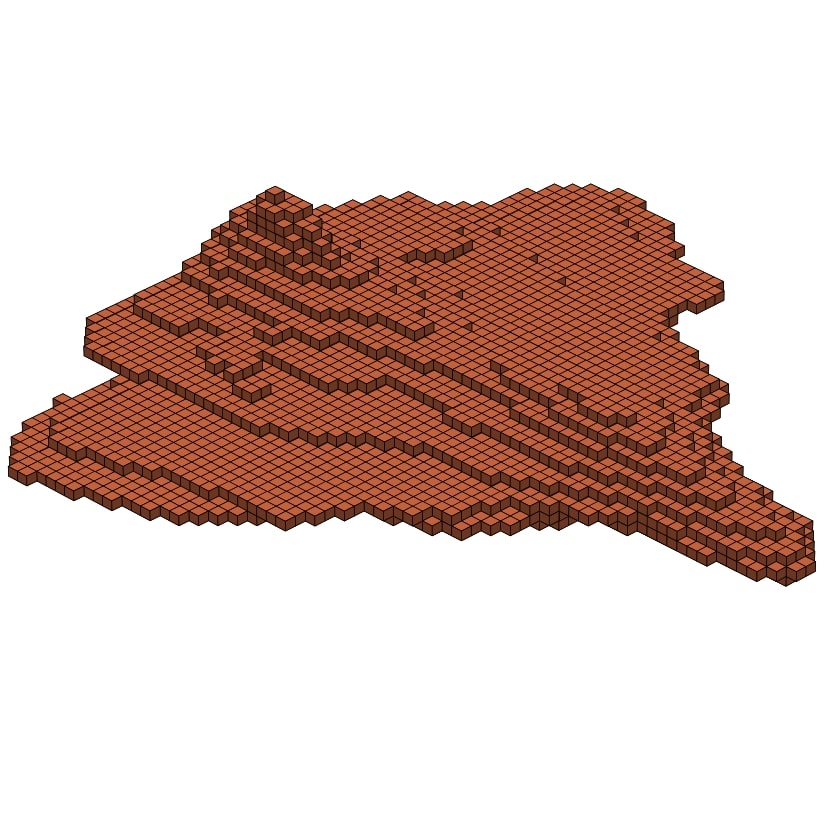} &
\includegraphics[width=0.2\linewidth]{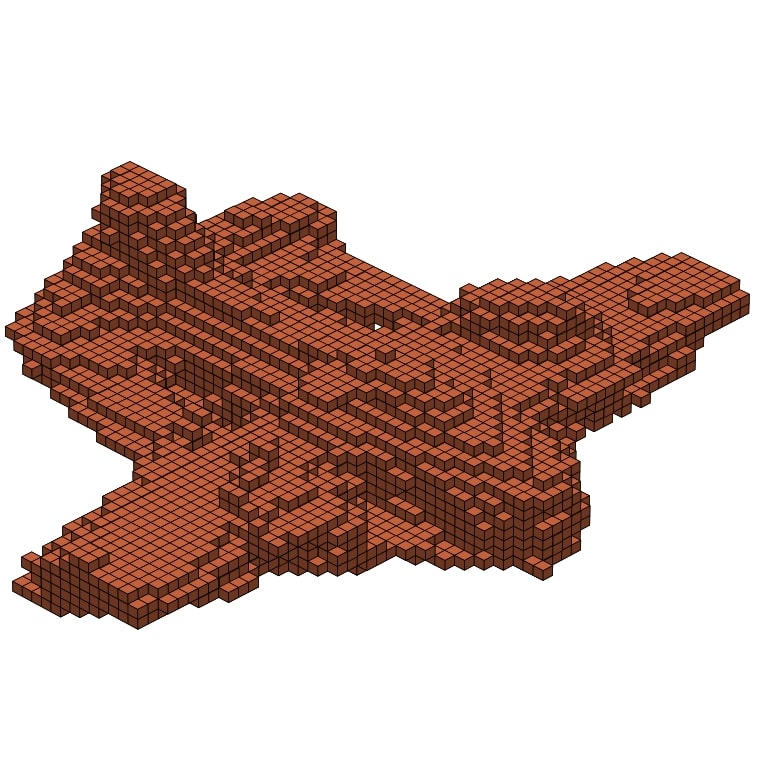} \\
``An airplane'' & ``A fighter plane'' 9/9 & ``A seaplane'' 2/9 
\end{tabular}
}
\caption{Images shown to the crowd workers in the human evaluation.  The first column shows results generated using the ShapeNet(v2) category name.  The second column shows examples of models which the crowd workers found easiest to identify based on the detailed text prompt and the third column shows the hardest.  The fraction of the nine crowd workers who chose the model generated with the detailed text prompt is also shown.}
\label{fig:detailed_human_eval}
\end{center}
\end{figure*}

In the human perceptual evaluation described in section 4.3 of the main paper, crowd workers recruited through Amazon Mechanical Turk~\cite{mishra_2019} were shown pairs of images, one generated from the ShapeNet(v2) category name (see the first column of Figure \ref{fig:detailed_human_eval})  and the other from a detailed text prompt containing either subcategory or attribute information.  The crowd workers were shown the detailed text prompt and asked to identify which of the two images it best describes.  Nine crowd workers viewed each image pair and we record the number of times the model from the detailed text prompt is selected.   For each detailed text prompt, this gives us a score from 0 to 9 indicating how effectively Clip-Forge can produce distinctive shapes which differ from the ShapeNet categories in a way which humans find semantically meaningful.  In Table~\ref{tab:text_query} the human evaluation scores are shown as colors for each query text for which the evaluation was conducted.  Figure \ref{fig:detailed_human_eval} shows a few examples in more detail.  The second column of Figure \ref{fig:detailed_human_eval} shows text prompts which produced distinctive shapes and the third column shows cases where the shapes were not as easily identified based on the text.  
We see that when the prompt elicited a very distinctive shape (`A monster truck'', ``A fighter plane'') a high fraction of the human raters were able to identify the correct model.   
In some cases the low score reflects a lack of resolution (for example  ``A swivel chair'', ``A billiard table'' and ``A seaplane'').   In the case of ``A wheelchair'', Clip-Forge was unable to generate round wheels, but as the bottom of the legs were joined up this gave enough of an impression of wheels for humans to select the model.  In the case of ``A muscle car'' Clip-Forge attempted to create the shape of a low form of a sports car, however the shape was not far enough from the generic car for the crowd workers to select it.

{
    \small
    \bibliographystyle{ieee_fullname}
    \bibliography{macros,main}
}


\end{document}